\documentclass[letterpaper]{article}
\usepackage[margin=1in,dvips]{geometry}
\usepackage{graphicx,psfrag,amsmath,amsthm,amssymb}
\usepackage{natbib}
\usepackage{url,color}
\usepackage{algorithm,algorithmic}

\newcommand{\B}{\ensuremath{\mathbf{B}}}
\newcommand{\C}{\ensuremath{\mathbf{C}}}
\newcommand{\D}{\ensuremath{\mathbf{D}}}

\newcommand{\G}{\ensuremath{\mathbf{G}}}

\newcommand{\LL}{\ensuremath{\mathbf{L}}}

\newcommand{\PP}{\ensuremath{\mathbf{P}}}

\newcommand{\RR}{\ensuremath{\mathbf{R}}}

\newcommand{\W}{\ensuremath{\mathbf{W}}}

\newcommand{\Y}{\ensuremath{\mathbf{Y}}}
\newcommand{\Z}{\ensuremath{\mathbf{Z}}}

\renewcommand{\c}{\ensuremath{\mathbf{c}}}

\newcommand{\f}{\ensuremath{\mathbf{f}}}

\newcommand{\q}{\ensuremath{\mathbf{q}}}

\newcommand{\x}{\ensuremath{\mathbf{x}}}
\newcommand{\y}{\ensuremath{\mathbf{y}}}
\newcommand{\z}{\ensuremath{\mathbf{z}}}
\newcommand{\0}{\ensuremath{\mathbf{0}}}
\newcommand{\1}{\ensuremath{\mathbf{1}}}

\newcommand{\bbR}{\ensuremath{\mathbb{R}}}

\newcommand{\calO}{\ensuremath{\mathcal{O}}}

\newcommand{\norm}[1]{\left\lVert#1\right\rVert}

\newcommand{\caja}[4][1]{{%
    \renewcommand{\arraystretch}{#1}%
    \begin{tabular}[#2]{@{}#3@{}}%
      #4%
    \end{tabular}%
    }}

{%
\begin{list}{#1}{
\vspace{-\topsep}
\vspace{-\partopsep}
\setlength{\itemindent}{0cm}
\setlength{\rightmargin}{0cm}
\setlength{\listparindent}{0cm}
\settowidth{\labelwidth}{#1}
\setlength{\leftmargin}{\labelwidth}
\addtolength{\leftmargin}{\labelsep}
\setlength{\itemsep}{0cm}
}%
}%
{%
\end{list}
\vspace{-\topsep}
\vspace{-\partopsep}
}

{\begin{enumerate}%
}%
{\end{enumerate}}

\newcommand{\diagop}{\operatorname{diag}}
\newcommand{\diag}[1]{\ensuremath{\diagop\left(#1\right)}}
\newcommand{\traceop}{\operatorname{tr}}
\newcommand{\trace}[1]{\ensuremath{\traceop\left(#1\right)}}

\theoremstyle{plain}

\newtheorem*{lemma*}{Lemma}

\newtheorem*{prop*}{Proposition}

\theoremstyle{definition}

\newtheorem*{defn*}{Definition}

\newtheorem*{exmp*}{Example}

\newtheorem*{conj*}{Conjecture}

\theoremstyle{remark}

\newtheorem*{rmk*}{Remark}

\newcommand{\gausskerd}[3]{G\biggl( \norm{\frac{#1-#2}{#3}}^2 \biggr)}

\newcommand{\gausskert}[3]{G( \norm{(#1-#2)/{#3}}^2 )}

\graphicspath{{grf/}}

\bibpunct[, ]{(}{)}{;}{a}{,}{,} 

\title{The Laplacian $K$-Modes Algorithm for Clustering}
\author{
  Weiran Wang \hspace{5ex} Miguel \'A. Carreira-Perpi\~n\'an \\
  Electrical Engineering and Computer Science, University of California, Merced \\
  {\url{http://eecs.ucmerced.edu}}
}
\date{Jun 15, 2014}

\AtBeginDvi{}

\begin{document}

\maketitle

\begin{abstract}
  
  In addition to finding meaningful clusters, centroid-based clustering algorithms such as $K$-means or mean-shift should ideally find centroids that are valid patterns in the input space, representative of data in their cluster. This is challenging with data having a nonconvex or manifold structure, as with images or text. We introduce a new algorithm, Laplacian $K$-modes, which naturally combines three powerful ideas in clustering: the explicit use of assignment variables (as in $K$-means); the estimation of cluster centroids which are modes of each cluster's density estimate (as in mean-shift); and the regularizing effect of the graph Laplacian, which encourages similar assignments for nearby points (as in spectral clustering). The optimization algorithm alternates an assignment step, which is a convex quadratic program, and a mean-shift step, which separates for each cluster centroid. The algorithm finds meaningful density estimates for each cluster, even with challenging problems where the clusters have manifold structure, are highly nonconvex or in high dimension. It also provides centroids that are valid patterns, truly representative of their cluster (unlike $K$-means), and an out-of-sample mapping that predicts soft assignments for a new point.
  
\end{abstract}

\section{Introduction}
\label{s:intro}

\begin{figure}[t]
  \centering
  \begin{tabular}{@{\hspace{0\linewidth}}c@{\hspace{0\linewidth}}c@{\hspace{0\linewidth}}c@{\hspace{0\linewidth}}c@{\hspace{0\linewidth}}c@{\hspace{0\linewidth}}c@{\hspace{0\linewidth}}c@{\hspace{0.00\linewidth}}c@{\hspace{0\linewidth}}c@{\hspace{0\linewidth}}c@{\hspace{0\linewidth}}}
    \multicolumn{7}{@{}c@{}}{\dotfill data\dotfill} & $K$-means & $K$-modes & GMS \\
    \includegraphics[width=0.10\linewidth,bb=275 367 327 419,clip]{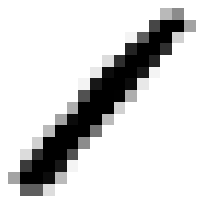} &
    \includegraphics[width=0.10\linewidth,bb=275 367 327 419,clip]{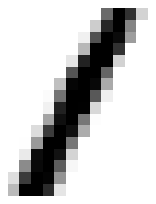} &
    \includegraphics[width=0.10\linewidth,bb=275 367 327 419,clip]{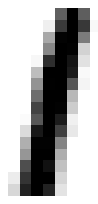} &
    \includegraphics[width=0.10\linewidth,bb=275 367 327 419,clip]{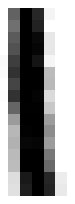} &
    \includegraphics[width=0.10\linewidth,bb=275 367 327 419,clip]{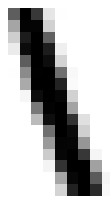} &
    \includegraphics[width=0.10\linewidth,bb=275 367 327 419,clip]{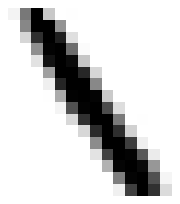} &
    \includegraphics[width=0.10\linewidth,bb=275 367 327 419,clip]{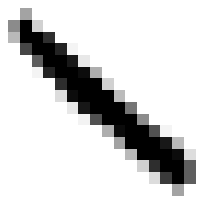} &
    \includegraphics[width=0.10\linewidth,bb=275 367 327 419,clip]{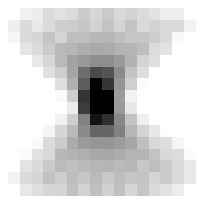} & 
    \includegraphics[width=0.10\linewidth,bb=275 367 327 419,clip]{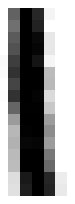} & 
    \includegraphics[width=0.10\linewidth,bb=275 367 327 419,clip]{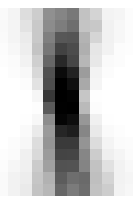} 
  \end{tabular}
  \caption{A cluster of 7 rotated-1 USPS digit images and the centroids found by $K$-means, $K$-modes (both with $K=1$) and mean-shift (with $\sigma$ so there is one mode). Example taken from \citet{CarreirWang13a}.}
  \label{f:rotated1}
\end{figure}

Given a dataset $\x_1,\dots,\x_N \in \bbR^D$, centroid-based clustering algorithms such as $K$-means \citep{Bishop06a} and mean-shift \citep{FukunagHostet75a,Cheng95a,Carreir00b,ComanicMeer02a} estimate a representative $\c_k \in \bbR^D$ of each cluster $k$ in addition to assigning data points to clusters. Besides finding meaningful clusters, we would ideally like to find centroids that are valid patterns in the input space, representative of data in their cluster. This is challenging with data having a nonconvex or manifold structure, as with images or text. Fig.~\ref{f:rotated1} illustrates this with a single cluster consisting of continuously rotated digit-1 images. Since these images represent a nonconvex cluster in the high-dimensional pixel space, their mean (which averages all orientations) is not a valid digit-1 image, which makes the centroid not interpretable and hardly representative of a digit 1. Mean-shift does not work well either: to produce a single mode, a large bandwidth is required, which makes the mode lie far from the manifold; a smaller bandwidth does produce valid digit-1 images, but then multiple modes arise for the same cluster, and under mean-shift they define each a cluster.

Forcing the centroids to be exemplars is often regarded as a way to ensure the centroids are valid patterns. Although there exist exemplar-based or $K$-medoids clustering algorithms \citep{KaufmanRousseeuw90a,Bishop06a,Hastie_09a} which constrain centroids to be points from the dataset (``exemplars'') and often minimize a $K$-means type of objective function with a possibly non-Euclidean distance, such algorithms are typically slow because updating centroid $\c_k$ requires testing all pairs of points in cluster $k$. Besides, the exemplars themselves are often noisy and thus not that representative of their neighborhood.

Remarkably, no algorithm that partitions the data using exactly $K$ meaningful modes existed in the literature until the \emph{$K$-modes algorithm} proposed by \citet{CarreirWang13a}. This combines the idea of clustering through binary assignment variables with the idea that high-density points are representative of a cluster. Each centroid found by the $K$-modes algorithm is the mode of a kernel density estimate defined by data points in each cluster. As a result, the centroids average out noise or idiosyncrasies that exist in individual data points and are representative of their cluster and neighborhood. This can be seen from the $K$-modes centroid for the rotated digit-1 problem in Fig.~\ref{f:rotated1}. $K$-modes was also shown to have nice properties such as being more robust to mis-specification of the bandwidth and to outliers. Its optimization procedure is also very efficient.

One important disadvantage of $K$-modes is that it uses the same assignment rule as $K$-means (each point is assigned to its closest centroid in Euclidean distance), so it can only find convex clusters (a Voronoi tessellation). Therefore, like $K$-means, it cannot handle clusters with nonconvex shapes or manifold structure, unlike mean-shift or spectral clustering \citep{ShiMalik00a}. The main contribution of this paper is to solve this issue, while keeping the nice properties that $K$-modes does have. The key idea is to modify the $K$-modes objective function such that the assignment rule becomes much more flexible. We then give an alternating optimization procedure to find the assignments and the modes. The resulting \emph{Laplacian $K$-modes} algorithm is able to produce for each cluster a nonparametric density and a mode as valid representative (like $K$-modes), to separate nonconvex shaped clusters (like mean-shift and spectral clustering), and to give soft assignment of data points to each cluster. Yet, all of these merits are achieved at a reasonable computational cost, and the algorithm works well with high-dimensional data.

\section{Related work}
\label{s:related}

\subsection{Centroids-based algorithms}

Given the number of clusters $K$, $K$-means \citep{Bishop06a} minimizes the objective
\begin{align}
  \label{e:kmeans-objfcn}
  \min_{\Z,\C} &\quad { \sum\nolimits^K_{k=1}{\sum\nolimits^N_{n=1}{z_{nk} \norm{\x_n-\c_k}^2}} } \\
  \text{s.t.} &\quad z_{nk} \in\{0,1\},\; \sum\nolimits^K_{k=1}{z_{nk}} = 1,\; n=1,\dots,N\notag
\end{align}
where $\Z = (z_{nk})$ are binary assignment variables (of point $n$ to cluster $k$) and $\C = (\c_1,\dots,\c_K)$ are centroids living in $\bbR^D$. At an optimum, centroid $\c_k$ is the mean of the points in its cluster.

Given a bandwidth $\sigma > 0$, Gaussian mean-shift \citep{FukunagHostet75a,Cheng95a,Carreir00b,ComanicMeer02a} defines a kernel density estimate (kde)
\begin{equation}
  \label{e:kde}
  p(\x) = \frac{1}{N} \sum^N_{n=1}{G\bigl(\norm{(\x-\x_n)/\sigma}^2\bigr)}
\end{equation}
with kernel $G(t) \propto \smash{e^{-t/2}}$, and applies the iteration (started from each data point):
\begin{equation}
  \label{e:GMS}
  p(n|\x) = \frac{\exp{\bigl(-\frac{1}{2} \norm{(\x-\x_n)/\sigma}^2\bigr)}}{\sum^N_{n'=1}{\exp{\bigl(-\frac{1}{2} \norm{(\x-\x_{n'})/\sigma}^2\bigr)}}}, \quad \x \leftarrow \f(\x) = \sum\nolimits^N_{n=1}{p(n|\x) \x_n}
\end{equation}
which converges to a mode (local maximum) of $p$ from nearly any initial \x\ \citep{Carreir07a}. Each mode is the centroid for one cluster, which contains all data points that converge to its mode. The user parameter is the bandwidth $\sigma$, which determines the resulting number of clusters implicitly.

Both algorithms have well-known pros and cons. $K$-means tends to define round clusters; mean-shift can obtain clusters of arbitrary shapes and has been very popular in low-dimensional clustering applications such as image segmentation \citep{ComanicMeer02a}, but does not work well in high dimension due to the scarcity of data and difficulty in obtaining a good kde. Both algorithms suffer from outliers, which can move centroids outside their cluster in $K$-means or create singleton modes in mean-shift. Computationally, $K$-means is much faster than mean-shift, at $\calO(KND)$ and $\calO(N^2D)$ per iteration, respectively, particularly with large datasets (in fact, accelerating mean-shift has been a topic of active research, e.g.\ \citet{Carreir06a,Yuan_10a}). Mean-shift does not require a value of $K$, which is sometimes convenient, but it is often desirable to force an algorithm to produce exactly $K$ clusters (e.g.\ in figure-ground separation ($K=2$) or in medical image analysis, where one may know or estimate the number of organs sought), which is not straightforward for mean-shift to achieve.

\subsubsection{The $K$-modes algorithm}

We now briefly summarize the original $K$-modes algorithm \citep{CarreirWang13a}. Its objective function
\begin{align}
  \label{e:kmodes}
  \max_{\Z,\C} &\quad \sum\nolimits_{n=1}^N \sum\nolimits_{k=1}^K z_{nk} \gausskerd{\x_n}{\c_k}{\sigma} \\
  \text{ s.t. } &\quad z_{nk} \in\{0,1\},\; \sum\nolimits^K_{k=1}{z_{nk}} = 1,\; n=1,\dots,N \notag
\end{align}
can be seen as the sum of a kde defined by each cluster separately. It is convenient to solve this problem with alternating optimization over $\Z$ and $\C$. For fixed $\C$, the optimization over $\Z$ decouples over each point, and $\x_n$ is assigned to cluster $l=\smash{ \arg\max\nolimits_k\; \gausskert{\x_n}{\c_k}{\sigma} }= \arg\min\nolimits_k\; \norm{\x_n-\c_k}$ due to the discrete constraints of the problem. For fixed $\Z$, the optimization over $\C$ decouples over each cluster and we have a separate unconstrained maximization for each centroid, of the form $ L(\c_k) = \smash{\sum^N_{n=1}{ z_{nk} \gausskert{\x_n}{\c_k}{\sigma} }}$, which is proportional to the cluster's kde (this is why each centroid is truly a mode), and can be done with mean-shift updates as in eq.~\eqref{e:GMS}. The cost per outer iteration of this procedure is $\calO(KND)$, which mainly comes from computing distances between data points and centroids. Since each step is strictly feasible and decreases the objective or leaves it unchanged, this converges to a local optimum in a finite number of outer-loop steps if the \C-step is exact.

\subsection{Laplacian smoothing and learning soft  assignments}

Obtaining hard assignments by optimizing over a discrete cluster indicator matrix is usually difficult, because interesting objective functions are typically NP-hard. Spectral clustering algorithms \citep{ShiMalik00a,YuShi03a} avoid this difficulty by first approximating the solution using eigenvectors of the normalized graph Laplacian. However, since the eigenvectors do not readily provide valid assignments, these algorithms need to run another clustering algorithm (usually $K$-means) on the eigenvectors to obtain actual partitions of the data---a post-processing step that is somewhat artificial and can introduce multiple local optima. In Laplacian $K$-modes, we relax $\Z$ to be a stochastic matrix, so our $\Z$-step results from a convex QP and provides soft assignments of points to clusters, which may also be used as posterior probabilities.

Laplacian smoothing has also been used in combination with nonnegative matrix factorization (NMF) for clustering \citep{Cai_11b}.  NMF learns a decomposition of the input data matrix where both basis and coefficients are nonnegative, and tends to produce a parts-based representation of the data \citep{LeeSeung99a}, though this is not always so. \citet{Cai_11b} add a Laplacian smoothing term regarding the coefficient matrix to the NMF objective function, so that data points that are close in input space are encouraged to have similar representations using the basis set. $K$-means is then applied to the learned coefficients matrix to obtain a final partition of the data. Like spectral clustering, this algorithm does not directly optimize over the assignments, but obtains them in a post-processing step.

There has been recent work in clustering that directly optimizes over a stochastic assignment matrix. \citet{Arora_11a} optimize over a stochastic matrix $\PP$ such that $\PP \PP^\top$ best approximates a rescaled similarity matrix. However, the optimization problem has multiple solutions which are related by rotations. Therefore, they propose to exploit the geometry of the problem using a rotation-based algorithm, which is straightforward for up to $K=4$ clusters but requires an optimization procedure to computing the projection onto the probability simplex for $K>4$ clusters. The idea of AnchorGraphs \citep{Liu_10c} is used by \citet{YangOja12a} to approximate the affinities between data points through a two-step Data-Cluster-Data (DCD) random walk. They then minimize the generalized KL divergence between a given sparse affinity matrix and the affinities obtained from DCD. In this formulation, the matrix containing probabilities of points moving to the (augmented) cluster nodes is stochastic. These approaches are related to Laplacian $K$-modes in that they optimize some objective over the assignment probabilities. But our Laplacian $K$-modes algorithm also obtains prototypical centroids, does not have the issue of rotational equivalence of \citet{Arora_11a}, and makes use of the efficient projection onto the probability simplex to deal with any number of clusters.

\section{Algorithm}
\label{s:alg}

\subsection{The Laplacian $K$-modes algorithm}

We change the assignment rule of $K$-modes to handle more complex shaped clusters based on two ideas: (1) the observation that \emph{nearby data points should have similar assignments}; and (2) the use of \emph{soft assignments}, which allows more flexibility in the clusters and simplifies the optimization. We first build a graph (e.g.\ $k$-nearest-neighbor graph) on the dataset, and let $w_{mn}$ be an affinity (e.g.\ binary, heat kernel) between $\x_m$ and $\x_n$. We then add to the $K$-modes objective function a Laplacian smoothing term $\smash{\frac{\lambda}{2}\sum_{m=1}^N \sum_{n=1}^N w_{mn} \norm{\z_m-\z_n}^2}$ to be minimized, where $\z_n=[z_{n1},\dots,z_{nK}]^\top$ is the assignment vector of $\x_n$, $n=1,\dots,N$, to each of the $K$ clusters, and $\lambda \ge 0$ is a trade-off parameter. The assignments are now continuous variables, but constrained to be positive and sum to $1$. Thus, $z_{nk}$ can be considered as the probability of assigning $\x_n$ to cluster $k$ (soft assignment). Thus, the \emph{Laplacian $K$-modes} objective function is:
\begin{align}
    \min_{\Z,\C} &\quad  \frac{\lambda}{2}\sum_{m=1}^N\sum_{n=1}^N w_{mn} \norm{\z_m-\z_n}^2 - \sum_{n=1}^N \sum_{k=1}^K z_{nk} \gausskerd{\x_n}{\c_k}{\sigma} \label{e:lapkmodes1} \\
    \mbox{s.t.} &\quad \sum_{k=1}^K z_{nk}=1, \; n=1,\dots,N,  \notag \\
    &\quad z_{nk}\ge 0, \; n=1,\dots,N,\; k=1,\dots,K. \notag
  \end{align}
We can rewrite this objective in matrix form:
\begin{align}
  \label{e:lapkmodes2}
  \min_{\Z,\C} &\quad \lambda \trace{\Z^\top \LL \Z}  - \trace{\B^\top \Z} \\
  \mbox{s.t.} &\quad \Z \1_K = \1_N,\quad \Z \ge \0 \notag
\end{align}
where $\LL = \D - \W$ is the graph Laplacian for the affinity matrix $\W = (w_{nm})$ and degree matrix $\D = \diag{\smash{\sum_{n=1}^N w_{mn}}}$, $\B = (b_{nk})$ is an $N\times K$ matrix containing data-centroid affinities $b_{nk}= \smash{ \gausskert{\x_n}{\c_k}{\sigma} }$, $n=1,\dots,N$, $k=1,\dots,K$, $\1_K$ is a $K$ dimensional vector of $1$s and $\ge$ means elementwise comparison. Other variations of the graph Laplacian can also be used (e.g.\ the normalized Laplacian), see \citet{Luxbur07a}. The constraint on $\Z$ shows it is a stochastic matrix. We can obtain a hard clustering if desired by assigning each point to the cluster with highest assignment value.

\subsubsection{Special cases of the hyperparameters $(\lambda,\sigma)$}

In Laplacian $K$-modes, in addition to $K$ there are two user parameters: $\lambda$ controls the smoothness of the assignment, and $\sigma$ controls the smoothness of the kde defined on each cluster. Consider first the case of $\lambda=0$, where Laplacian $K$-modes becomes the original $K$-modes algorithm. \citet{CarreirWang13a} already noted that the $K$-modes algorithm has two interesting limit cases: it becomes $K$-means when $\sigma\rightarrow\infty$, and a form of $K$-medoids when $\sigma\rightarrow 0$, since the centroids are driven towards data points. In both cases the assignments are hard (1-out-of-$K$ coding). The case when $\lambda\rightarrow\infty$ makes the first term in eq.~\eqref{e:lapkmodes1} dominant and forces all connected points to have identical assignments, which is not interesting for the purpose of clustering. Therefore, the most interesting behavior of the algorithm is for intermediate $\lambda$. Finally, another interesting special case of Laplacian $K$-modes corresponds to $\lambda>0$ and $\sigma\rightarrow\infty$, which we call \emph{Laplacian $K$-means}, and which seems to be a new algorithm as well.

\subsection{Optimization procedure for Laplacian $K$-modes}
\label{s:opt}

To solve~\eqref{e:lapkmodes1}, we use alternating optimization over $\C$ and $\Z$, which takes advantage of the problem's structure.

\paragraph{$\C$-step}

For fixed $\Z$, we are only concerned with the second term of \eqref{e:lapkmodes1} which is the $K$-modes objective. Therefore, our step over $\C$ is identical to that of $K$-modes: it decouples over clusters and we apply mean-shift to solve for each $\c_k$ separately. The cost of this step is $\calO(KND)$.

\paragraph{$\Z$-step}

Unlike in $K$-modes, our $\Z$-step no longer decouples, which means we have to solve for $NK$ variables all together. Since the graph Laplacian \LL\ is positive semidefinite, the problem over \Z\ is a convex quadratic program (QP). While we could apply a standard QP algorithm, such as an interior point method, we provide here an algorithm that is very simple (no parameters to set), efficient and that scales well to real problems where the number of points $N$ or the number of clusters $K$ is very large. The solution is based on the gradient proximal algorithm used by \citet{BeckTeboul09a}. Their general framework solves convex problems of the form $\min\nolimits_\x f(\x) = g(\x)+h(\x)$, where $g$ is convex and has Lipschitz continuous gradient (with constant $L$), and $h$ is convex but not necessarily differentiable. The gradient proximal algorithm iteratively updates the variables by first taking a gradient step of the first function and then projecting it with the second function, i.e., $\x_{\tau+1} = { \arg\min\nolimits_\y \; {\frac{L}{2} \norm{\smash{\y-(\x_{\tau}-\frac{1}{L}\nabla g(\x_{\tau}))}}^2 + h(\y)} }$. It can be proven that the algorithm converges in objective function value with rate $\calO(1/\tau)$ (where $\tau$ is the iteration counter) with a \emph{constant stepsize} $\frac{1}{L}$, and using Nesterov's acceleration scheme improves the rate to $\calO(1/\tau^2)$.

To apply this framework to our $\Z$-step, we make the identification that $g$ is our smooth quadratic objective function, which has continuous gradient with $L=2\lambda M$ being the (smallest) Lipschitz constant, where $M$ is the largest eigenvalue of $\LL$, and $h$ is the indicator function of the probability simplex. Consequently, our proximal step is computing the Euclidean projection of the gradient step onto the probability simplex. Note that computing the Euclidean projection onto the $K$-dimensional simplex is itself a quadratic program. Fortunately, there exists an efficient algorithm which computes the exact projection with $\calO(K\log K)$ time complexity \citep{Duchi_08a,WangCarreir13a}.

\begin{algorithm}[t]
  \caption{Accelerated gradient projection for the $\Z$ step.}
  \label{alg:gradproj}
  \renewcommand{\algorithmicrequire}{\textbf{Input:}}
  \renewcommand{\algorithmicensure}{\textbf{Output:}}
  \begin{algorithmic}[1]
    \REQUIRE Initial $\Z_0 \in\RR^{N\times K}$, $s=\frac{1}{2\lambda M}$, $M=$ largest eigenvalue of graph Laplacian $\LL$.
    \STATE Set $\Y_1 = \Z_0,\ t_1 = 1,\ \tau=1$.
    \REPEAT
    \STATE Compute gradient at $\Y_\tau$: $\G_\tau = 2\lambda\LL\Y_\tau-\B$
    \STATE $\Z_\tau =$ simplex projection of each row of $\Y_\tau-s\G_\tau$
    \STATE $t_{\tau+1} = (1+\sqrt{1+4t_\tau^2})/2$
    \STATE $\Y_{\tau+1} = \Z_\tau + \big(\frac{t_\tau-1}{t_{\tau+1}}\big)(\Z_\tau-\Z_{\tau-1})$
    \STATE $\tau=\tau+1$
    \UNTIL convergence
    \ENSURE $\Z_\tau$ is the solution of the $\Z$-step.
  \end{algorithmic}
\end{algorithm}

We provide the accelerated gradient projection algorithm for our $\Z$-step in Algorithm~\ref{alg:gradproj}. Notice the graph Laplacian is sparse and its largest eigenvalue $M$ can be obtained efficiently (e.g.\ by power iterations). Therefore the constant stepsize $s$ can be easily determined right after constructing the graph Laplacian. Compared to a pure gradient projection algorithm, the additional computational effort of the acceleration scheme in maintaining an auxiliary sequence \Y\ (lines 5--6 of Algorithm~\ref{alg:gradproj}) is minimal, and we clearly observe an improved convergence behavior in experiments.

Each iteration of Algorithm~\ref{alg:gradproj} costs $\calO(NK\rho+NK\log K)$, where $\rho$ is the neighborhood size in constructing $\LL$ (or the number of nonzero entries in each row). The first term accounts for computing the gradient and the second term accounts for projecting each row of $\Z$ onto the probability simplex. Notice how, although it is solving a large QP, the cost per iteration of our $\Z$-step is independent of the input dimensionality $D$ (in contrast, the $\C$-step has time complexity $\calO(KND)$). Despite its sublinear convergence rate, the algorithm has a clear advantage in its simplicity: it does not require any line search or costly matrix operation, and it is very easy to implement.

\subsubsection{Convergence properties}

In the $\C$-step, each mean-shift update increases the density of the cluster kde (or leaves it unchanged) and its convergence rate to a mode is linear in general \citep{Carreir07a}. In the $\Z$-step, the accelerated gradient projection converges theoretically at $\calO(1/\tau^2)$ rate where $\tau$ is the iteration counter, although this algorithm seems to perform much better than the theoretical guarantee in practice \citep{BeckTeboul09a}. We alternate the \C\ and \Z\ steps until a convergence criterion is satisfied (e.g.\ the change to the variables is below some threshold). Notice both steps use iterative procedures, so the number of iterations depends on the convergence accuracy. Let $\epsilon_1$ and $\epsilon_2$ be the optimization precision for the $\C$-step and $\Z$-step, respectively, then one alternation of our algorithm costs $(D\log(1/\epsilon_1)+\rho/\sqrt{\epsilon_2}) KN$. We found empirically that moderate accuracy and few iterations suffice for good clustering results, and our algorithm scales well due to its low per-iteration cost. In an efficient implementation, both steps can be inexact (e.g.\ each could run for a fixed, small number of iterations). Since the \Z-step algorithm is feasible, exiting it early produces valid assignments.

\subsubsection{Homotopy algorithm}

As with $K$-means and $K$-modes, the Laplacian $K$-modes objective function has local optima, which are caused by the nonlinear, kde term. One strategy to find a good optimum consists of first finding a good optimum for $K$-means and then run a homotopy algorithm initialized there. We can construct a homotopy by varying continuously $\lambda$ from $0$ and $\sigma$ from $\infty$, which corresponds to $K$-means, to their target values $(\lambda^*,\sigma^*)$. In practice, we follow this path approximately, by running some iterations of the fixed-($\lambda$,$\sigma$) Laplacian $K$-modes algorithm for each value of ($\lambda$,$\sigma$). As is well known with homotopy techniques, this tends to find better optima than starting directly at the target value $(\lambda^*,\sigma^*)$. A good optimum for $K$-means can be obtained by picking the best of several random restarts, or by using the $K$-means++ initialization strategy, which has approximation guarantees \citep{ArthurVassil07a}.

\subsubsection{Setting the hyperparameters}

We believe that, in an unsupervised setting, the user should be able to explore different scenarios and so the algorithm should have a small number of intuitive hyperparameters to control this---rather than automatically guessing, say, the number of clusters, which often is not uniquely defined for a dataset. Laplacian $K$-modes has 3 intuitive hyperparameters, which allow a user to explore different scenarios: more or less clusters ($K$), different scales ($\sigma$), and degree of membership smoothness ($\lambda$). Typical values are around $\lambda=1$, which allows some amount of propagation among neighbors and thus nonconvex clusters, and $\sigma$ obtained from a kde bandwidth formula \citep{WandJones94a}, such as the average distance to the 7th nearest neighbor gives a reasonable density \citep{ZelnikPerona05a}. (An even better option is to use an adaptive kde, where each point has a different bandwidth, obtained using ``entropic affinities'' \citep{HintonRoweis03a,VladymCarreir13a}. Here, the bandwidth of each data point is computed so as to produce an effective number of neighbors $k$ set by the user.) These hyperparameters values usually produce good clustering results and provide a starting point for improvement. In the homotopy algorithm, the path of $\sigma$ or $\lambda$ values should be followed slowly so we end in a good minimum. In practice, one changes the parameter geometrically in as many steps as one can afford computationally.

In a supervised setting, the hyperparameters can be selected with a validation set using the out-of-sample mapping for Laplacian $K$-modes (section~\ref{s:out-of-sample}).

\subsection{Out-of-sample problem}
\label{s:out-of-sample}

We now consider the out-of-sample problem, that is, given an unseen test point $\x \in \bbR^D$, we wish to find a meaningful assignment $\z(\x)$ to the clusters found during training. A natural and efficient way to do this is to solve a problem of the same form as~\eqref{e:lapkmodes1} with a dataset consisting of the original training set augmented with \x, but keeping \Z\ and \C\ fixed to the values obtained during training (this avoids having to solve for all points again). After dropping constant terms, this is equivalent to the following problem:
\begin{align*}
  \min_{\z} &\quad \lambda \sum\nolimits_{n=1}^N w_{n} \norm{\z-\z_n}^2 - \sum\nolimits_{k=1}^K z_{k} \gausskerd{\x}{\c_k}{\sigma} \\
  \mbox{s.t.} &\quad z_{k}\ge 0,\ k=1,\dots,K, \; \sum\nolimits_{k=1}^K{z_k} = 1
\end{align*}
where $w_n$ is the affinity between test point $\x$ and training point $\x_n$.  The above problem can be further reduced to the following quadratic program:
\begin{equation}
  \label{e:obj-osp}
  \textstyle
  \min_{\z} \quad \frac{1}{2} \norm{\z-(\bar{\z}+\gamma\q)}^2  \;\text{ s.t.}\quad \z^\top \1_K = 1, \; \z \ge \0
\end{equation}
where the expressions for $\bar{\z}$, $\q = [q_1,\dots,q_K]^\top$ and $\gamma$ are as follows:
\begin{align*}
  \bar{\z} = \sum_{n=1}^N \frac{w_n}{\sum\limits_{n'=1}^N{w_{n'}}} \z_n, \ q_k = \frac{G ( \norm{(\x-\c_k)/\sigma}^2 )}{\sum\limits_{k'=1}^K G ( \norm{(\x-\c_{k'})/\sigma}^2 ) }, \ \gamma = \frac{ \sum\limits_{k=1}^K{ G ( \norm{(\x-\c_{k})/\sigma}^2 ) } }{2\lambda \sum\limits_{n=1}^N{w_n}}.
\end{align*}
Thus, the out-of-sample solution is the projection of the $K$-dimensional vector $\bar{\z}+\gamma\q$ onto the probability simplex. The computational cost is $\calO(ND)$, dominated by the cost of $\bar{\z}$, since the simplex projection costs $\calO(K \log{K})$.

The solution has an intuitive interpretation, consisting of the linear combination of two terms, each a valid assignment vector (having positive elements that sum to $1$). The Laplacian term, $\bar{\z}$, is the weighted average of the neighboring training points' assignments, and results in nonconvex clusters. The kde term, $\q$, assigns a point based on its distances (posterior probabilities) to the centroids, and results in convex clusters. These two distinct assignment rules are combined using a weight $\gamma$ to give the final assignment. Essentially, $\x$ is assigned to cluster $k$ with high probability if its nearby points are assigned to it ($\bar{z}_k$ is large) or if it is close to $\c_k$ ($q_k$ is large). Although defined variationally, the out-of-sample mapping is just as useful as a closed-form expression: computationally it does not require an iterative procedure, and the interpretation above also illuminates the meaning of the Laplacian in the training objective~\eqref{e:kmeans-objfcn}. In fact, iterating the out-of-sample mapping sequentially over the training points gives another (slower) way to solve the \Z-step, i.e., alternating optimization over $\z_1,\dots,\z_N$.

Finally, Table~\ref{t:clustercomp} compares Laplacian $K$-modes with other popular clustering algorithms (see section~\ref{s:opt} for the complexity analysis).

\begin{table}[t]
  \centering
  \caption{Comparison of properties of different clustering algorithms.}
  \label{t:clustercomp}
  \begin{tabular}{@{}l|cccccc@{}}
    \hline
    & $K$-means & $K$-medoids & Mean-shift & \caja{c}{c}{Spectral\\ clustering} &  $K$-modes & \caja{c}{c}{Laplacian\\ $K$-modes} \\
    \hline
    Centroids     & \caja[0.7]{c}{c}{likely \\ invalid}   & ``valid'' & ``valid'' & N/A & valid & valid \\
    Nonconvex clusters  & no        & depends & yes & yes & no & yes \\
    Density       & no        & no & yes & no & yes & yes \\
    Assignment    & hard      & hard & hard & hard & hard & soft \\
    Cost per iteration & $KND$ & $KN^2D$ & $N^2D$ & $N^2\sim N^3$ & $KND$ & \caja{c}{c}{$(D\log(1/\epsilon_1)$\\$+\rho/\sqrt{\epsilon_2}) KN$} \\
    \hline
  \end{tabular}
\end{table}

\section{Experiments}
\label{s:expts}

\subsection{Illustrative experiments}

\subsubsection{Spirals dataset}

\begin{figure}[t]
  \centering
  \psfrag{s}[r][r]{\raisebox{-1.3ex}{$\sigma\rightarrow$}\hspace{2ex}}
  \begin{tabular}{@{}c@{\hspace{0\linewidth}}c@{\hspace{0\linewidth}}c@{\hspace{0\linewidth}}c@{\hspace{0\linewidth}}@{}}
    $K$-modes & \multicolumn{3}{@{}c@{}}{\dotfill Laplacian $K$-modes \dotfill} \\
    \includegraphics[width=0.25\linewidth]{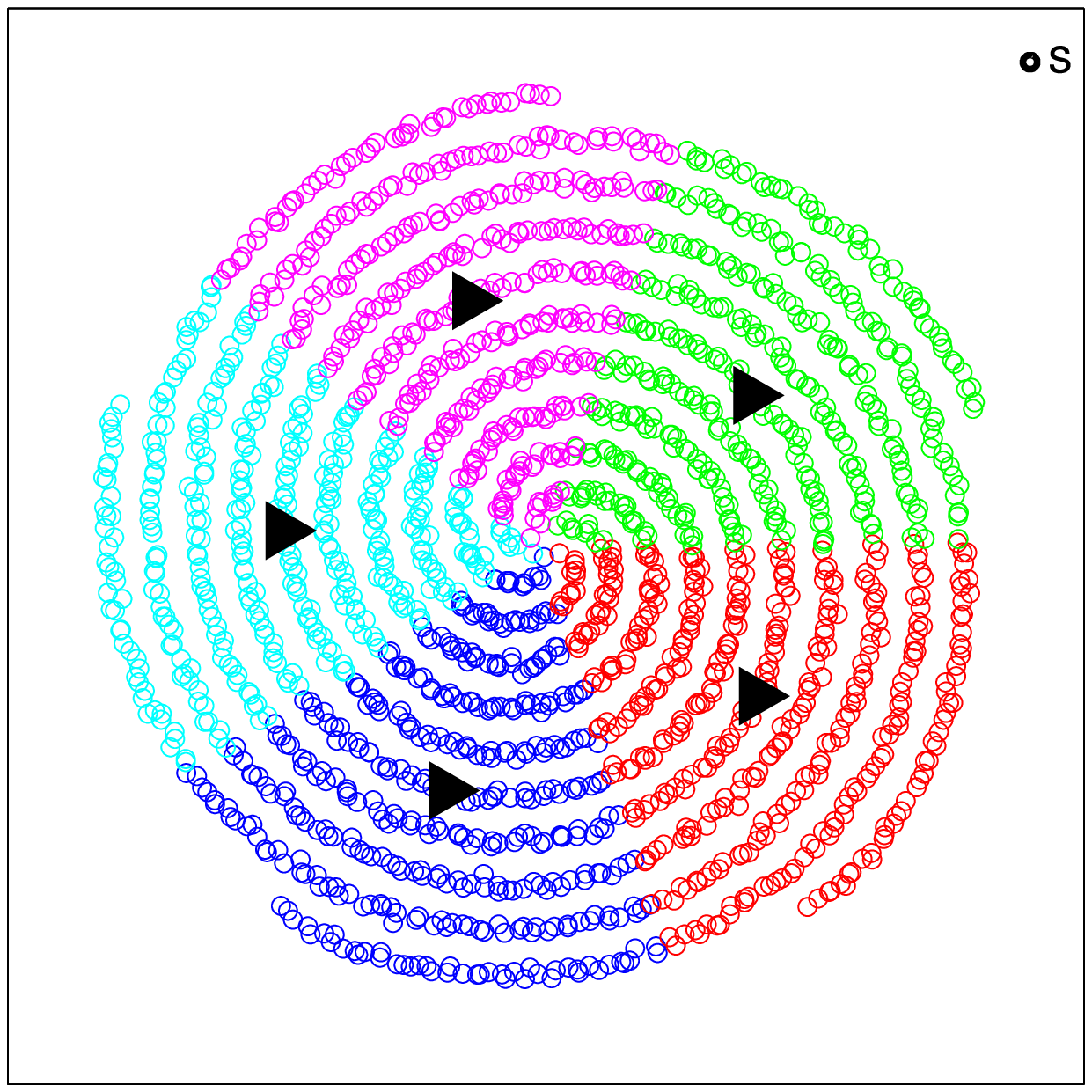} &
    \includegraphics[width=0.25\linewidth]{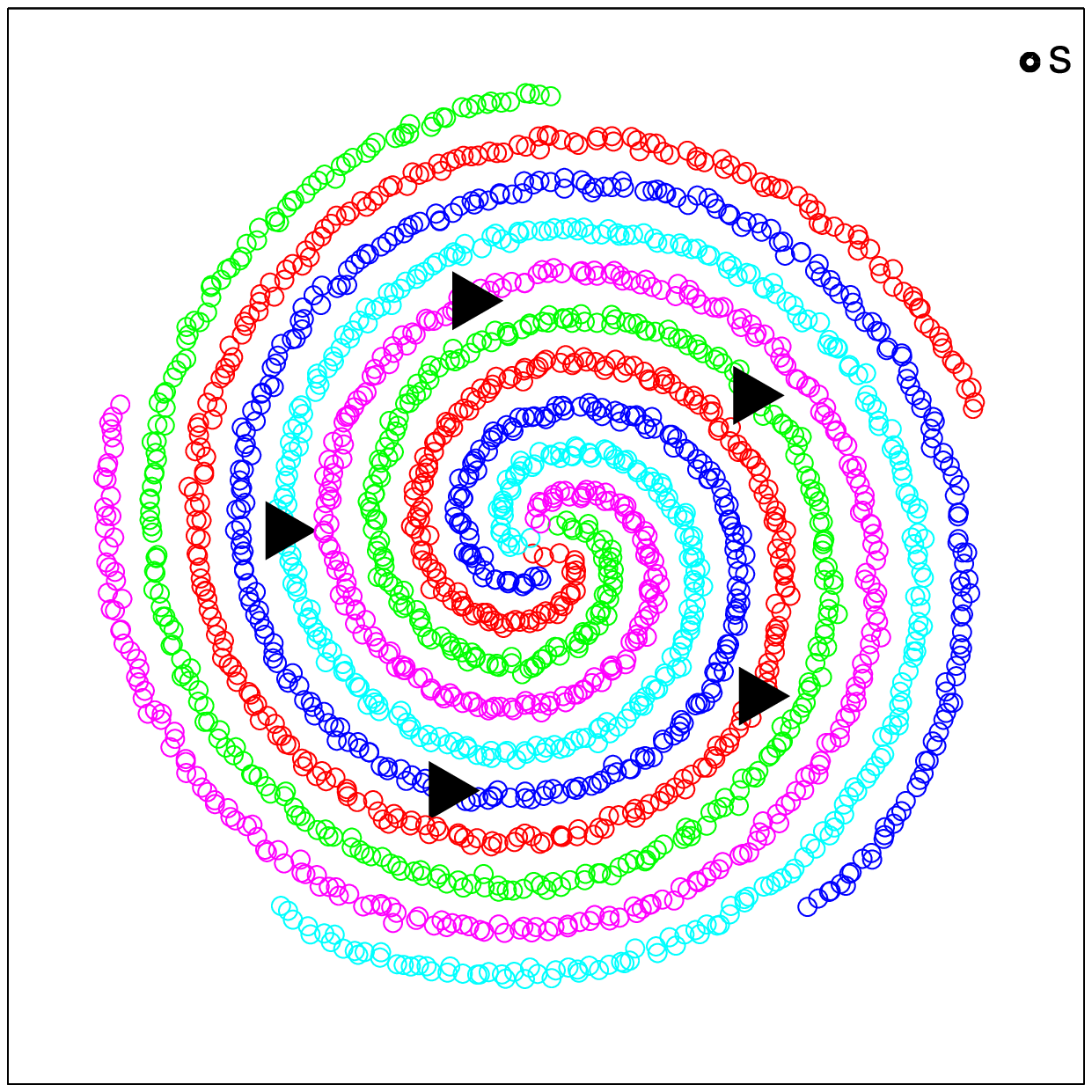} &
    \includegraphics[width=0.25\linewidth]{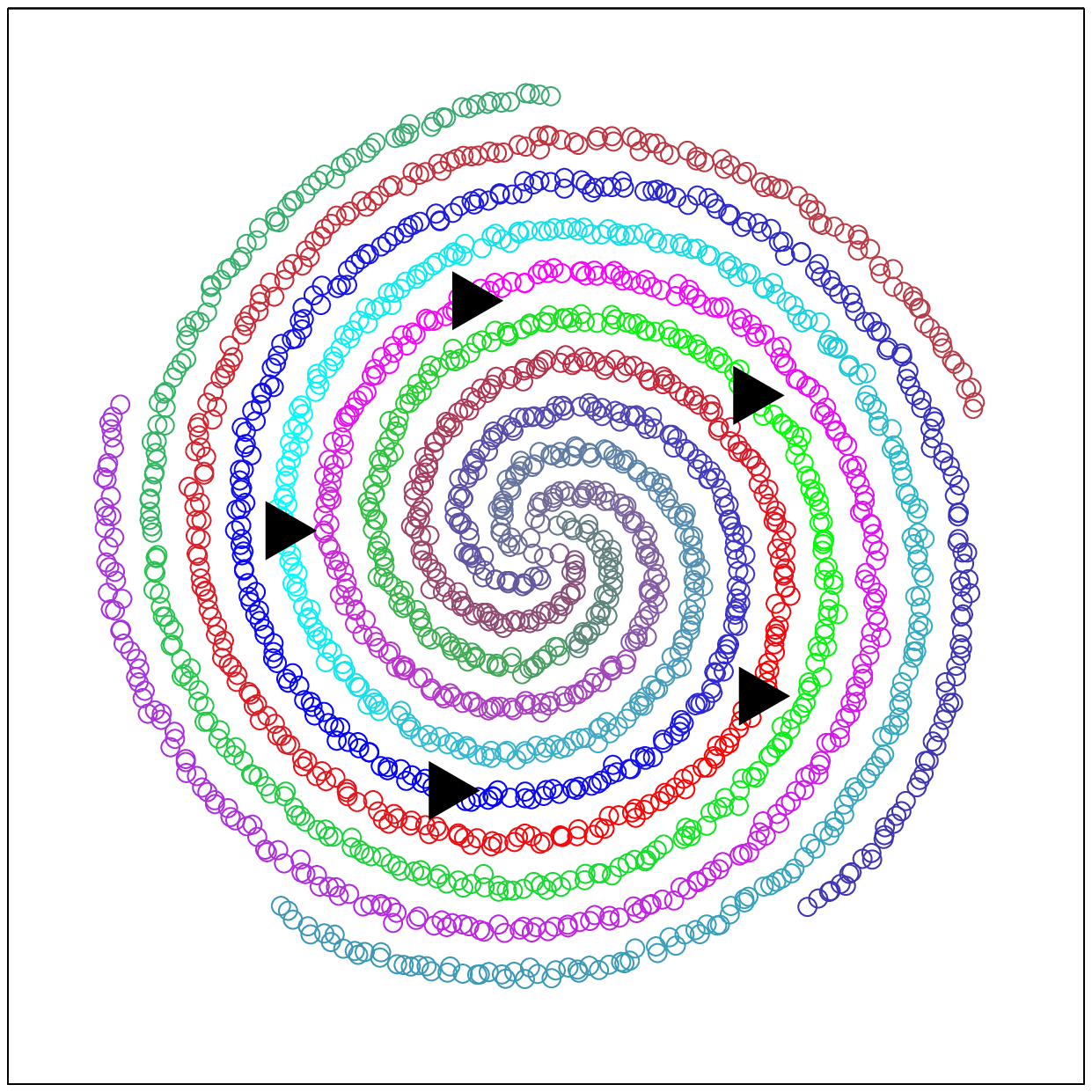} &
    \includegraphics[width=0.25\linewidth]{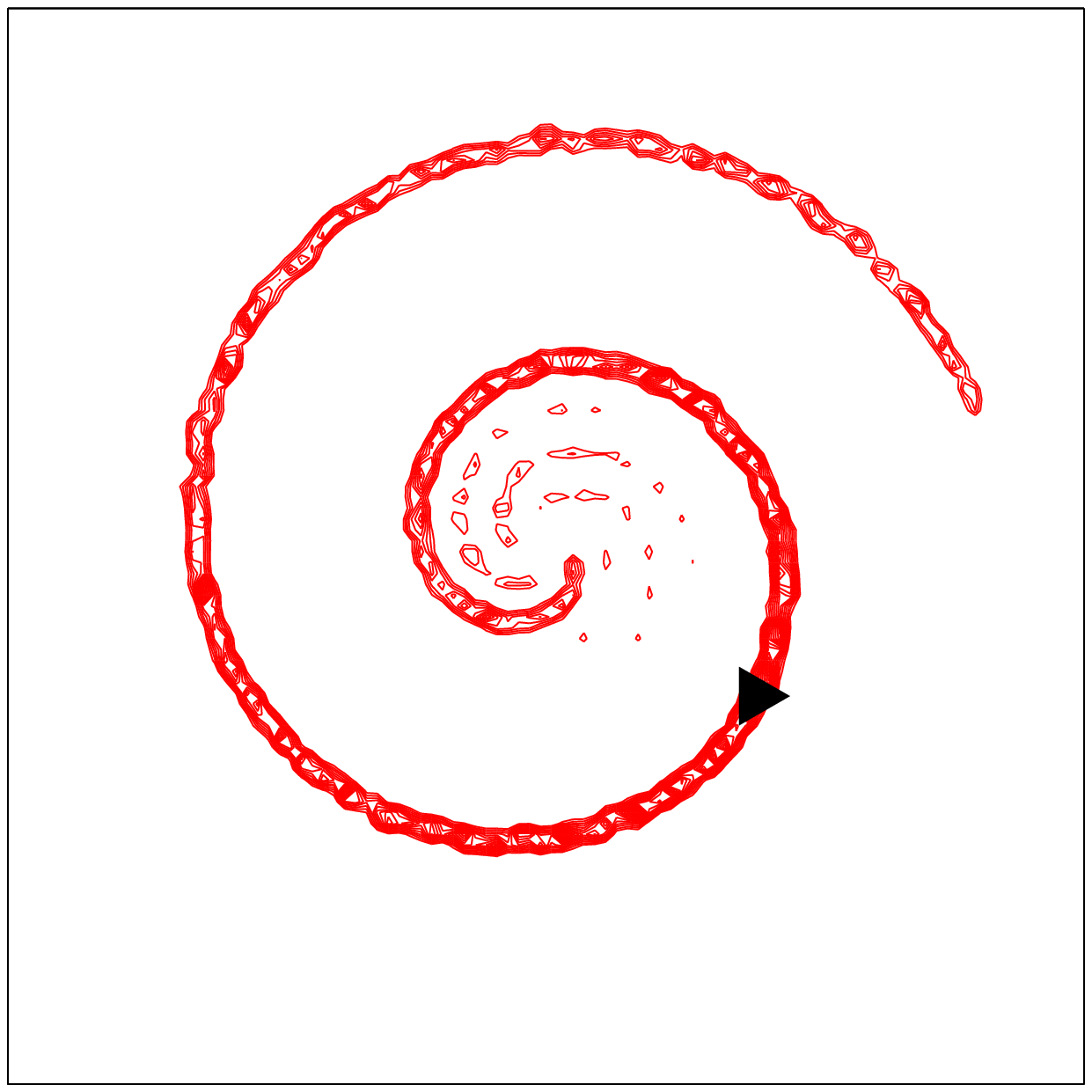}
  \end{tabular}
  \caption{Synthetic dataset of $5$-spirals. From left to right: $K$-modes clustering ($\lambda=0$, $\sigma=0.2$, the circle at the top right corner has a radius of $\sigma$); Laplacian $K$-modes clustering ($\lambda=100$, $\sigma=0.2$); Laplacian $K$-modes assignment probabilities; contours of the kde of the ``red'' cluster.}
  \label{f:5spirals}
\end{figure}

We first demonstrate the power of Laplacian smoothing. The 2D dataset in Fig.~\ref{f:5spirals} consist of $5$ spirals where each spiral contains $400$ points (denoted by $\circ$). The natural way of partitioning this dataset into $K=5$ groups is to assign points of each spiral into a separate cluster. Due to the nonconvex shape of the spirals, the ideal result can not be possibly achieved by $K$-modes (plot 1, we color each point differently according to the cluster it is assigned to) or $K$-means (not shown, result similar to $K$-modes), even though the $K$-modes centroids (denoted by $\blacktriangleright$) are lying on each spiral and are valid representatives of the dataset. We then build a $5$-nearest-neighbor graph on this dataset using heat kernel weighting and run Laplacian $K$-modes using the $K$-means result as initialization. We achieve a perfect separation of the spirals and one valid centroid for each spiral in a few steps of our alternating optimization scheme, as shown in plot 2. We show the assignment probabilities $\Z$ in plot 3, where each data point $\x_n$ is colored using a mixture of the 5 clusters' colors with its assignment probability $\z_n$ being the mixing coefficient. We show the contours of the kde defined on the ``red'' cluster in plot 4. In spite of the nonconvex, 1D manifold nature of the cluster, the kde is localized to the cluster and represents its shape and density well. This illustrates why using soft assignments gives more flexibility: the weights in the kde of each point vary, which allows for a more flexible kde. It is obvious that running mean-shift on this dataset with the same $\sigma$ will result in a large number of modes and therefore clusters. In contrast, the number of modes is fixed in  Laplacian $K$-modes and the algorithm will track one of the major modes in each cluster.

It is interesting to notice that because the kernel width $\sigma$ we use is quite small, only a small proportion of data points are close enough to centroids to have nonzero affinity. This implies that the $\B$ matrix in (5) of the main paper is quite sparse. Nonetheless, we achieve good assignment probabilities using the graph Laplacian, which propagates the sparse ``label'' information in $\B$ throughout the graph. This also partly explains the success of Laplacian smoothing in spectral clustering \citep{ShiMalik00a} and semi-supervised learning algorithms \citep{Zhu_03a,Belkin_06a}.

\subsubsection{Noisy two moons}

\begin{figure*}[t]
  \centering
  \begin{tabular}{@{}c@{\hspace{0\linewidth}}c@{\hspace{0\linewidth}}c@{\hspace{0\linewidth}}c@{\hspace{0\linewidth}}@{}}
    $K$-means & \multicolumn{3}{@{}c@{}}{\dotfill Laplacian $K$-modes \dotfill} \\
    \includegraphics[width=0.25\linewidth]{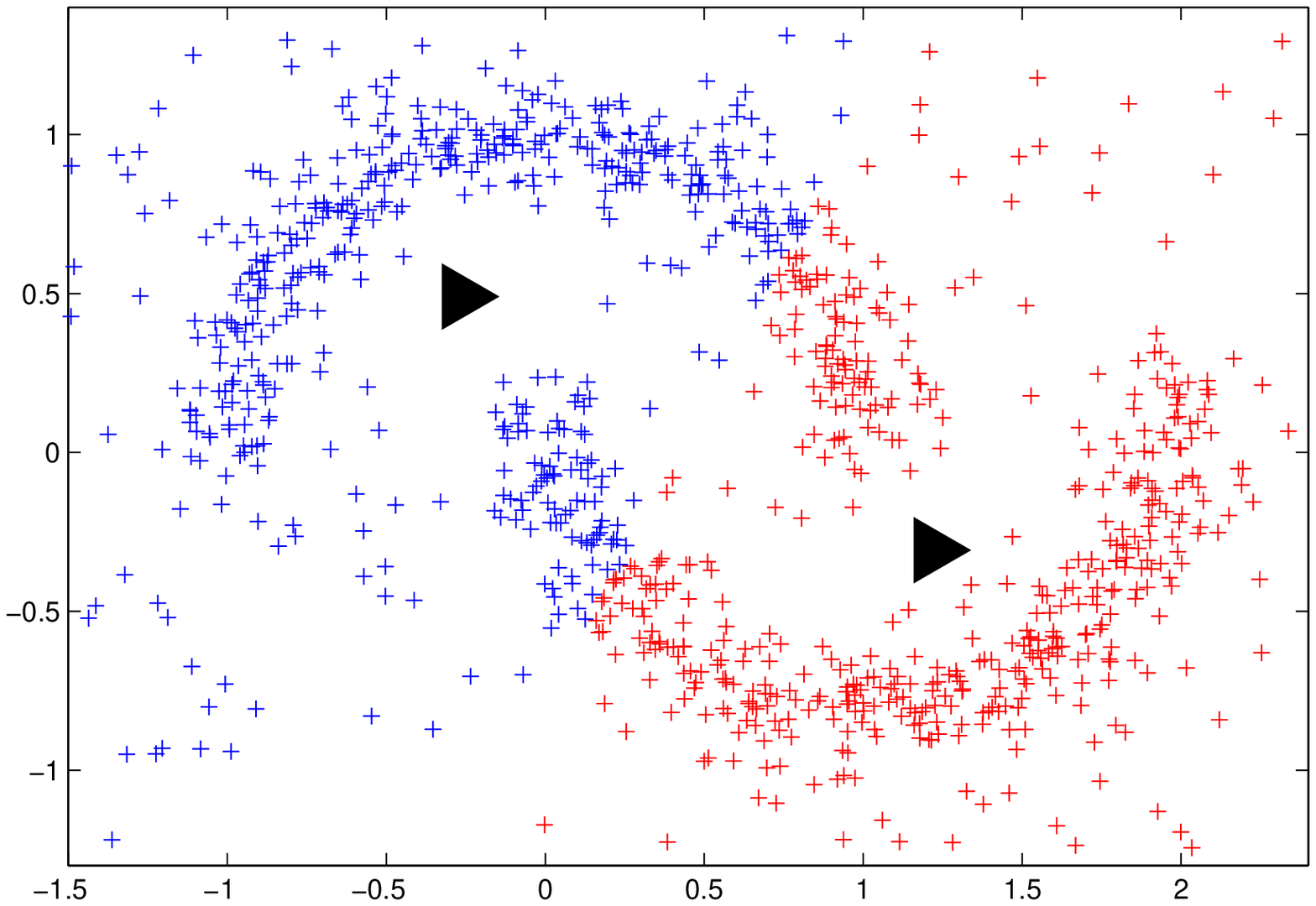} &
    \includegraphics[width=0.25\linewidth]{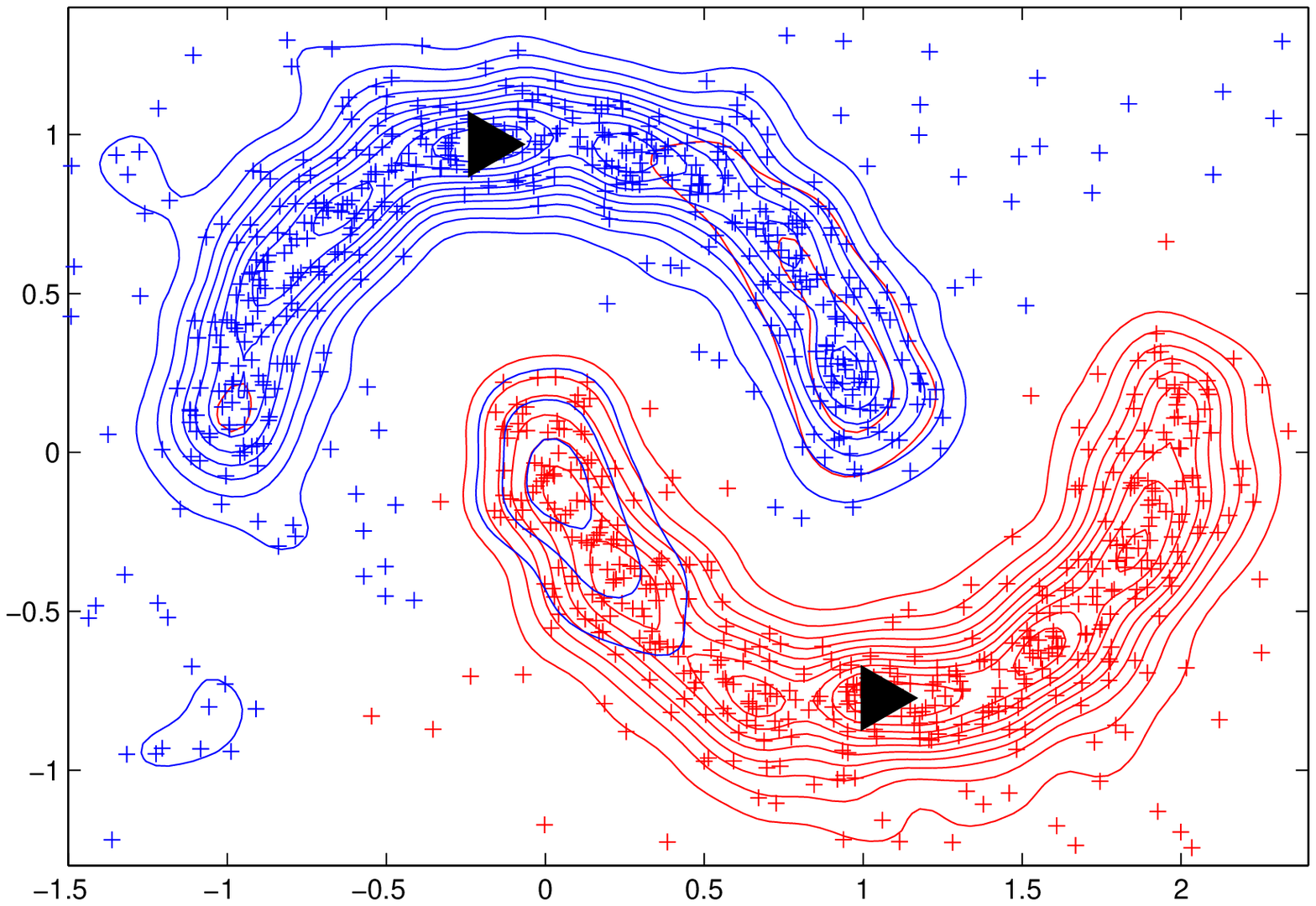} &
    \includegraphics[width=0.25\linewidth]{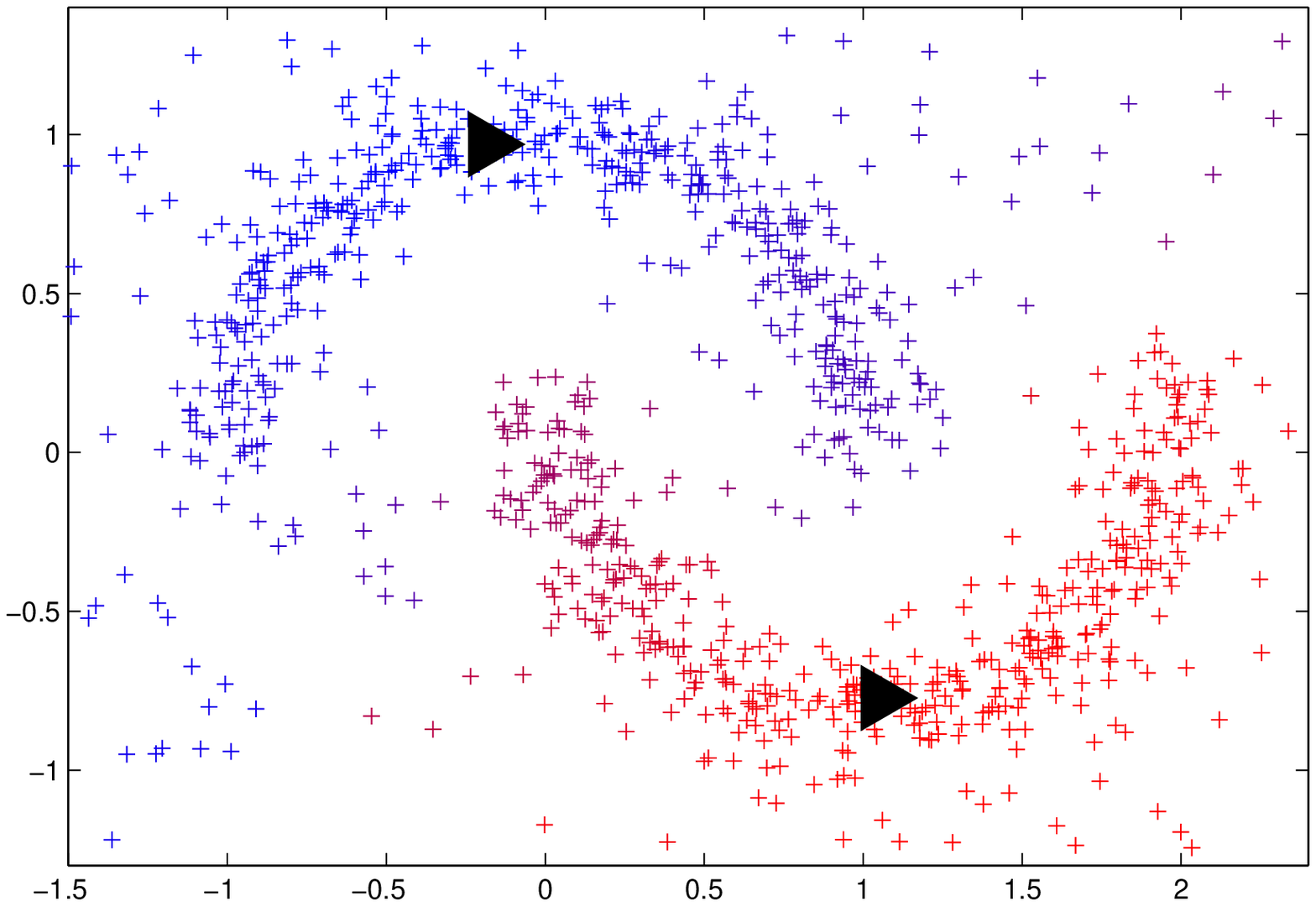} &
    \includegraphics[width=0.25\linewidth]{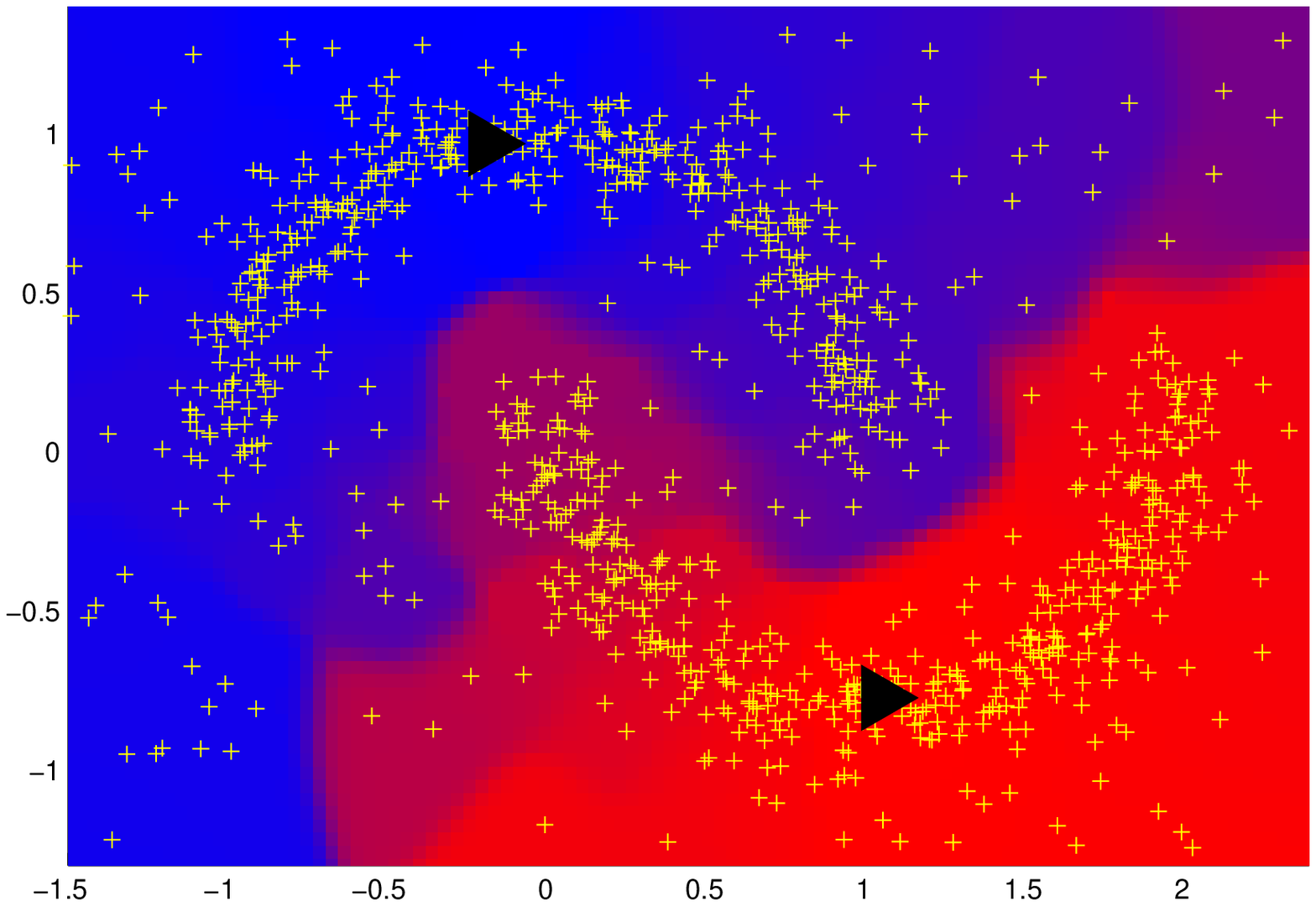}
  \end{tabular}
  \caption{Synthetic dataset of $2$-moons. We denote data points by $+$ and centroids by $\blacktriangleright$.  We run Laplacian $K$-modes in homotopy and show results at final parameter value ($\lambda=1$ and $\sigma=0.1$). From left to right: $K$-modes clustering ($\lambda=0$, $\sigma\rightarrow\infty$); Laplacian $K$-modes clustering and contours of kde of each cluster; Laplacian $K$-modes assignment probabilities for training set (used as mixing coefficients for coloring each training point); out-of-sample mapping in input space, colored in the same way as assignment probabilities of plot 3 (training points are now plotted in yellow). }
  \label{f:2moons}
\end{figure*}

We demonstrate Laplacian $K$-modes on the ``two-moons'' dataset in Fig.~\ref{f:2moons}. The dataset has two nonconvex, interleaved clusters (each has $400$ points) and we add many outliers ($200$ points) around them. The ``moons'' cannot be perfectly separated by either $K$-means (results shown in plot 1) or $K$-modes, since both define Voronoi tessellations. This problem is also difficult for hierarchical clustering because, as is well known, its major problem is that it creates connections between different clusters as the merging occurs. We build a $5$-nearest-neighbor graph on this dataset using heat kernel weighting, and run Laplacian $K$-modes from the $K$-means initialization. We run the homotopy version and reduce $\sigma$ from $5$ to $0.1$ in $10$ steps while fixing $\lambda=1$. The hard partition obtained, along with the two centroids and kde's for each cluster at $\sigma=0.1$ are given in plot 2. Even with the heavy noise and outliers, the ``inliers'' are still perfectly separated, the modes lie in high density areas and we obtain a good density estimate for each cluster. We show the assignment probabilities \Z\ in plot 3, colored using the same scheme as in the spirals example, where each data point $\x_n$ is colored using a mixture of the clusters' colors (red or blue) with its assignment probability $\z_n$ being the mixing coefficient. The assignment is certain near the centroids (purer color) and less crisp at boundaries and outliers (mixed color). Finally, the out-of-sample mapping in input space is shown in plot 4, where we compute out-of-sample assignments for a fine grid and color each grid point using the same scheme as in plot 3. We see clearly that the assignment rule is very different from the hard assignment of $K$-means. The mapping at each point combines the average assignment of nearby training points and the assignment from centroids, being able to model a complex shape.

\begin{figure}[t]
  \centering
  \begin{tabular}{@{}c@{\hspace{0.0\linewidth}}c@{\hspace{0.01\linewidth}}c@{\hspace{0.01\linewidth}}c@{}}
    Original image & Classification error (\%) & Normalized cut & Laplacian $K$-modes \\
    \includegraphics[width=0.24\linewidth]{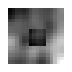} &
    \psfrag{sigma}[][]{$\sigma$}
    \psfrag{error}[][]{}
    \psfrag{Ncut}[l][l]{Normalized cut}
    \psfrag{Lap K-modes}[l][l]{Lap.\ $K$-modes}
    \raisebox{-5ex}{\includegraphics[width=0.28\linewidth,height=0.28\linewidth]{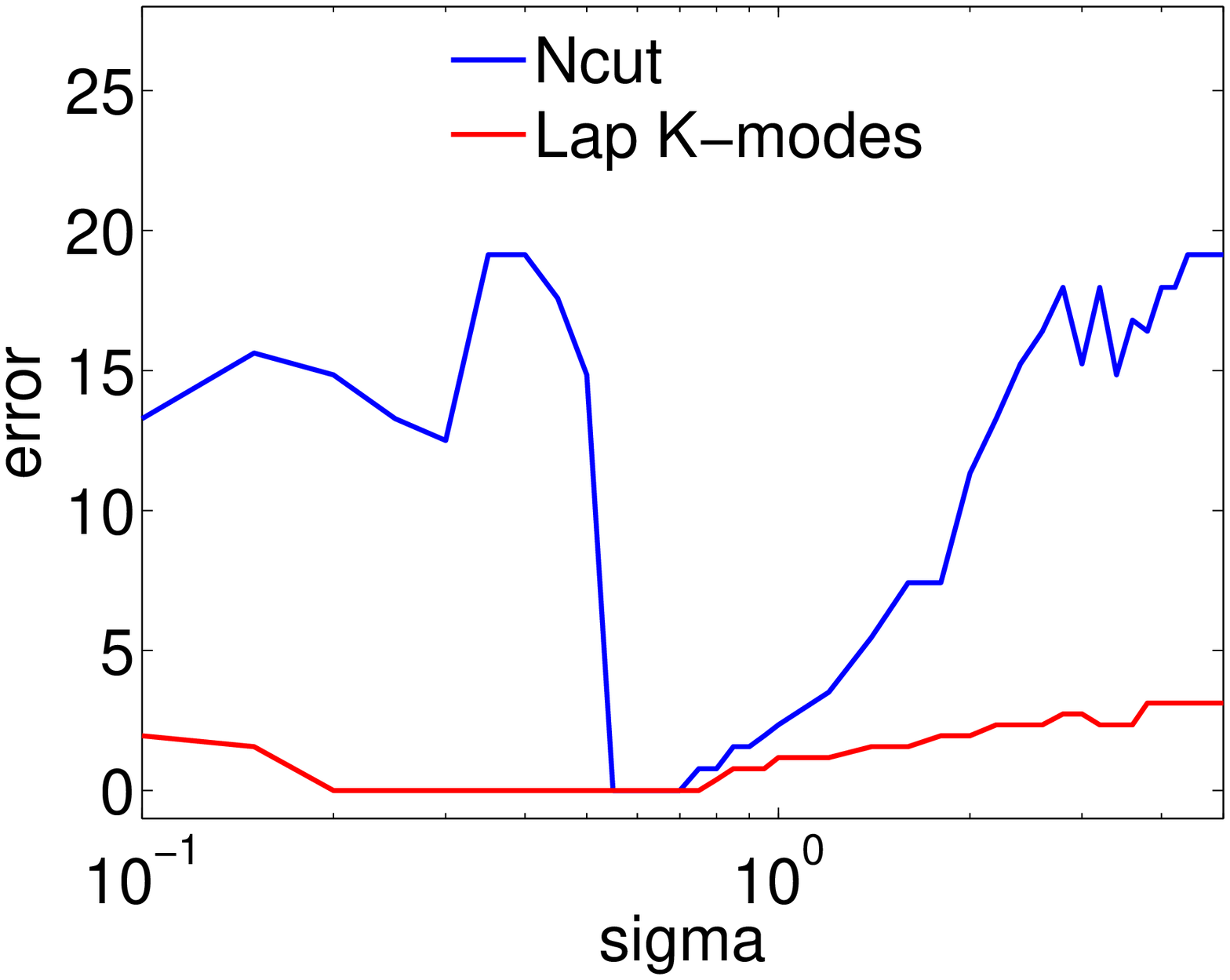}} &
    \includegraphics[width=0.23\linewidth]{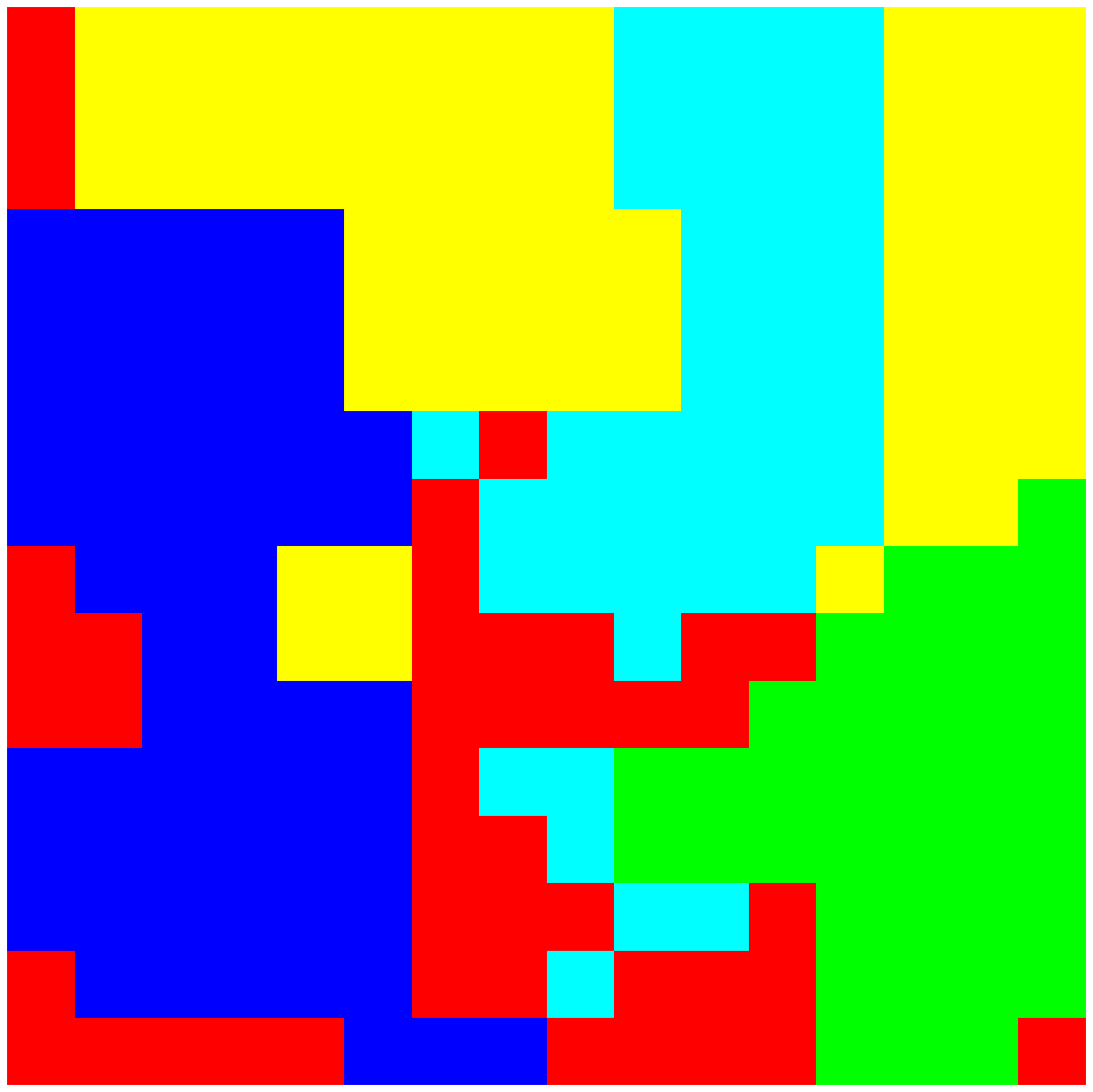}&
    \includegraphics[width=0.23\linewidth]{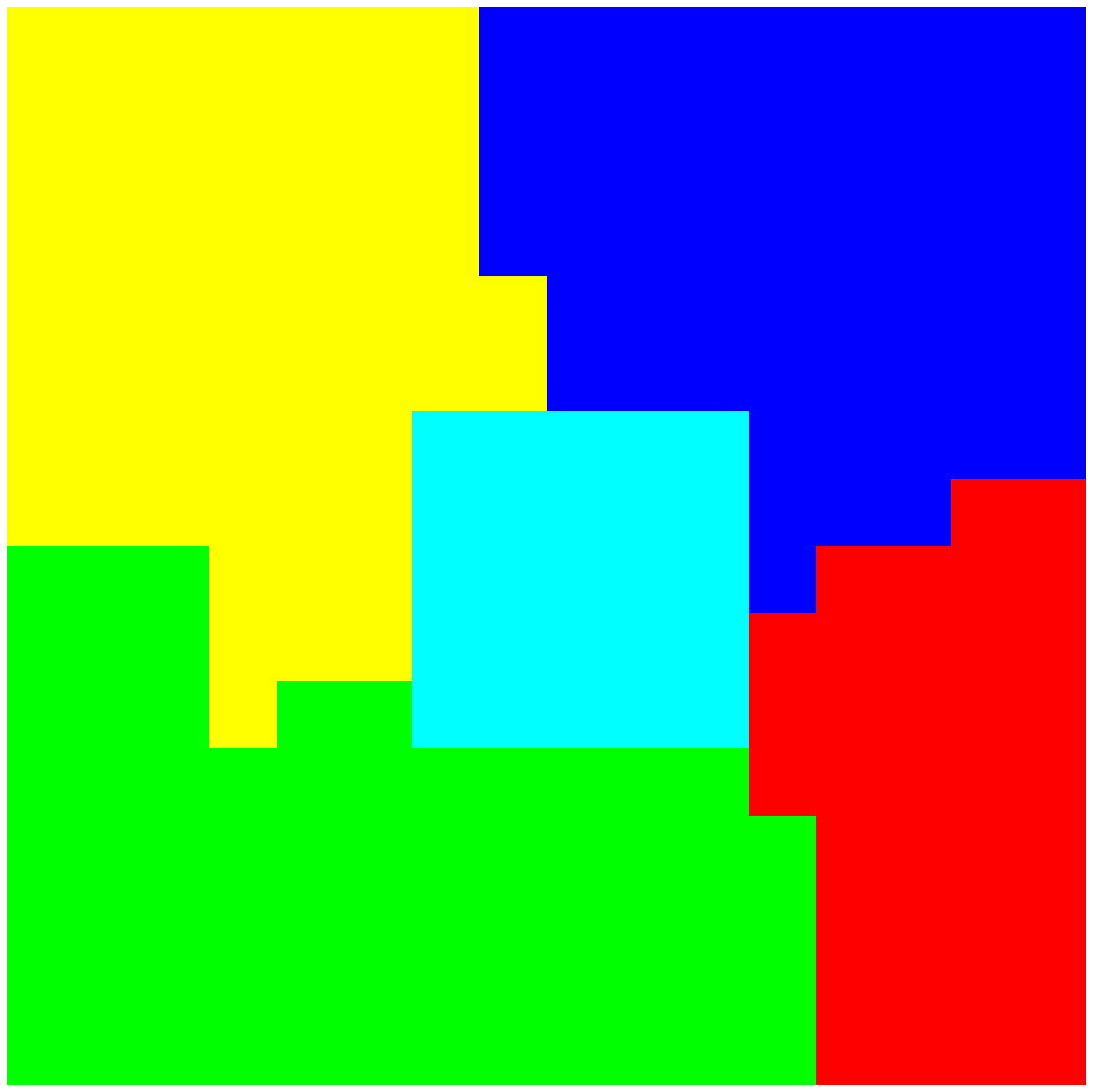}
  \end{tabular}
  \caption{Occluder segmentation result: original image; classification errors over the range of $\sigma$; segmentations of normalized cut and Laplacian $K$-modes at $\sigma=0.2$.}
  \label{f:occluder}
\end{figure}

\subsection{Figure-ground segmentation}

We consider the problem of segmenting an occluder from a textured background in a grayscale image. This problem has been shown to be difficult for spectral clustering \citep{ChennubJepson03a,CarreirZemel05a}, because of the intensity gradients between the occluder and the background (and within the background itself), which cause many graph edges to connect them, see the example in fig.~\ref{f:occluder}. We formalize it as a clustering problem and partition the pixels into $K=5$ clusters. We use for each pixel its 2D location and intensity value as features, and build a graph where each pixel is connected to the eight nearby pixels, with edges weighted using a heat kernel of width $\sigma$ ($\sigma$'s value equals that of the kde bandwidth for Laplacian $K$-modes). The goal is to have one of the clusters extract the occluder, which can then be separated from the background. To measure the performance, we choose the cluster that overlaps most with the occluder as positive prediction (the rest of the pixels are considered as background/negative prediction) and compute the classification error. Fig.~\ref{f:occluder} shows normalized cut \citep{YuShi03a} performs well for a narrow range of $\sigma$, while Laplacian $K$-modes (with fixed $\lambda=0.1$) has a much more stable performance when using the same graph: the range of good $\sigma$ values that produce a perfect segmentation is much wider. This shows the difference between our algorithm and spectral clustering: even though both algorithms impose smoothness on assignments, the graph Laplacian is only a regularization term in our model, and the kde term makes our algorithm more robust to the graph construction. As a more powerful algorithm, Laplacian $K$-modes has inherited the robustness properties from the original $K$-modes algorithm \citep{CarreirWang13a}.

\subsection{Clustering analysis}

\begin{table}[b]
  \centering
  \caption{Statistics (size, dimensionality, \# of classes) of three real world datasets.}
  \label{t:datasets}
  \begin{tabular}{|c||c|c|c|}
    \hline
    dataset & $N$ & $D$ & $K$ \\
    \hline
    MNIST & 2000 & 784 & 10 \\
    \hline
    COIL--20 & 1440 & 1024 & 20 \\
    \hline
    TDT2 & 9394 & 36771 & 30 \\
    \hline
  \end{tabular}
\end{table}

\begin{table}[bh!]
  \centering
  \caption{Clustering accuracy and normalized mutual information (\%) on 3 datasets.}
  \begin{tabular}{@{}c@{\hspace{1ex}}|c||c|c|c|c|c|c|c|c|@{}}
    \cline{2-9}
    & dataset & $K$-means   & $K$-modes  & GMS    &  NCut   & GNMF  & DCD  & Laplacian $K$-modes\\
    \cline{2-9}
    & MNIST   & 58.2        & 59.2      & 15.9   &  65.5   & 66.2  & 69.4 & \textbf{70.5} \\
    ACC \raisebox{0pt}[0pt][0pt]{$\Bigg\{\Bigg.$} & COIL--20& 66.5        & 67.2      & 27.2   &  79.0   & 75.3  & 71.5 & \textbf{81.0} \textbf{(81.5)}\\
    & TDT2    & 68.9        & 70.0      & N/A    &  88.4   & 88.6  & 55.1 & \textbf{91.4} \\
    \cline{2-9}
    & MNIST   & 53.3        & 53.6      & 6.51   & 66.9           & 64.9  & 65.6  & \textbf{68.8} \\
    NMI \raisebox{0pt}[0pt][0pt]{$\Bigg\{\Bigg.$} & COIL--20& 75.3        & 75.9      & 38.9   & \textbf{88.0}  & 87.5  & 77.6  & 87.3 \textbf{(88.0)}\\
    & TDT2    & 75.3        & 75.8      & N/A    & 83.7           & 83.7  & 68.6  & \textbf{88.8} \\
    \cline{2-9}
  \end{tabular}
  \label{t:expt-acc-nmi}
\end{table}

We report clustering statistics in datasets with known pattern class labels (which the algorithms did not use): (1) MNIST \citep{Lecun_98a}, which contains $28 \times 28$ grayscale handwritten digit images (we randomly sample 200 of each digit); (2) COIL--20 \citep{Nene_96a}, which contains $32\times 32$ grayscale images of 20 objects viewed from varying angles; (3) the NIST Topic Detection and Tracking (TDT2) corpus, which contains on-topic documents of different semantic categories (documents appearing in more than one category are removed and only the largest 30 categories are kept). Statistics of the datasets are collected in table~\ref{t:datasets}. Datasets (2) and (3) are the same as used by \citet{Cai_11b}, and we also use the same features: pixel values for (1) and (2), and TFIDF for (3).

We compare the following algorithms: $K$-means, initialized randomly; $K$-modes, a special case of Laplacian $K$-modes with $\lambda=0$; Gaussian mean-shift (GMS), we search for $\sigma$ that produces exactly $K$ modes; Normalized cut (NCut), one typical spectral clustering algorithm, we use the implementation of \citet{YuShi03a}; Graph regularized NMF (GNMF) proposed by \citet{Cai_11b}; Data-Cluster-Data random walk (DCD) proposed by \citet{YangOja12a}; and Laplacian $K$-modes, initialized from $K$-means. $K$ is set to the number of classes in the ground truth. All the datasets are normalized to have unit norm per sample.

Several algorithms use the graph Laplacian: for NCut, GNMF, and Laplacian $K$-modes, we build a 5-nearest-neighbor graph and use a binary weighting scheme to compute the graph Laplacian (as in \citealp{Cai_11b}); for DCD, we find that it achieves better performance using a graph built with a larger neighborhood size, so we let DCD select its optimal size in $\{5,10,20,30\}$. We run each algorithm with $20$ random restarts, letting them use optimal values for their respective hyperparameters (if they have any) based on a grid search, and report the best performance from different random restarts. Notice we do not use the out-of-sample mapping here because its does not exist for all algorithms we compare with. Clustering accuracy (ACC) and normalized mutual information (NMI), two widely used criteria \citep{Cai_11b,Arora_11a}, are used for evaluation. The results are given in table~\ref{t:expt-acc-nmi} (N/A means our GMS code ran out of memory).

It is clear that algorithms using Laplacian smoothing are in general superior than algorithms not using it, which demonstrates the importance of the graph Laplacian in separating nonconvex and manifold clusters. GMS performs poorly for the reasons described earlier. On all datasets, Laplacian $K$-modes achieves the best or close to best performance under both criteria. We find there exists a wide range of hyperparameters with which our algorithm gives very competitive performance. We are able to further improve our performance on COIL--20 using the homotopy technique described earlier: we fix $\lambda$ at $0.01$, and decrease $\sigma$ from $0.45$ to $0.1$ gradually in $7$ steps, initializing the algorithm for the current $\sigma$ value from the solution for the previous $\sigma$ value. This improved result is shown in parenthesis in table~\ref{t:expt-acc-nmi}.

\begin{figure}[t]
  \centering
  \begin{tabular}[b]{@{}c@{\hspace{0\linewidth}}c@{\hspace{0\linewidth}}c@{\hspace{0\linewidth}}c@{\hspace{0\linewidth}}c@{\hspace{0\linewidth}}c@{\hspace{0\linewidth}}c@{\hspace{0\linewidth}}c@{\hspace{0\linewidth}}c@{\hspace{0\linewidth}}c@{\hspace{0\linewidth}}c@{}}
    \rotatebox{90}{\small\hspace{1ex}$K$-means} &
    \includegraphics[width=0.095\linewidth,bb=142 226 494 578,clip]{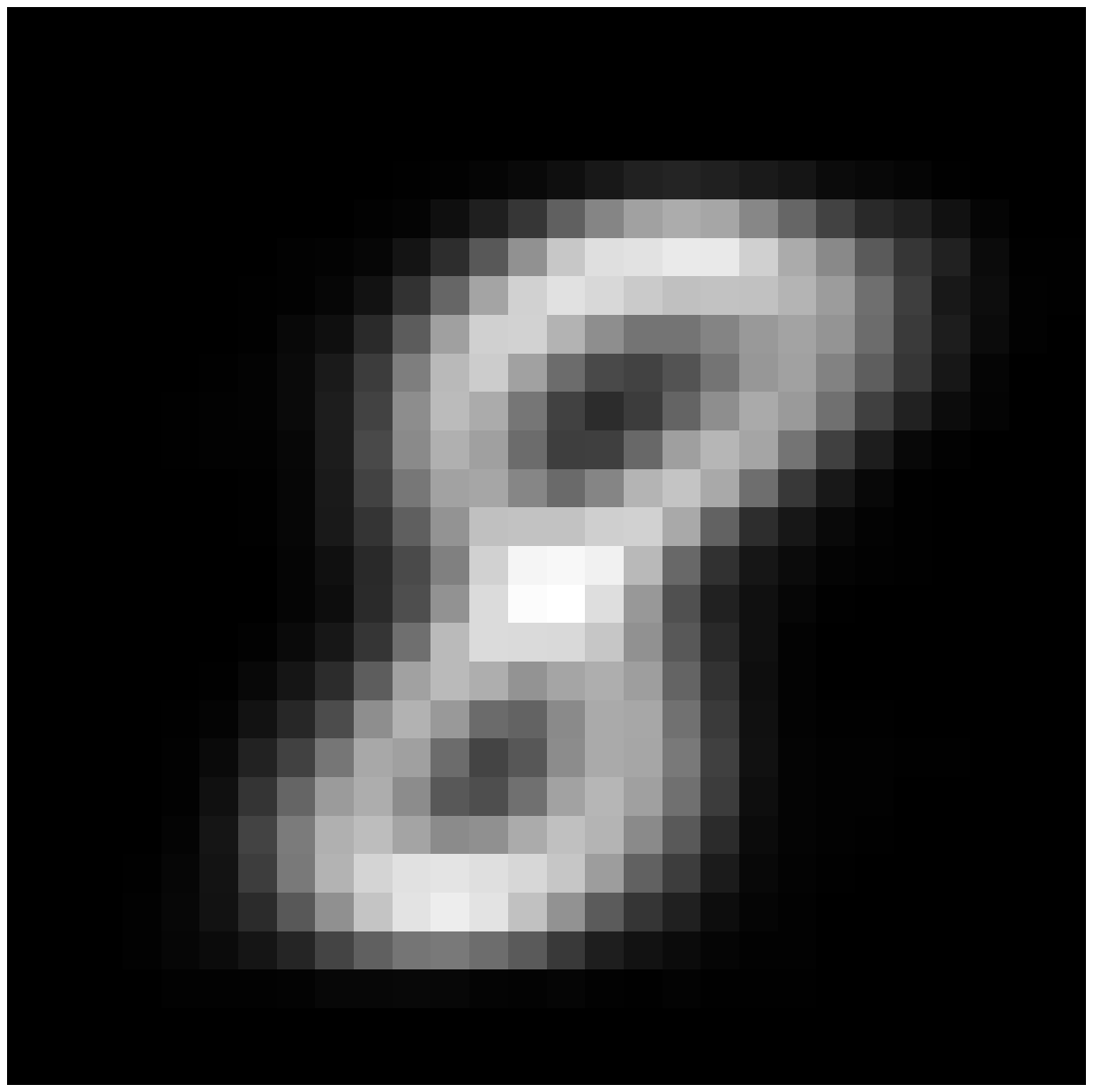} &
    \includegraphics[width=0.095\linewidth,bb=142 226 494 578,clip]{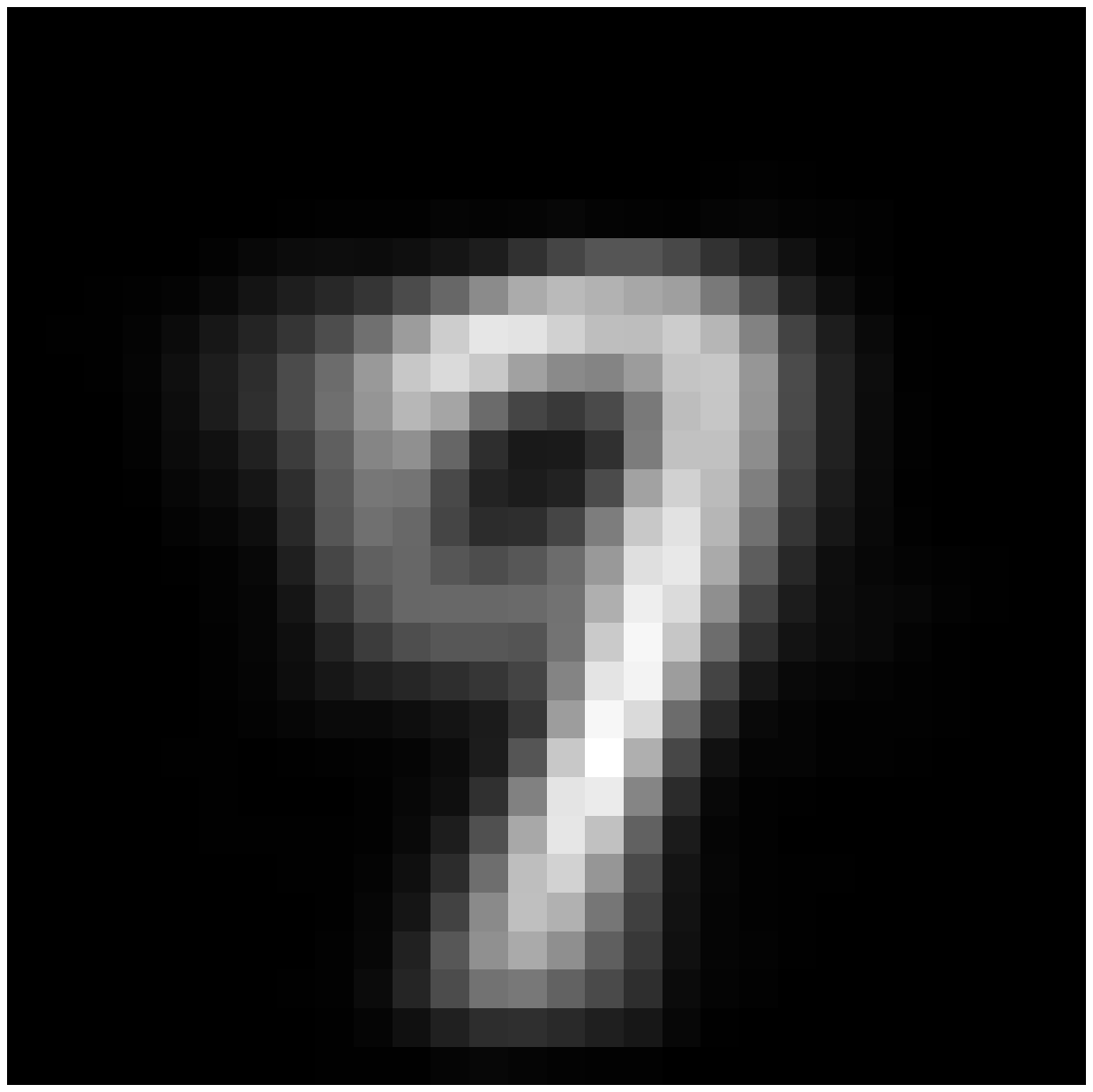} &
    \includegraphics[width=0.095\linewidth,bb=142 226 494 578,clip]{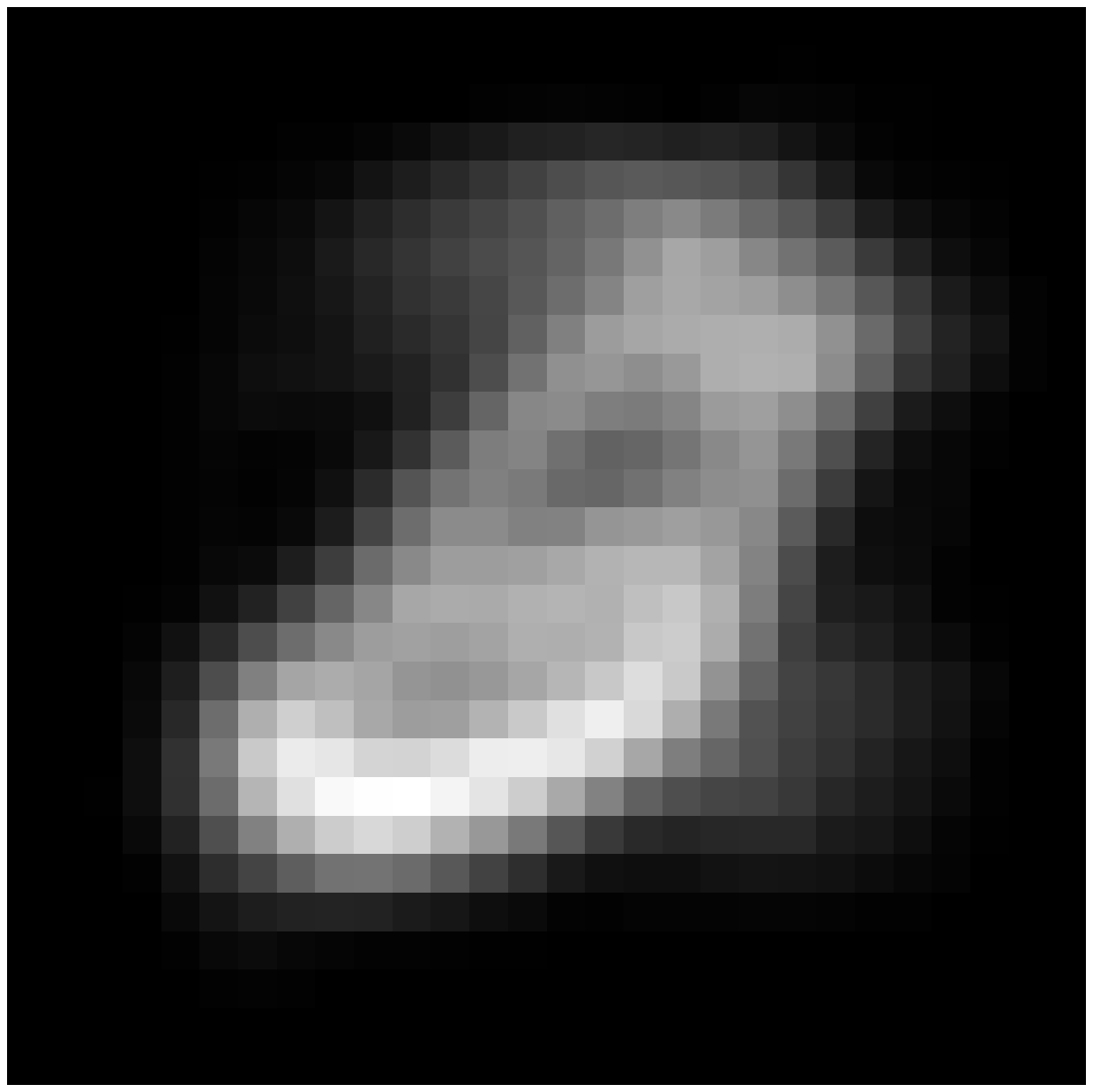} &
    \includegraphics[width=0.095\linewidth,bb=142 226 494 578,clip]{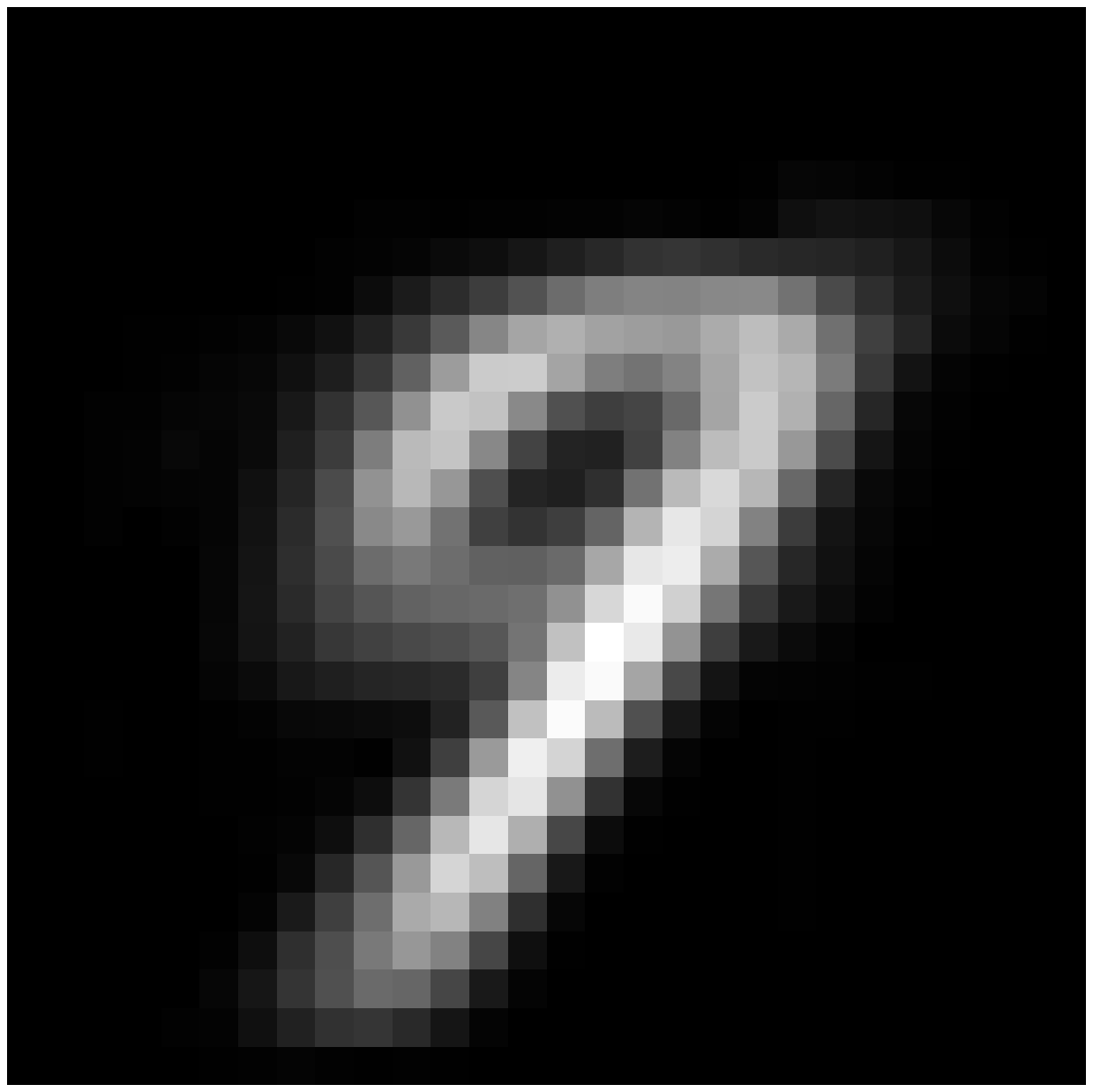} &
    \includegraphics[width=0.095\linewidth,bb=142 226 494 578,clip]{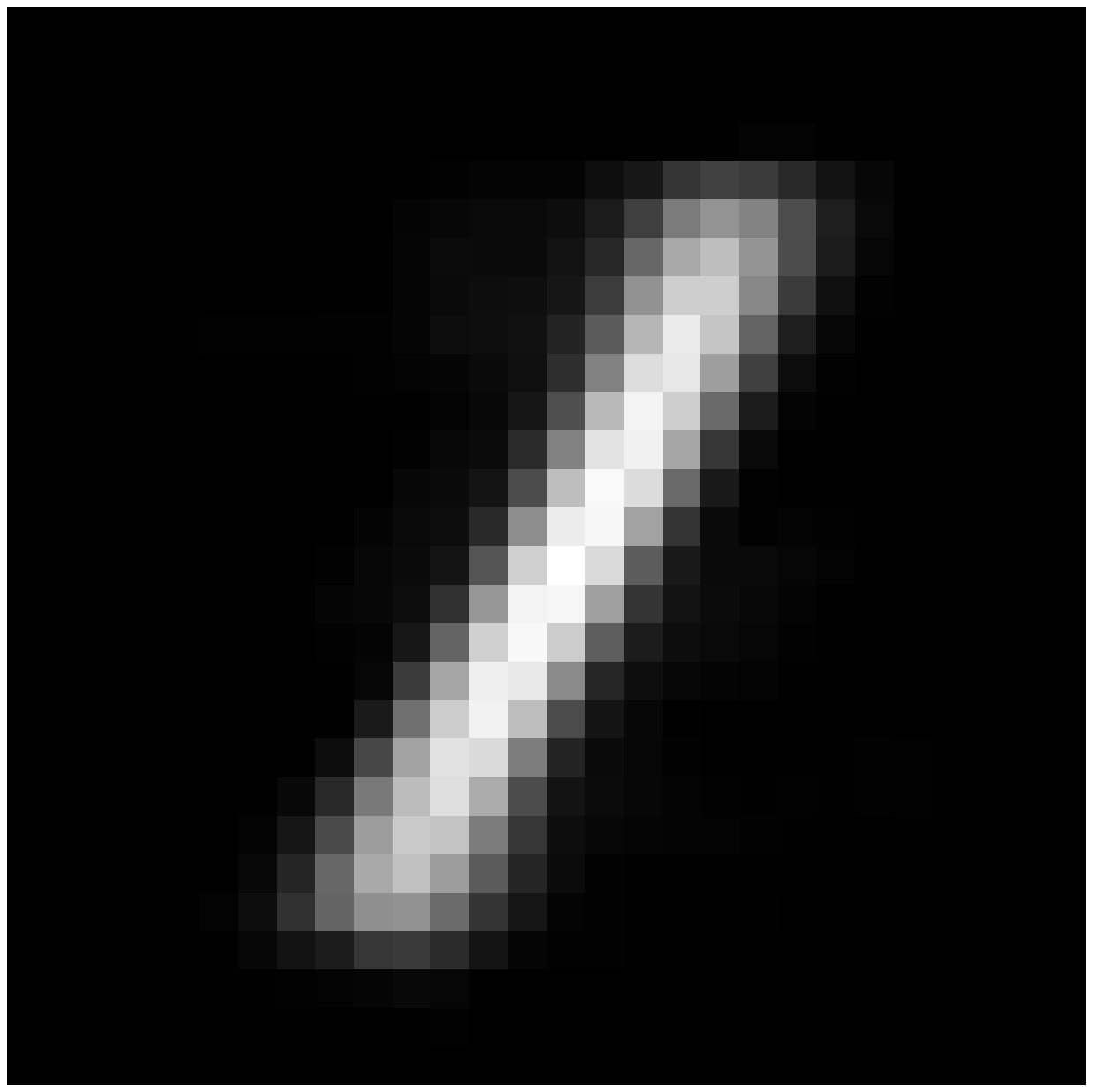} &
    \includegraphics[width=0.095\linewidth,bb=142 226 494 578,clip]{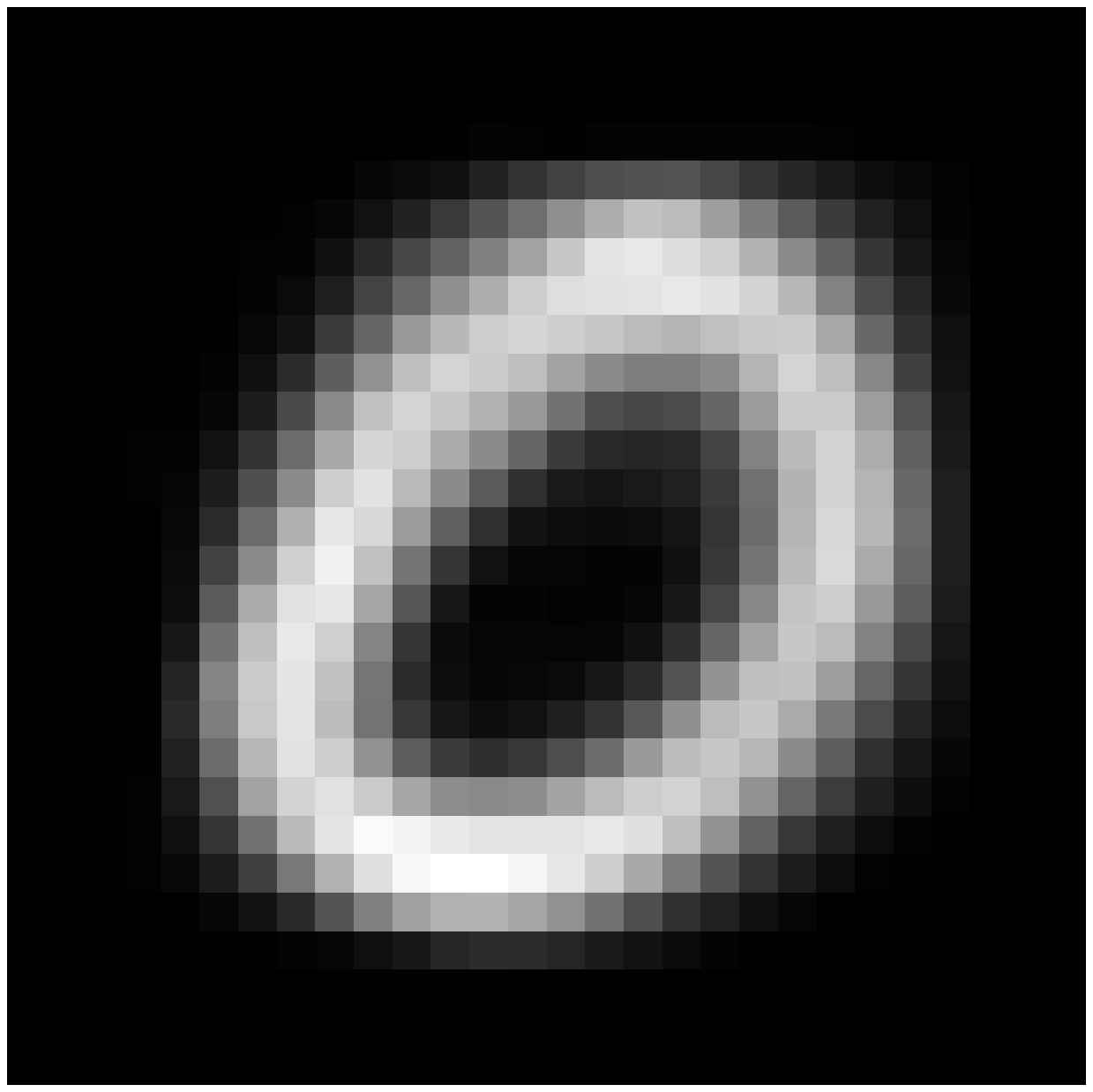} &
    \includegraphics[width=0.095\linewidth,bb=142 226 494 578,clip]{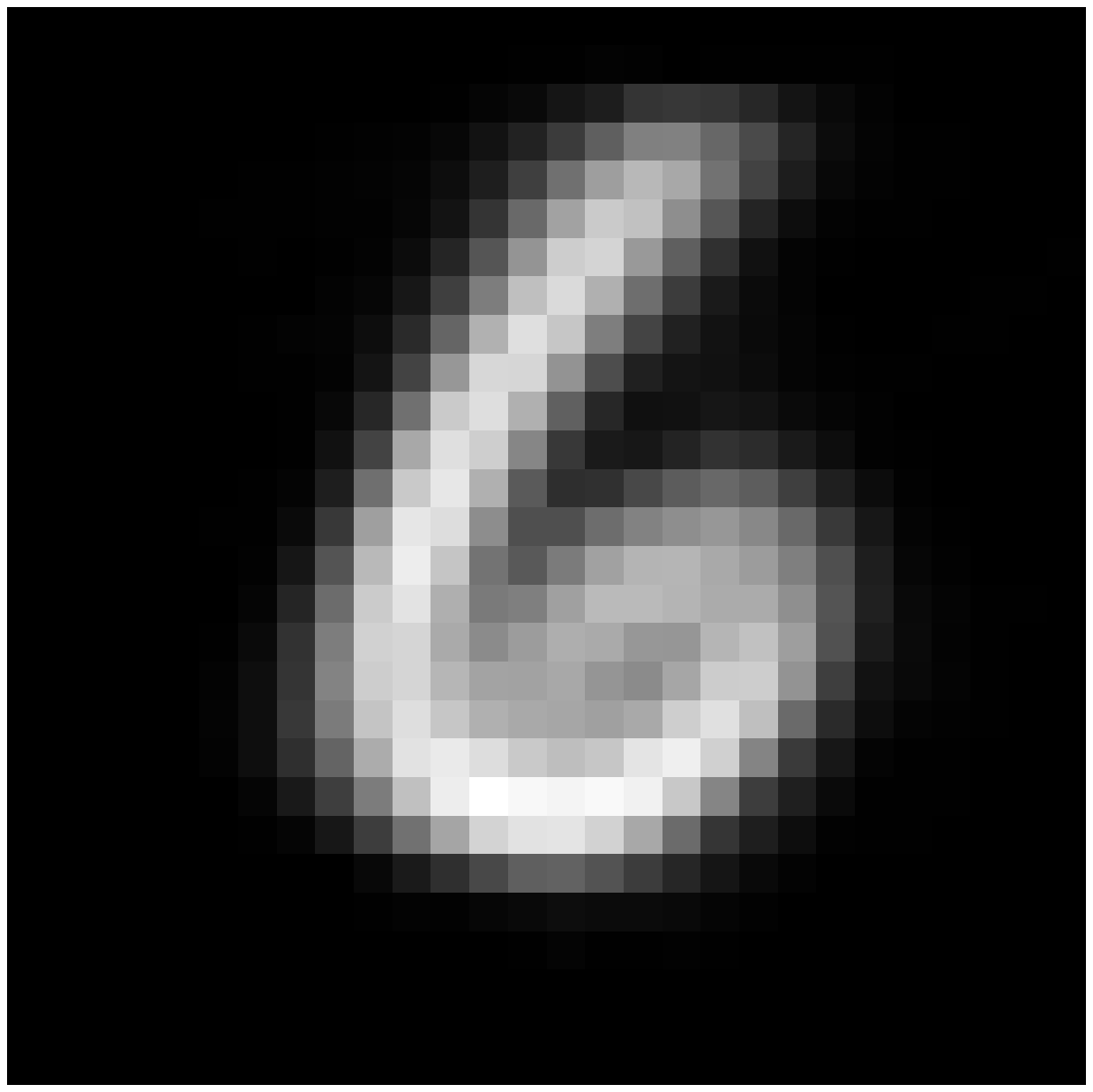} &
    \includegraphics[width=0.095\linewidth,bb=142 226 494 578,clip]{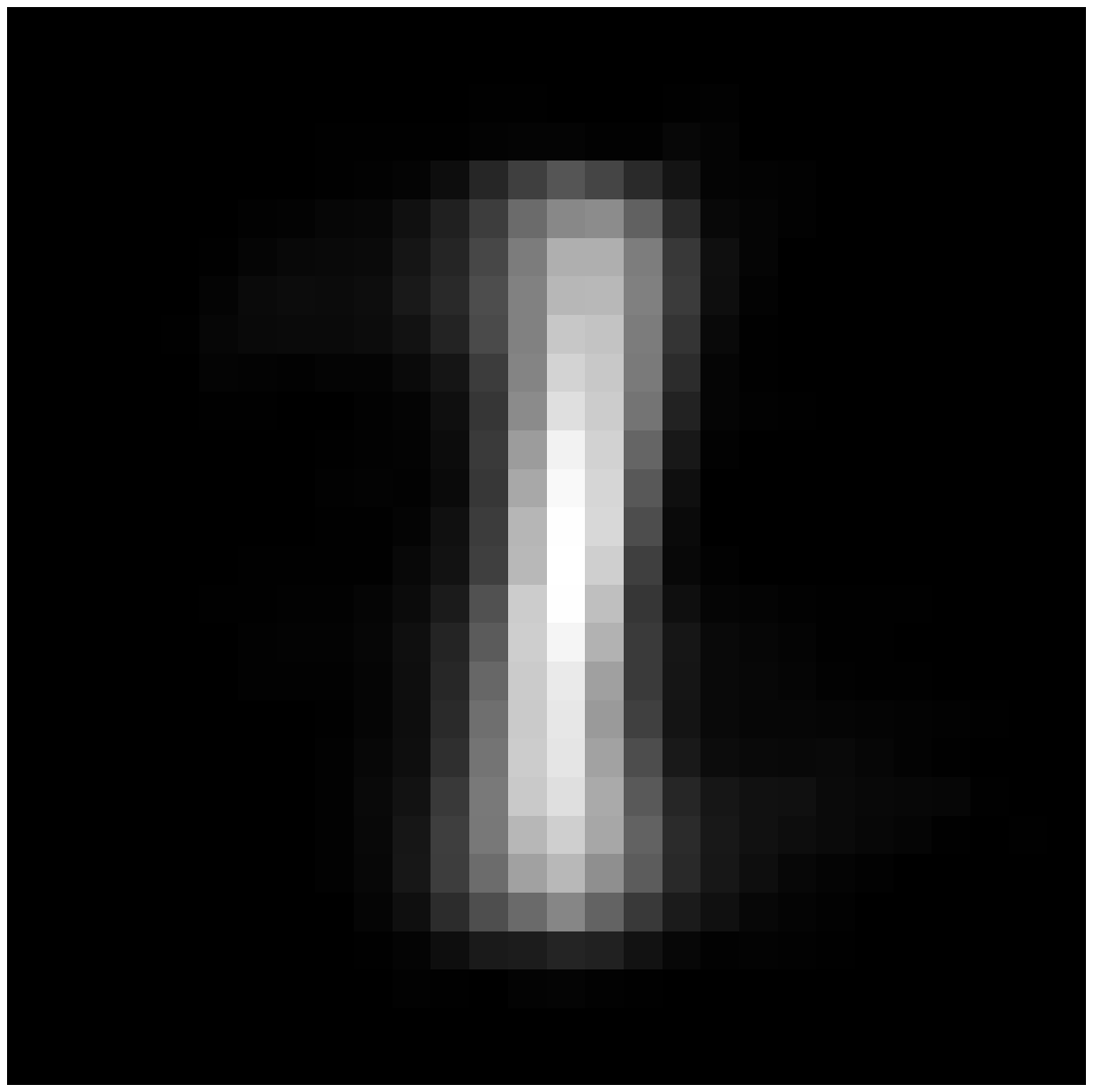} &
    \includegraphics[width=0.095\linewidth,bb=142 226 494 578,clip]{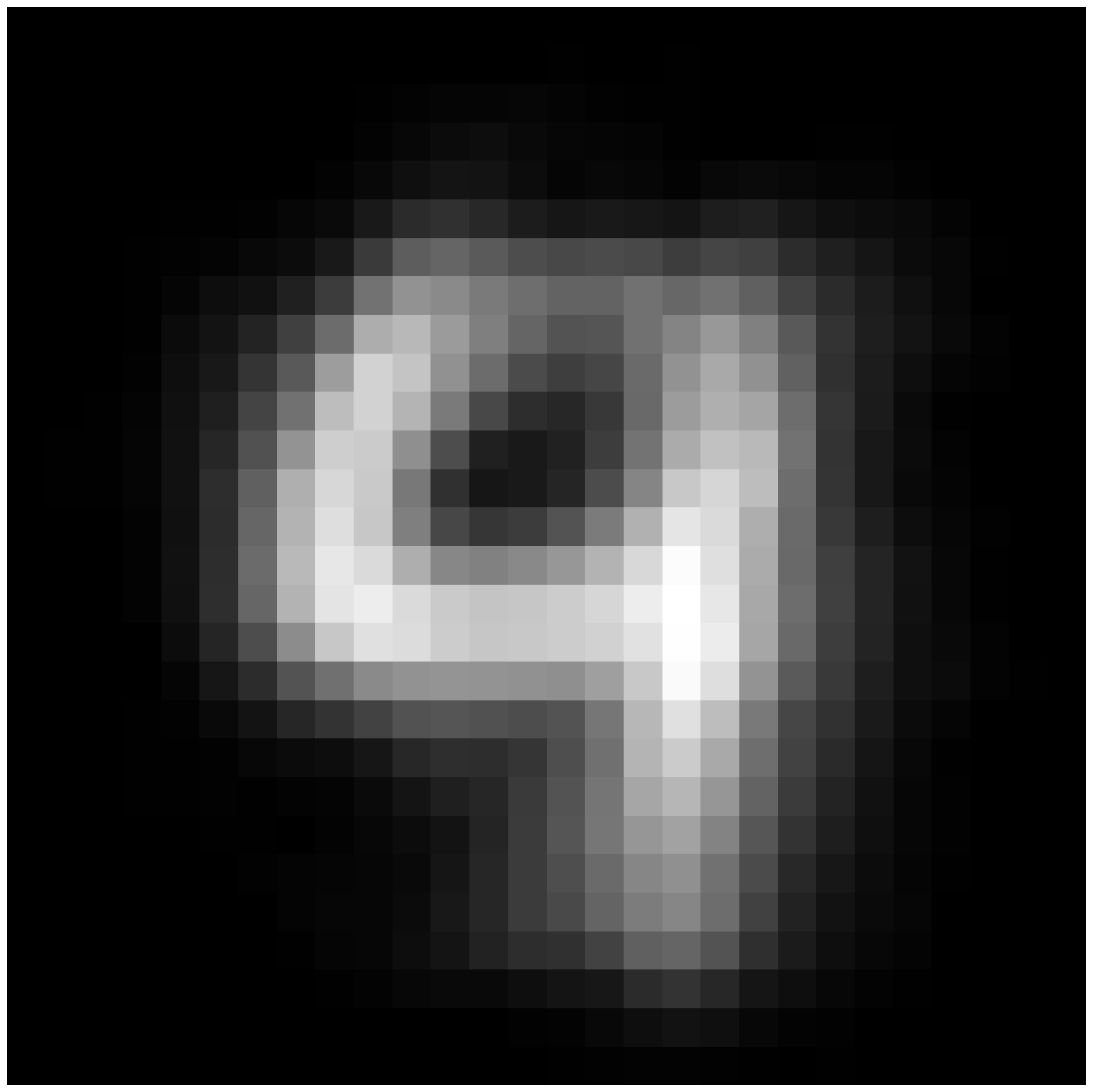} &
    \includegraphics[width=0.095\linewidth,bb=142 226 494 578,clip]{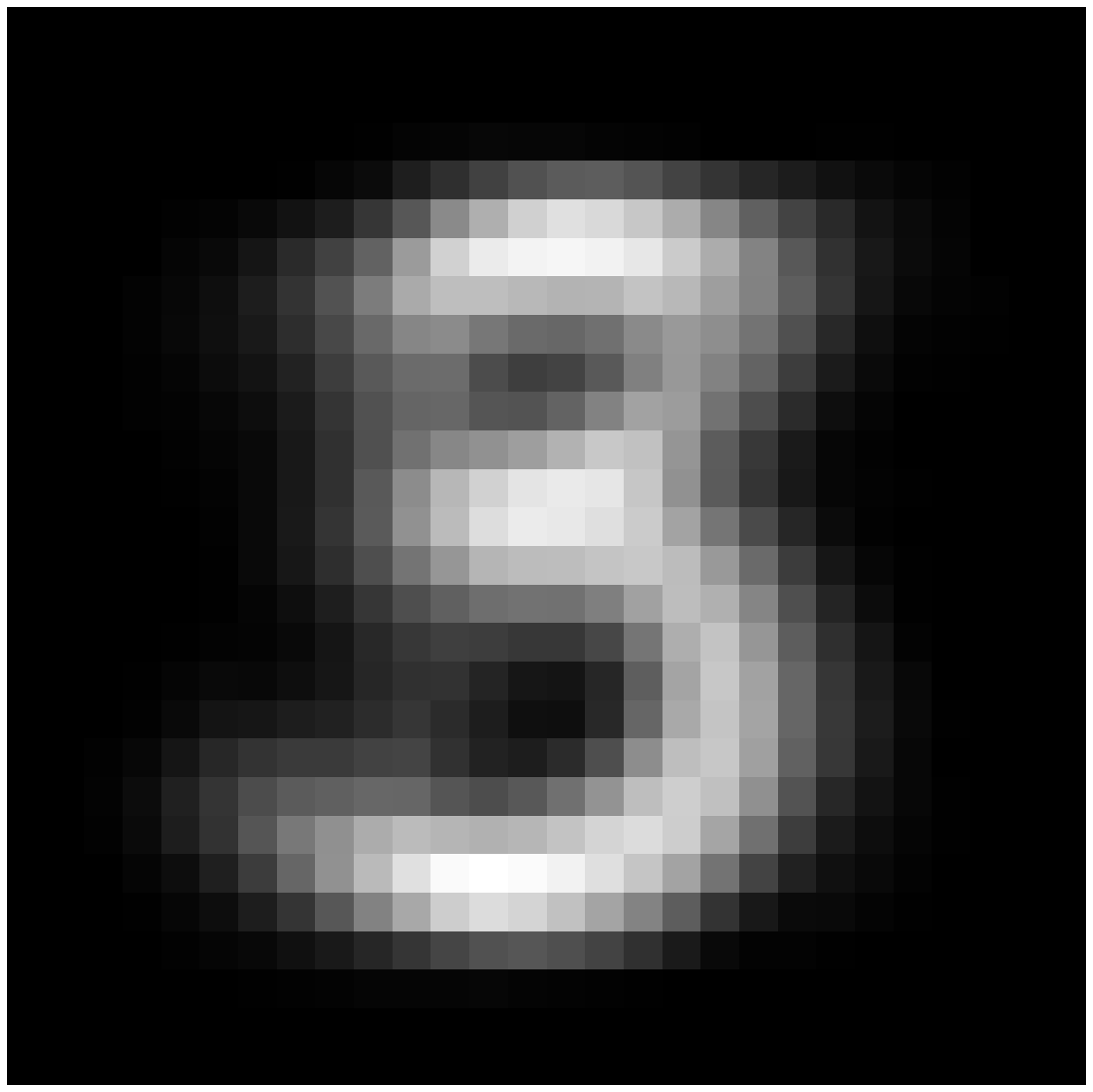} \\[-1ex]
    \rotatebox{90}{\small\hspace{0.5ex}$K$-modes} &
    \includegraphics[width=0.095\linewidth,bb=142 226 494 578,clip]{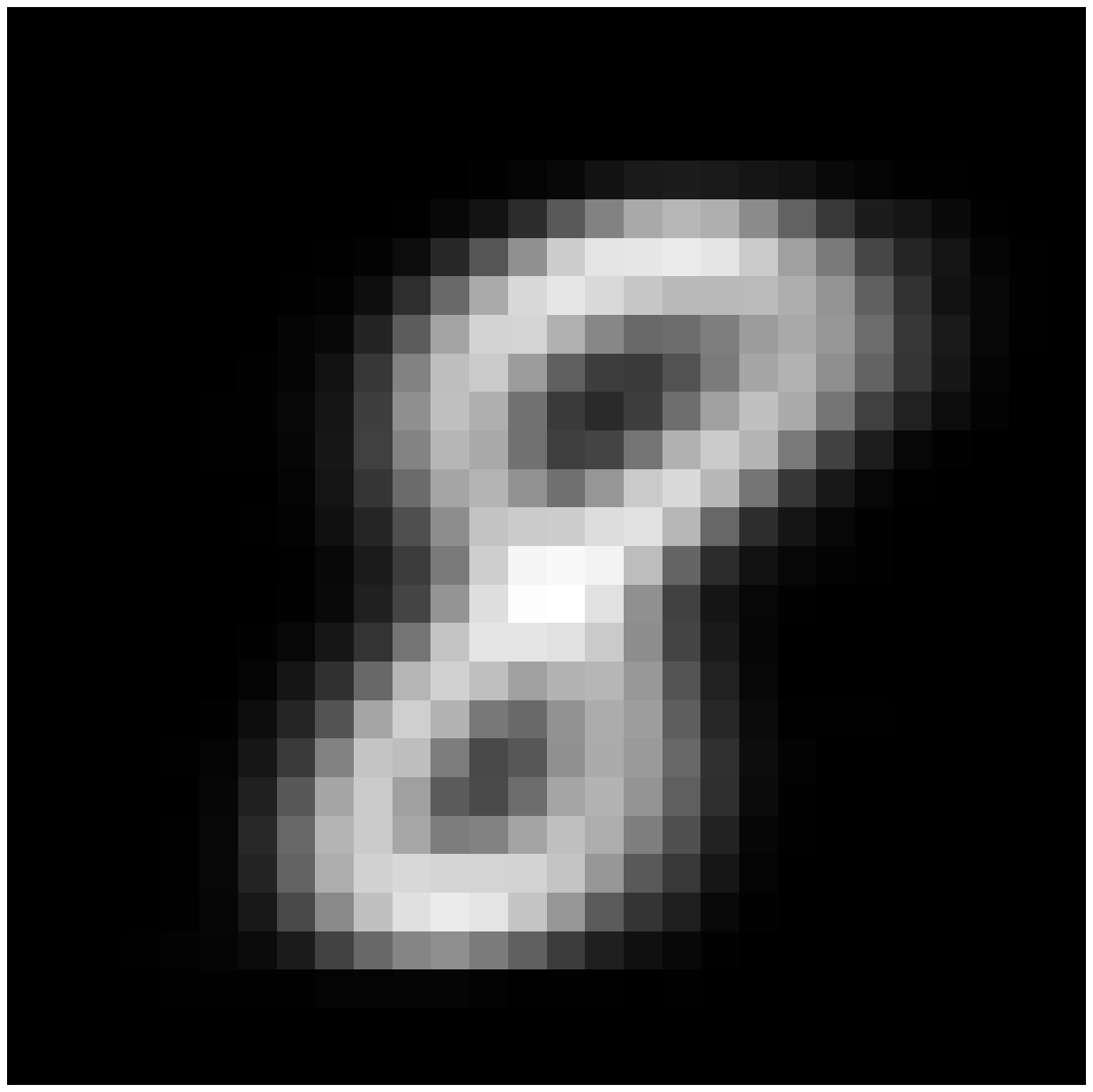} &
    \includegraphics[width=0.095\linewidth,bb=142 226 494 578,clip]{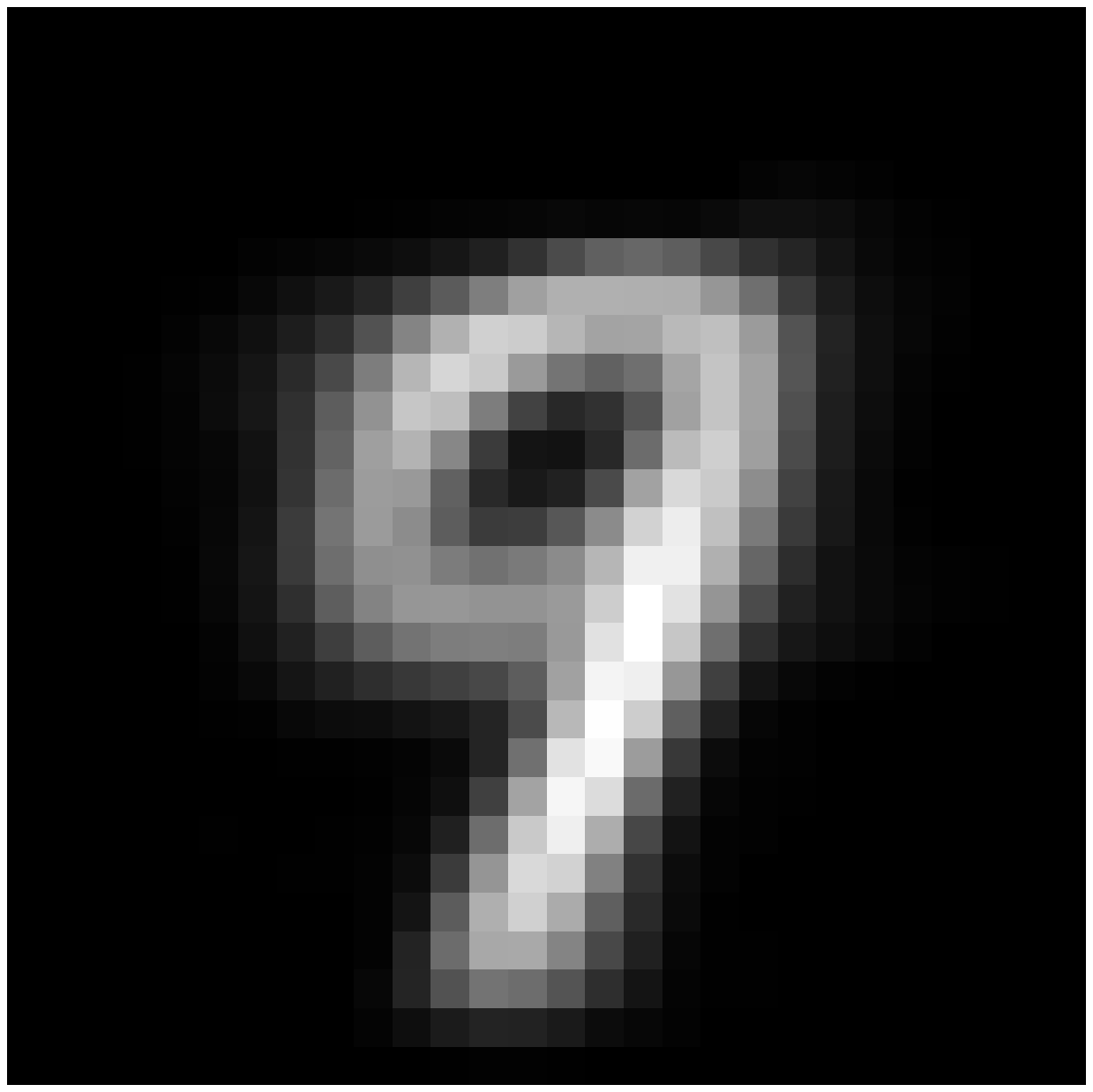} &
    \includegraphics[width=0.095\linewidth,bb=142 226 494 578,clip]{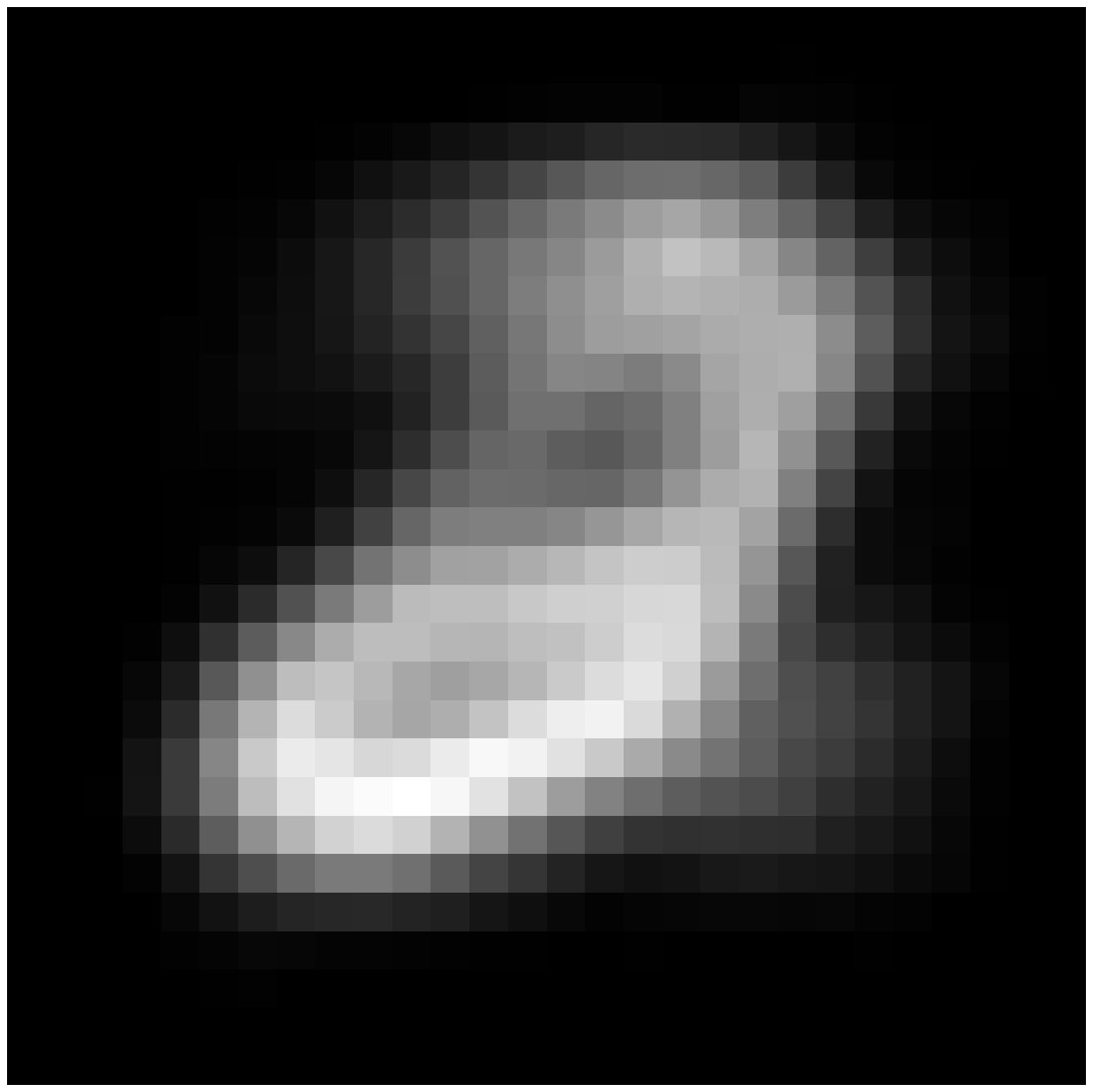} &
    \includegraphics[width=0.095\linewidth,bb=142 226 494 578,clip]{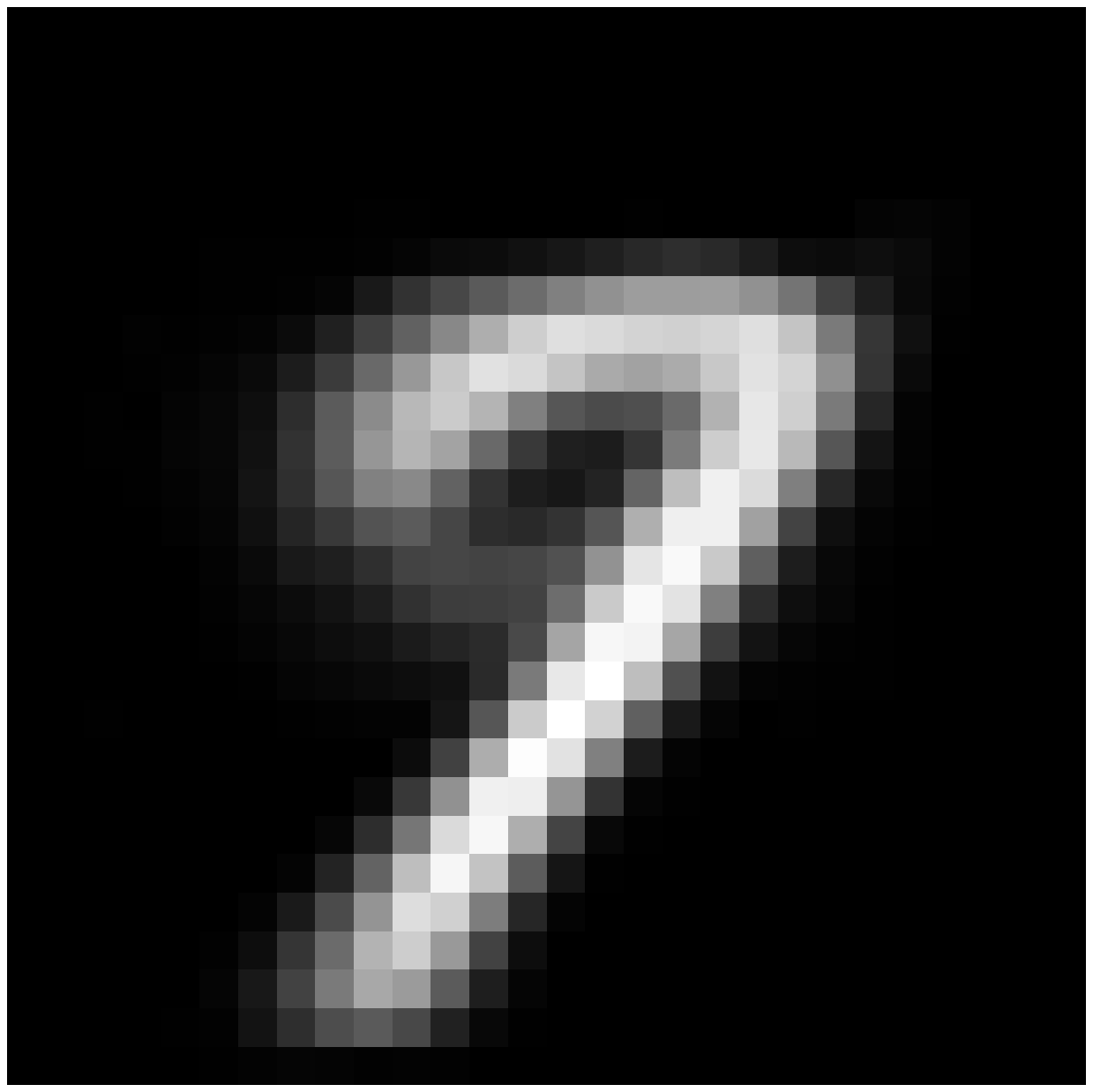} &
    \includegraphics[width=0.095\linewidth,bb=142 226 494 578,clip]{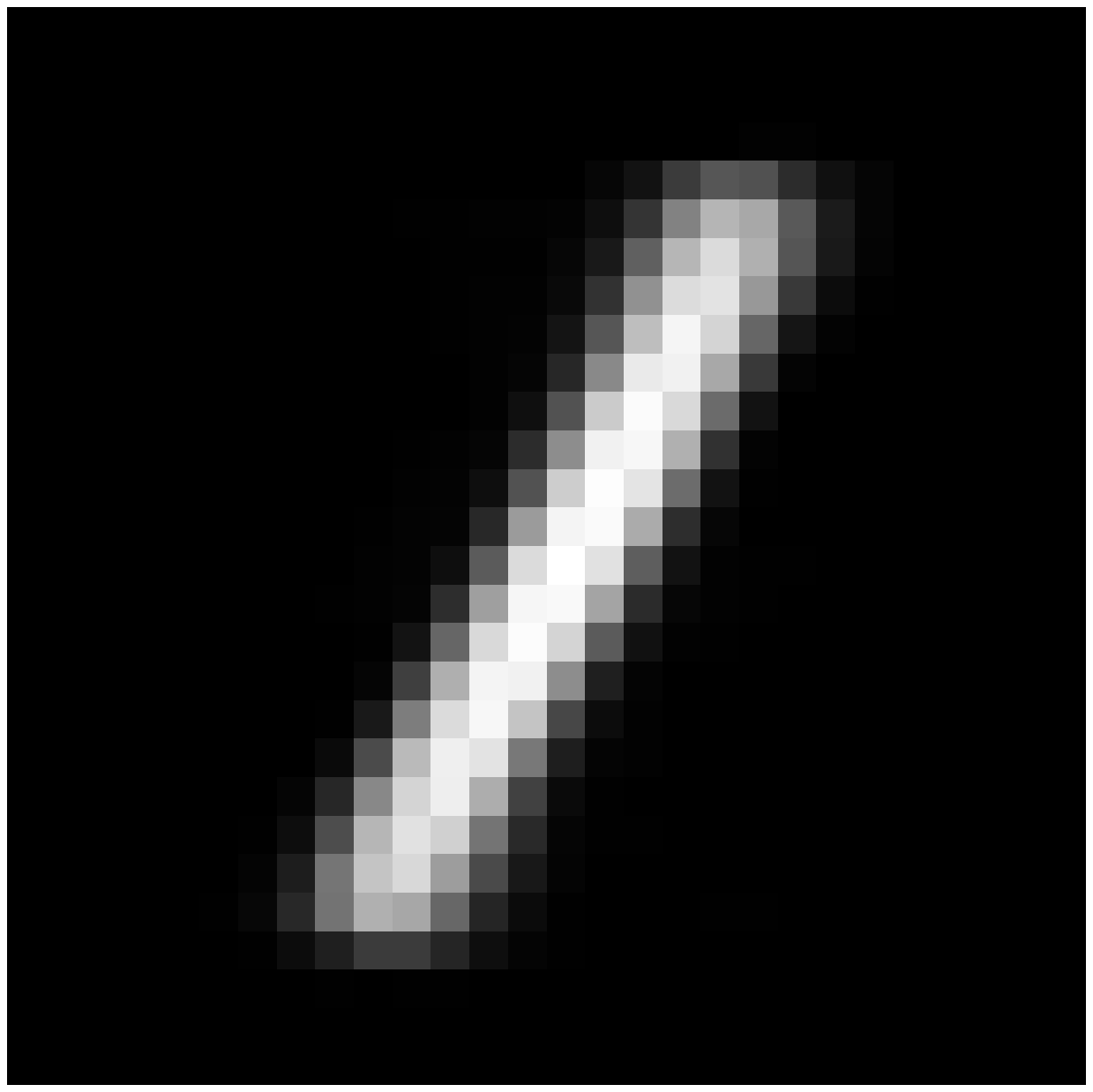} &
    \includegraphics[width=0.095\linewidth,bb=142 226 494 578,clip]{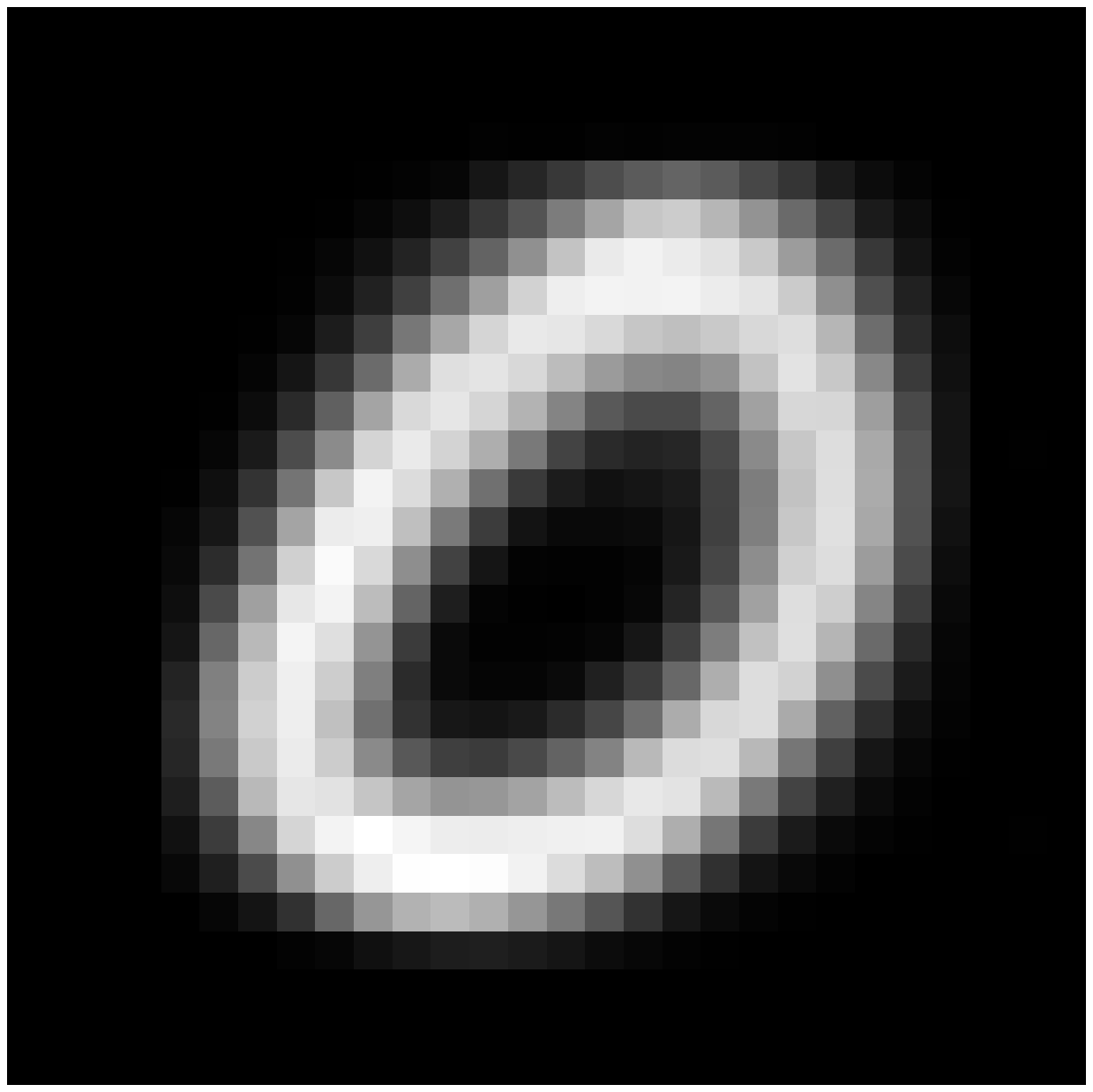} &
    \includegraphics[width=0.095\linewidth,bb=142 226 494 578,clip]{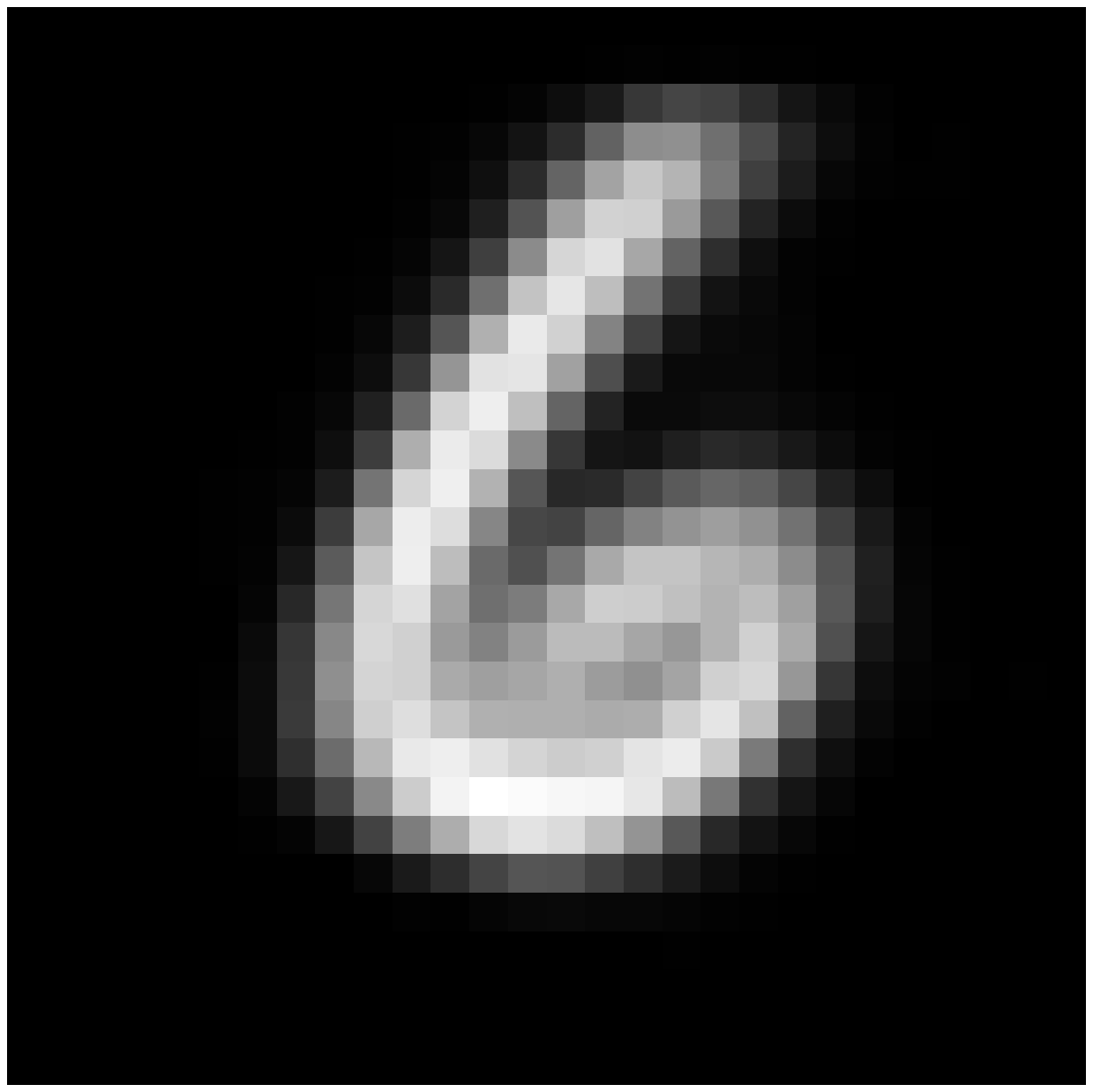} &
    \includegraphics[width=0.095\linewidth,bb=142 226 494 578,clip]{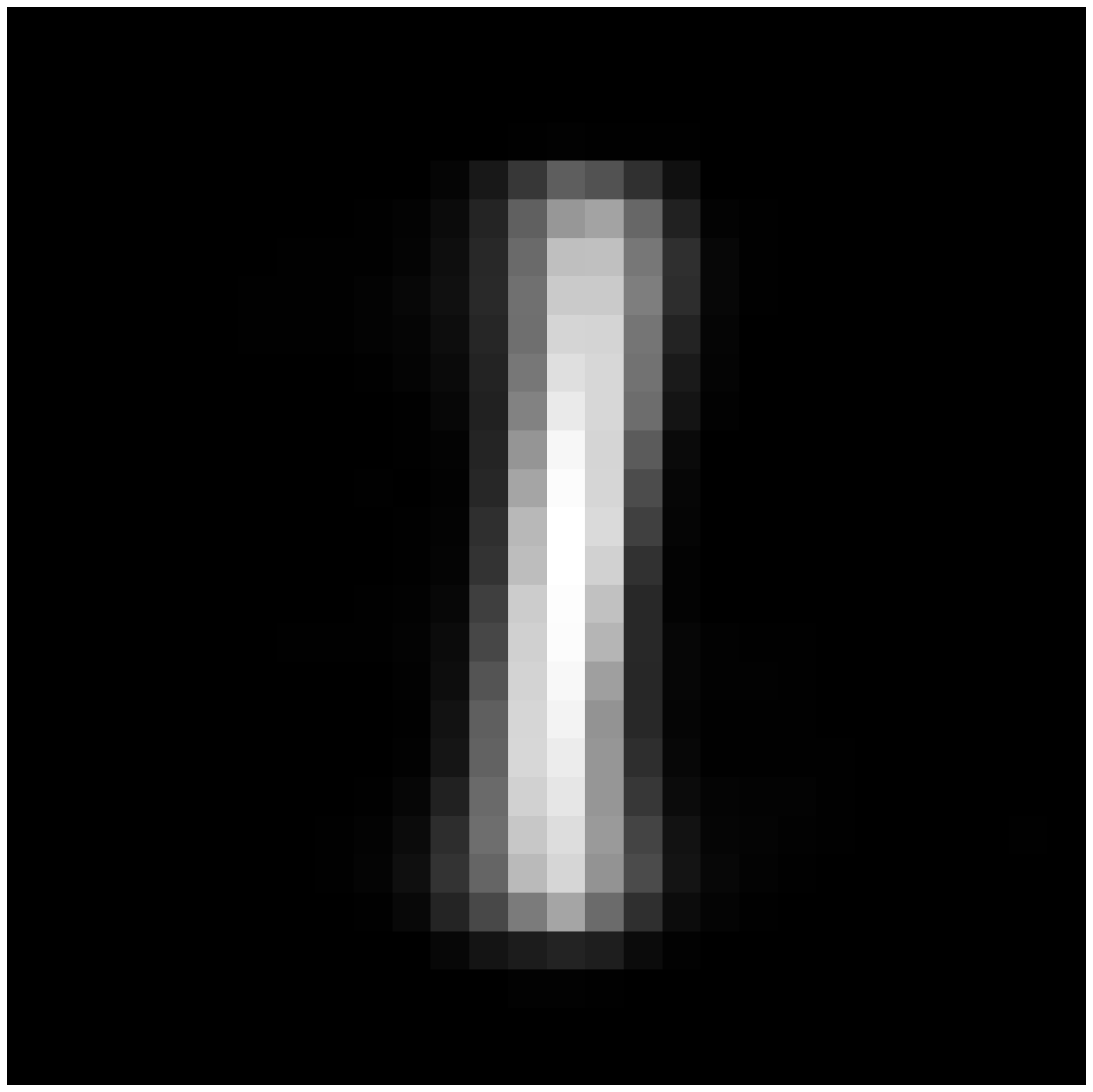} &
    \includegraphics[width=0.095\linewidth,bb=142 226 494 578,clip]{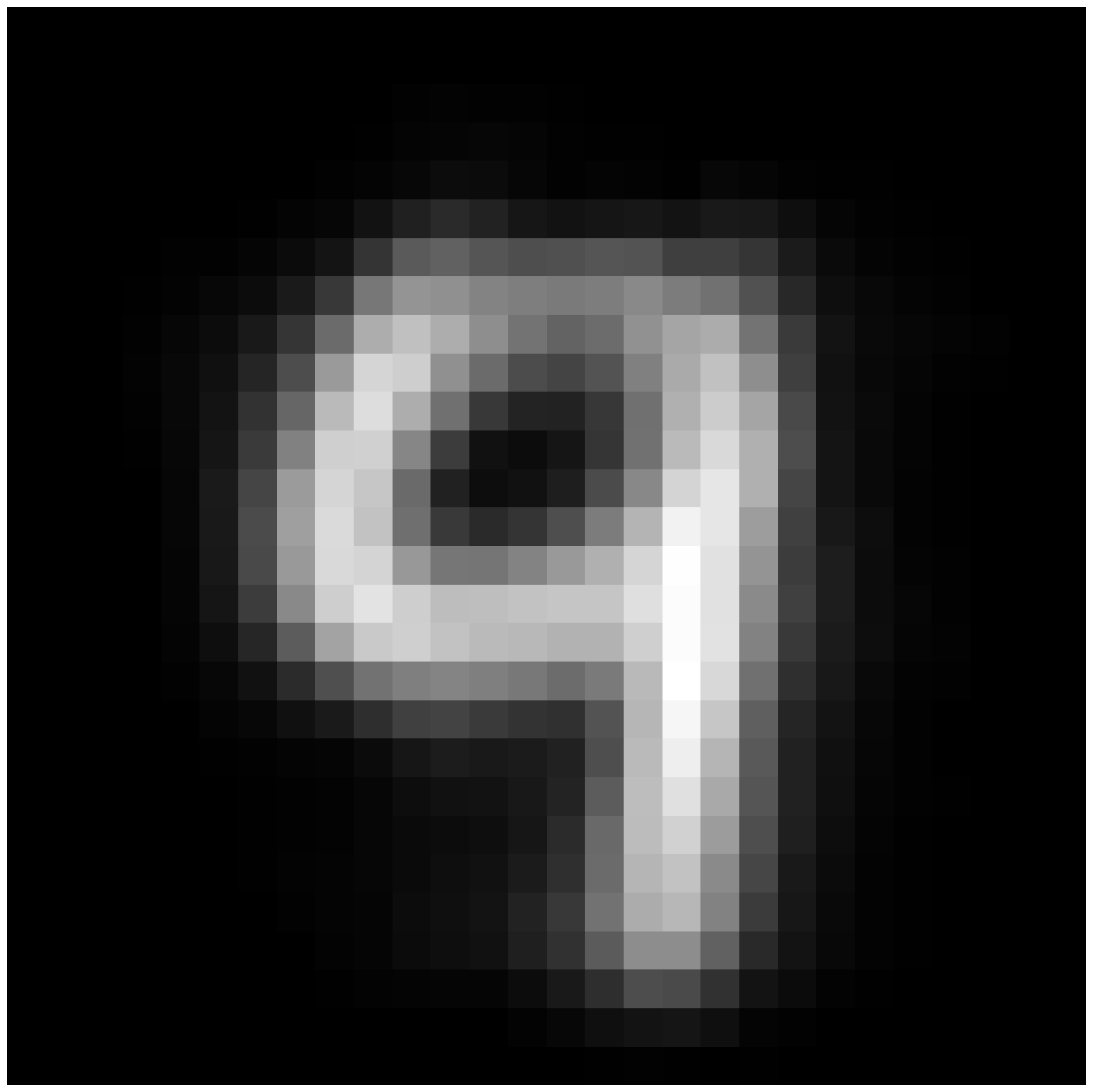} &
    \includegraphics[width=0.095\linewidth,bb=142 226 494 578,clip]{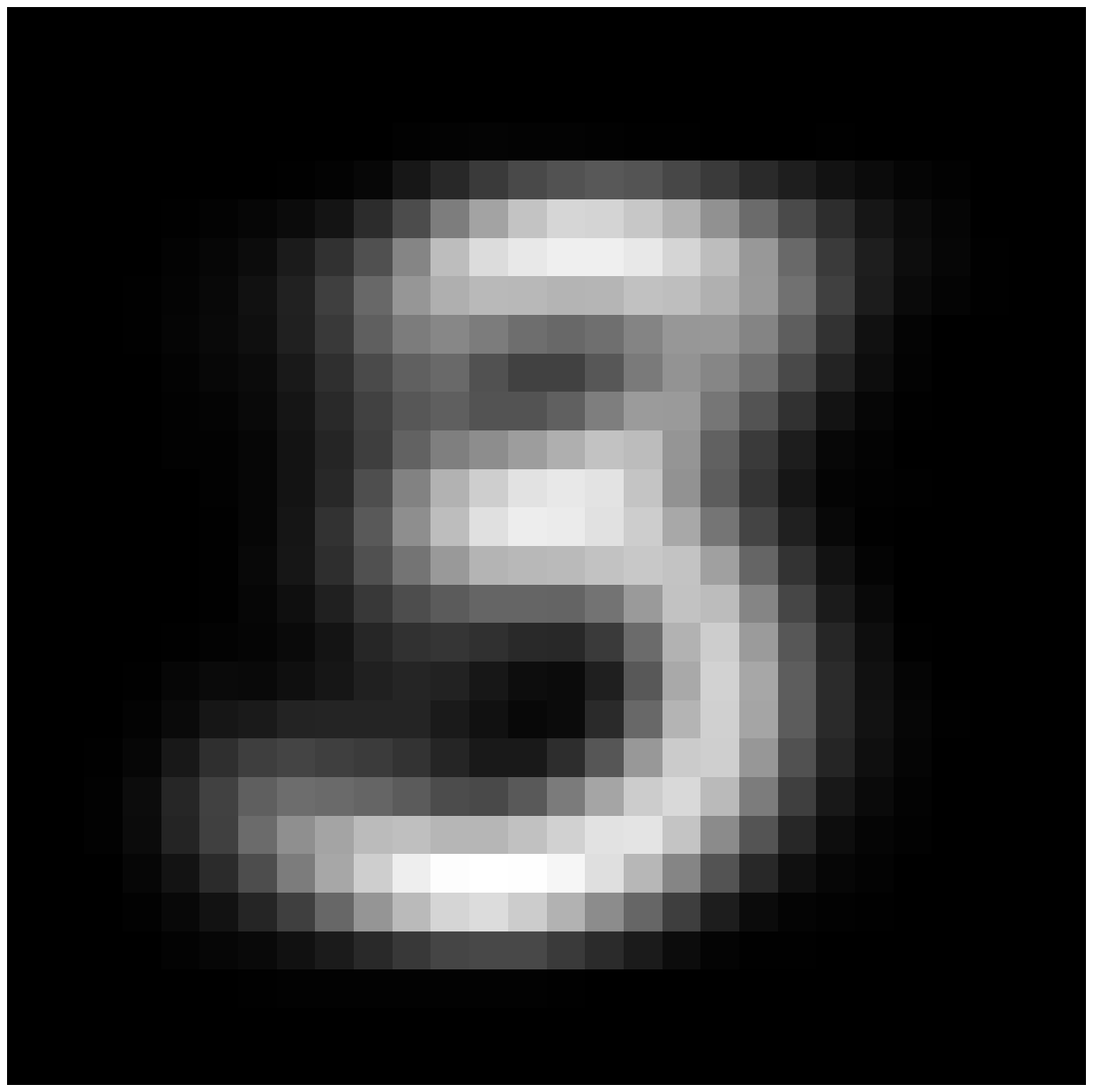} \\[-1ex]
    \rotatebox{90}{\small\hspace{0.5ex}\caja{c}{c}{Laplacian \\ $K$-modes}} &
    \includegraphics[width=0.095\linewidth,bb=142 226 494 578,clip]{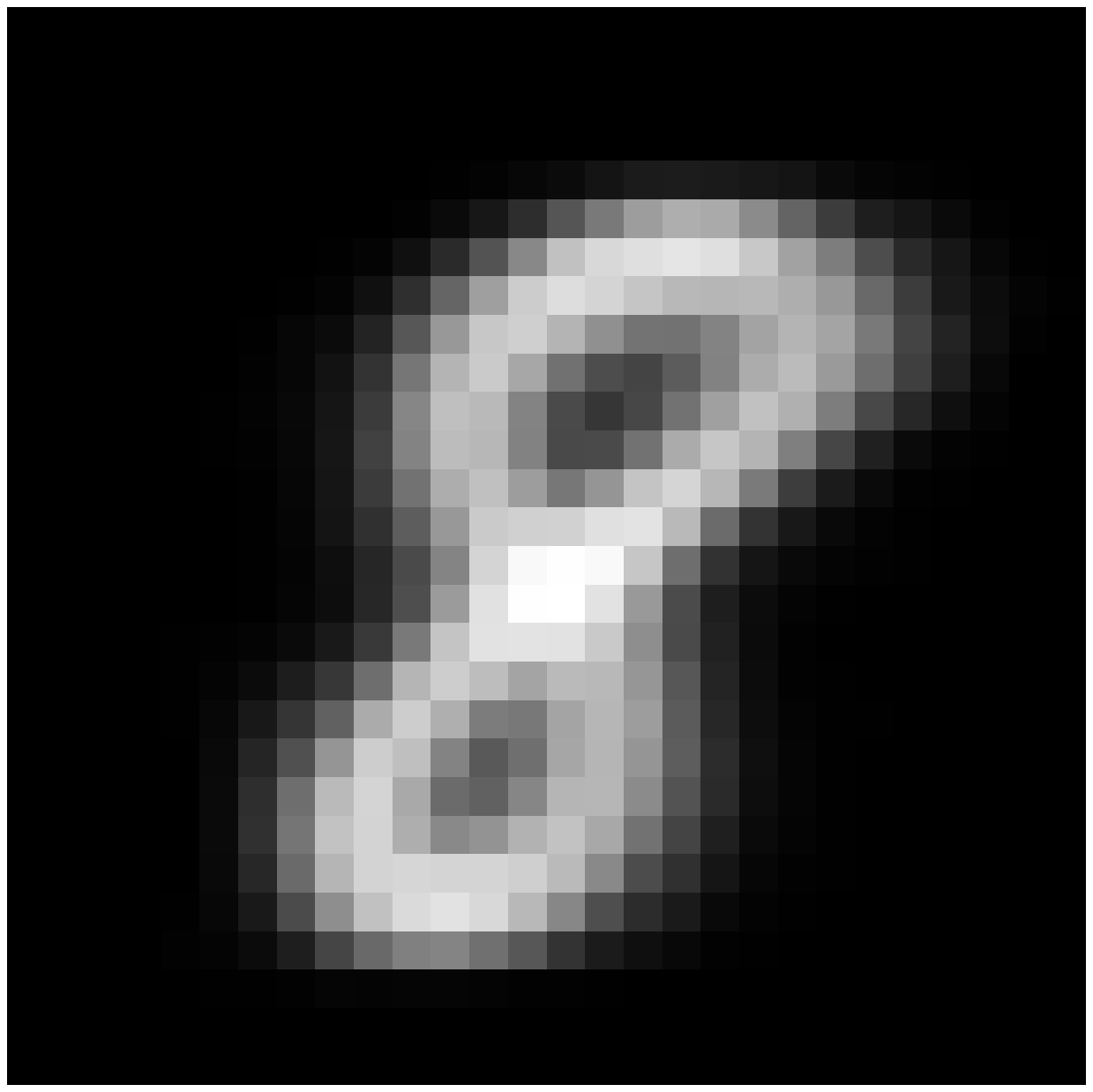} &
    \includegraphics[width=0.095\linewidth,bb=142 226 494 578,clip]{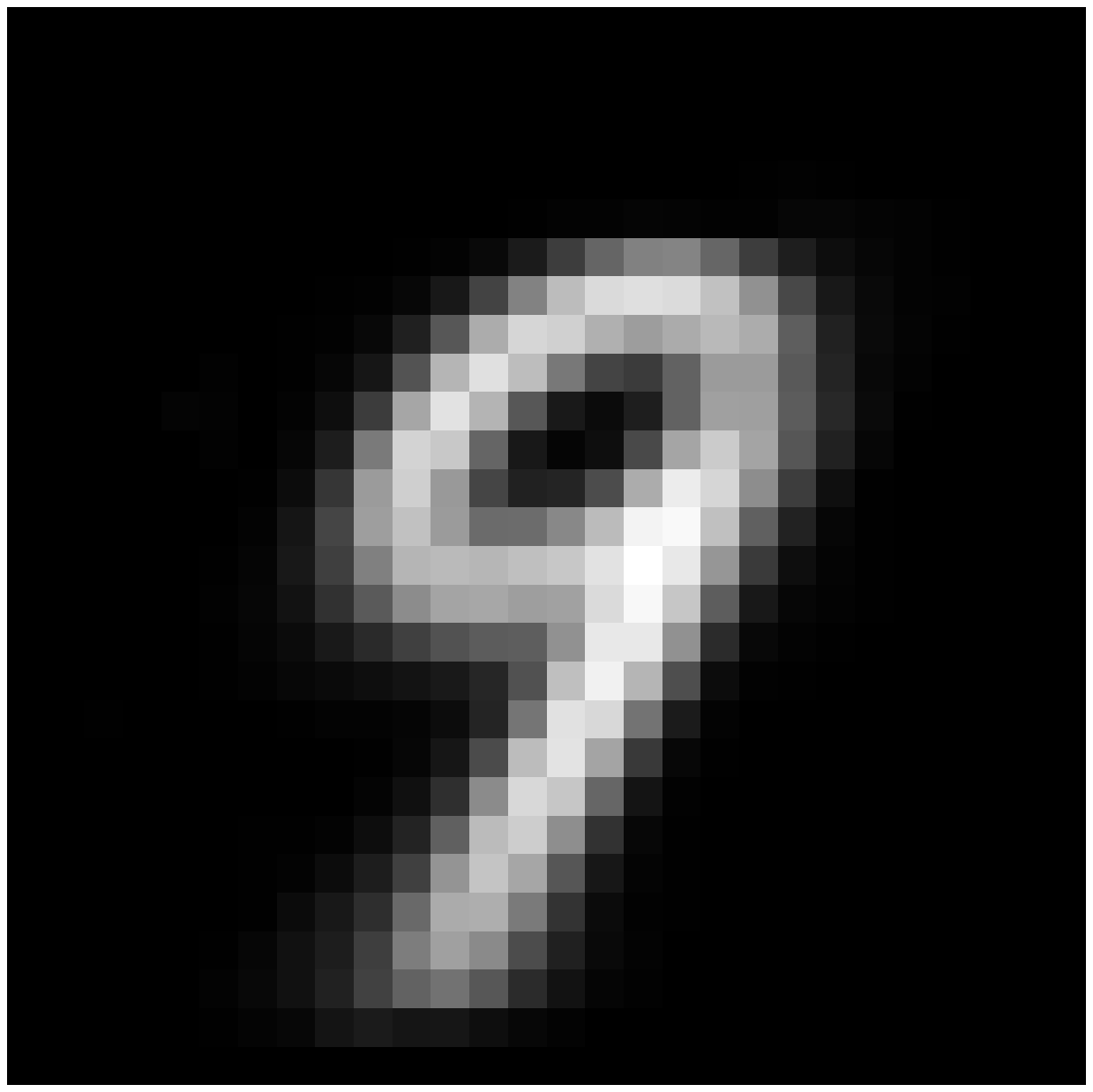} &
    \includegraphics[width=0.095\linewidth,bb=142 226 494 578,clip]{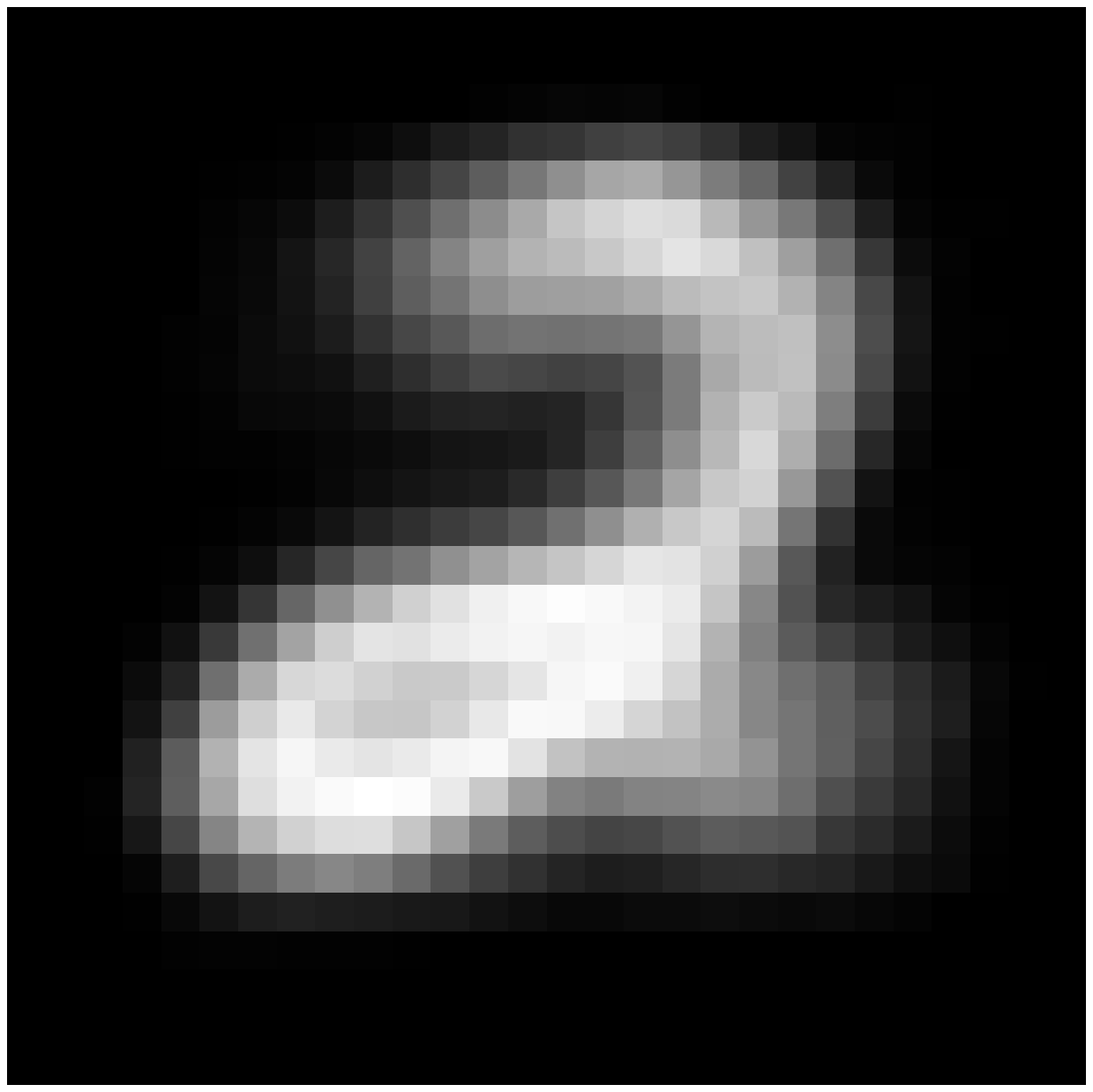} &
    \includegraphics[width=0.095\linewidth,bb=142 226 494 578,clip]{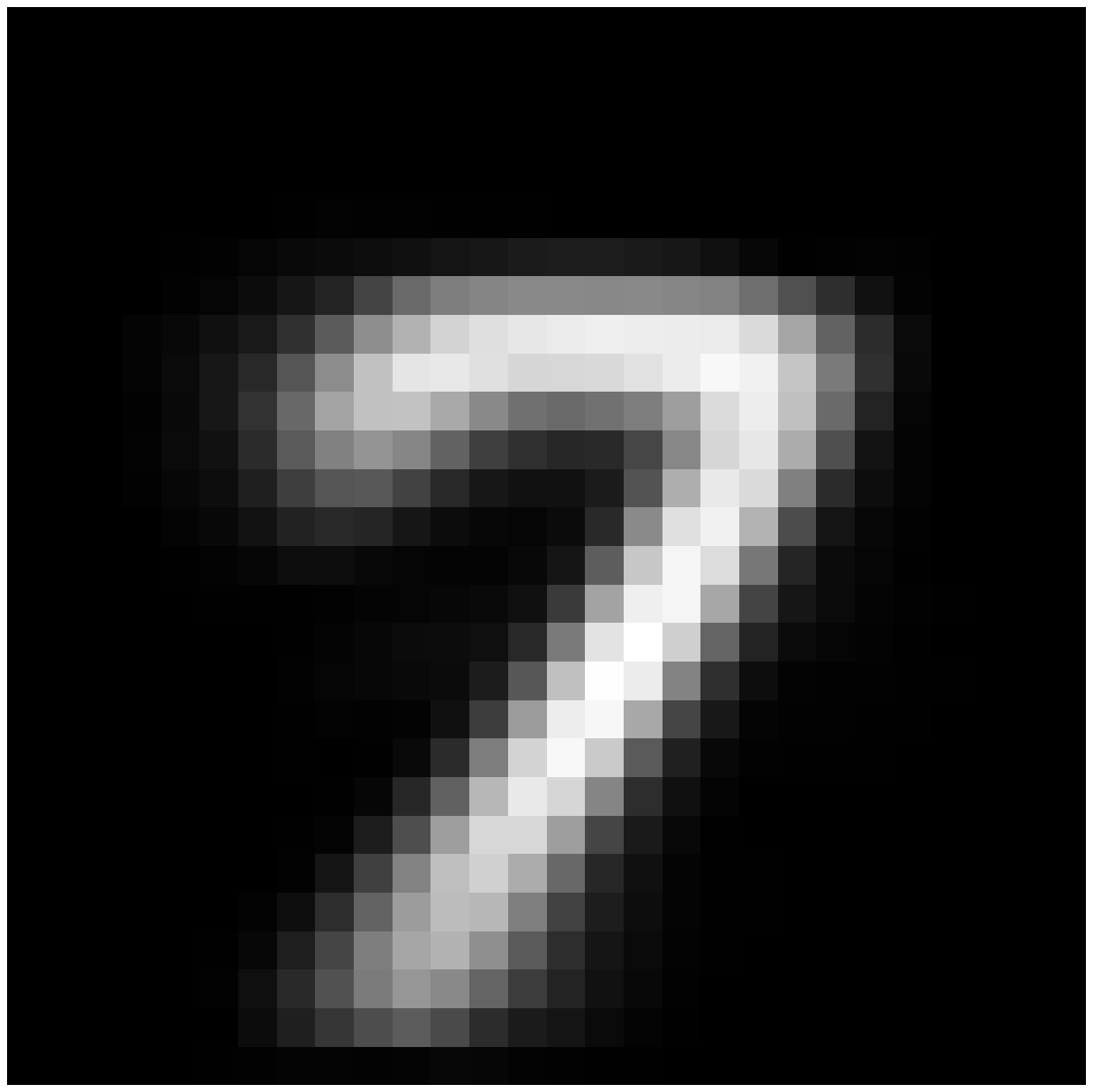} &
    \includegraphics[width=0.095\linewidth,bb=142 226 494 578,clip]{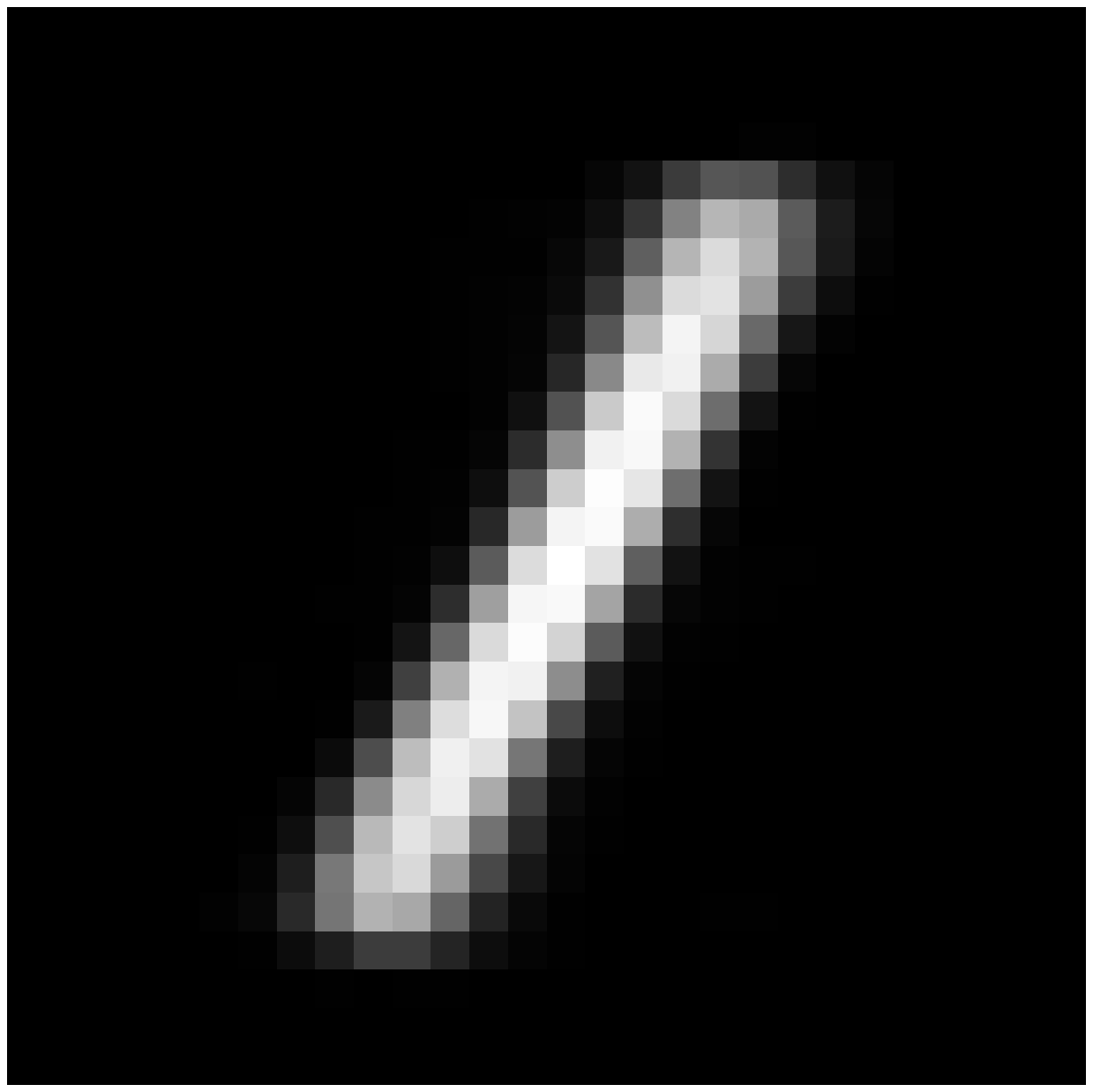} &
    \includegraphics[width=0.095\linewidth,bb=142 226 494 578,clip]{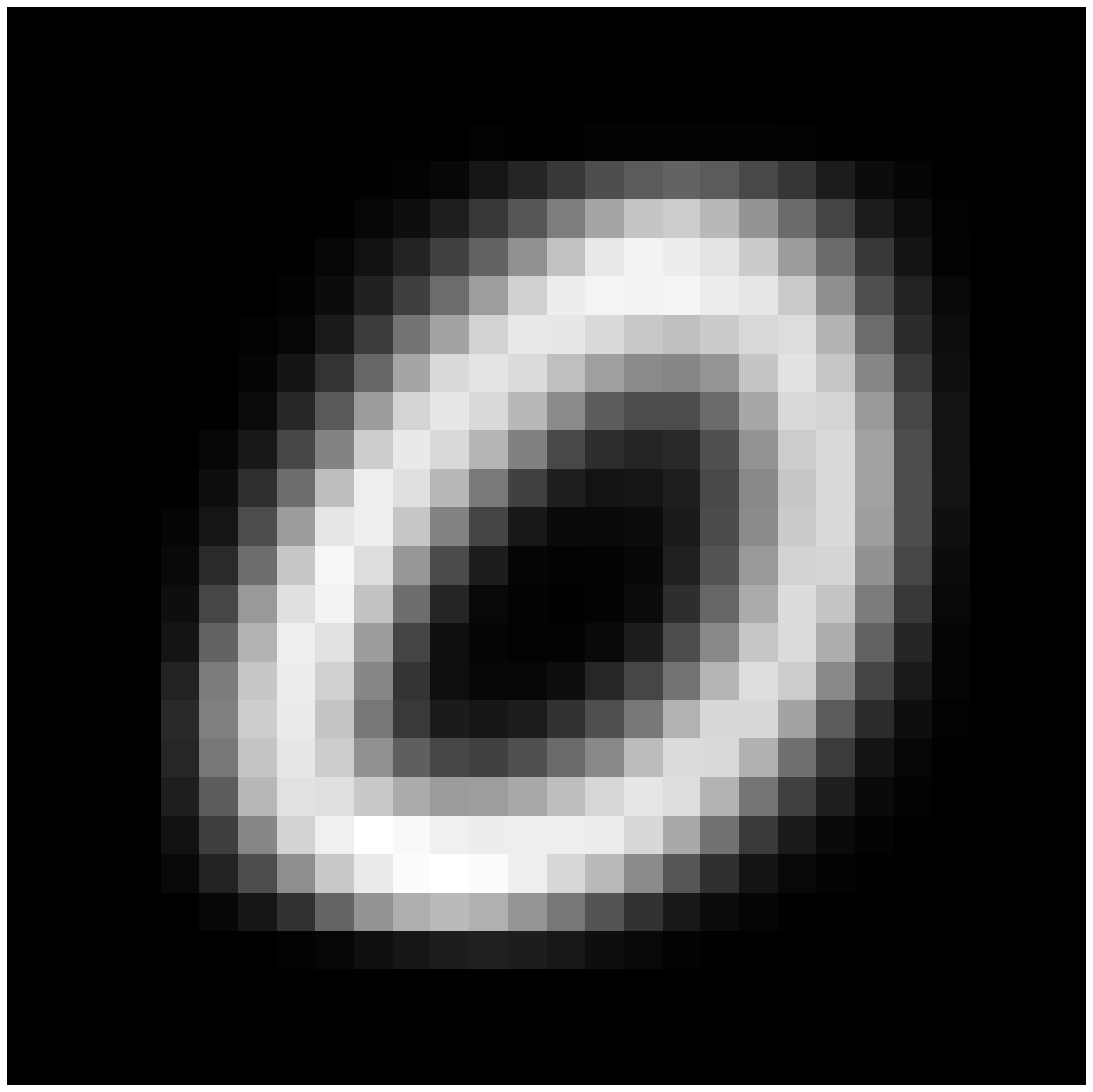} &
    \includegraphics[width=0.095\linewidth,bb=142 226 494 578,clip]{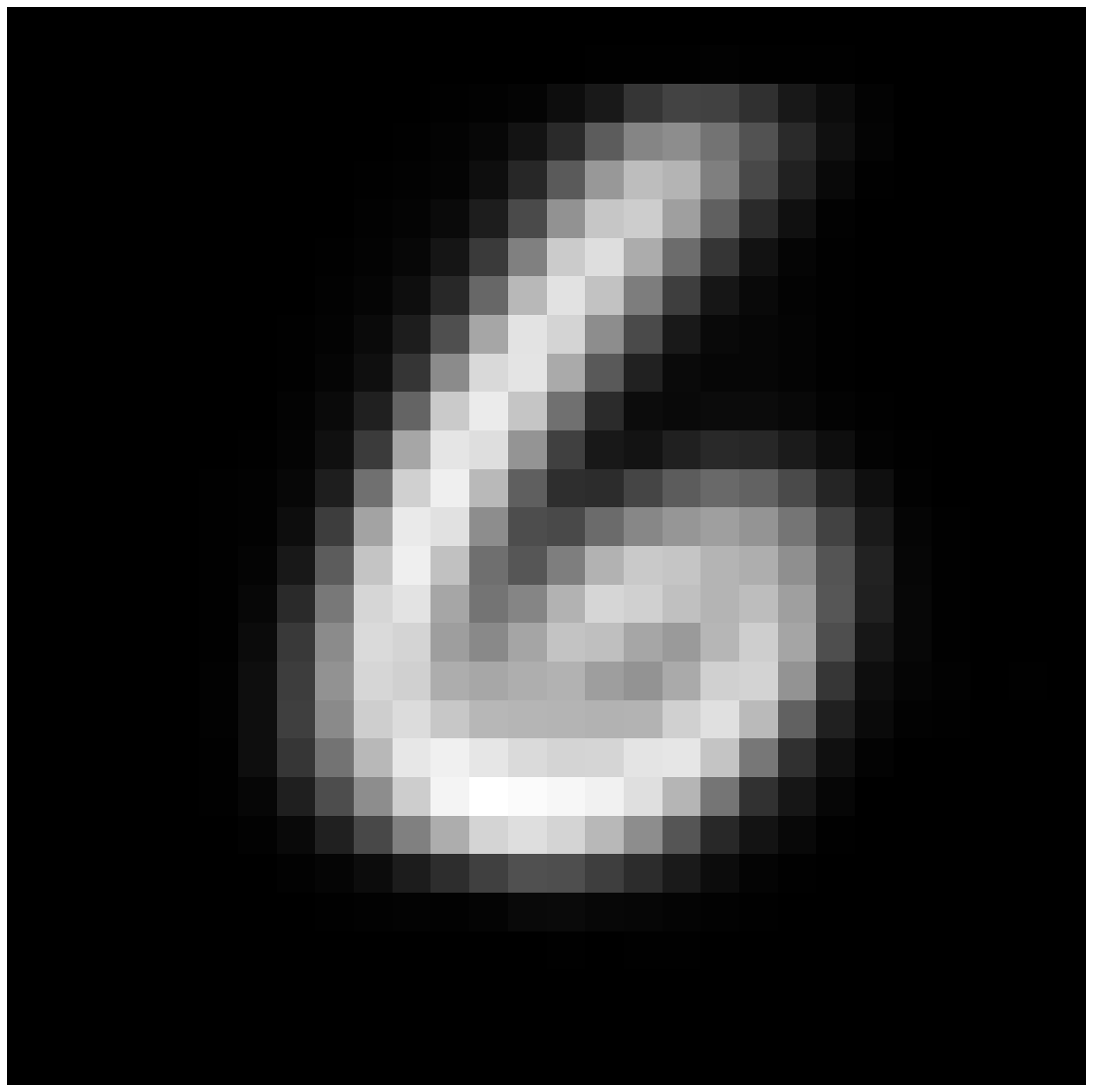} &
    \includegraphics[width=0.095\linewidth,bb=142 226 494 578,clip]{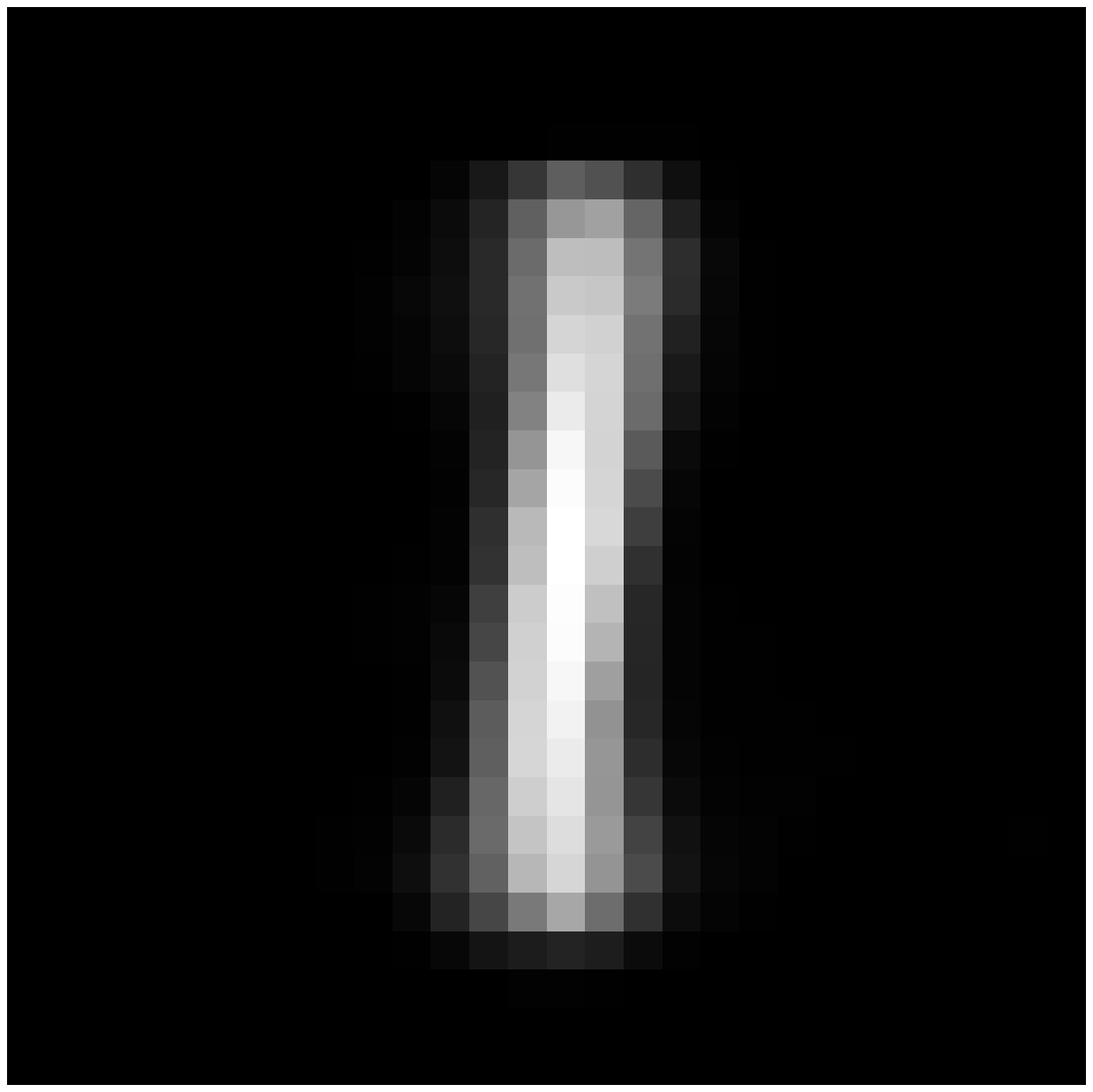} &
    \includegraphics[width=0.095\linewidth,bb=142 226 494 578,clip]{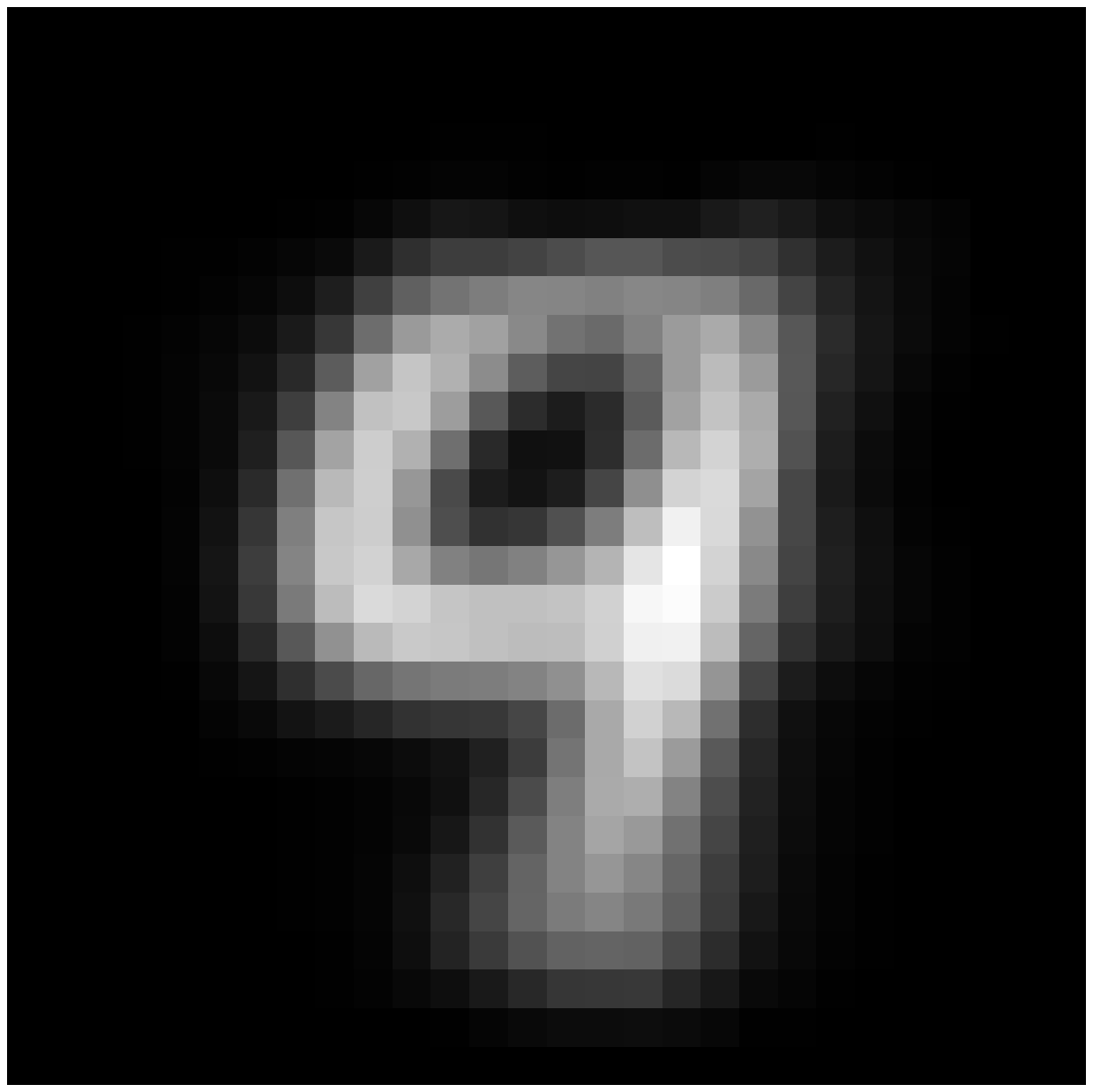} &
    \includegraphics[width=0.095\linewidth,bb=142 226 494 578,clip]{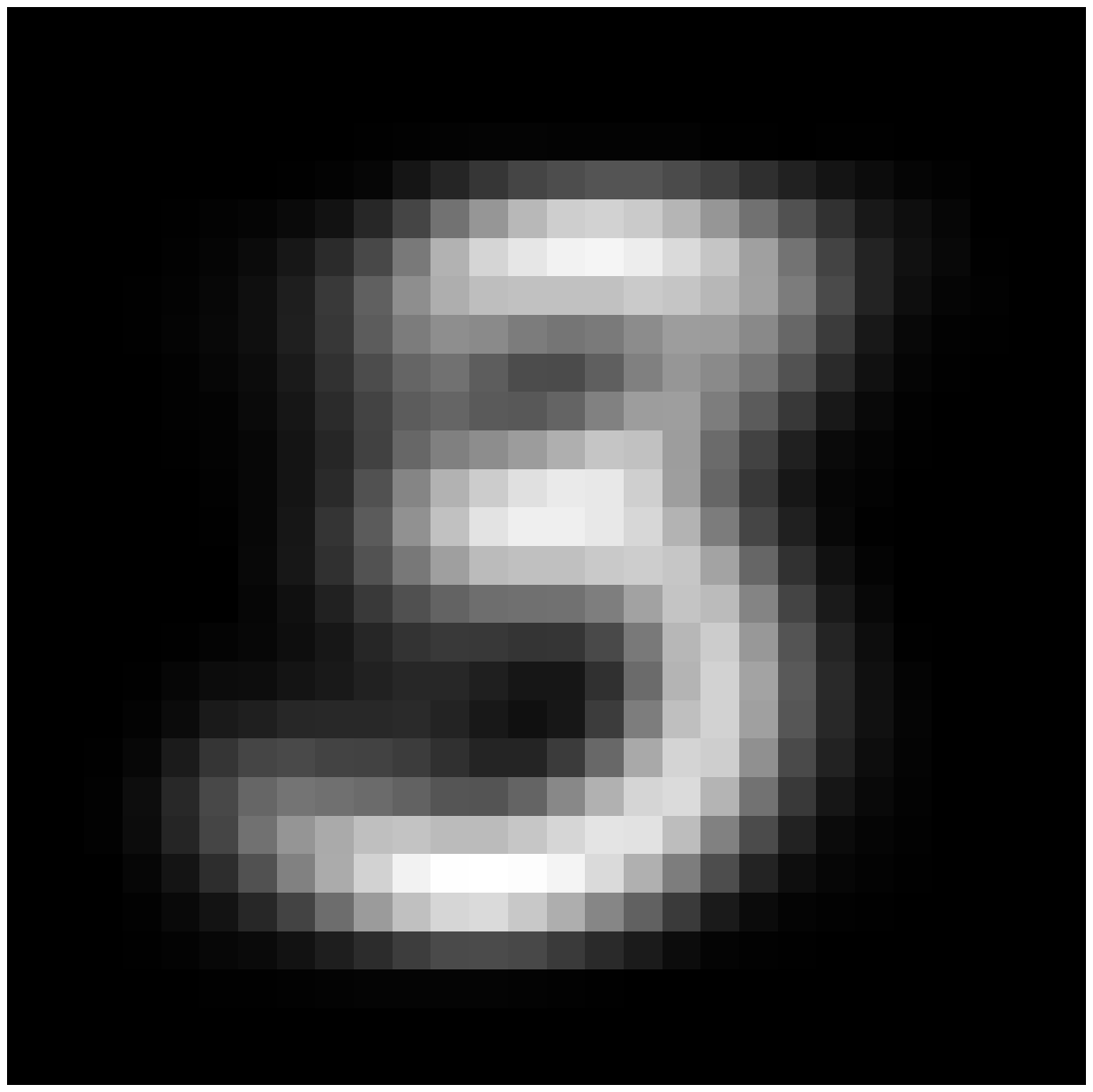} \\[-1ex]
    \rotatebox{90}{\small\hspace{2ex}\caja{c}{c}{Mean \\ shift}} &
    \includegraphics[width=0.095\linewidth,bb=142 226 494 578,clip]{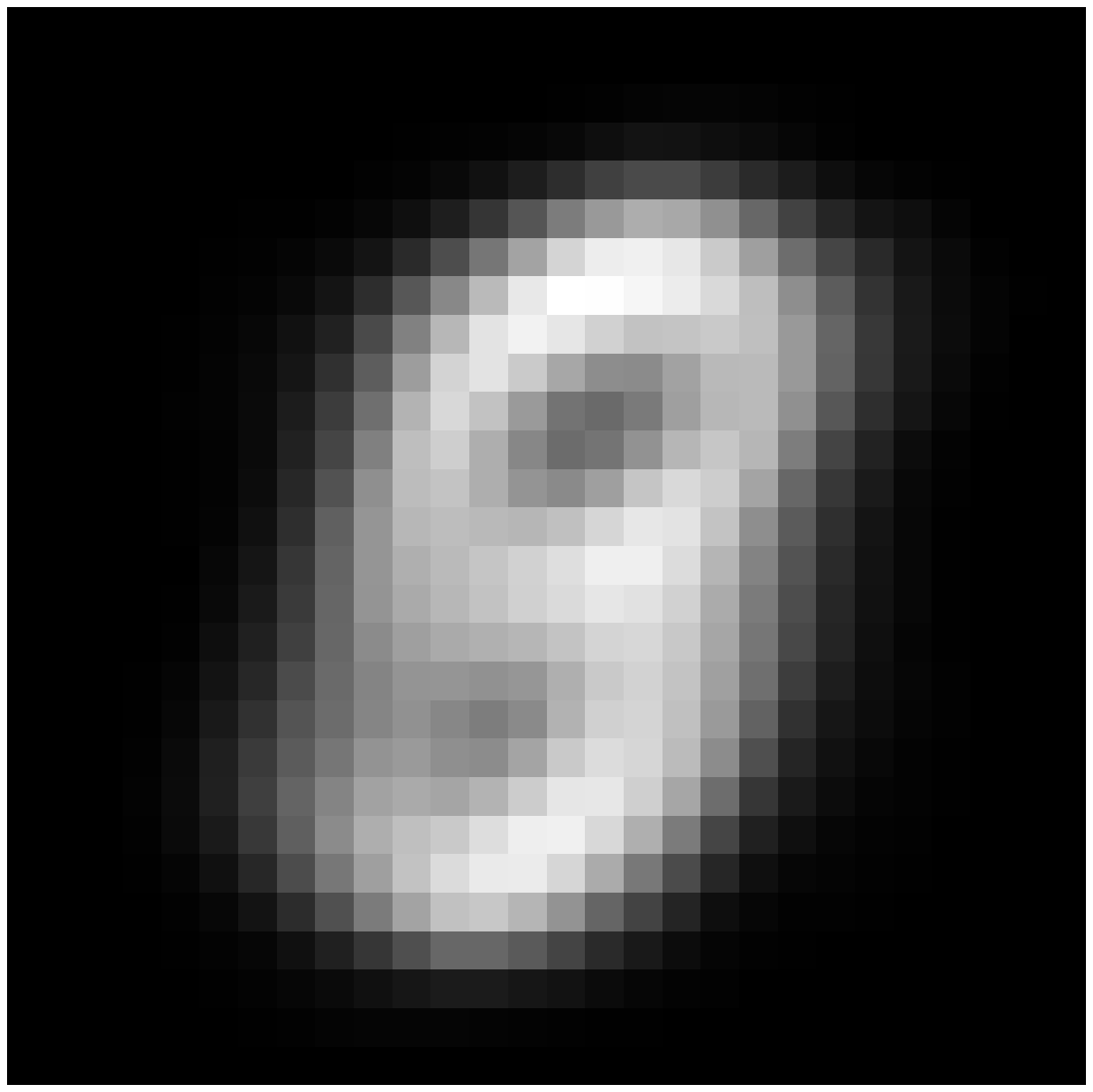} &
    \includegraphics[width=0.095\linewidth,bb=142 226 494 578,clip]{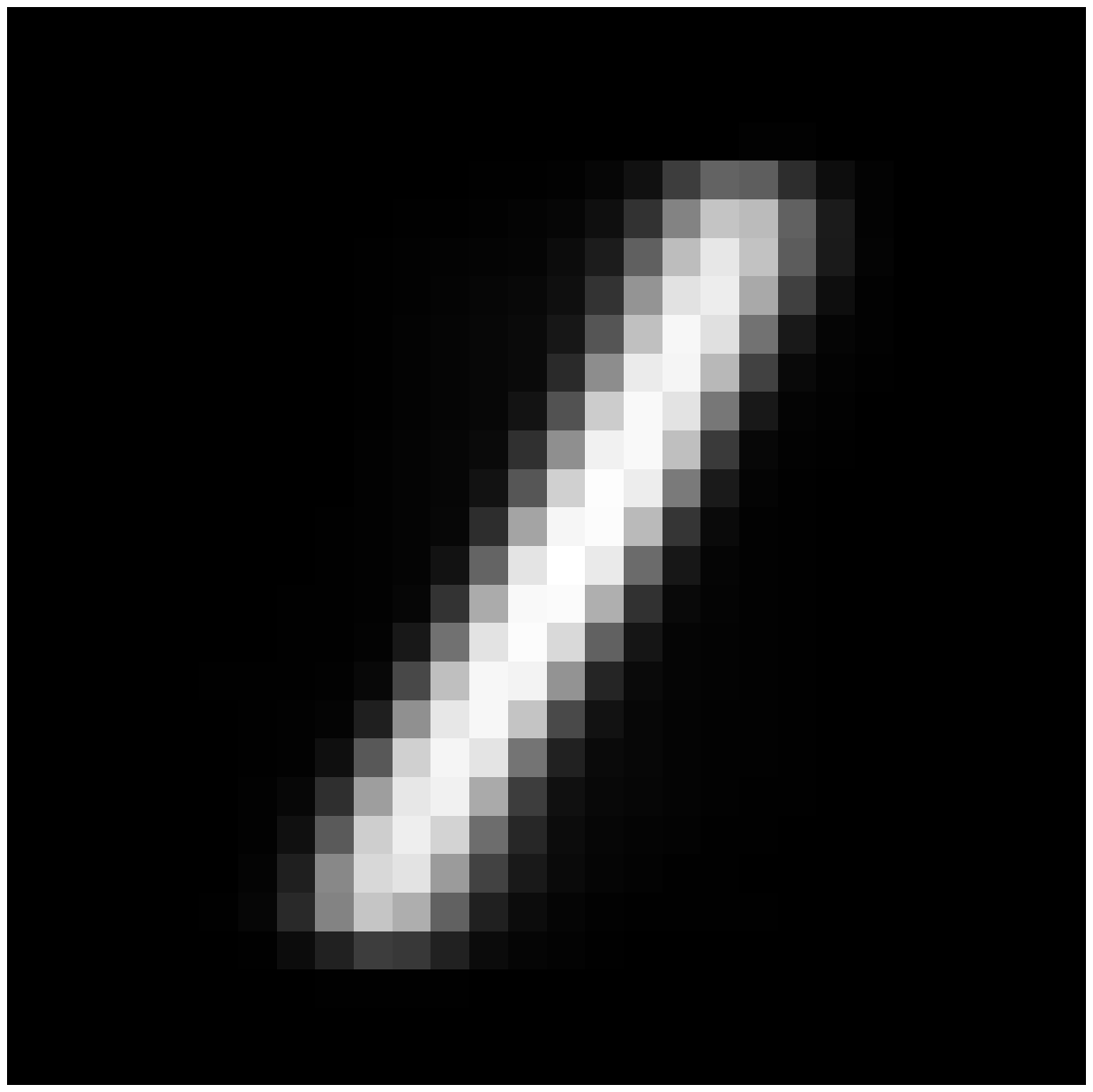} &
    \includegraphics[width=0.095\linewidth,bb=142 226 494 578,clip]{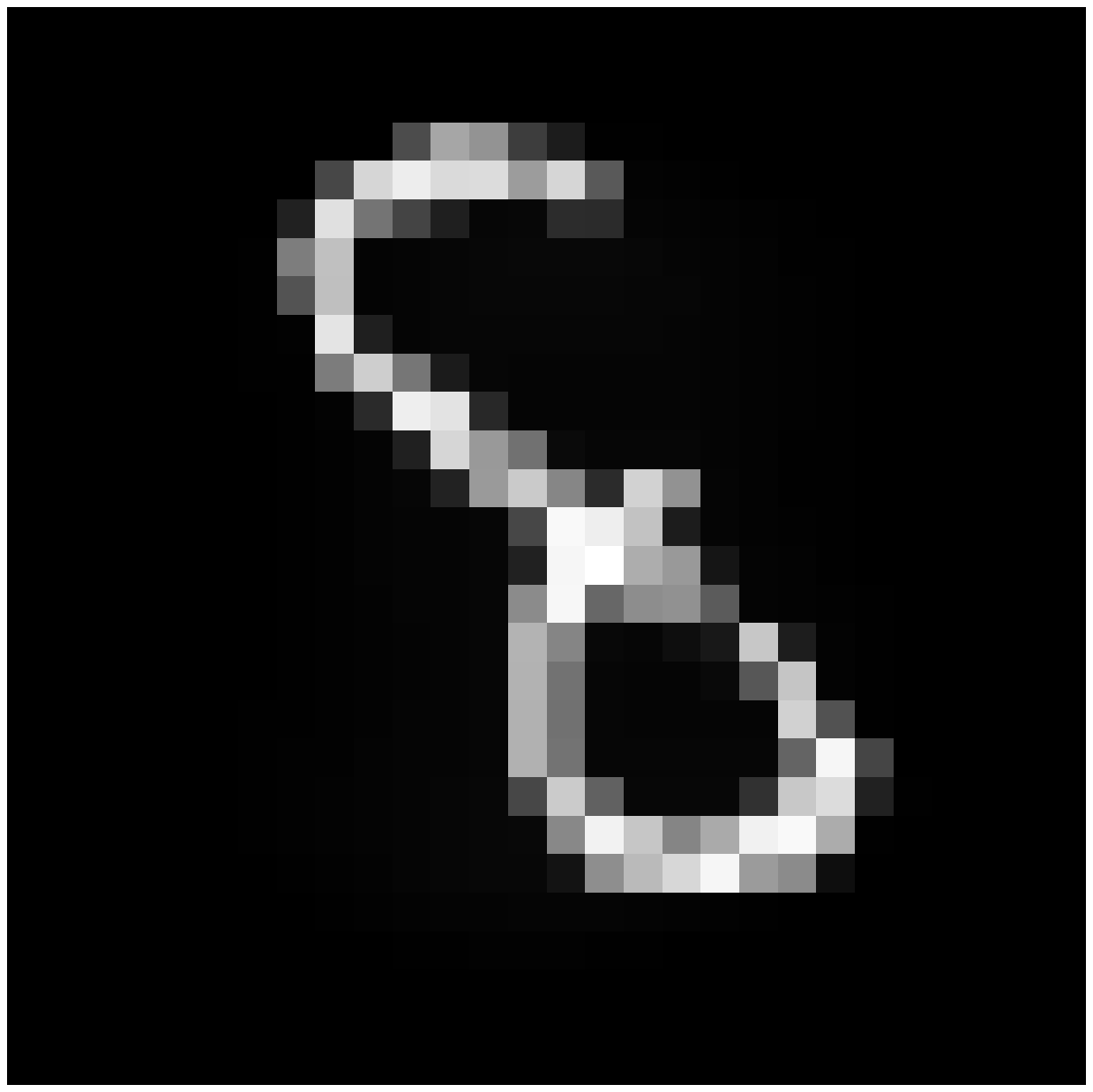} &
    \includegraphics[width=0.095\linewidth,bb=142 226 494 578,clip]{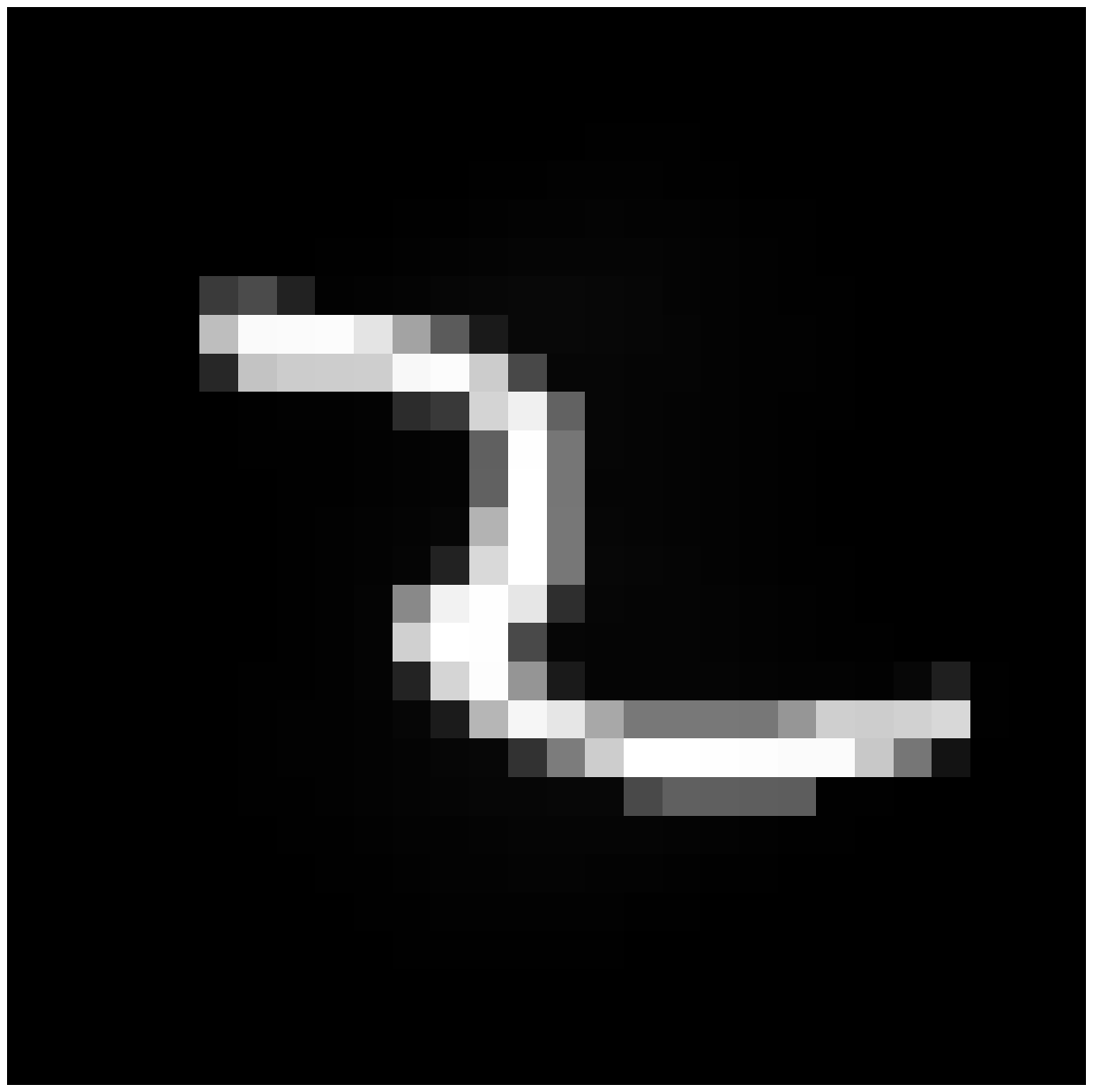} &
    \includegraphics[width=0.095\linewidth,bb=142 226 494 578,clip]{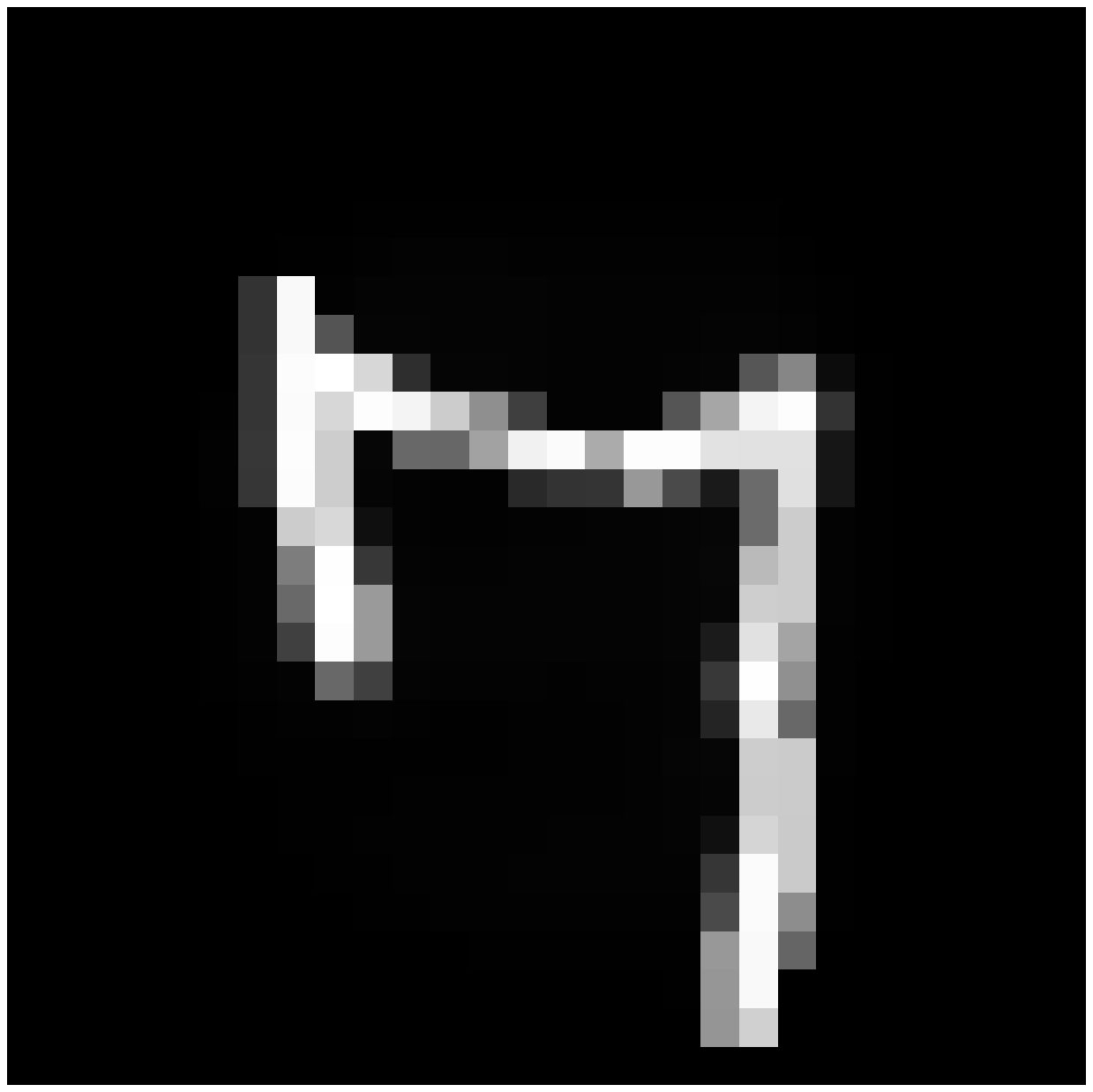} &
    \includegraphics[width=0.095\linewidth,bb=142 226 494 578,clip]{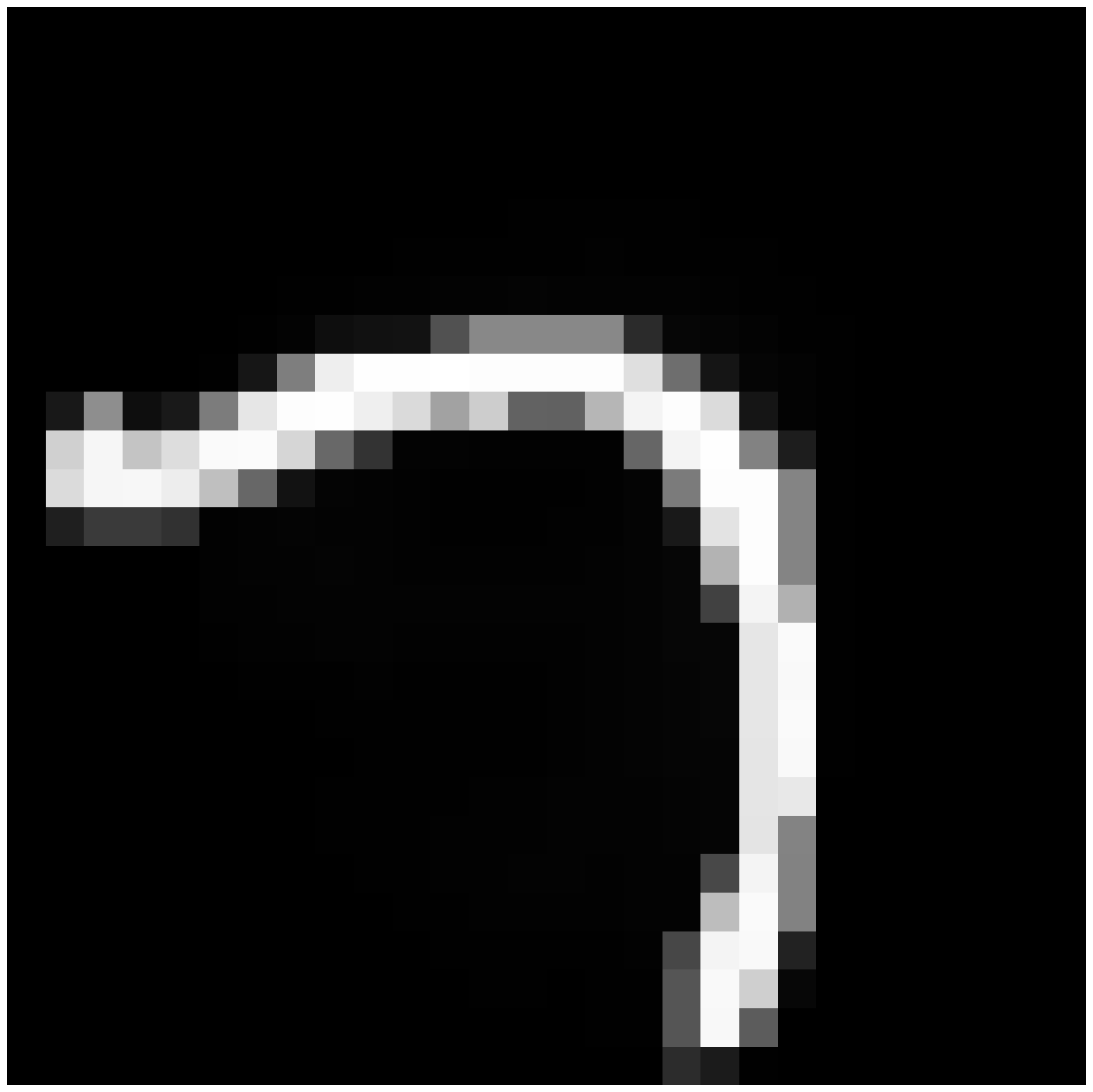} &
    \includegraphics[width=0.095\linewidth,bb=142 226 494 578,clip]{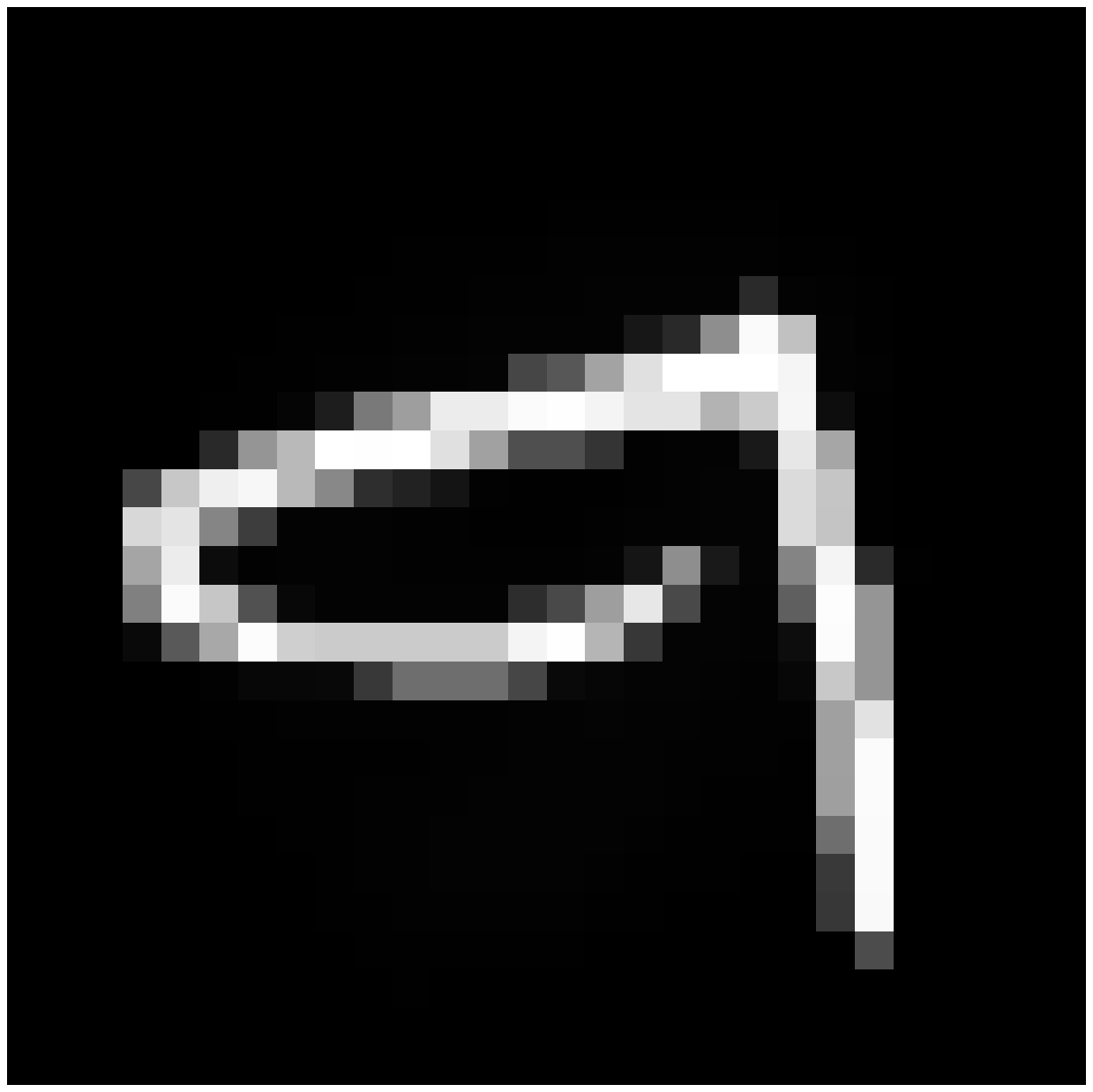} &
    \includegraphics[width=0.095\linewidth,bb=142 226 494 578,clip]{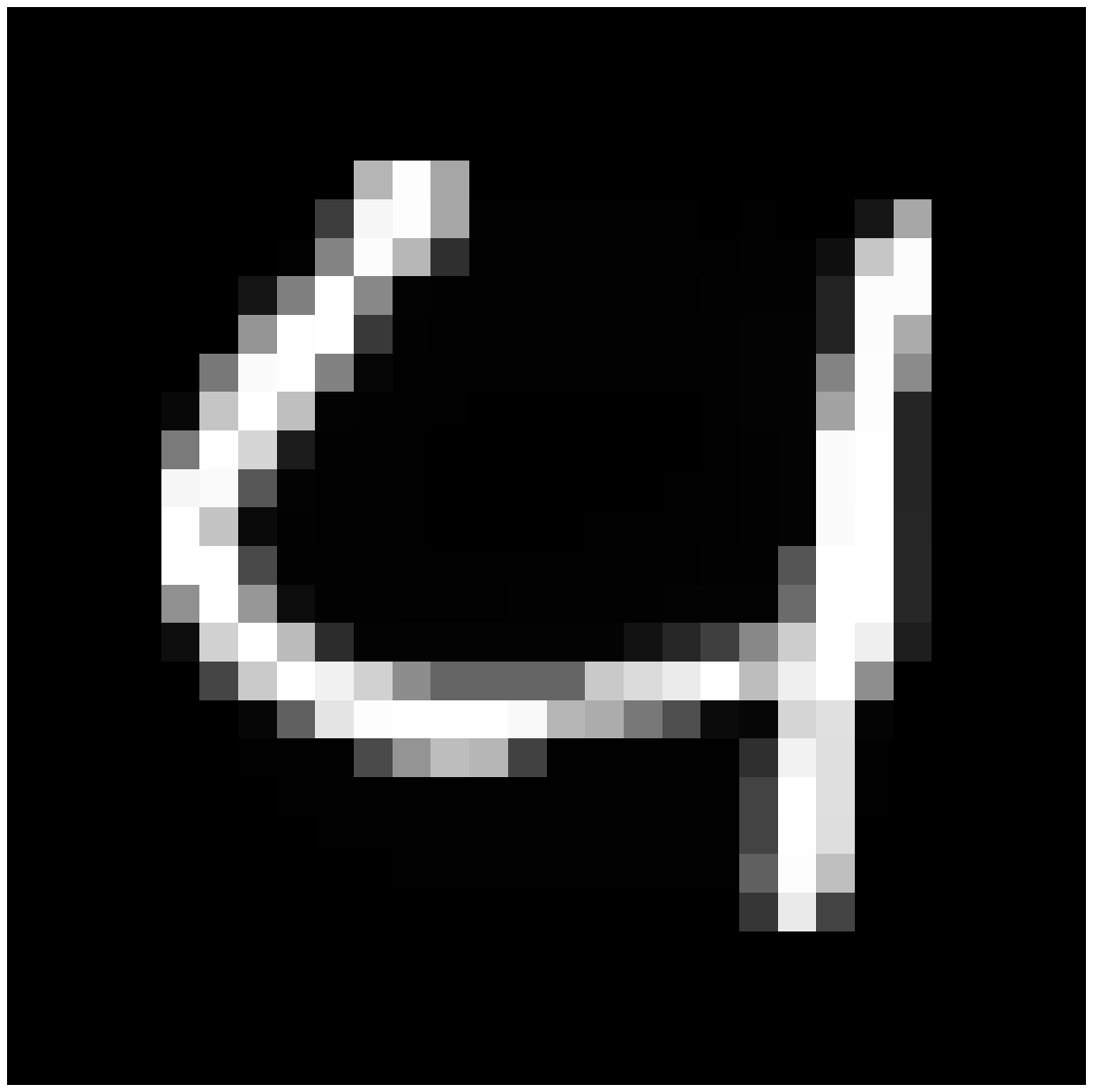} &
    \includegraphics[width=0.095\linewidth,bb=142 226 494 578,clip]{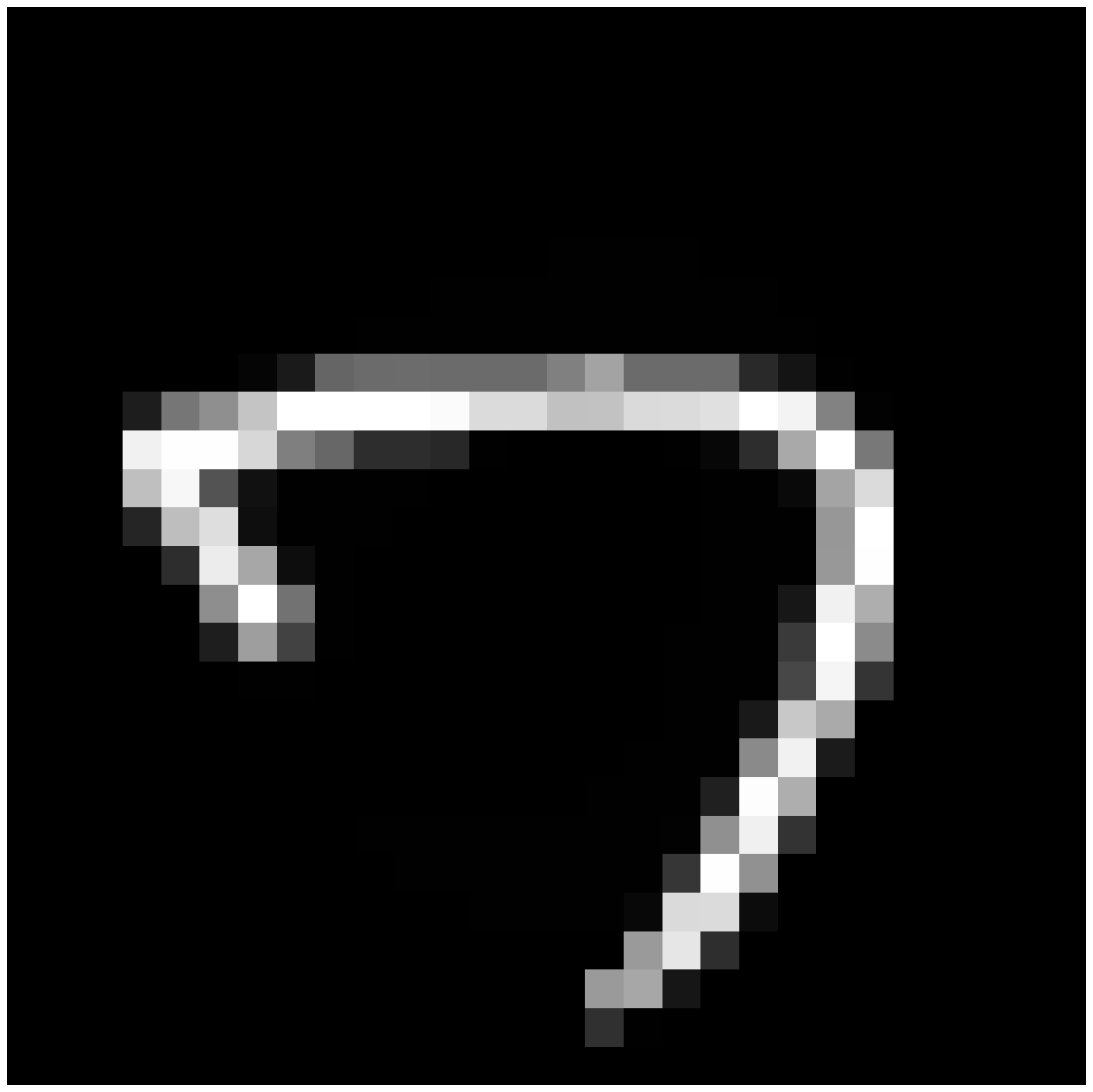} &
    \includegraphics[width=0.095\linewidth,bb=142 226 494 578,clip]{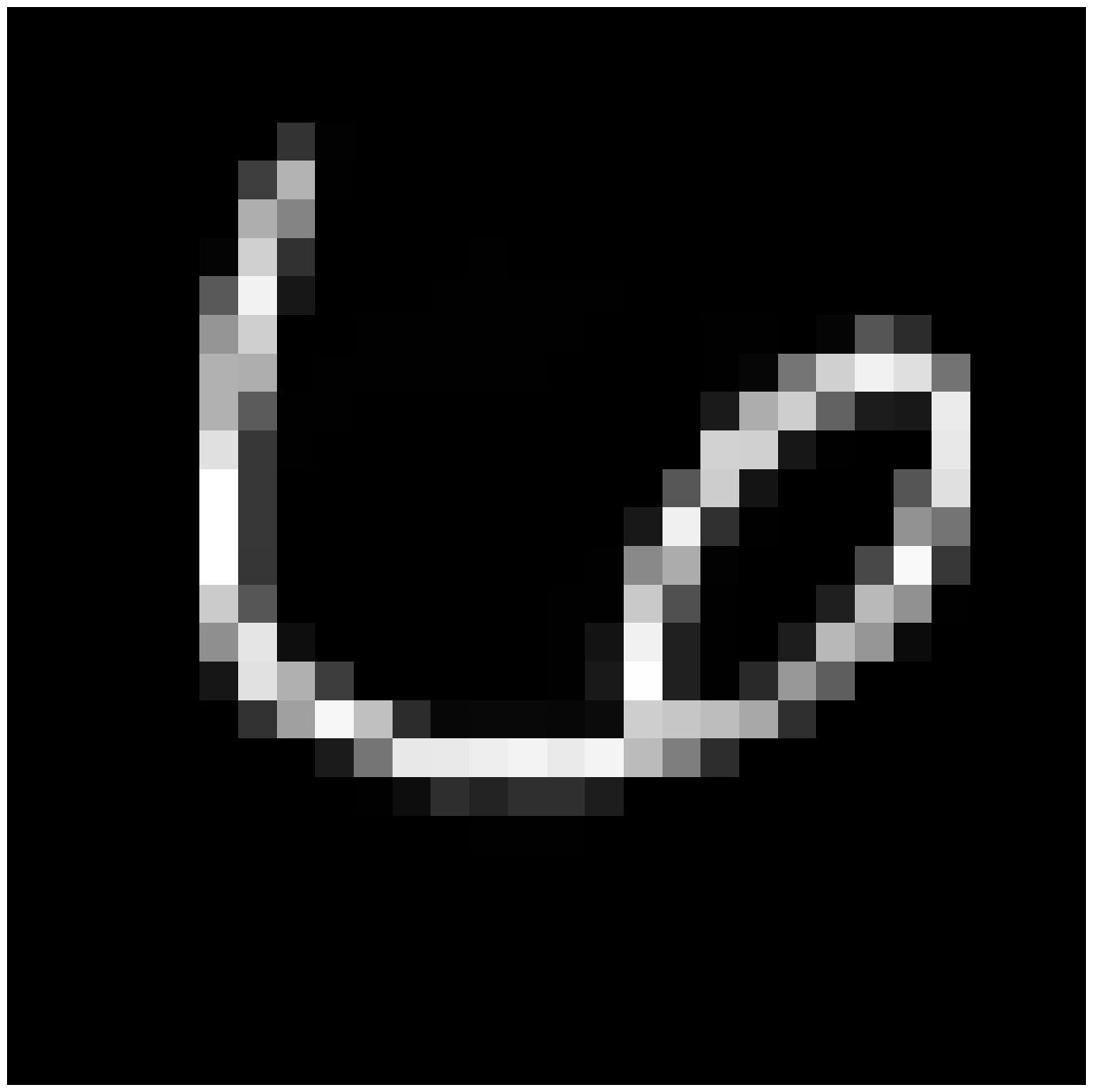}
  \end{tabular}
  \caption{Centroids found by different algorithms on MNIST. \emph{First row}: $K$-means ($\lambda=0$, $\sigma\rightarrow\infty$, ACC: 55.2\%, NMI: 50.2\%). \emph{Second row}: $K$-modes ($\lambda=0$, $\sigma=0.35$, ACC: 56.0\%, NMI: 50.6\%). \emph{Third row}: Laplacian $K$-modes ($\lambda=0.07$, $\sigma=0.35$, ACC: 70.5\%, NMI: 68.8\%). \emph{Fourth row}: mean-shift ($\sigma=0.2485$).}
  \label{f:mnist}
\end{figure}

\begin{figure}[t]
  \centering
  \begin{tabular}[b]{@{}c@{\hspace{0\linewidth}}c@{\hspace{0\linewidth}}c@{\hspace{0\linewidth}}c@{\hspace{0\linewidth}}c@{\hspace{0\linewidth}}c@{\hspace{0\linewidth}}c@{\hspace{0\linewidth}}c@{\hspace{0\linewidth}}c@{\hspace{0\linewidth}}c@{\hspace{0\linewidth}}c@{\hspace{0\linewidth}}c@{\hspace{0\linewidth}}c@{\hspace{0\linewidth}}c@{\hspace{0\linewidth}}c@{\hspace{0\linewidth}}c@{\hspace{0\linewidth}}c@{\hspace{0\linewidth}}c@{\hspace{0\linewidth}}c@{\hspace{0\linewidth}}c@{\hspace{0\linewidth}}c@{}}
    \rotatebox{90}{\tiny\hspace{1ex}\caja{c}{c}{$K$- \\ means}} &
    \includegraphics[width=0.048\linewidth,bb=142 226 494 578,clip]{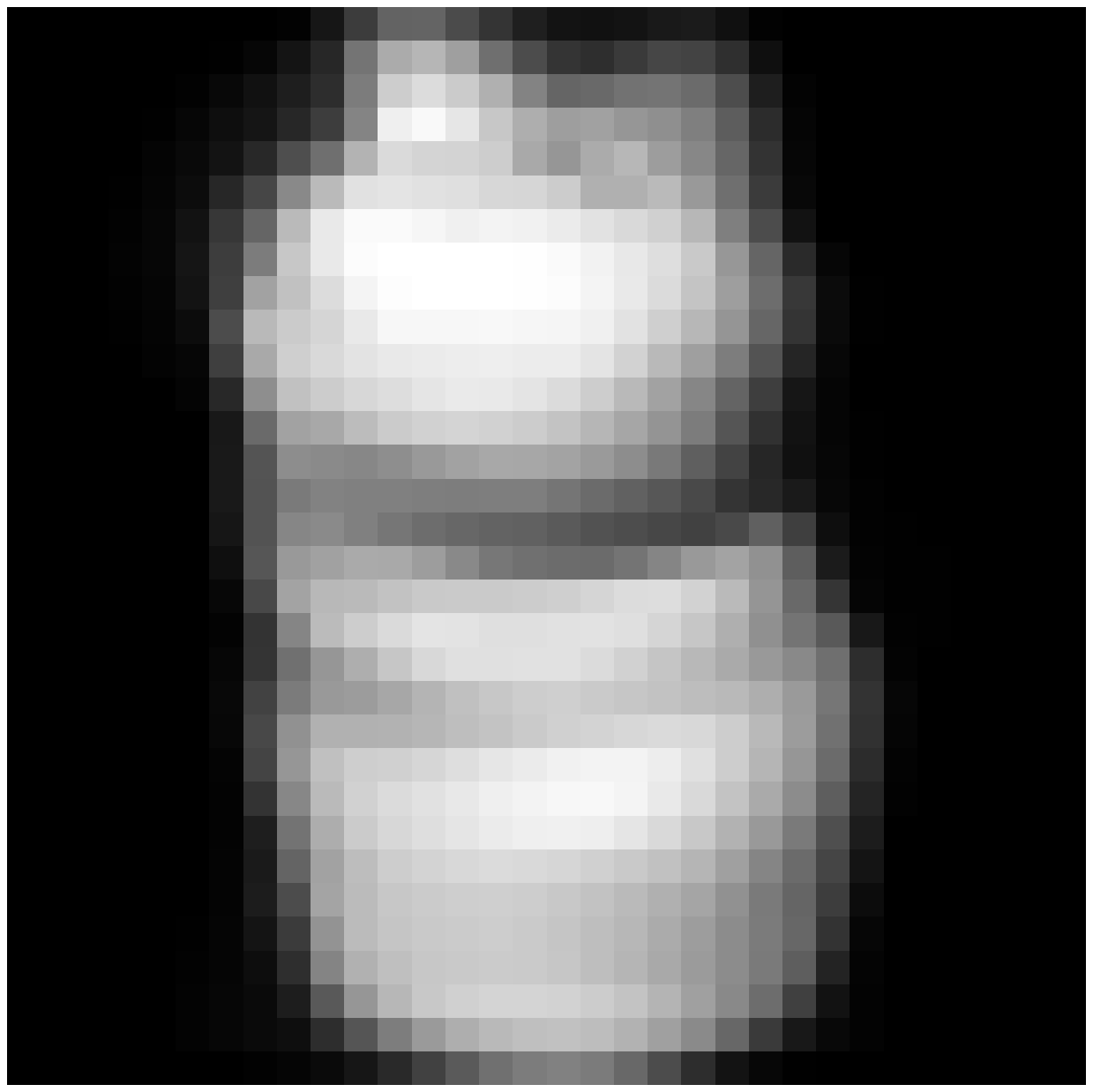} &
    \includegraphics[width=0.048\linewidth,bb=142 226 494 578,clip]{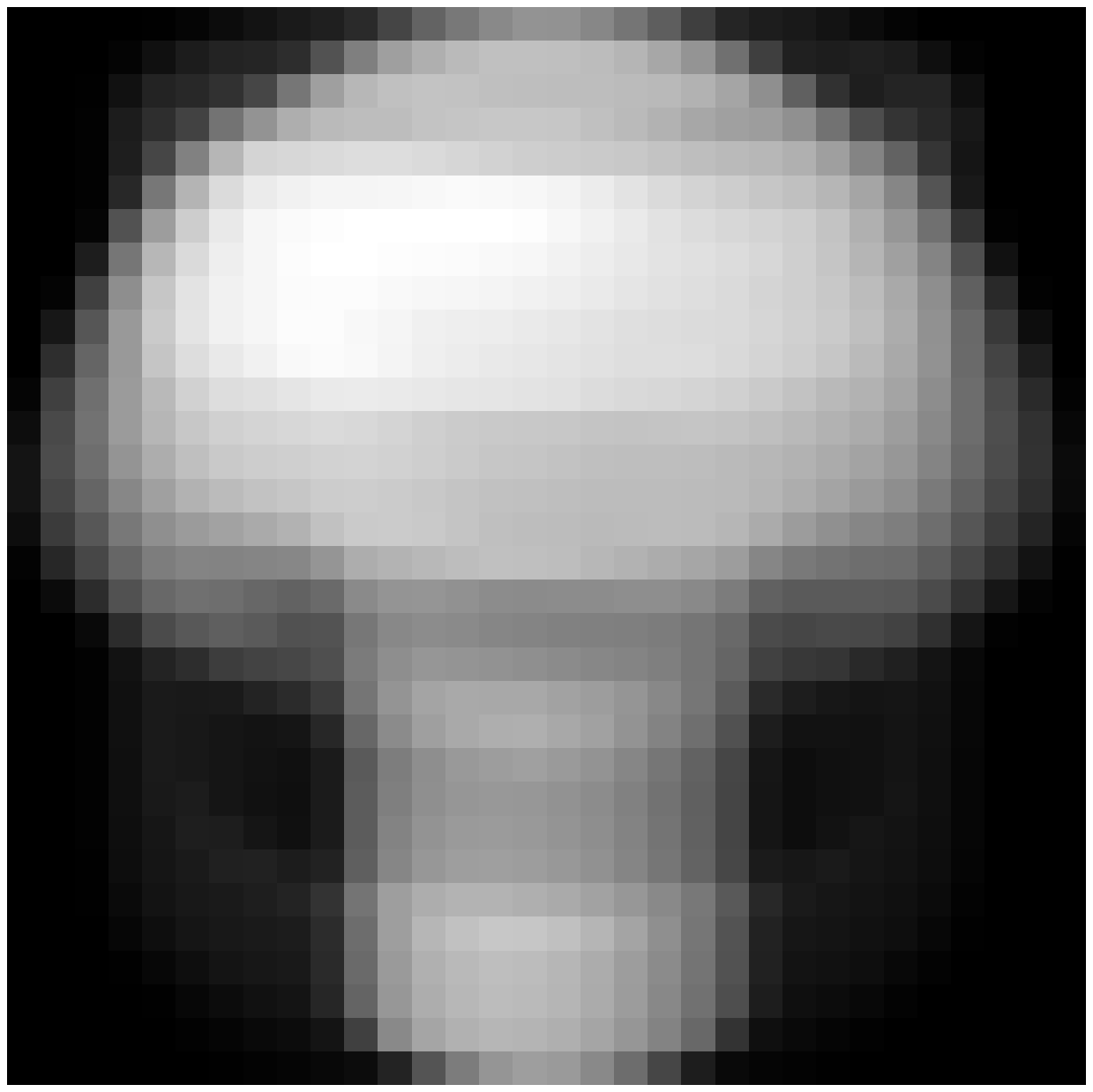} &
    \includegraphics[width=0.048\linewidth,bb=142 226 494 578,clip]{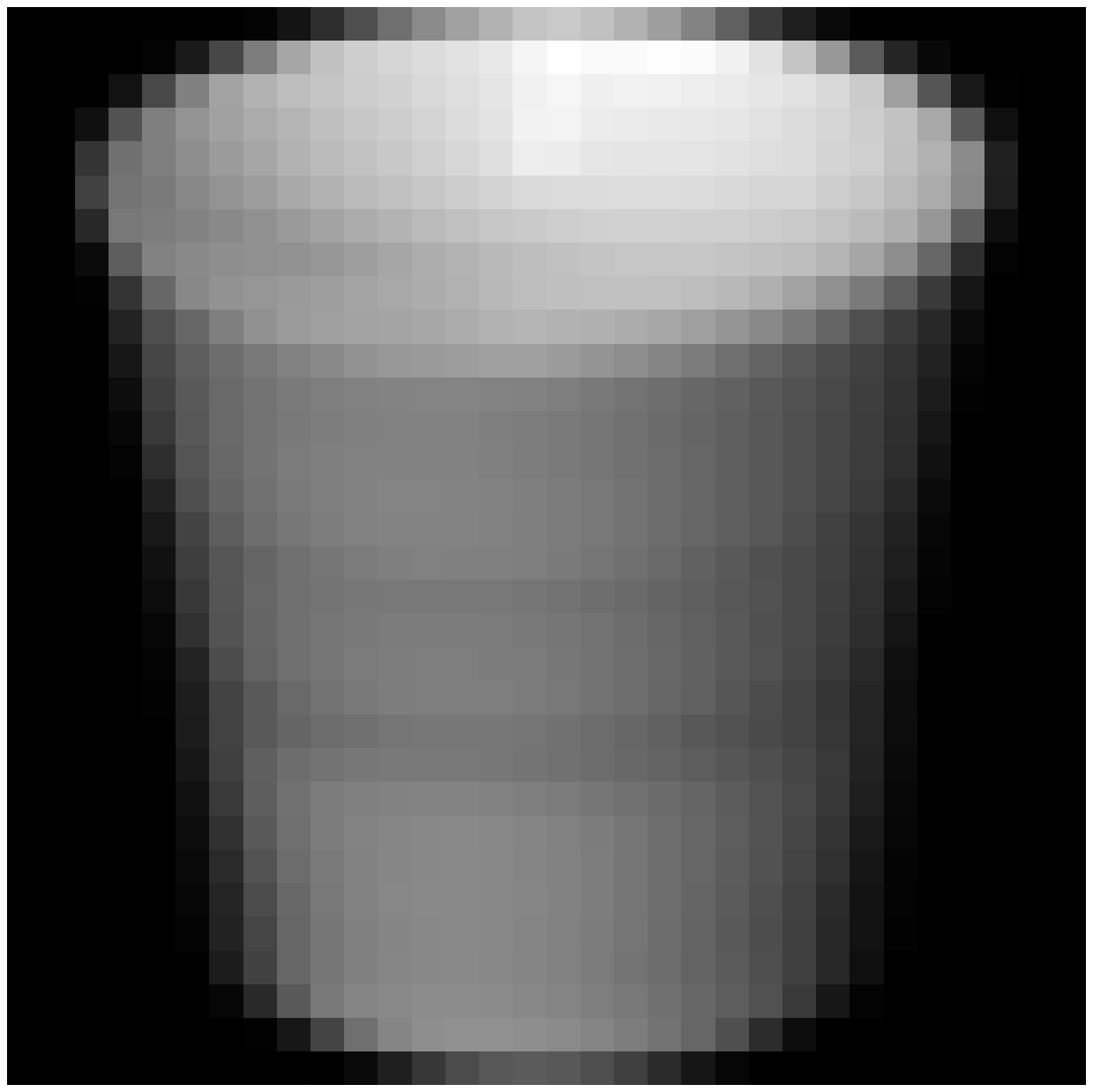} &
    \includegraphics[width=0.048\linewidth,bb=142 226 494 578,clip]{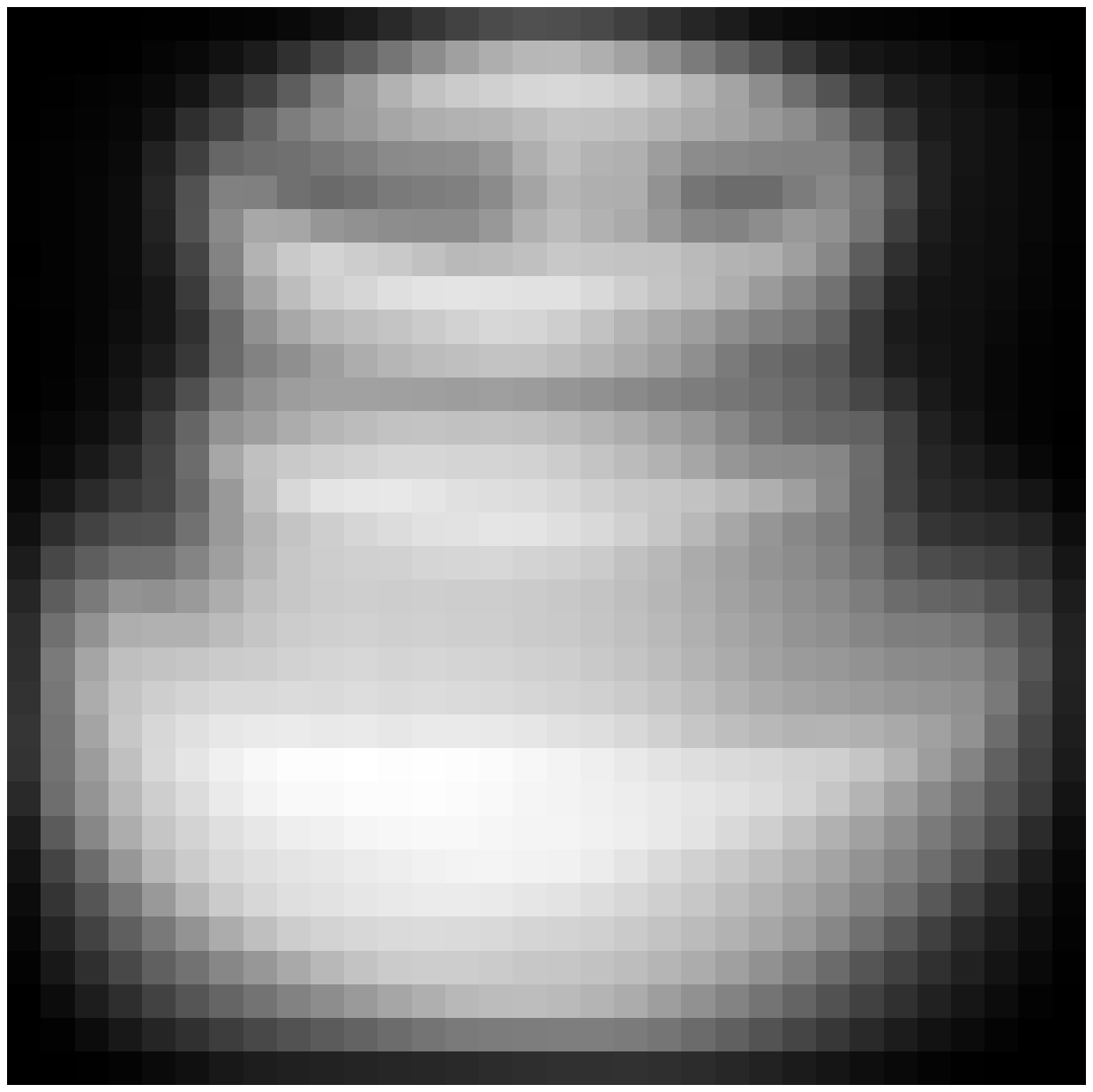} &
    \includegraphics[width=0.048\linewidth,bb=142 226 494 578,clip]{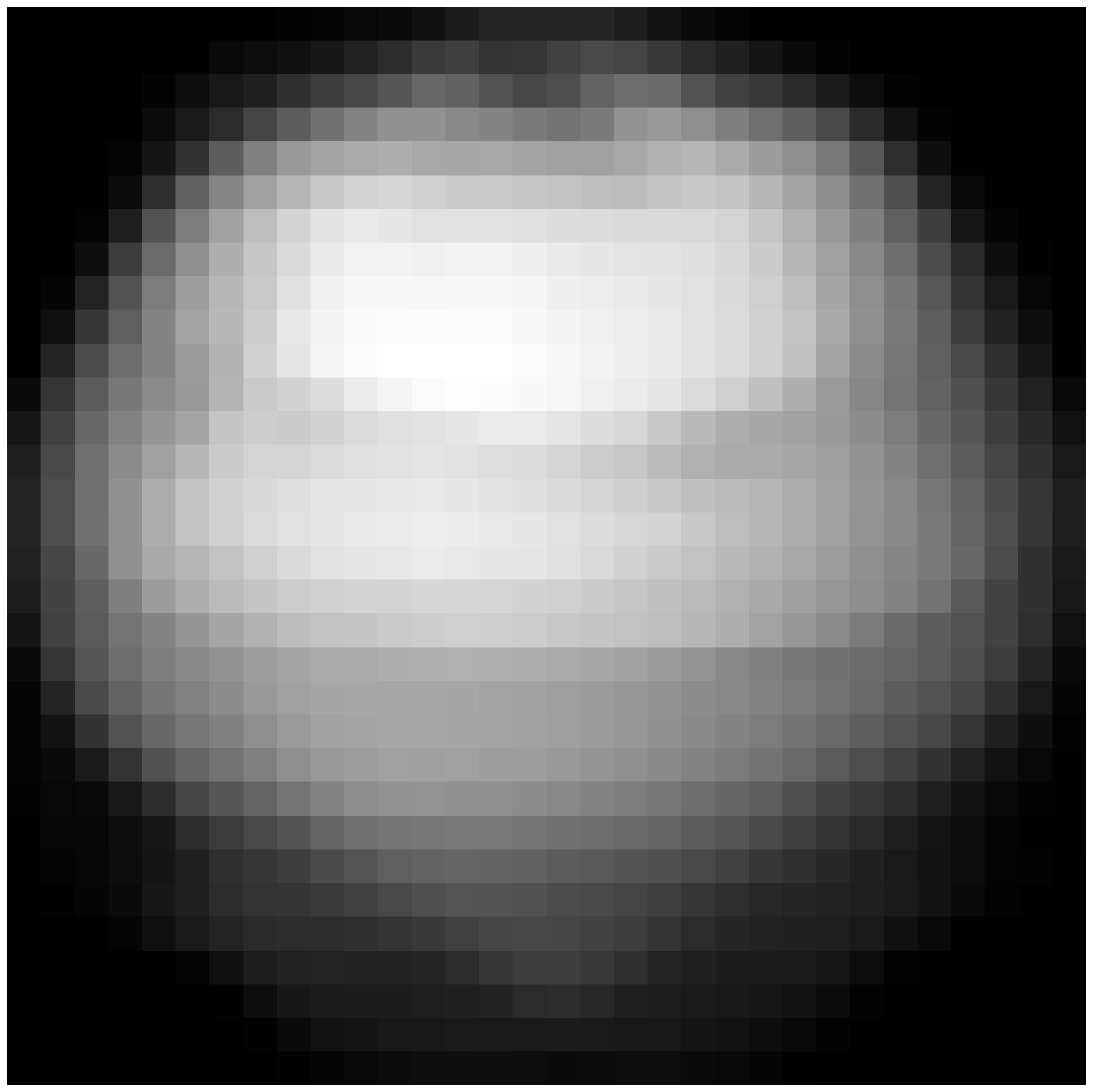} &
    \includegraphics[width=0.048\linewidth,bb=142 226 494 578,clip]{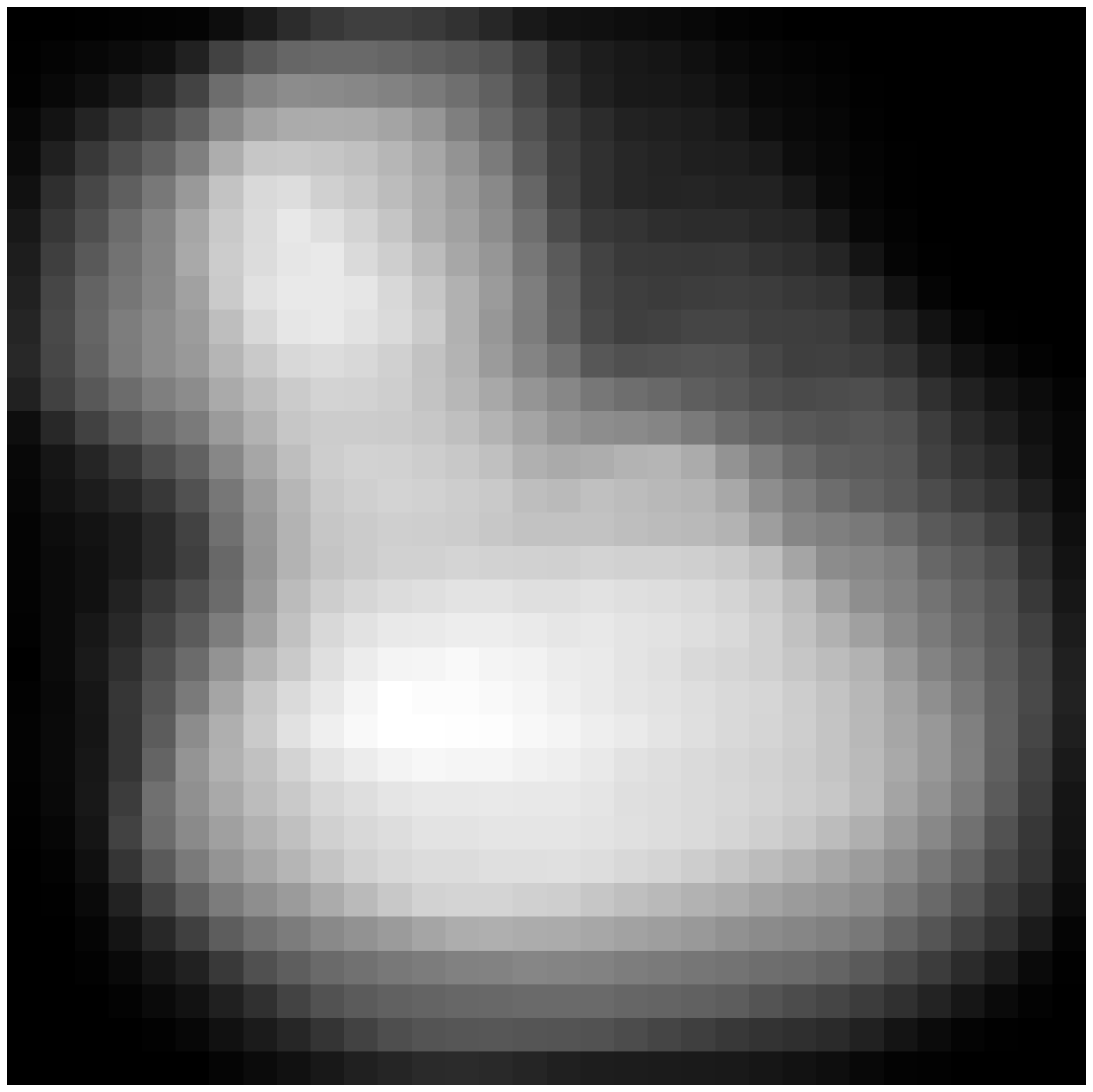} &
    \includegraphics[width=0.048\linewidth,bb=142 226 494 578,clip]{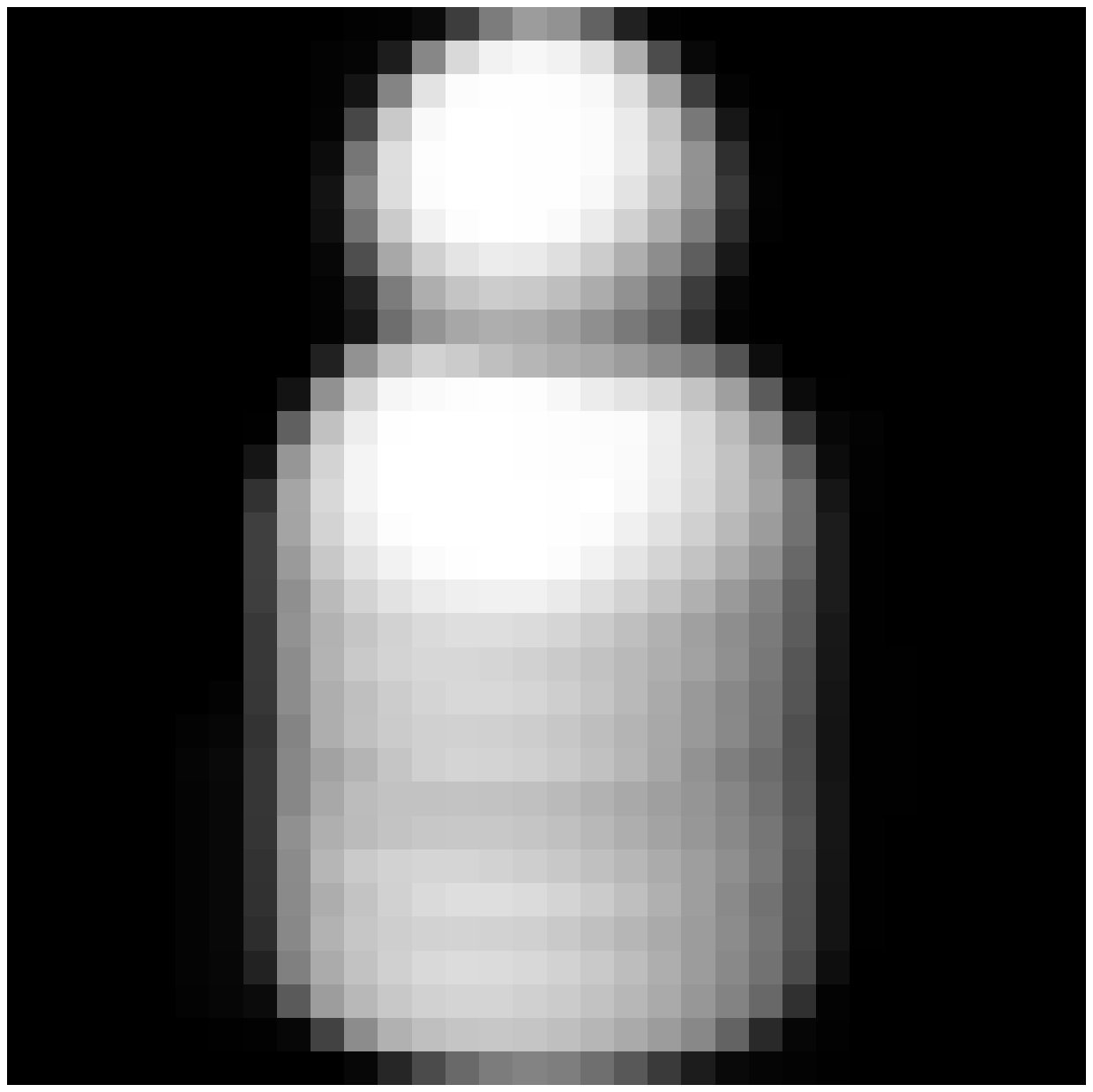} &
    \includegraphics[width=0.048\linewidth,bb=142 226 494 578,clip]{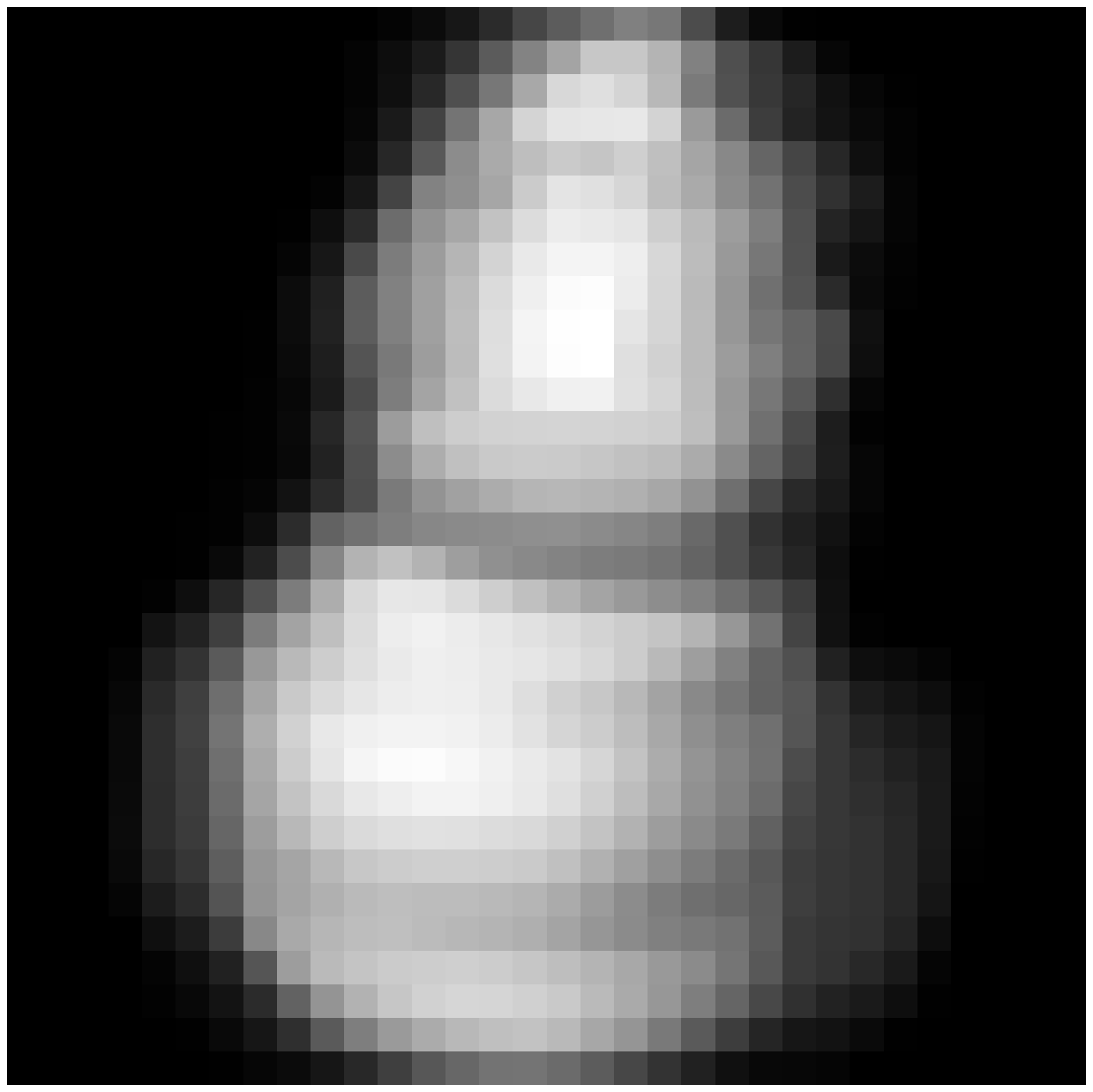} &
    \includegraphics[width=0.048\linewidth,bb=142 226 494 578,clip]{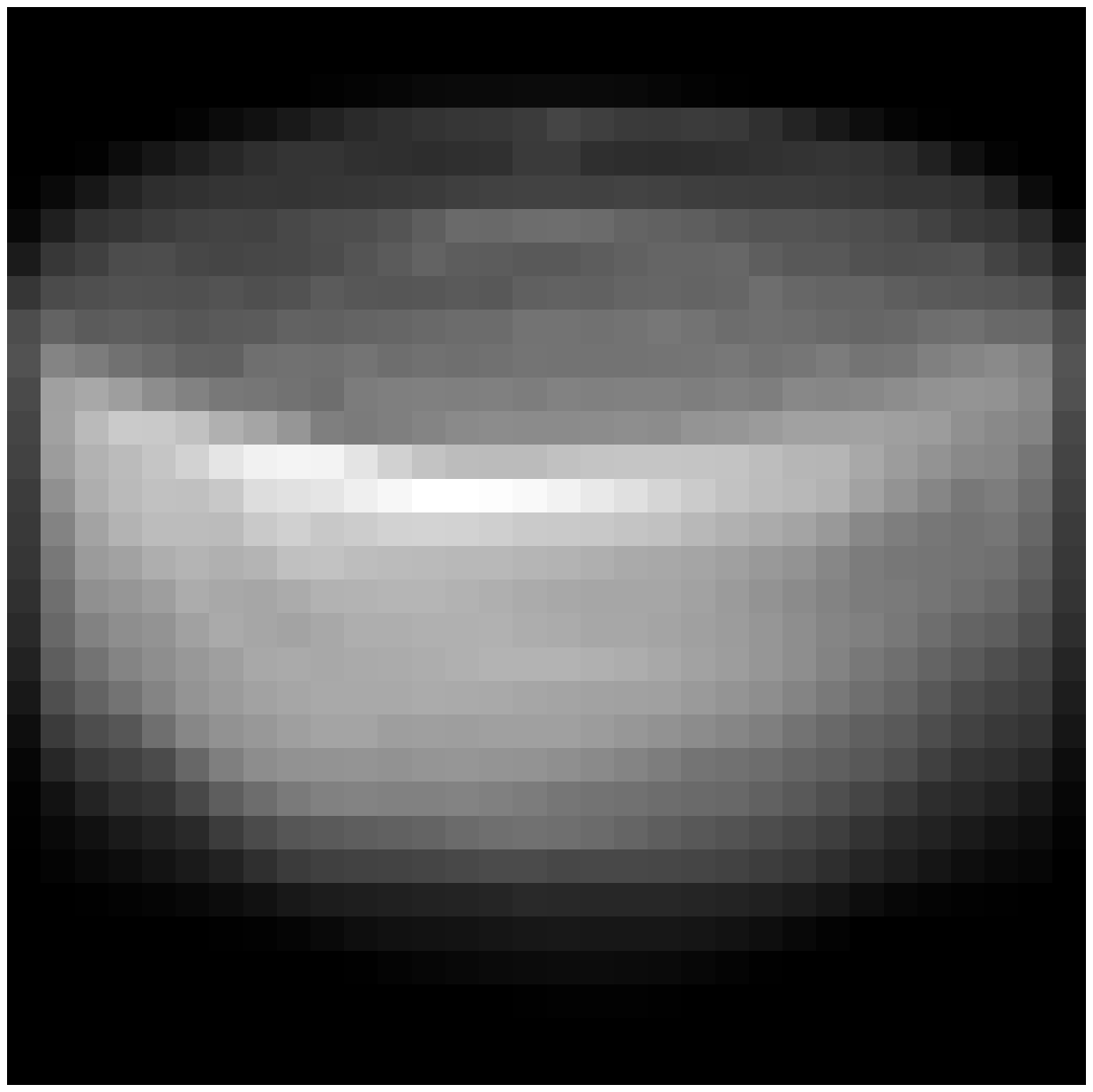} &
    \includegraphics[width=0.048\linewidth,bb=142 226 494 578,clip]{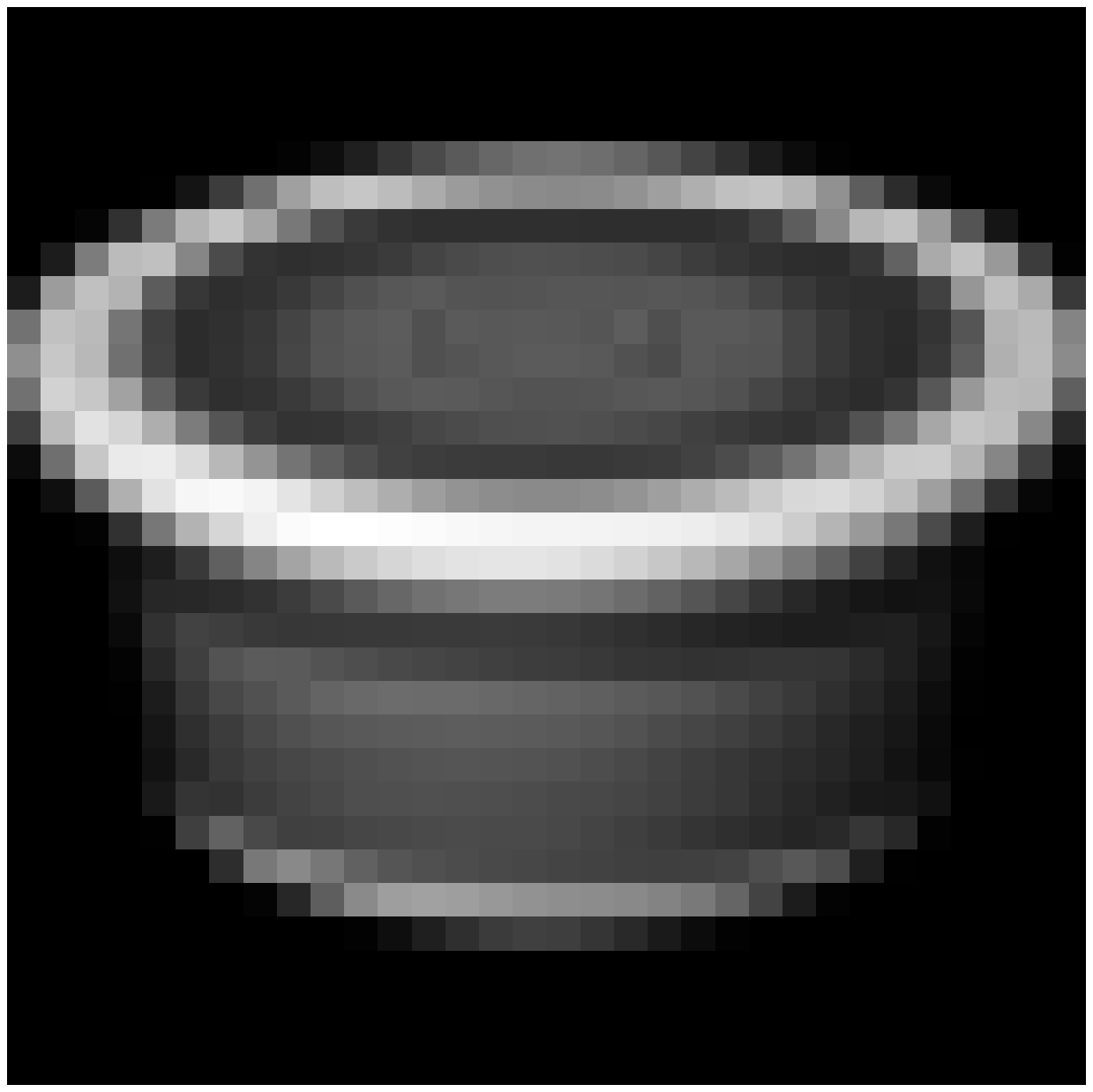} &
    \includegraphics[width=0.048\linewidth,bb=142 226 494 578,clip]{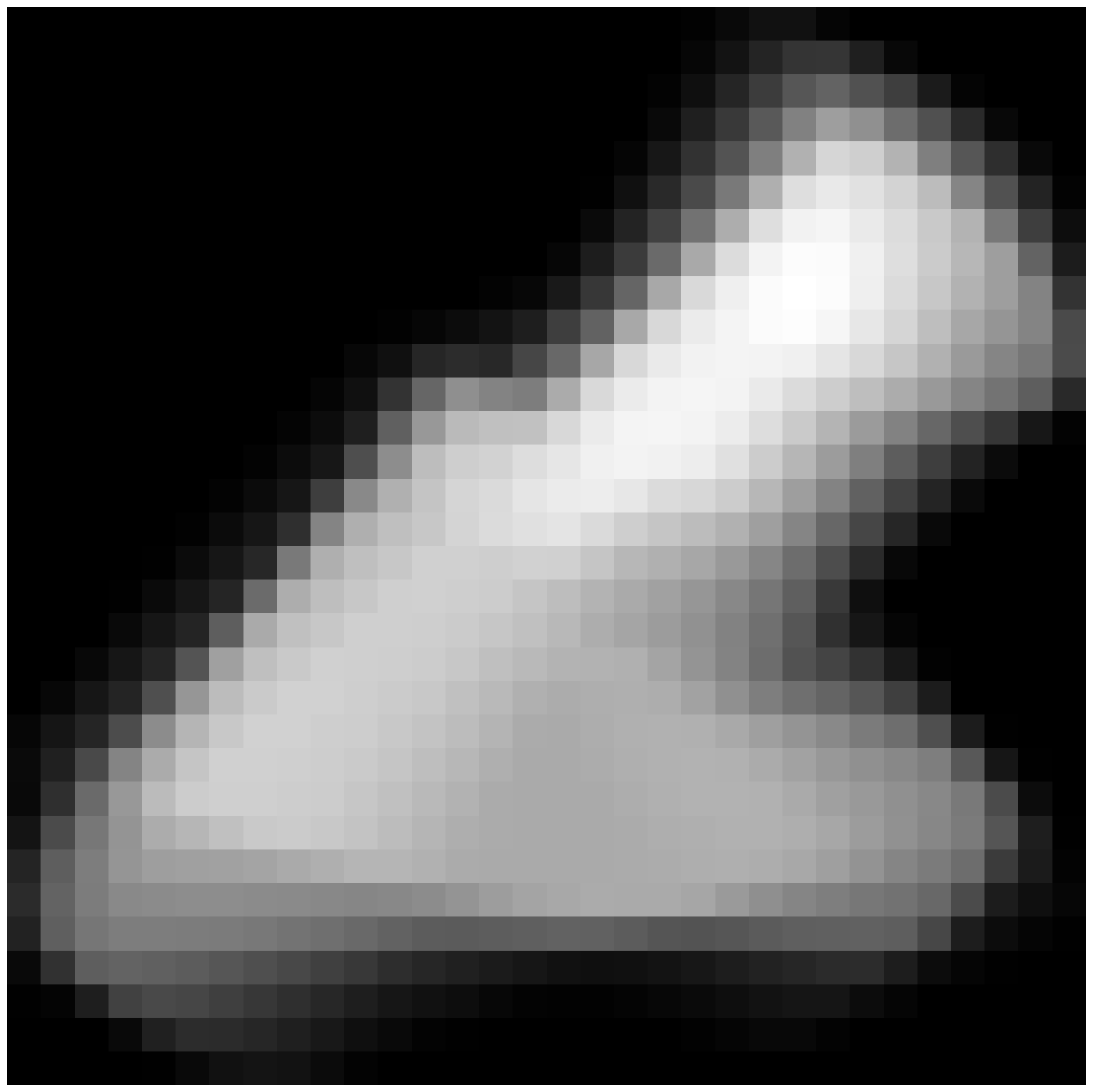} &
    \includegraphics[width=0.048\linewidth,bb=142 226 494 578,clip]{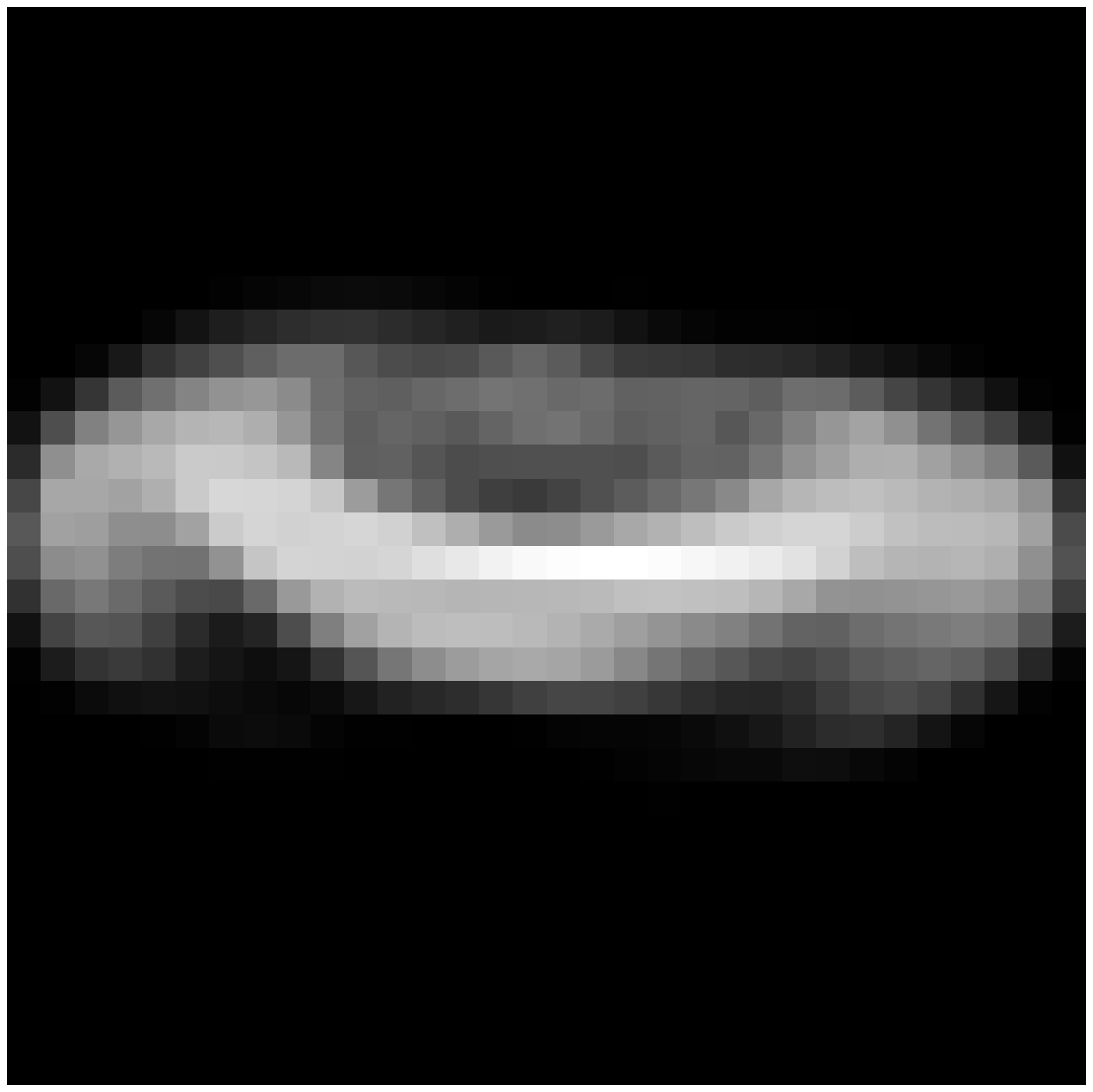} &
    \includegraphics[width=0.048\linewidth,bb=142 226 494 578,clip]{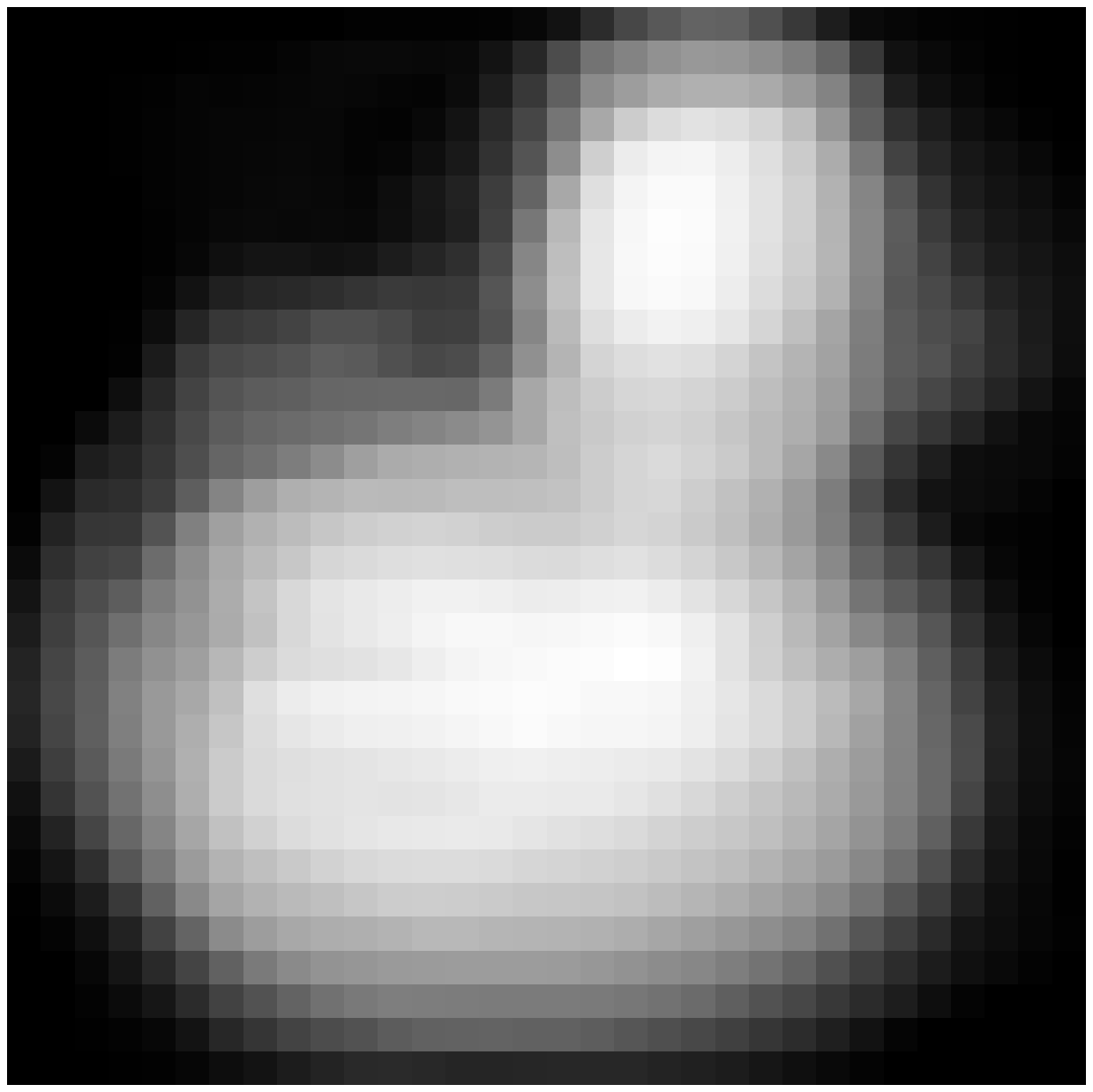} &
    \includegraphics[width=0.048\linewidth,bb=142 226 494 578,clip]{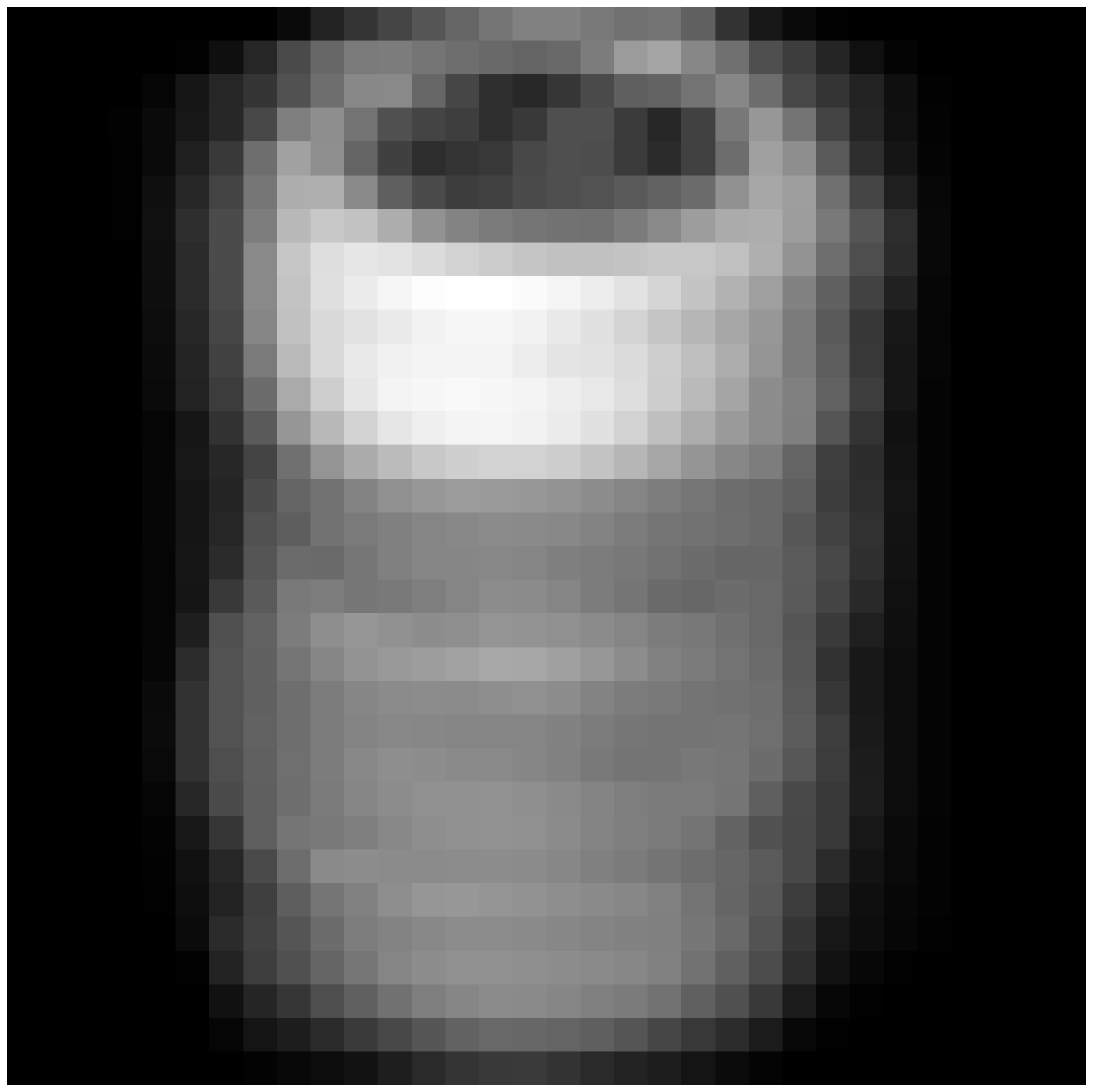} &
    \includegraphics[width=0.048\linewidth,bb=142 226 494 578,clip]{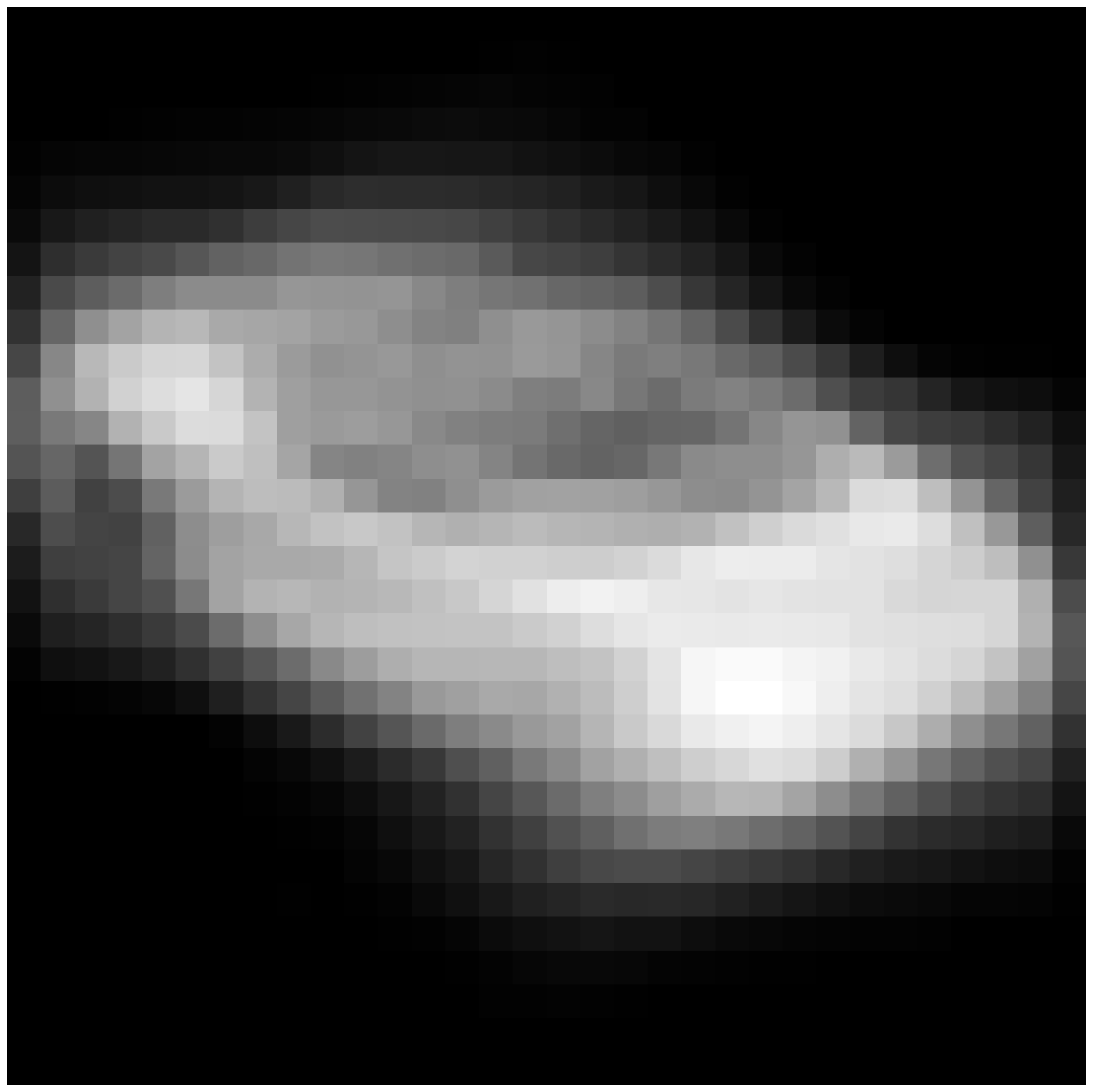} &
    \includegraphics[width=0.048\linewidth,bb=142 226 494 578,clip]{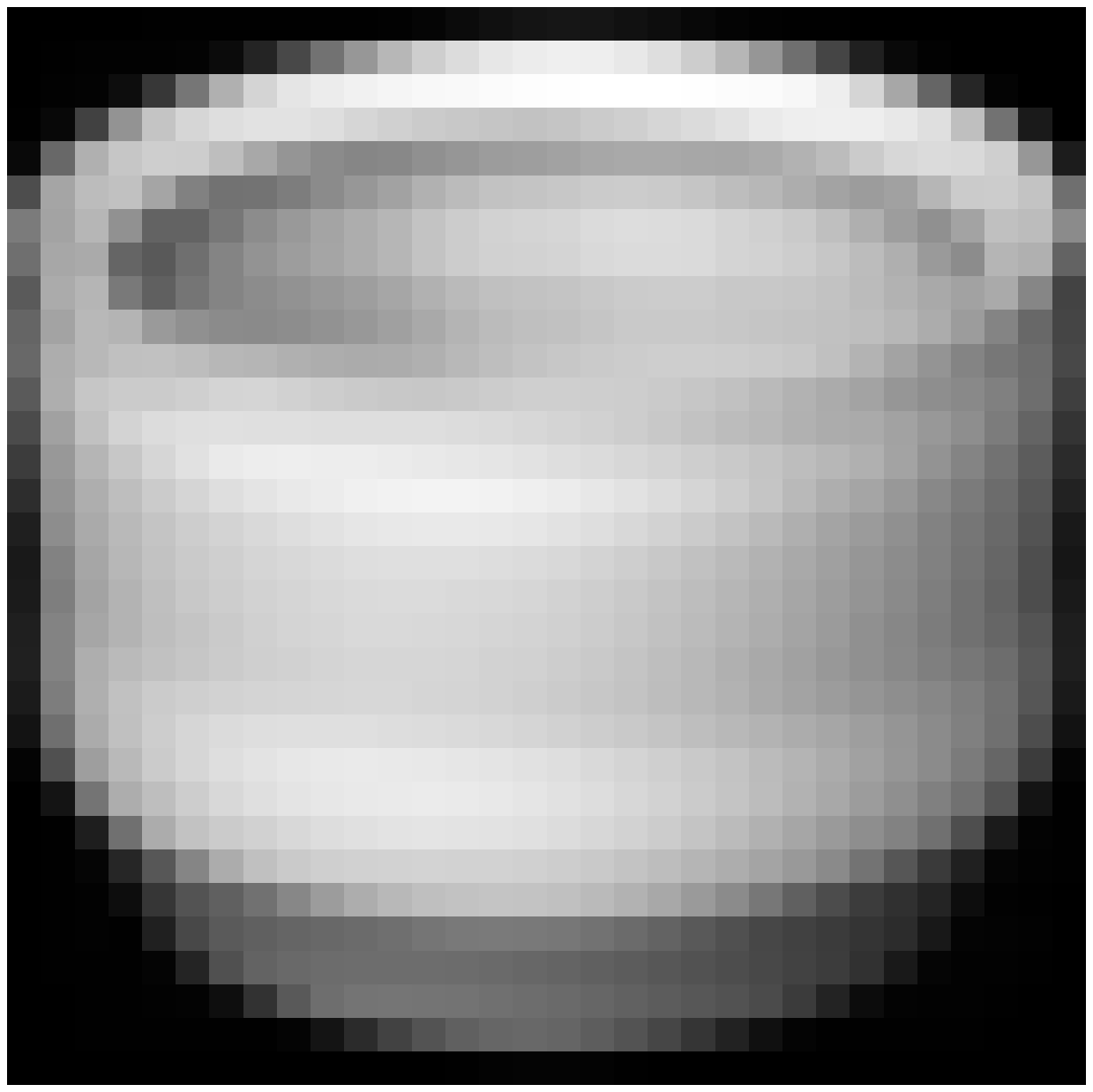} &
    \includegraphics[width=0.048\linewidth,bb=142 226 494 578,clip]{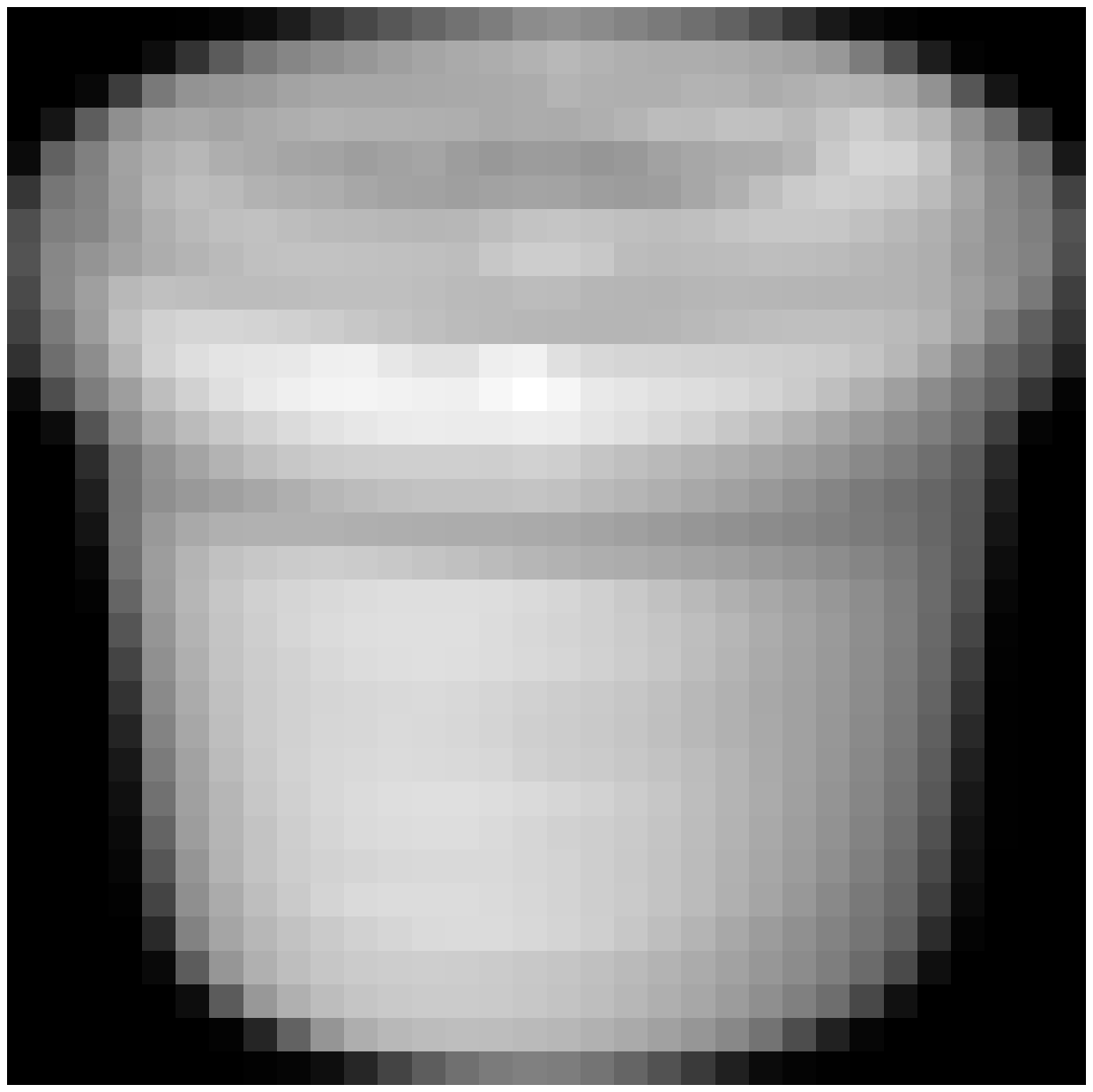} &
    \includegraphics[width=0.048\linewidth,bb=142 226 494 578,clip]{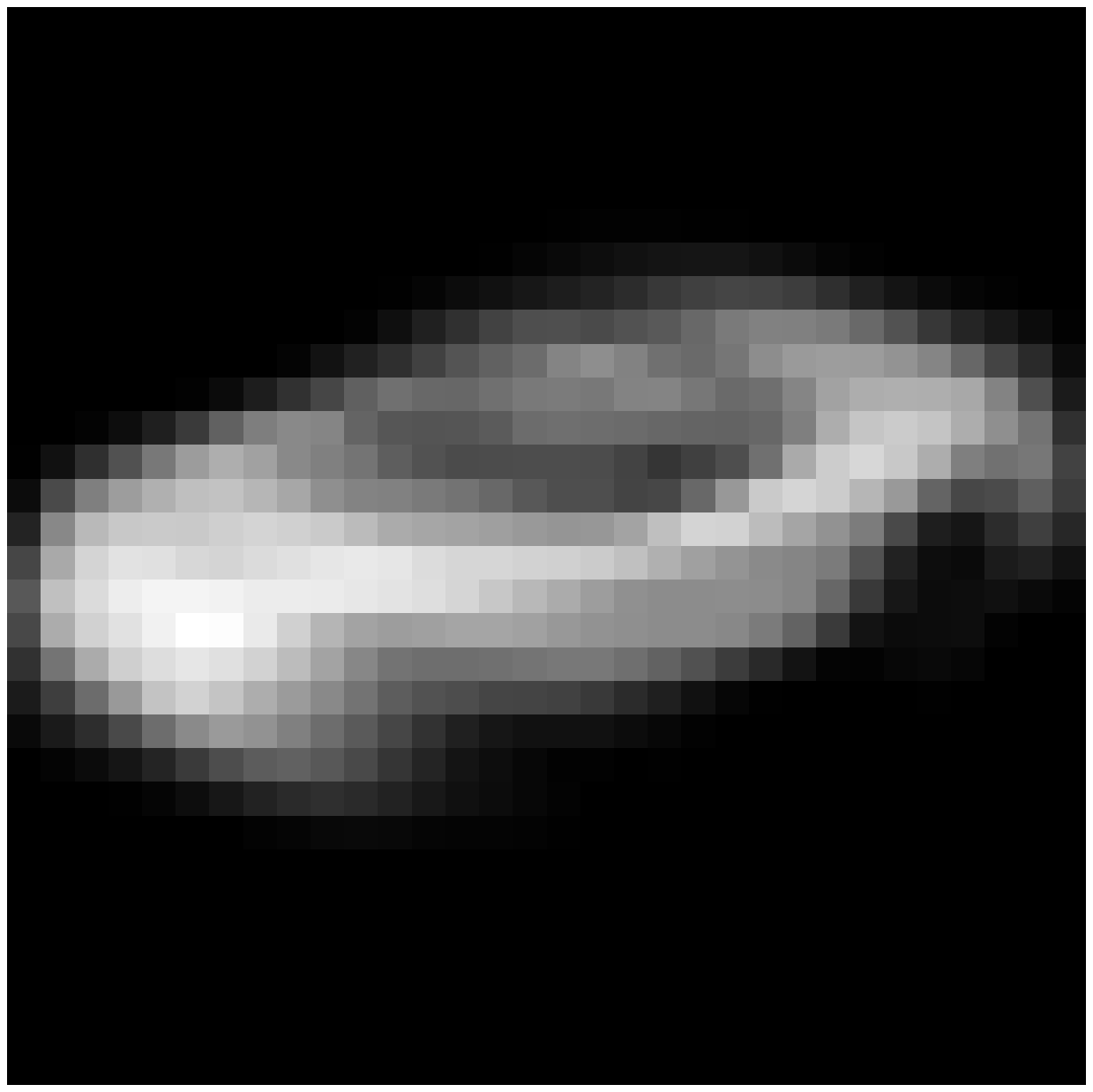} &
    \includegraphics[width=0.048\linewidth,bb=142 226 494 578,clip]{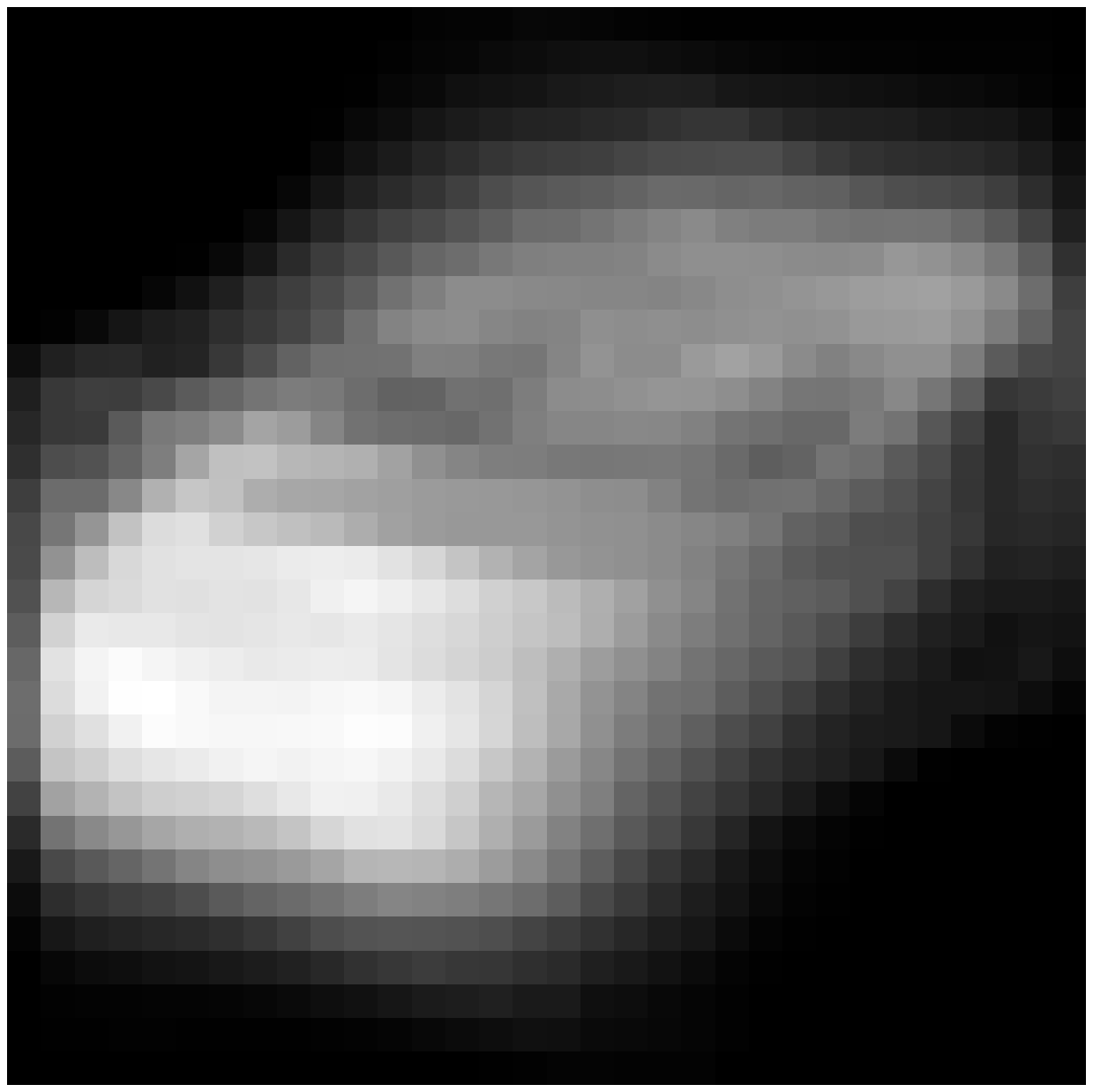} &
    \includegraphics[width=0.048\linewidth,bb=142 226 494 578,clip]{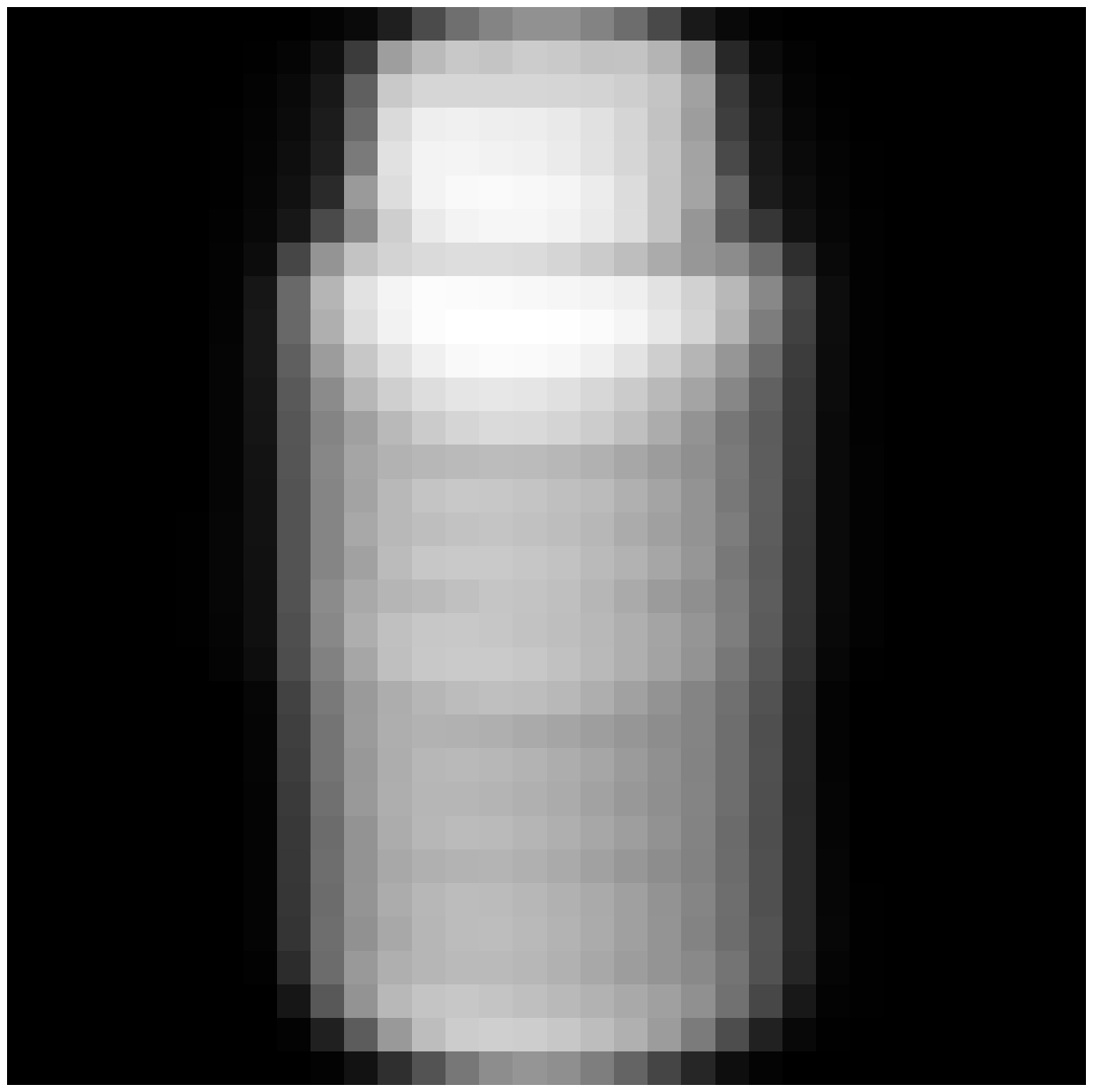} \\[-1ex]
    \rotatebox{90}{\tiny\hspace{0.5ex}\caja{c}{c}{$K$- \\ modes}} &
    \includegraphics[width=0.048\linewidth,bb=142 226 494 578,clip]{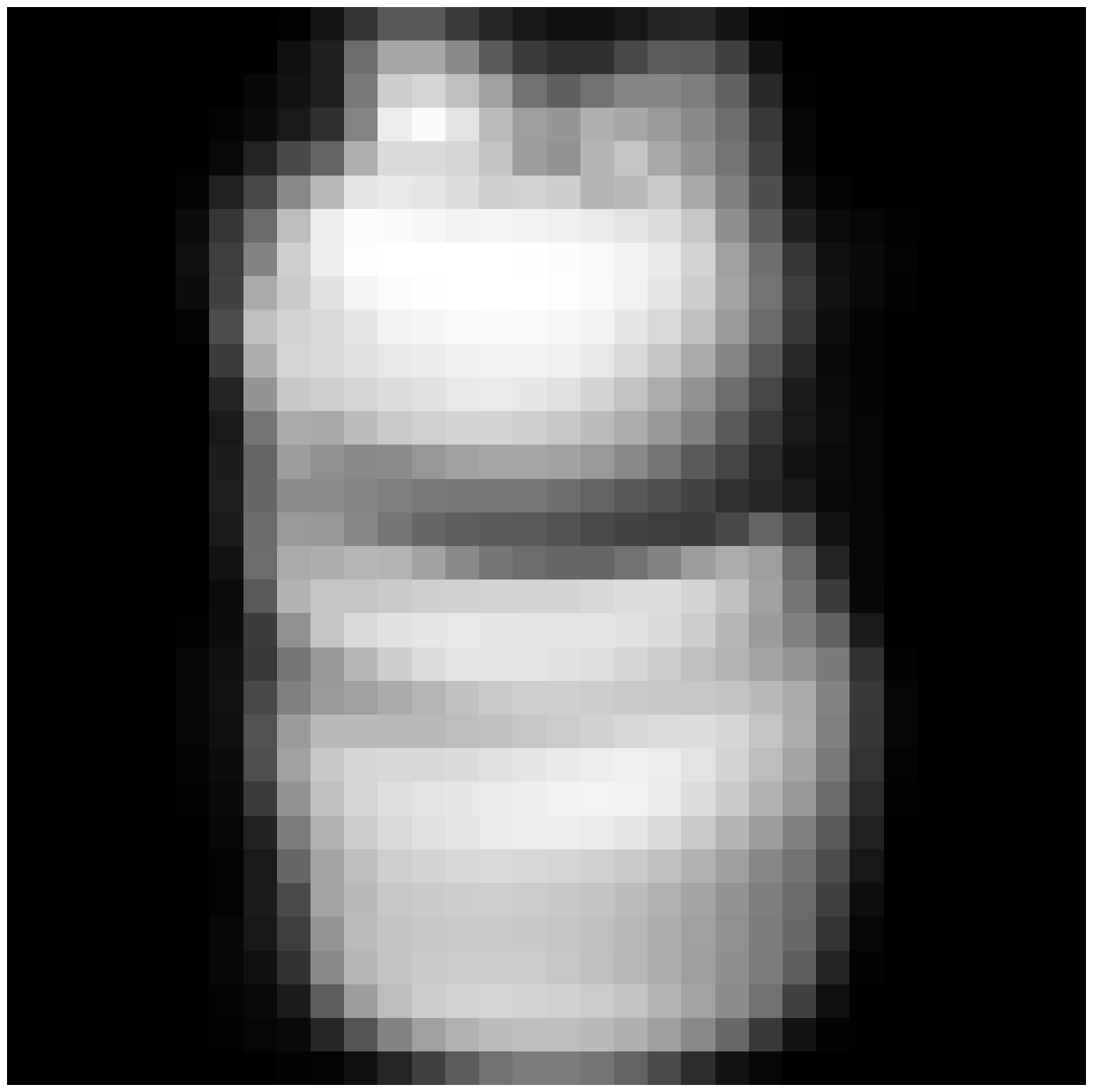} &
    \includegraphics[width=0.048\linewidth,bb=142 226 494 578,clip]{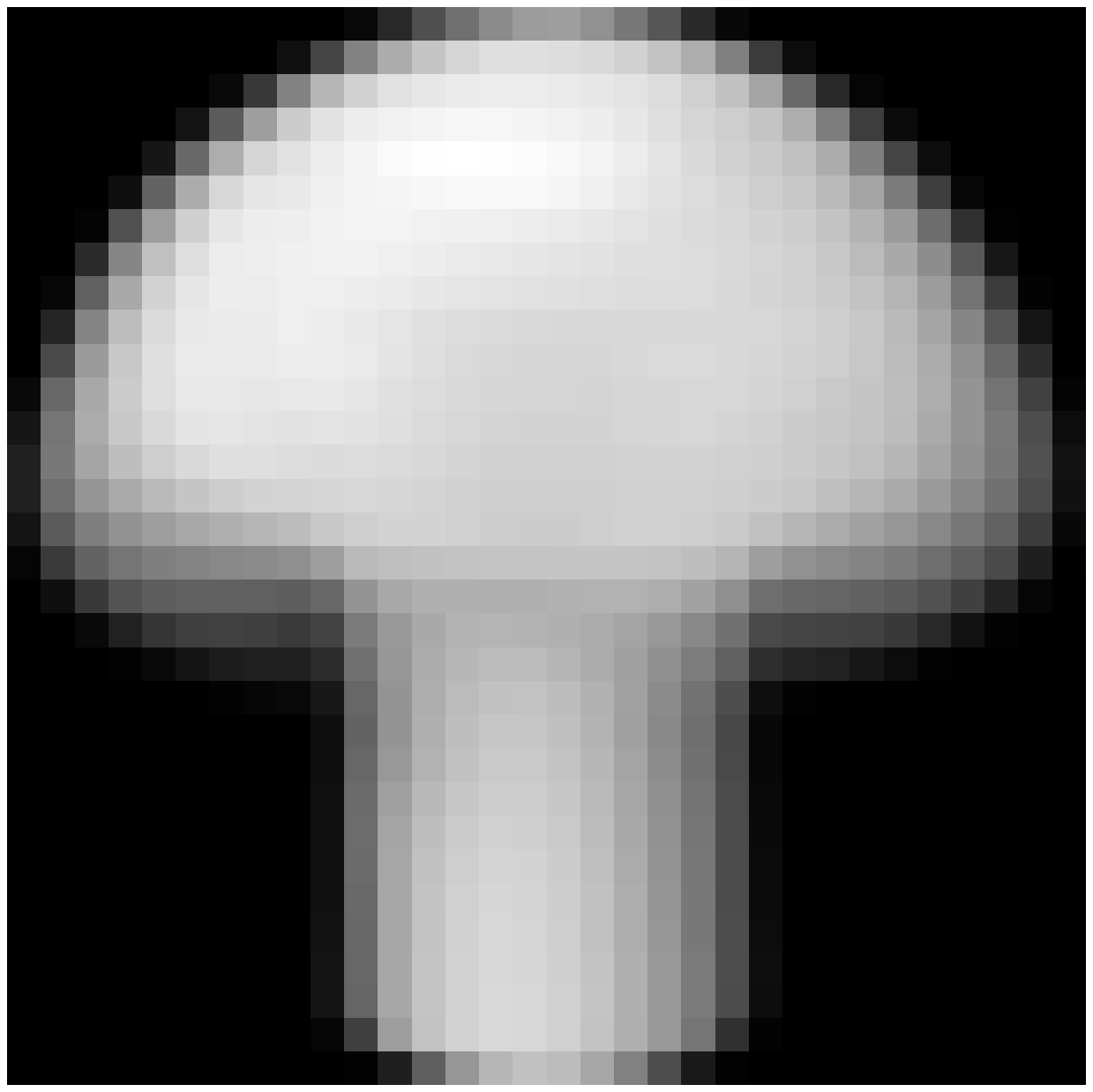} &
    \includegraphics[width=0.048\linewidth,bb=142 226 494 578,clip]{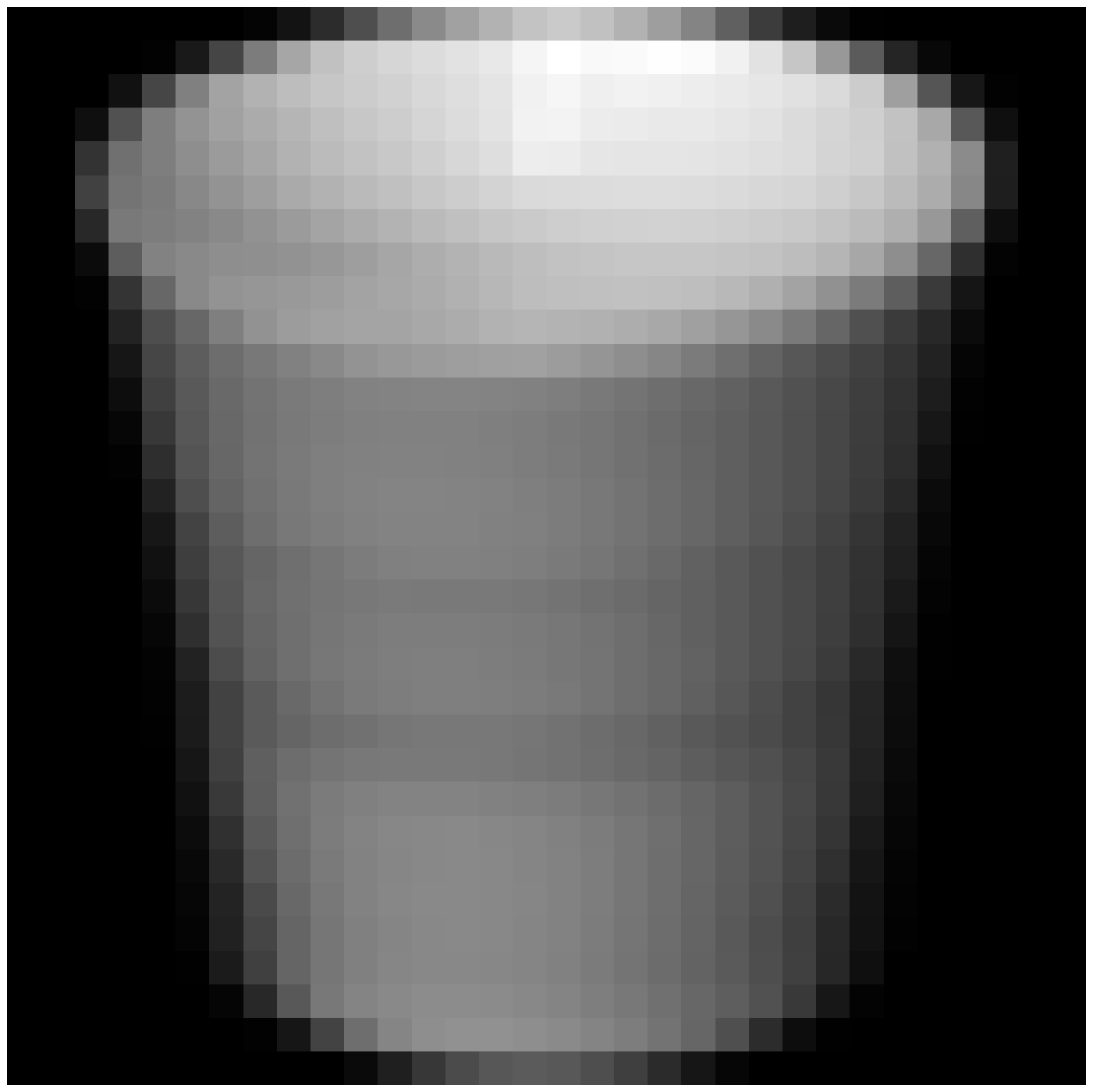} &
    \includegraphics[width=0.048\linewidth,bb=142 226 494 578,clip]{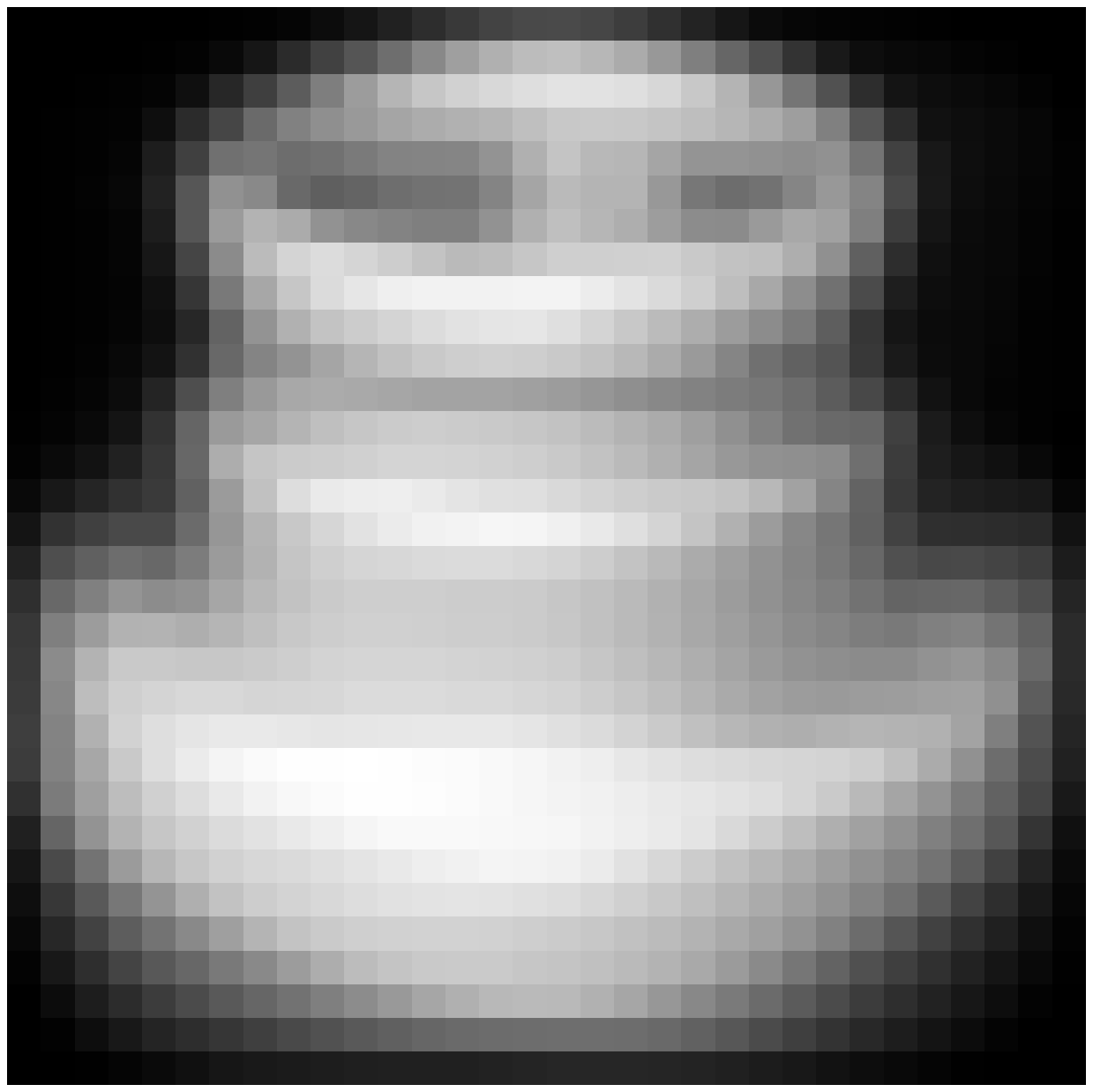} &
    \includegraphics[width=0.048\linewidth,bb=142 226 494 578,clip]{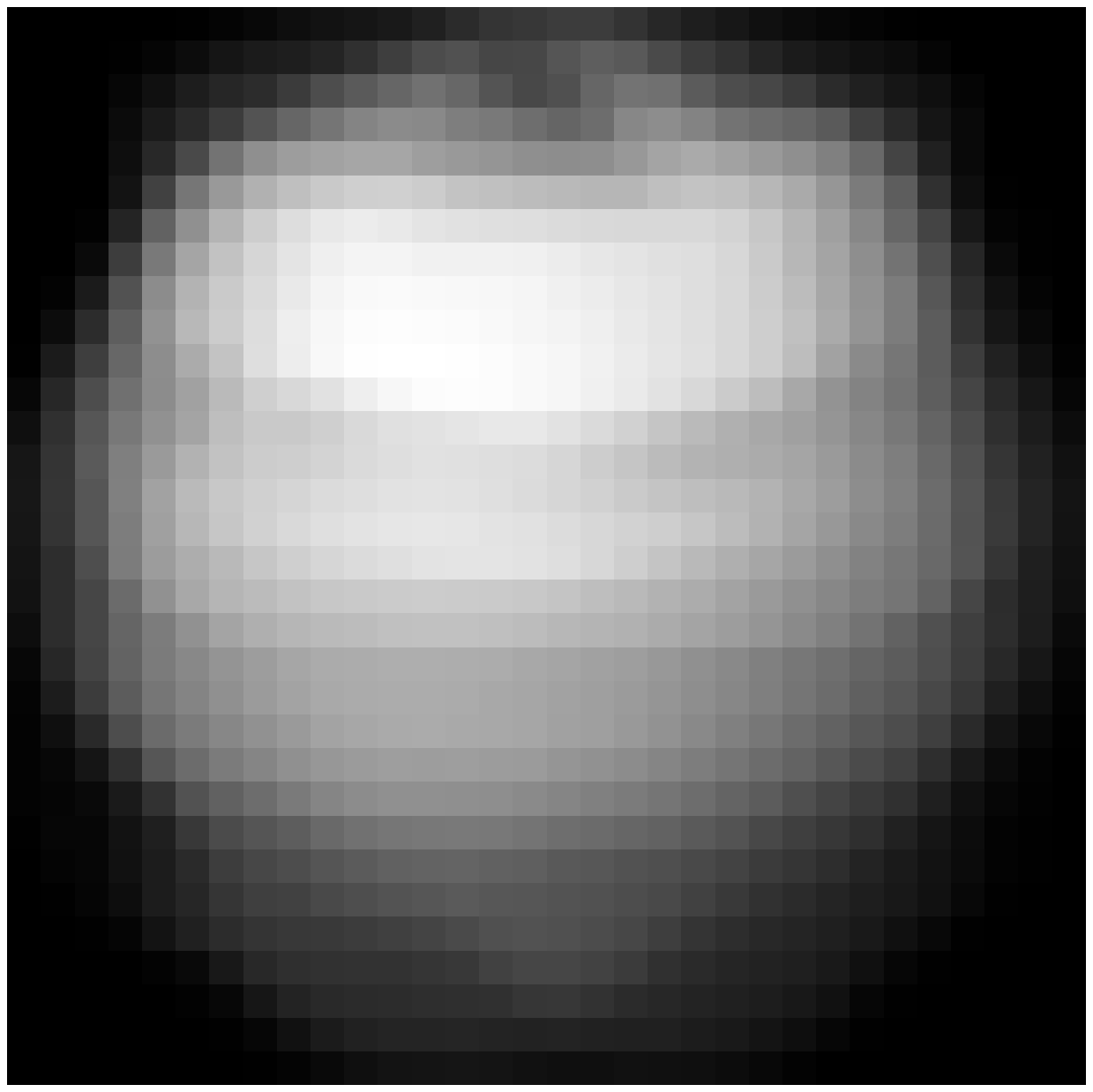} &
    \includegraphics[width=0.048\linewidth,bb=142 226 494 578,clip]{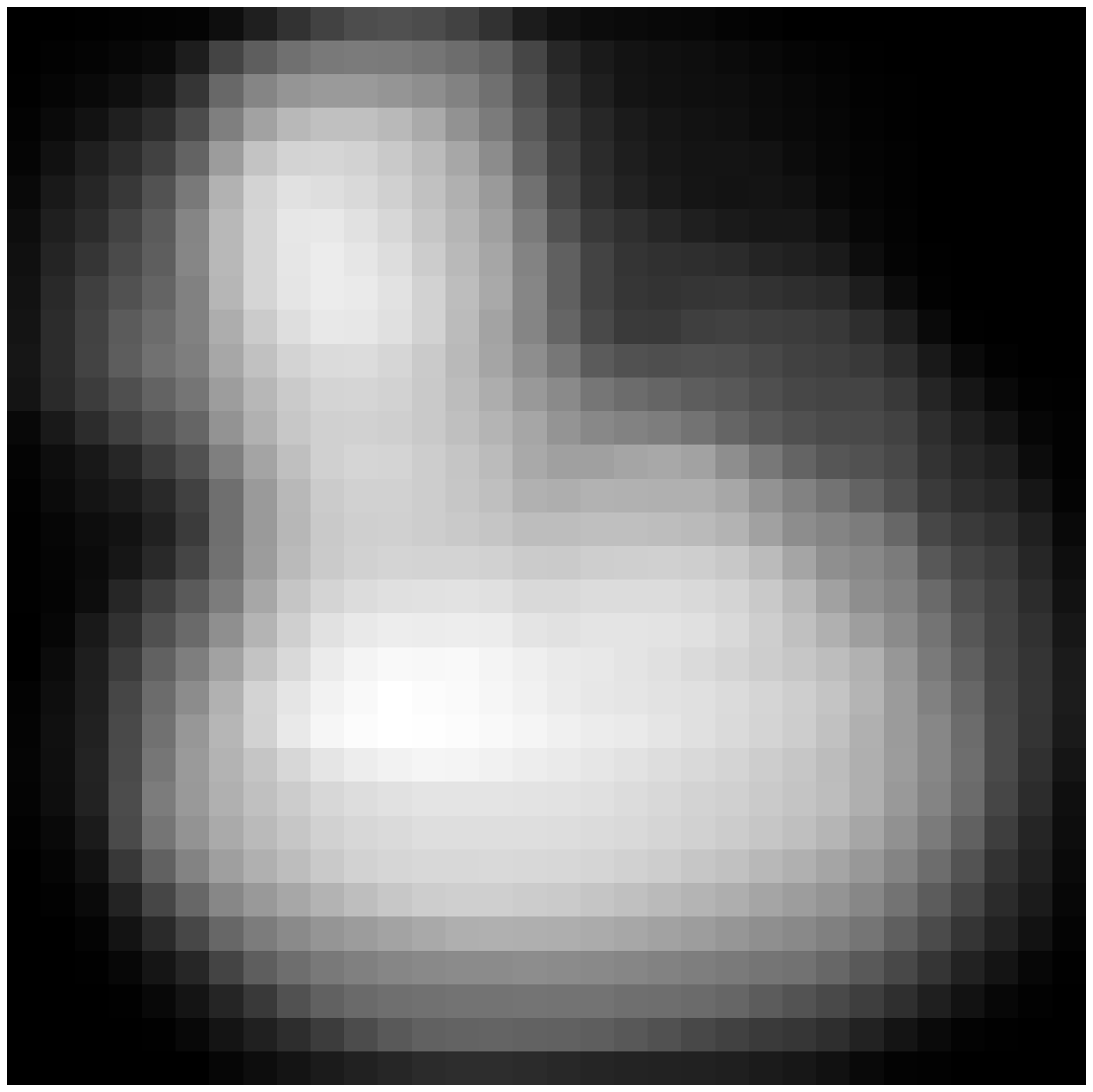} &
    \includegraphics[width=0.048\linewidth,bb=142 226 494 578,clip]{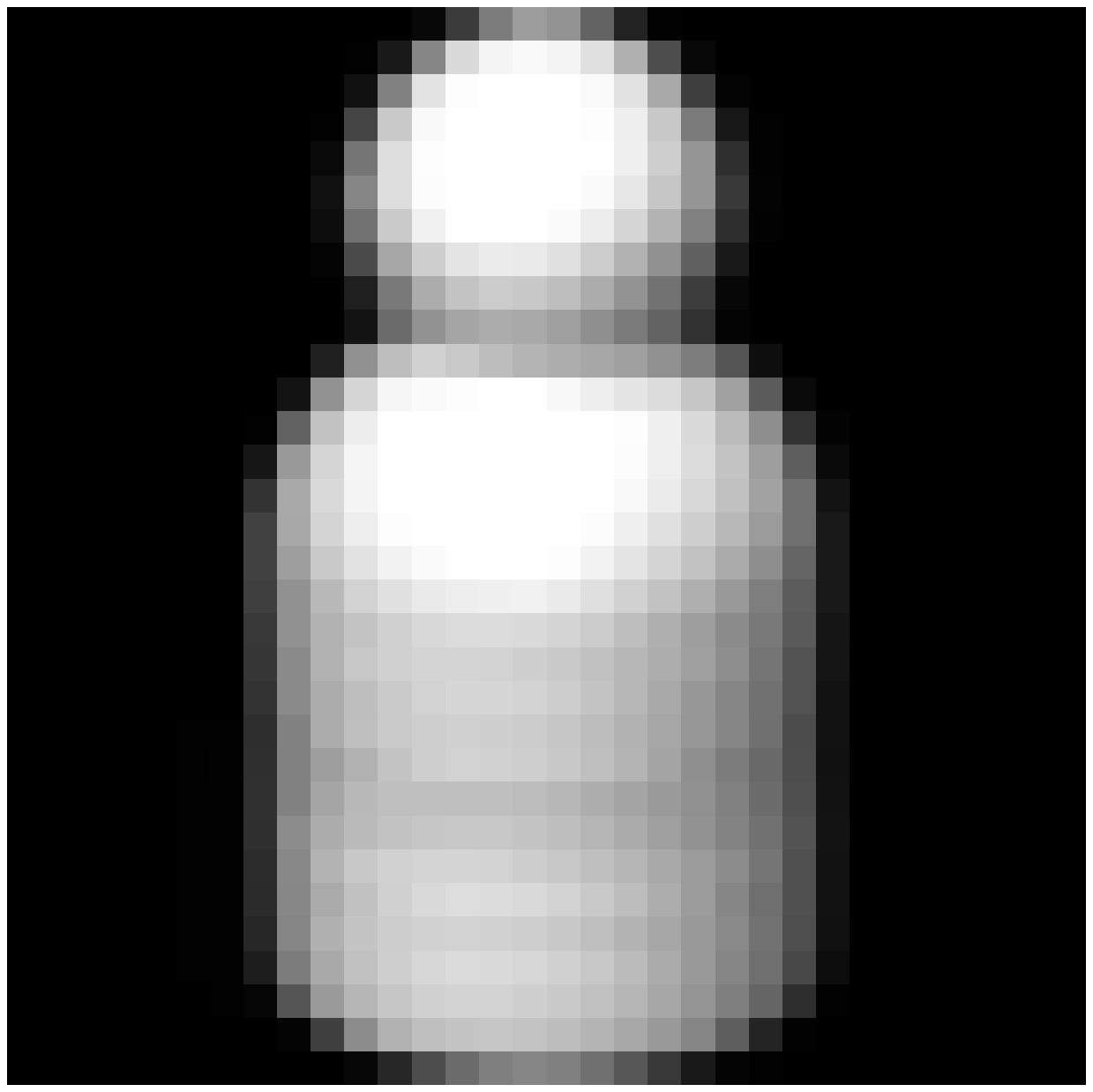} &
    \includegraphics[width=0.048\linewidth,bb=142 226 494 578,clip]{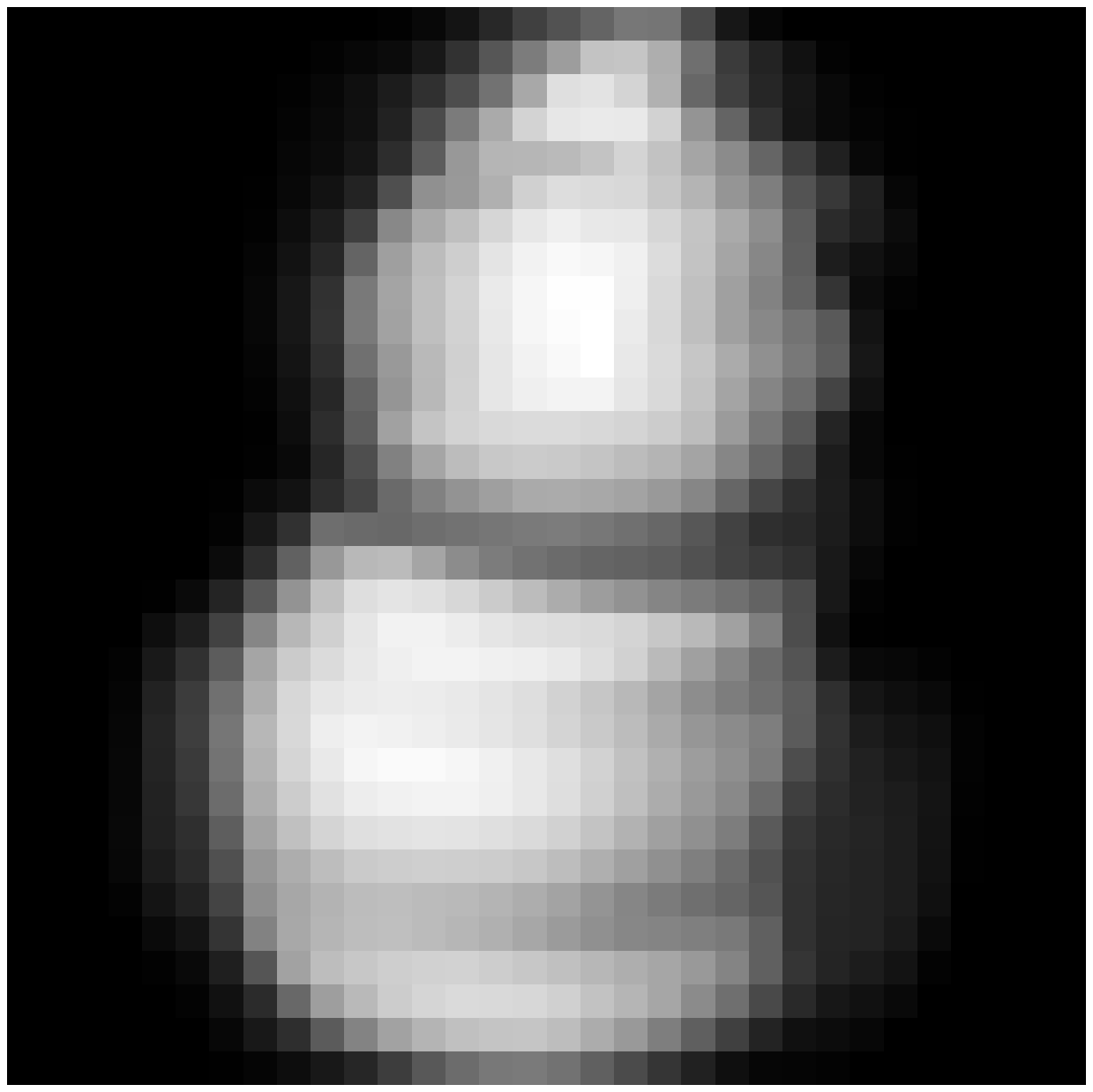} &
    \includegraphics[width=0.048\linewidth,bb=142 226 494 578,clip]{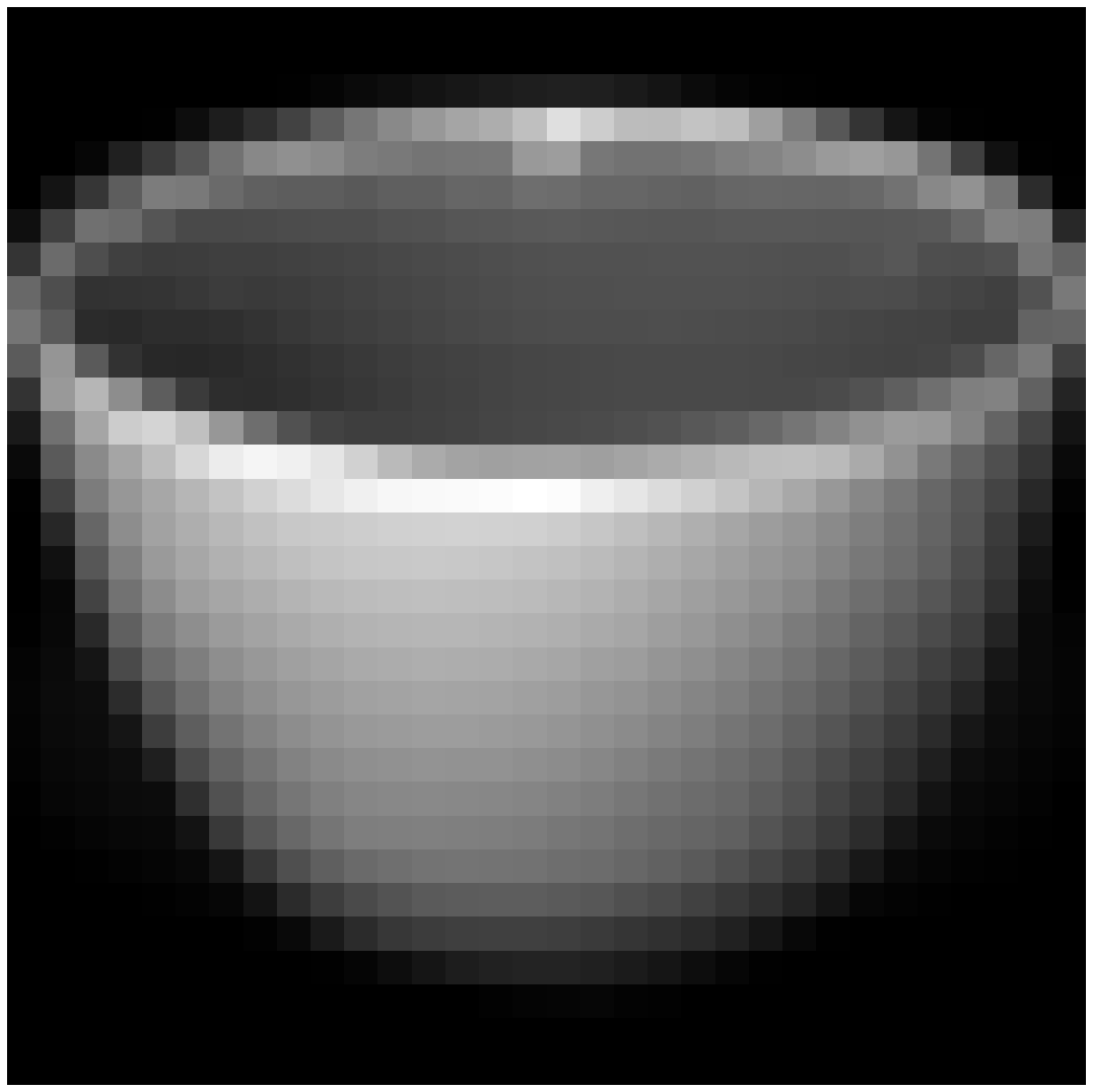} &
    \includegraphics[width=0.048\linewidth,bb=142 226 494 578,clip]{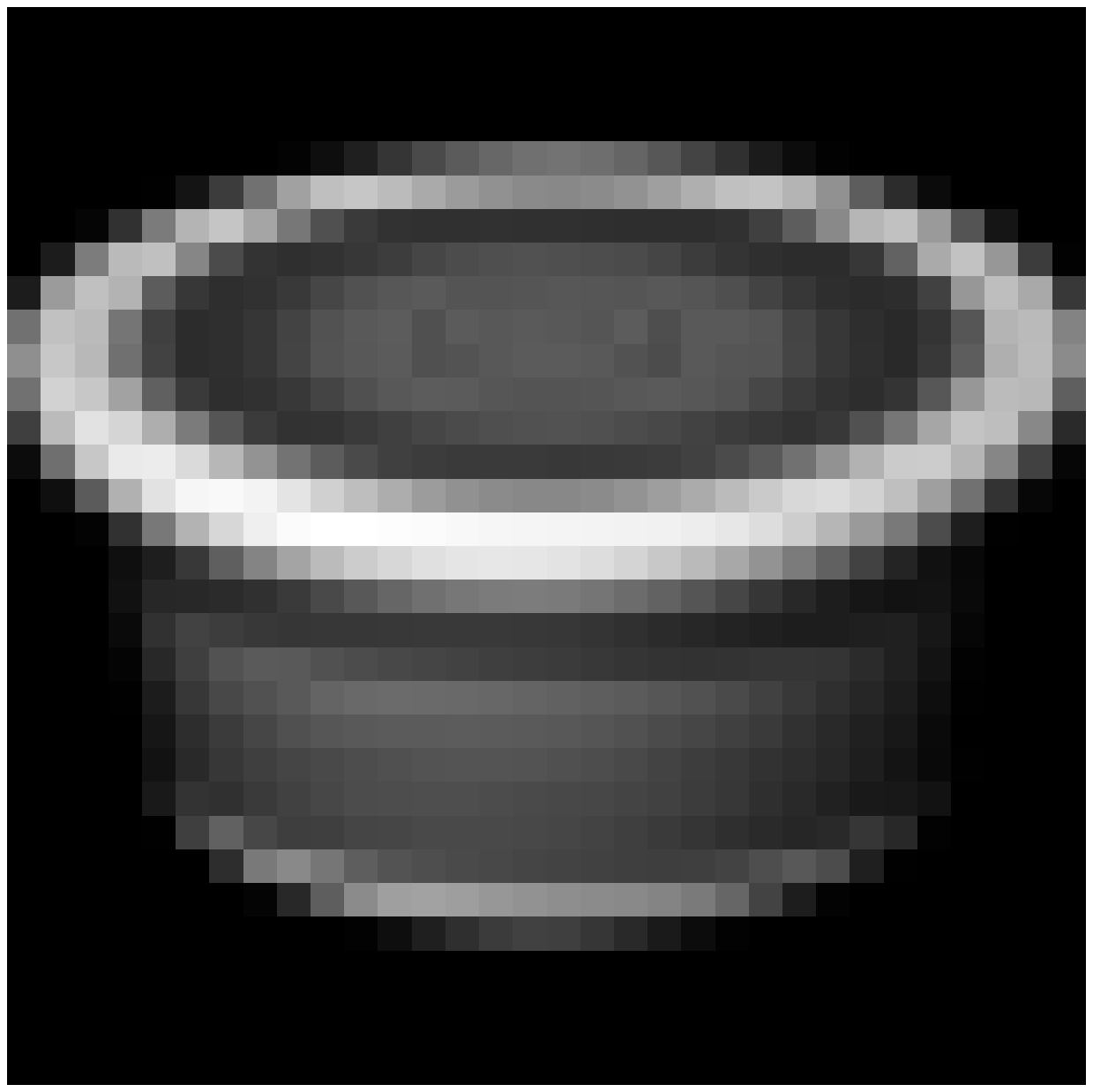} &
    \includegraphics[width=0.048\linewidth,bb=142 226 494 578,clip]{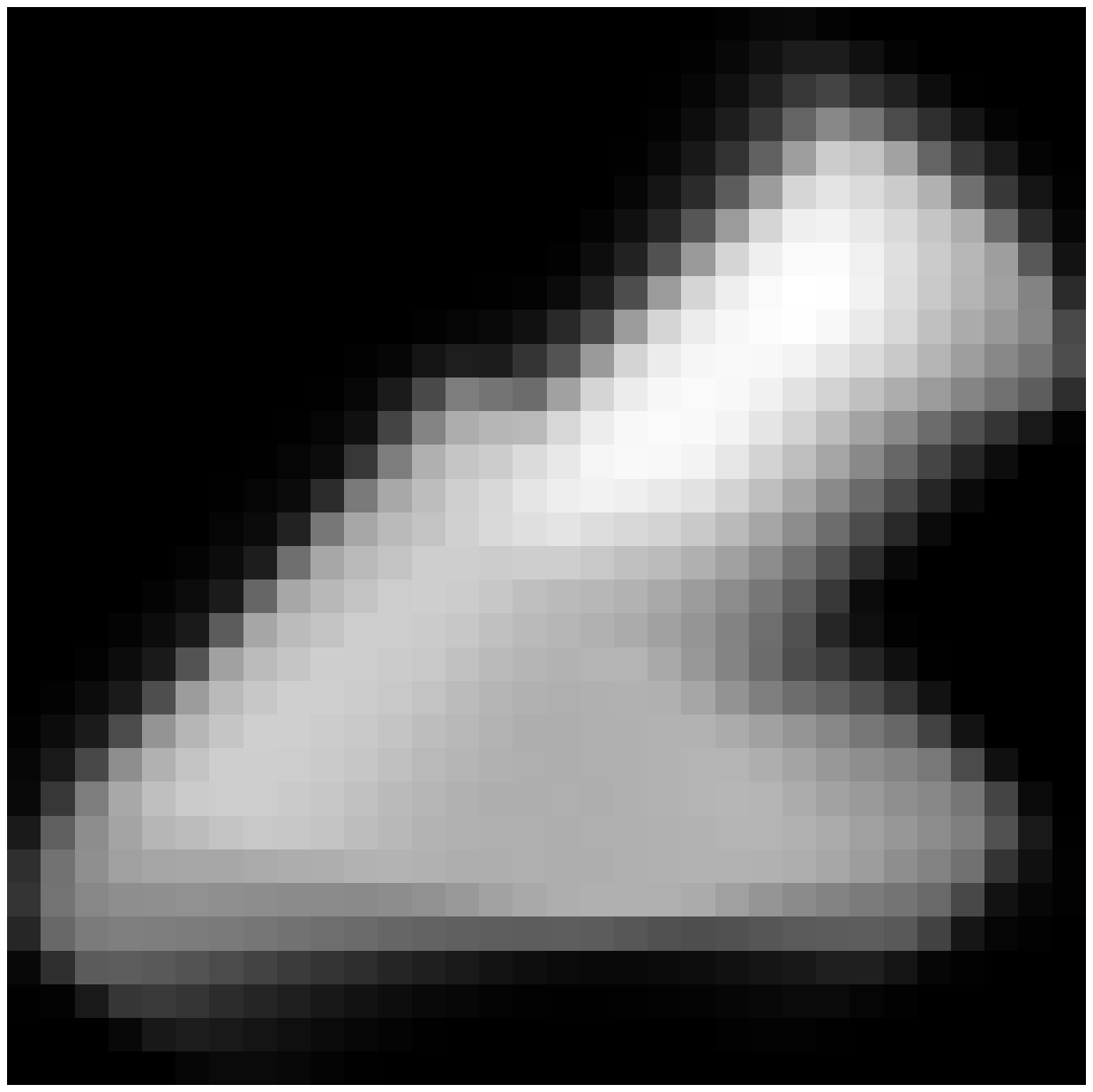} &
    \includegraphics[width=0.048\linewidth,bb=142 226 494 578,clip]{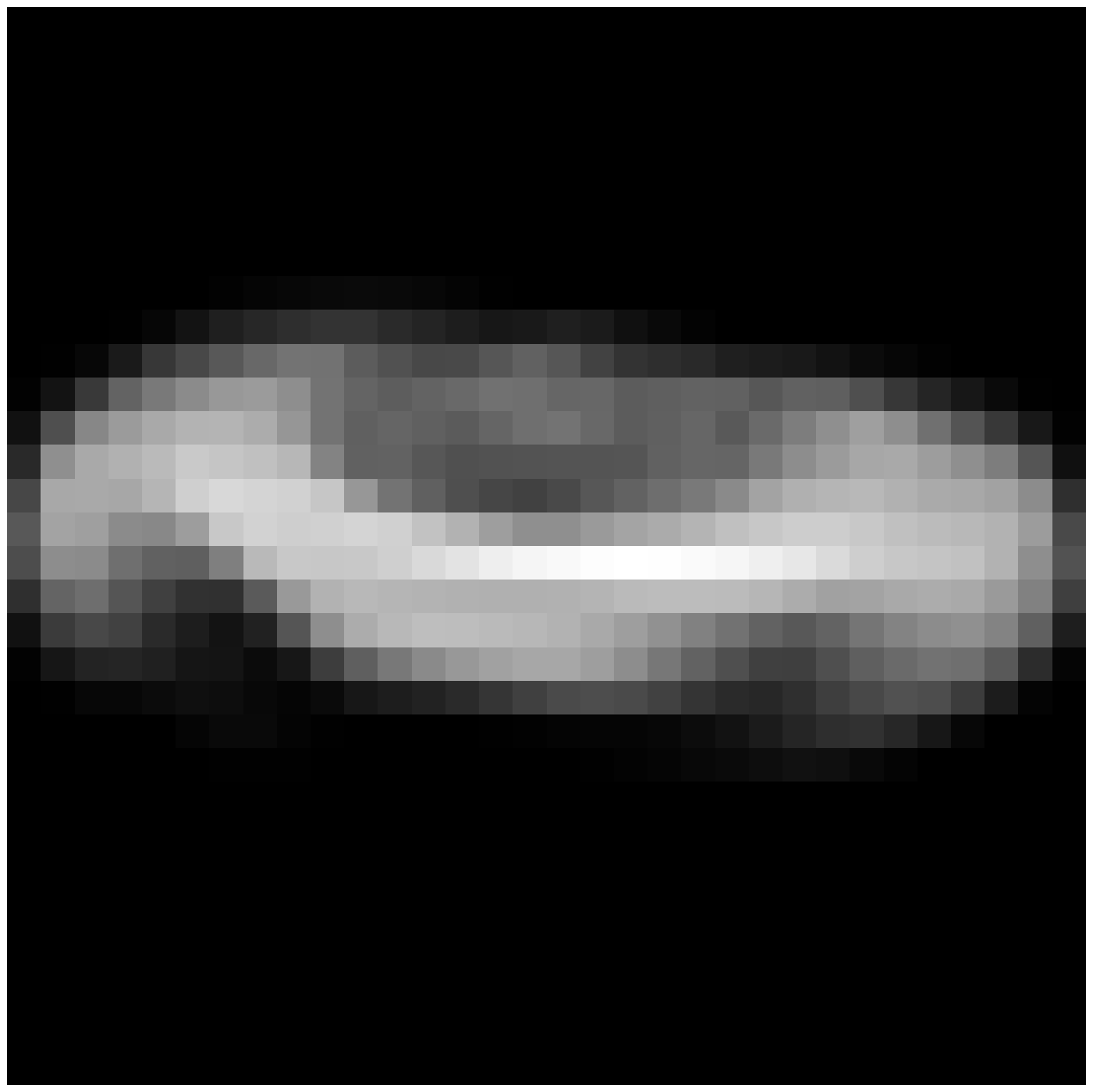} &
    \includegraphics[width=0.048\linewidth,bb=142 226 494 578,clip]{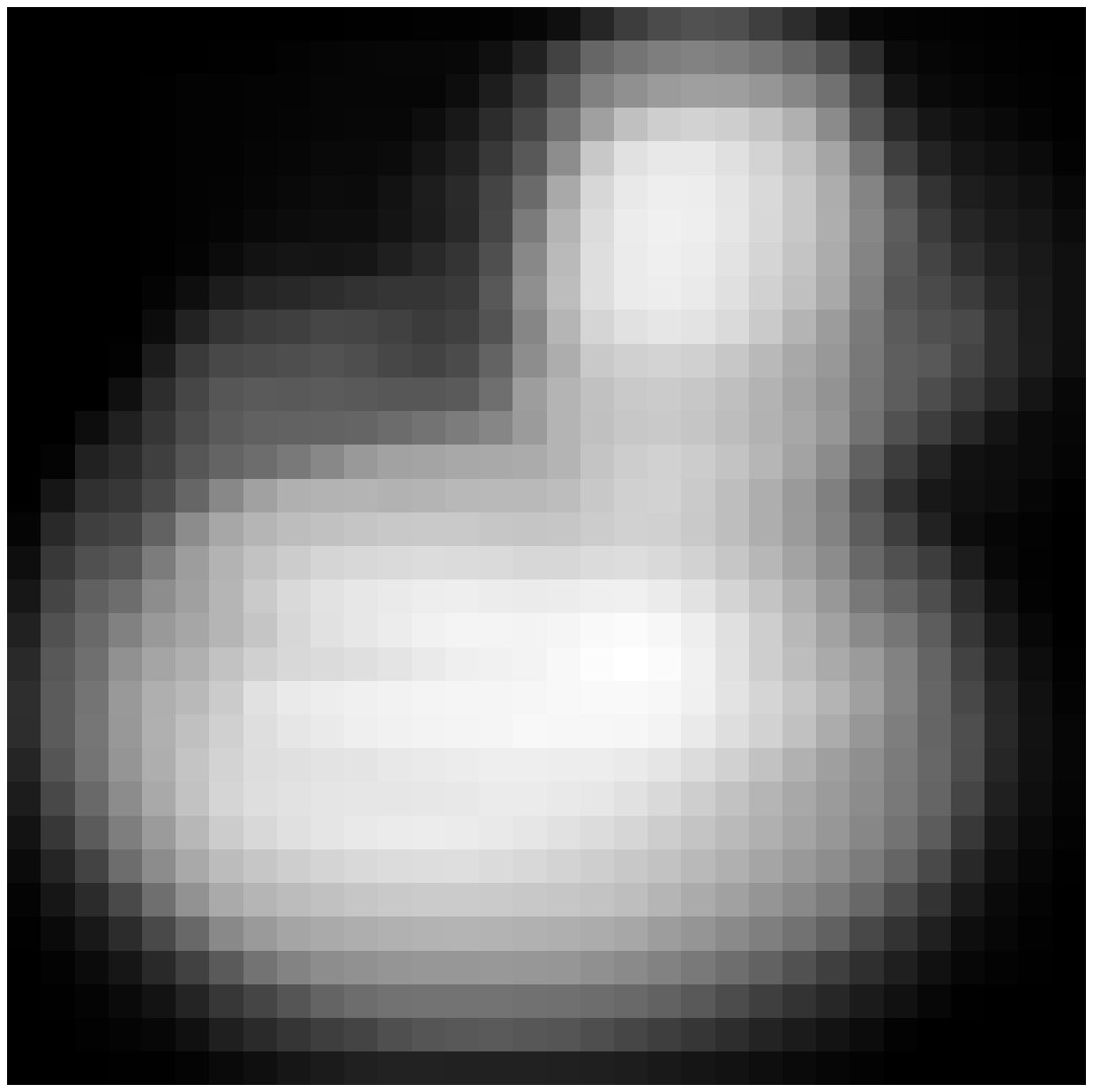} &
    \includegraphics[width=0.048\linewidth,bb=142 226 494 578,clip]{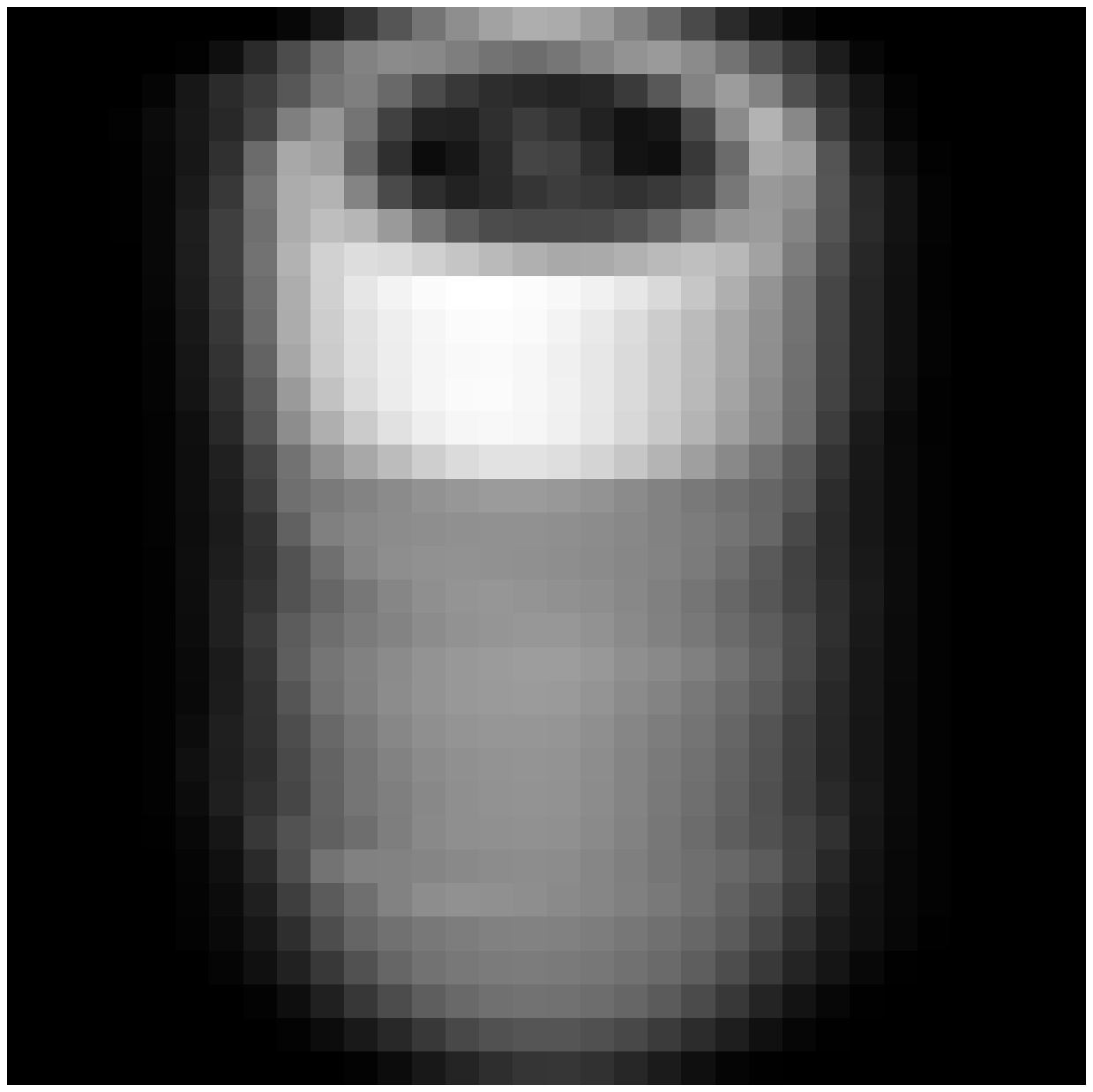} &
    \includegraphics[width=0.048\linewidth,bb=142 226 494 578,clip]{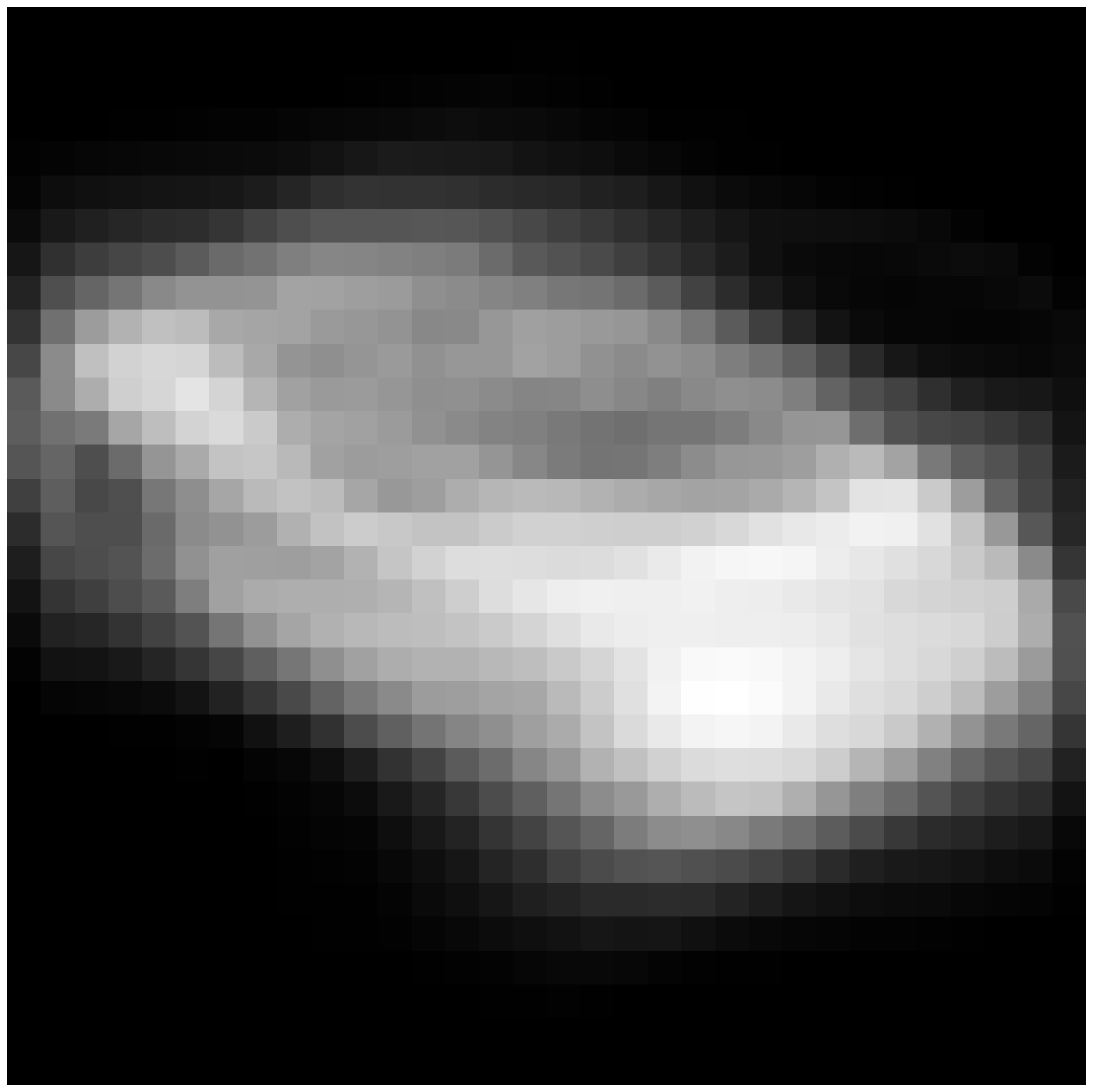} &
    \includegraphics[width=0.048\linewidth,bb=142 226 494 578,clip]{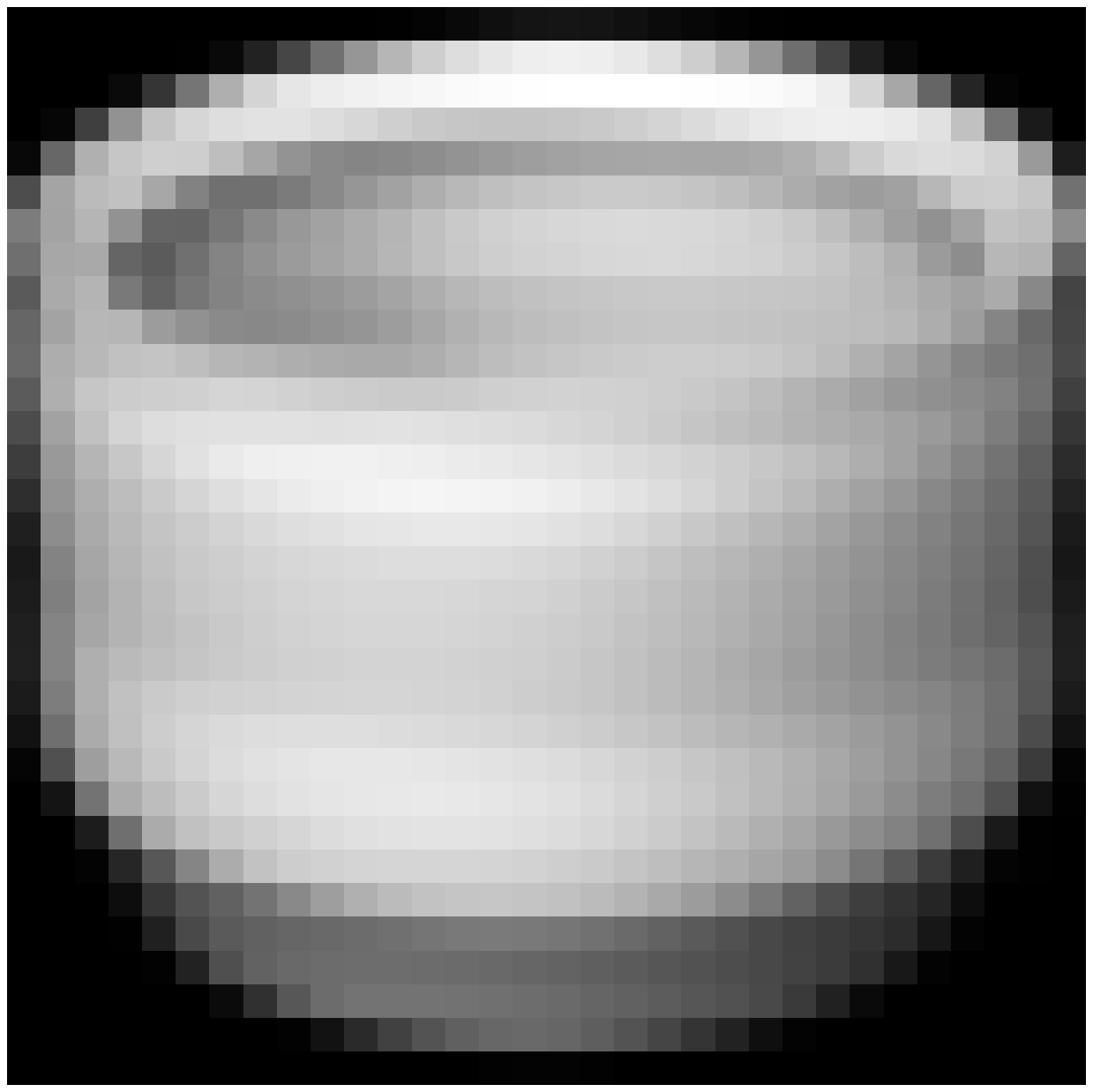} &
    \includegraphics[width=0.048\linewidth,bb=142 226 494 578,clip]{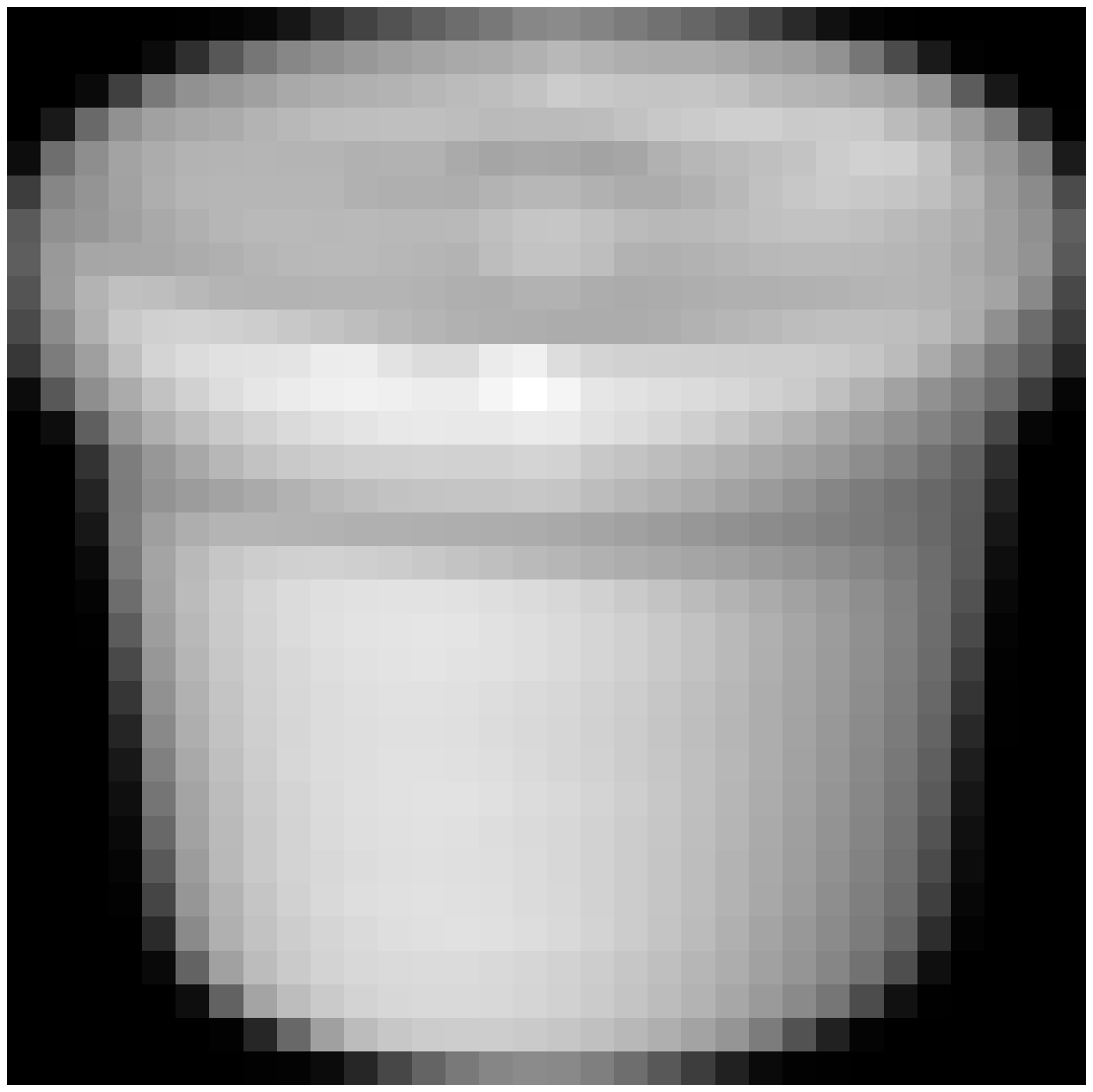} &
    \includegraphics[width=0.048\linewidth,bb=142 226 494 578,clip]{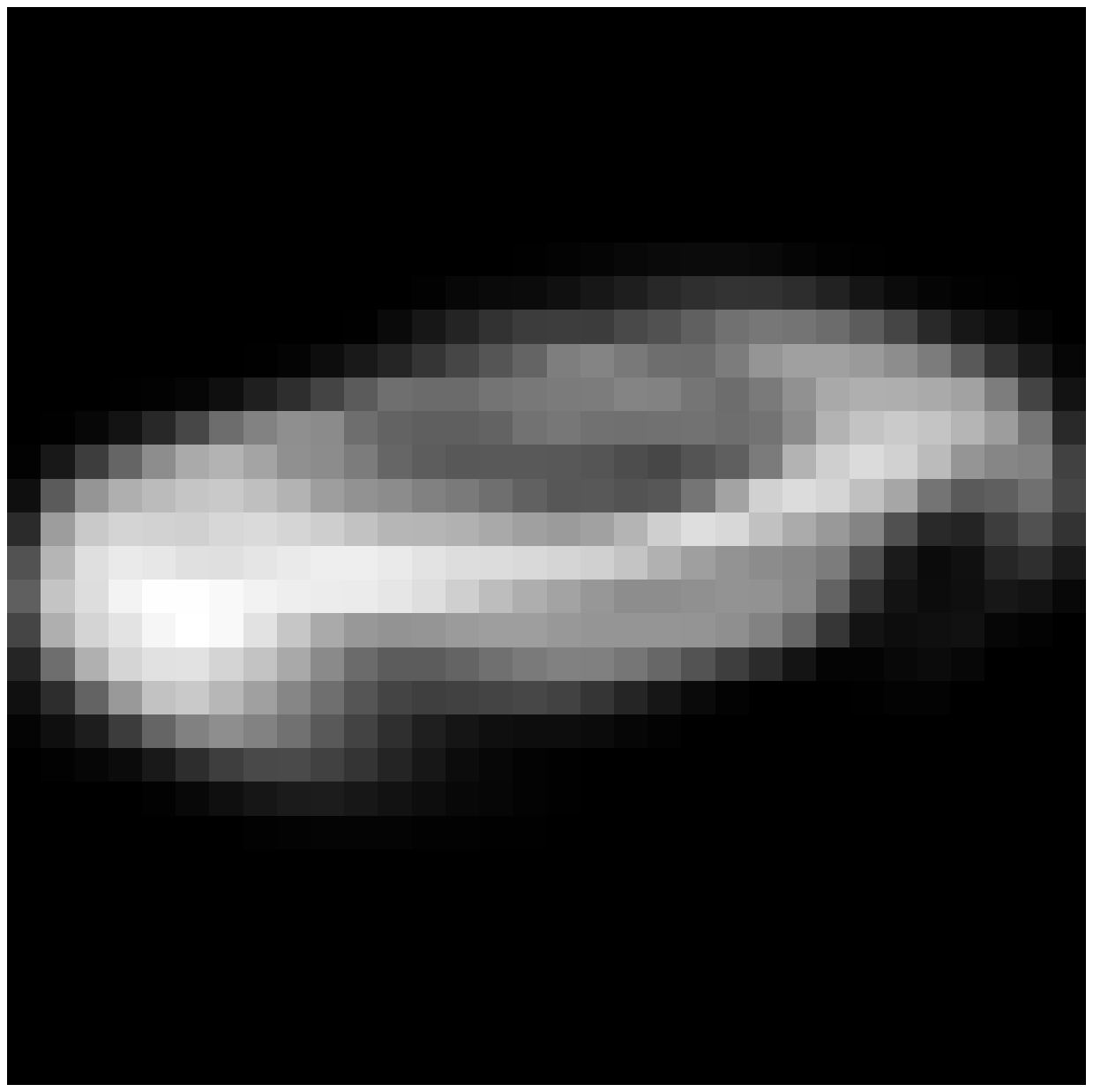} &
    \includegraphics[width=0.048\linewidth,bb=142 226 494 578,clip]{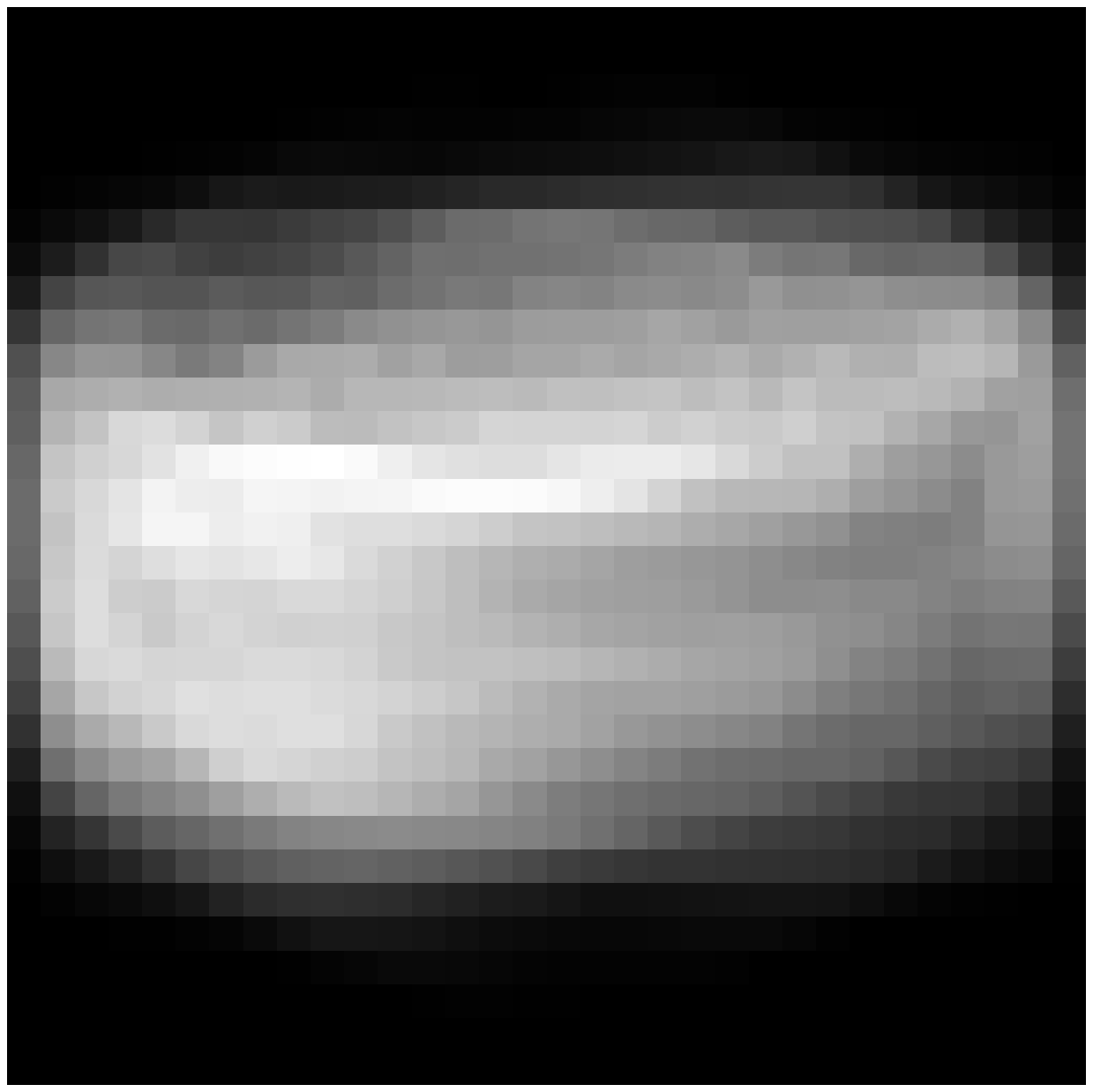} &
    \includegraphics[width=0.048\linewidth,bb=142 226 494 578,clip]{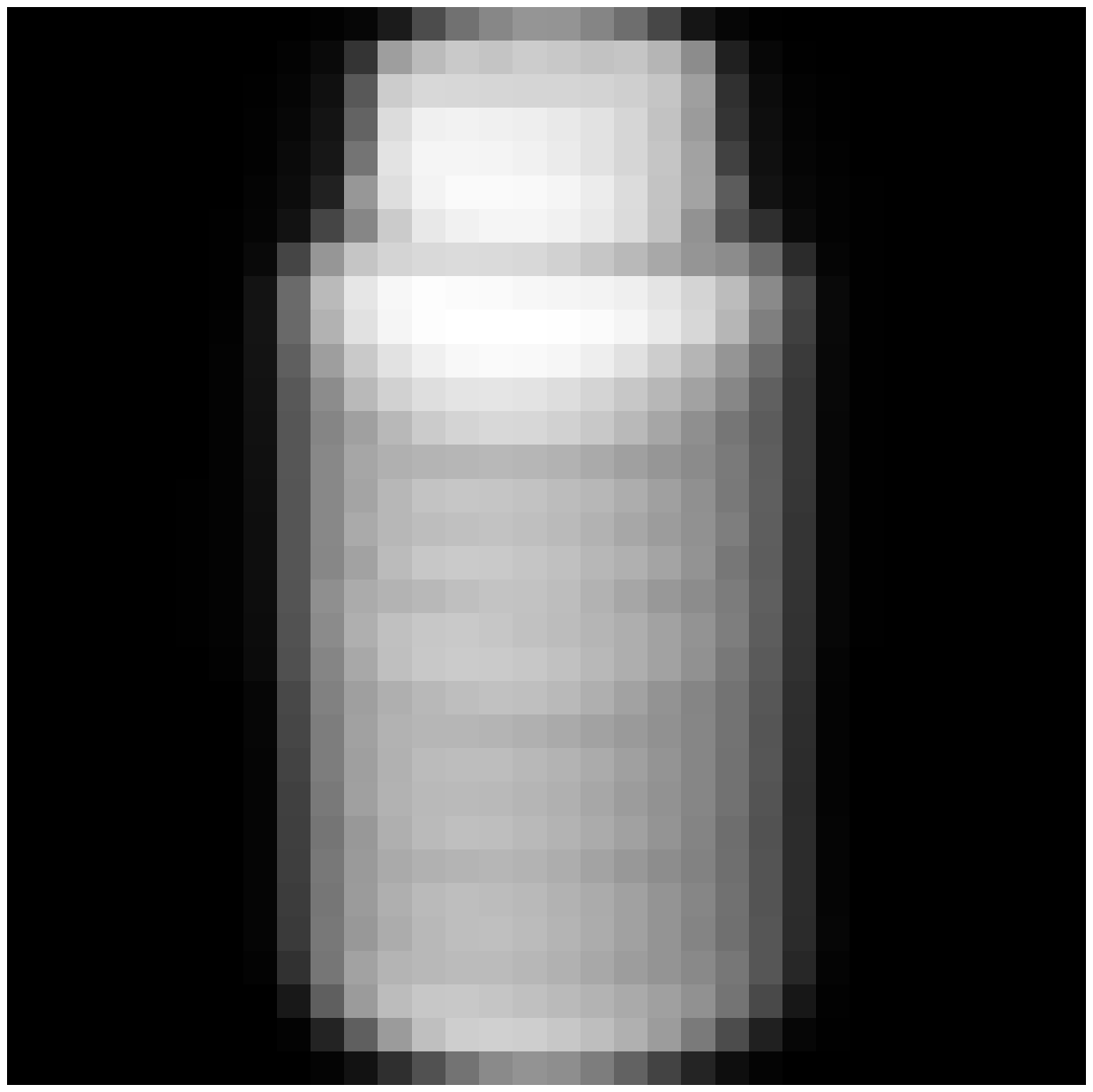}\\[-1ex]
    \rotatebox{90}{\tiny\caja{c}{c}{Lap $K$- \\ modes}} &
    \includegraphics[width=0.048\linewidth,bb=142 226 494 578,clip]{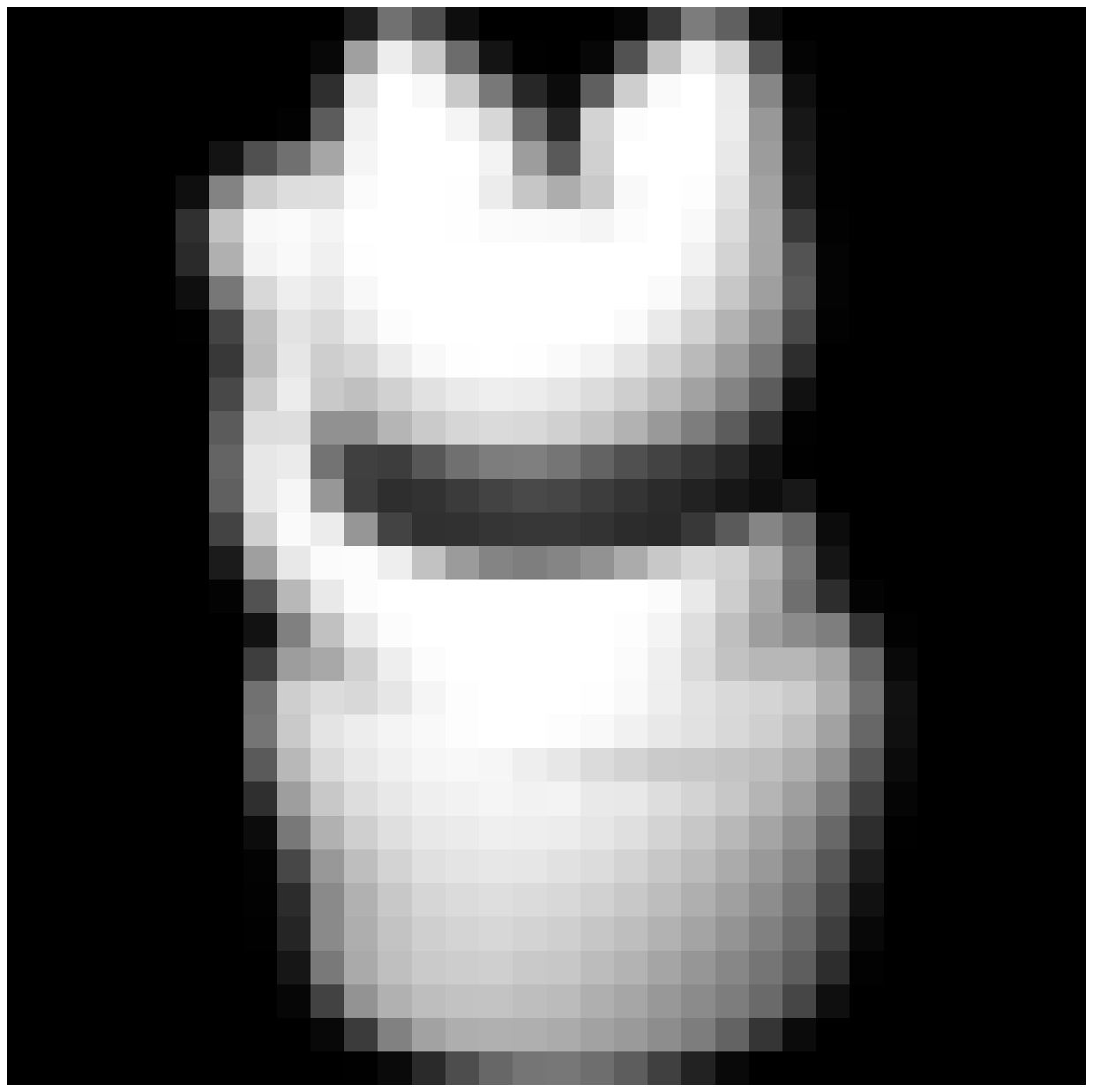} &
    \includegraphics[width=0.048\linewidth,bb=142 226 494 578,clip]{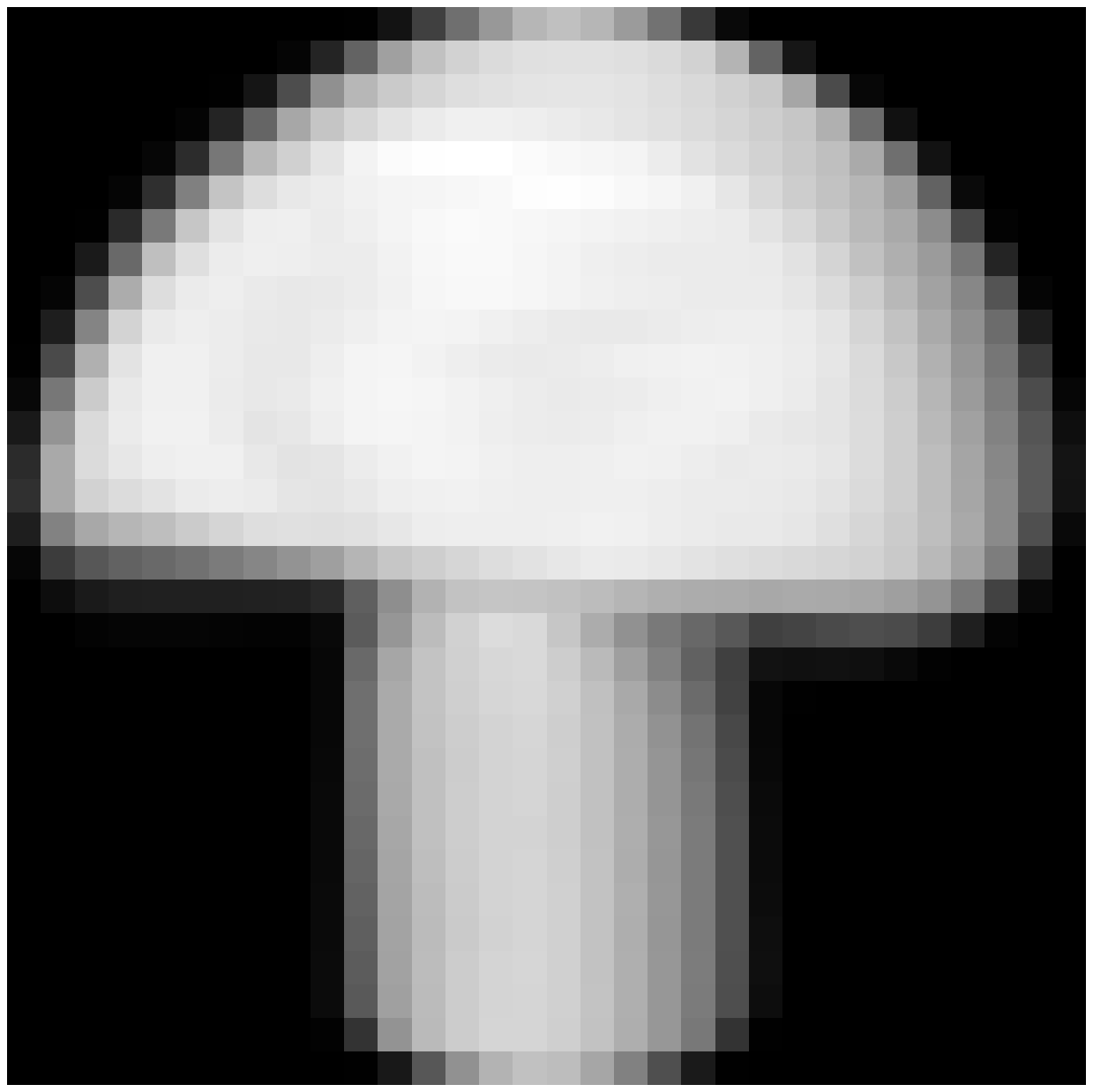} &
    \includegraphics[width=0.048\linewidth,bb=142 226 494 578,clip]{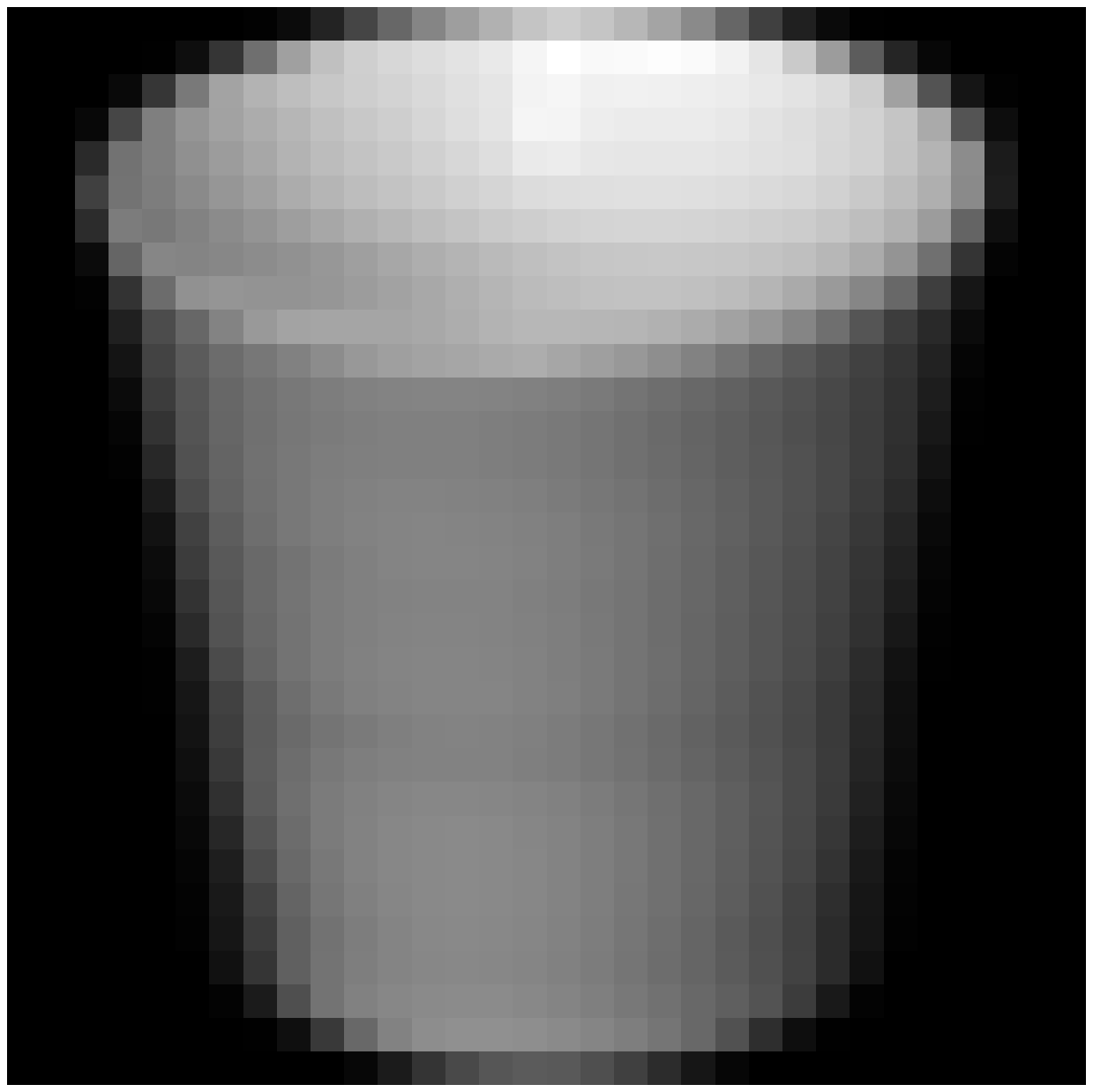} &
    \includegraphics[width=0.048\linewidth,bb=142 226 494 578,clip]{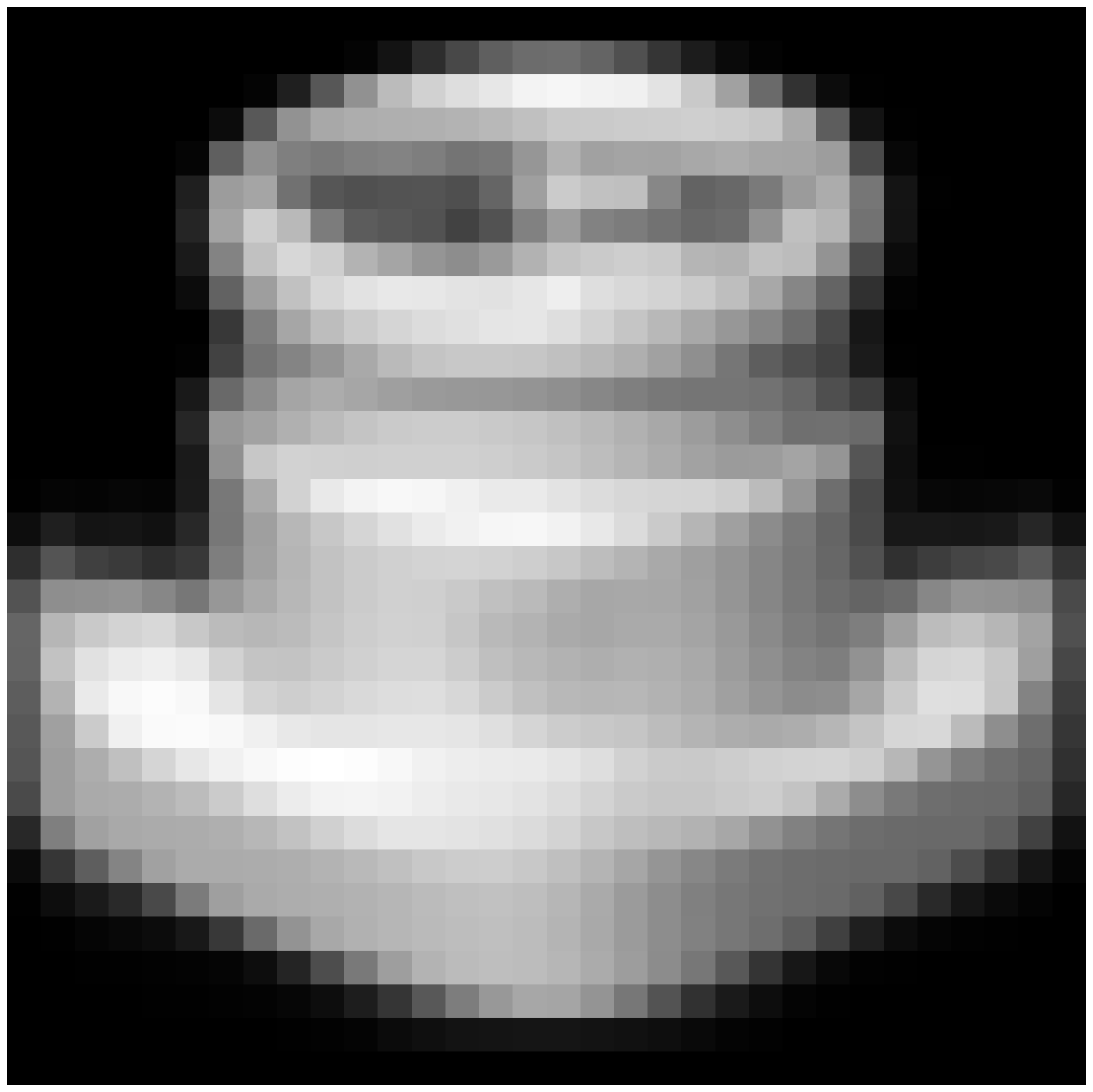} &
    \includegraphics[width=0.048\linewidth,bb=142 226 494 578,clip]{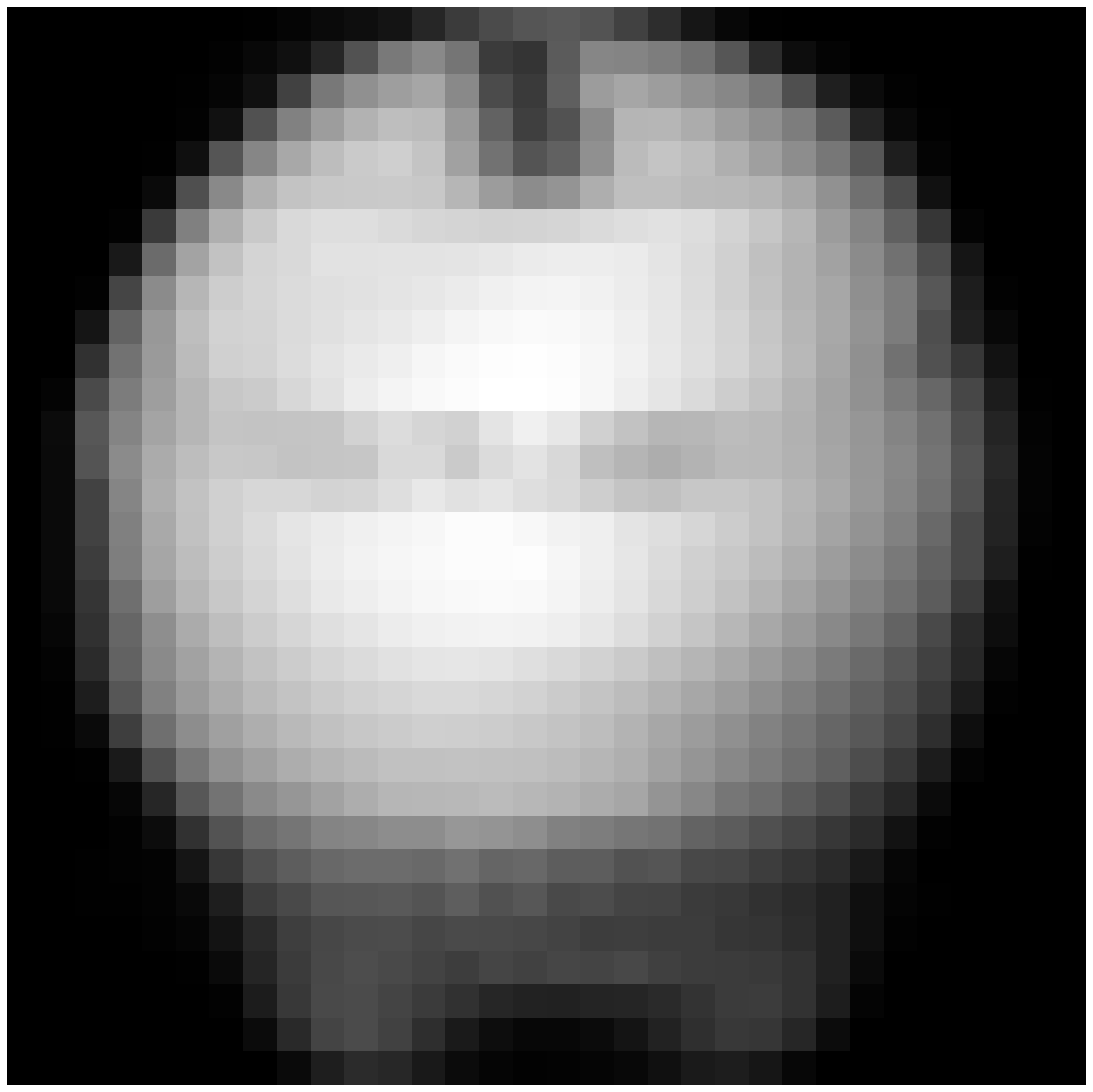} &
    \includegraphics[width=0.048\linewidth,bb=142 226 494 578,clip]{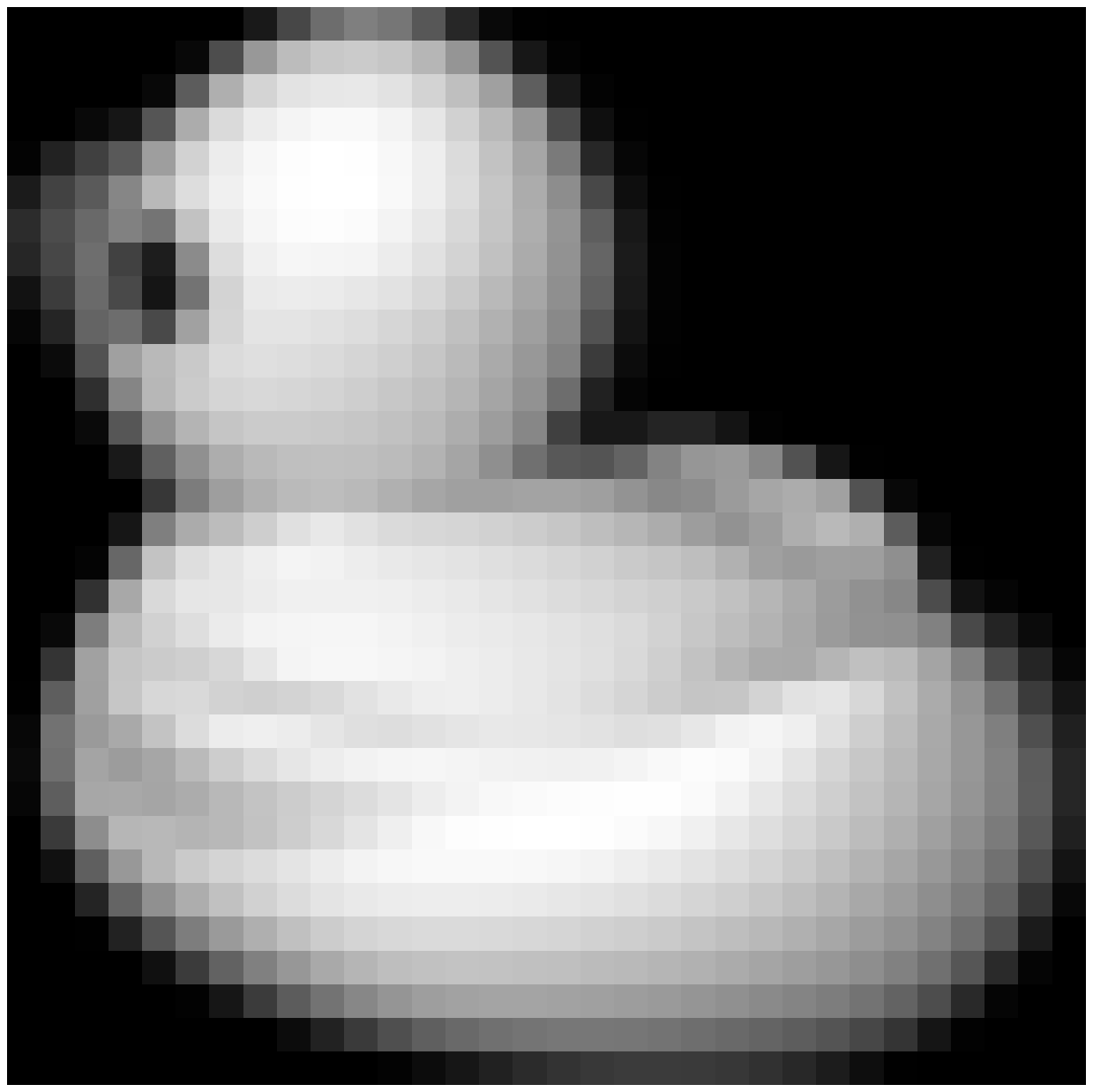} &
    \includegraphics[width=0.048\linewidth,bb=142 226 494 578,clip]{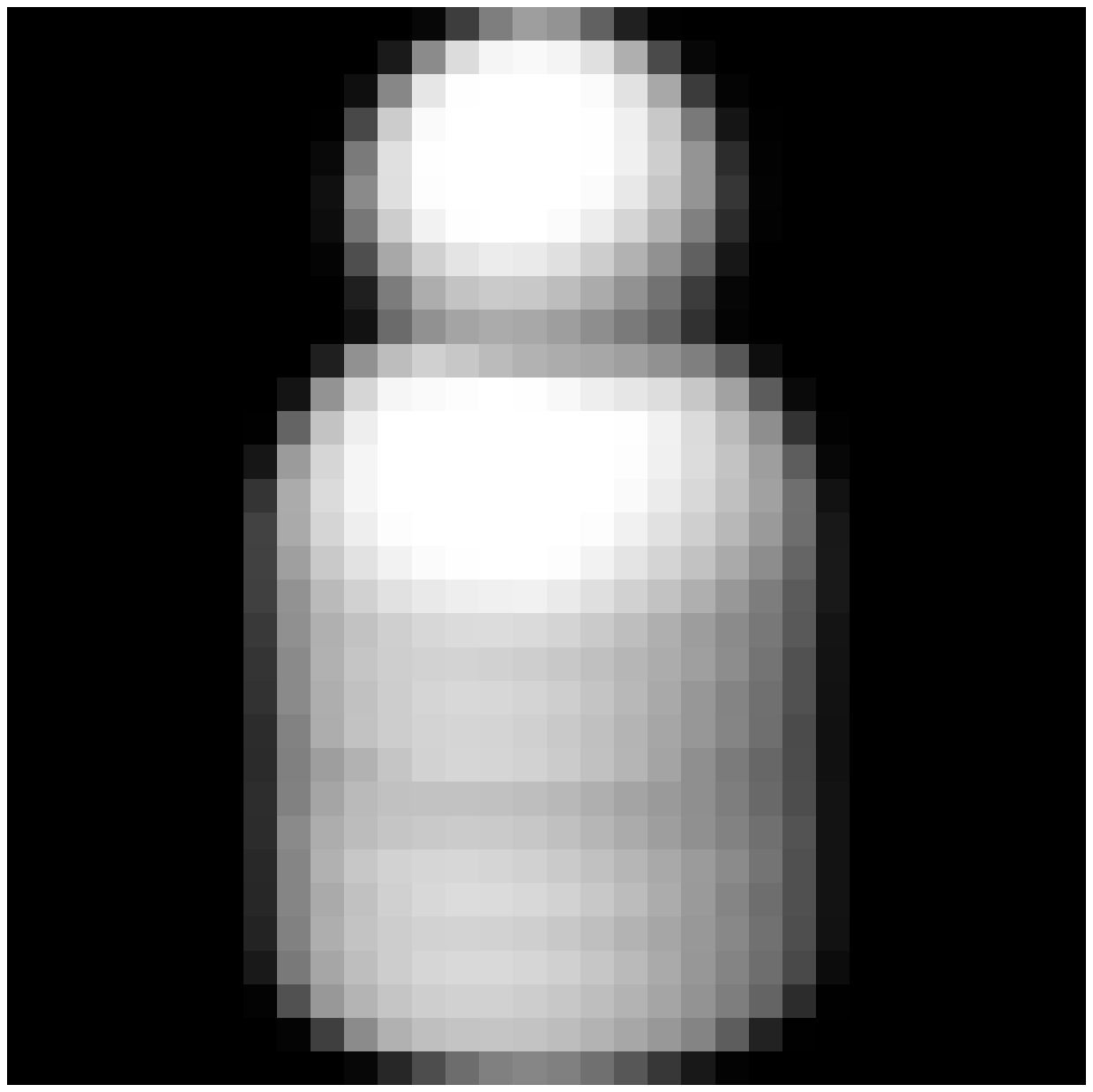} &
    \includegraphics[width=0.048\linewidth,bb=142 226 494 578,clip]{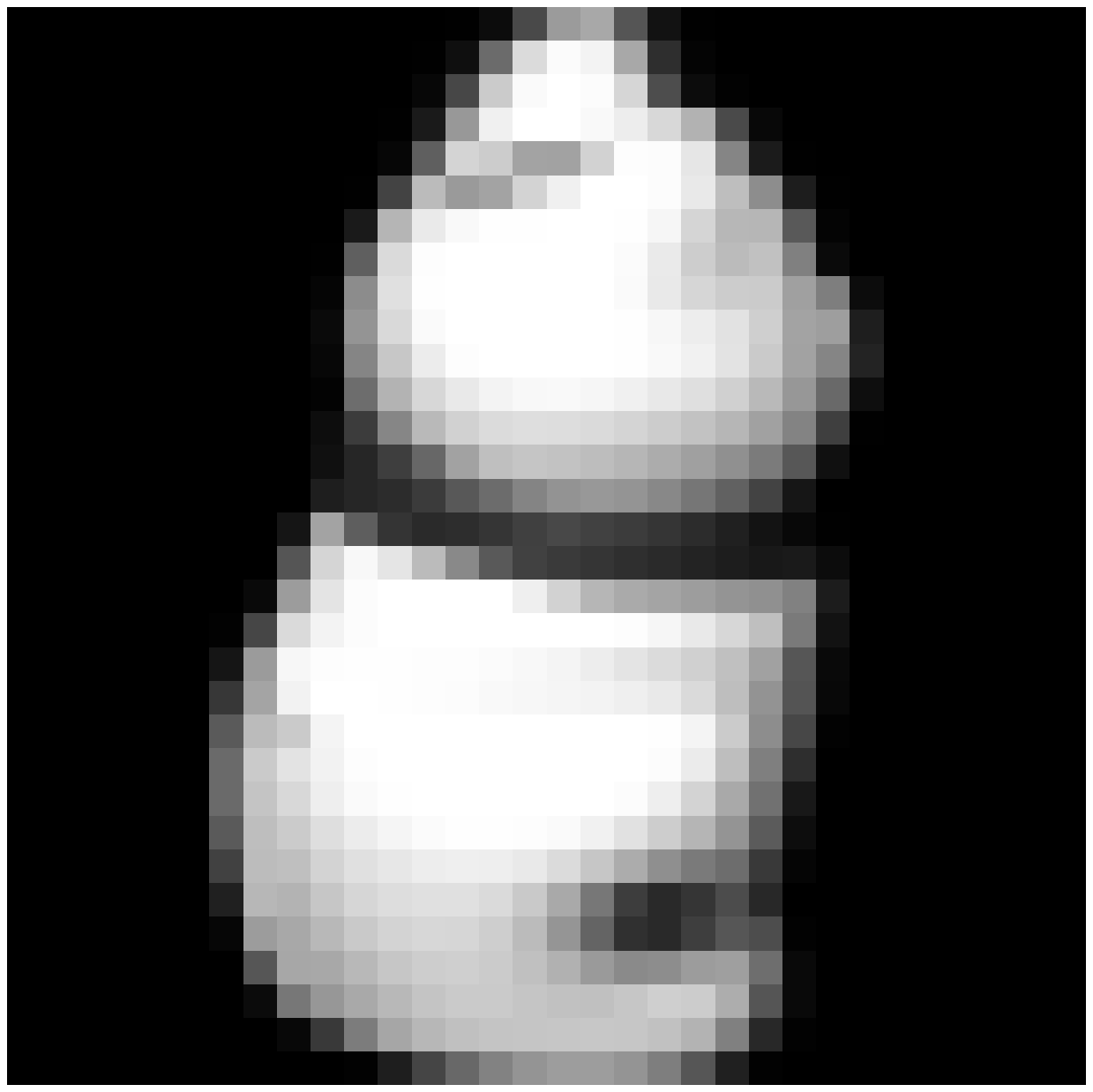} &
    \includegraphics[width=0.048\linewidth,bb=142 226 494 578,clip]{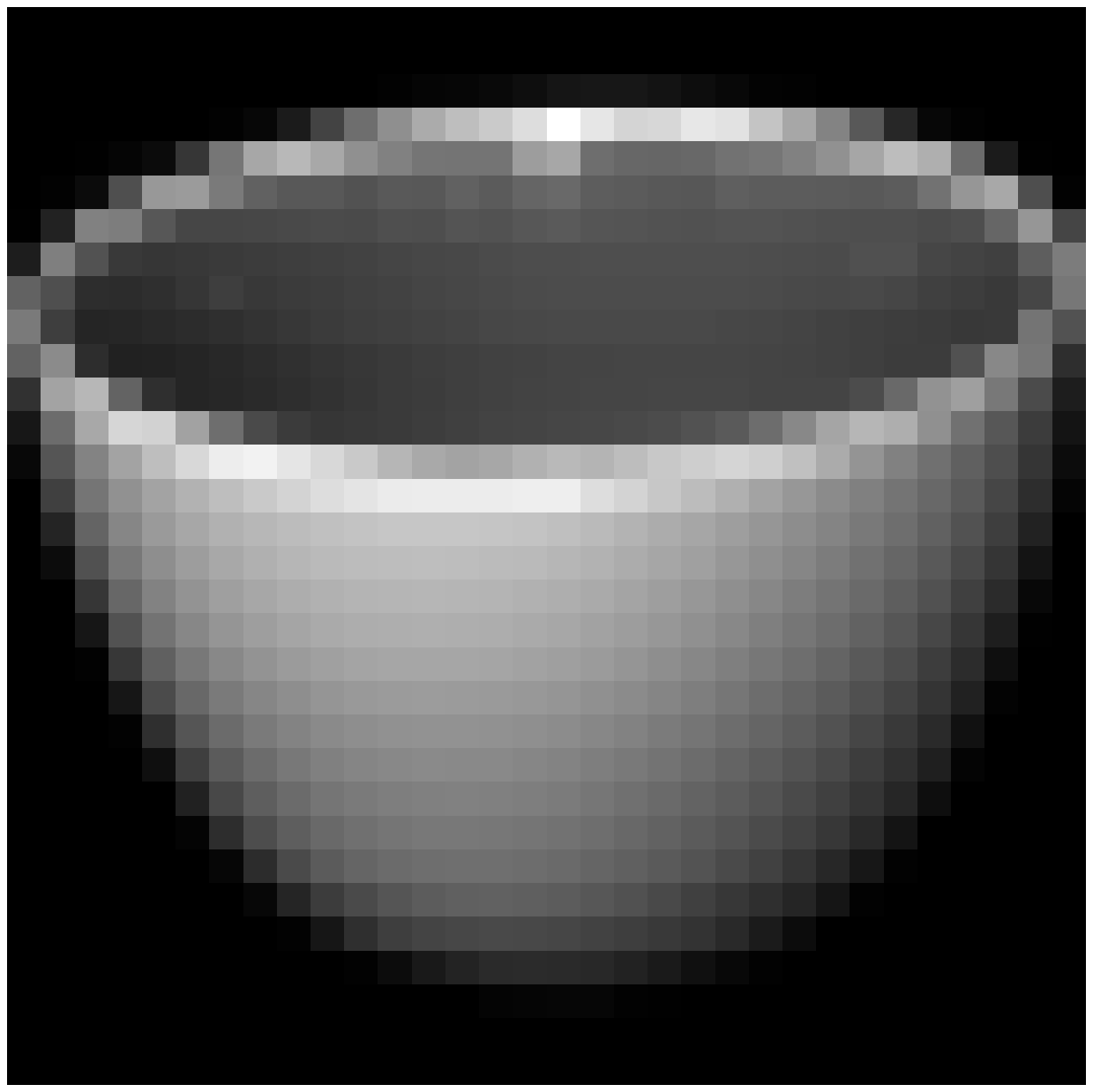} &
    \includegraphics[width=0.048\linewidth,bb=142 226 494 578,clip]{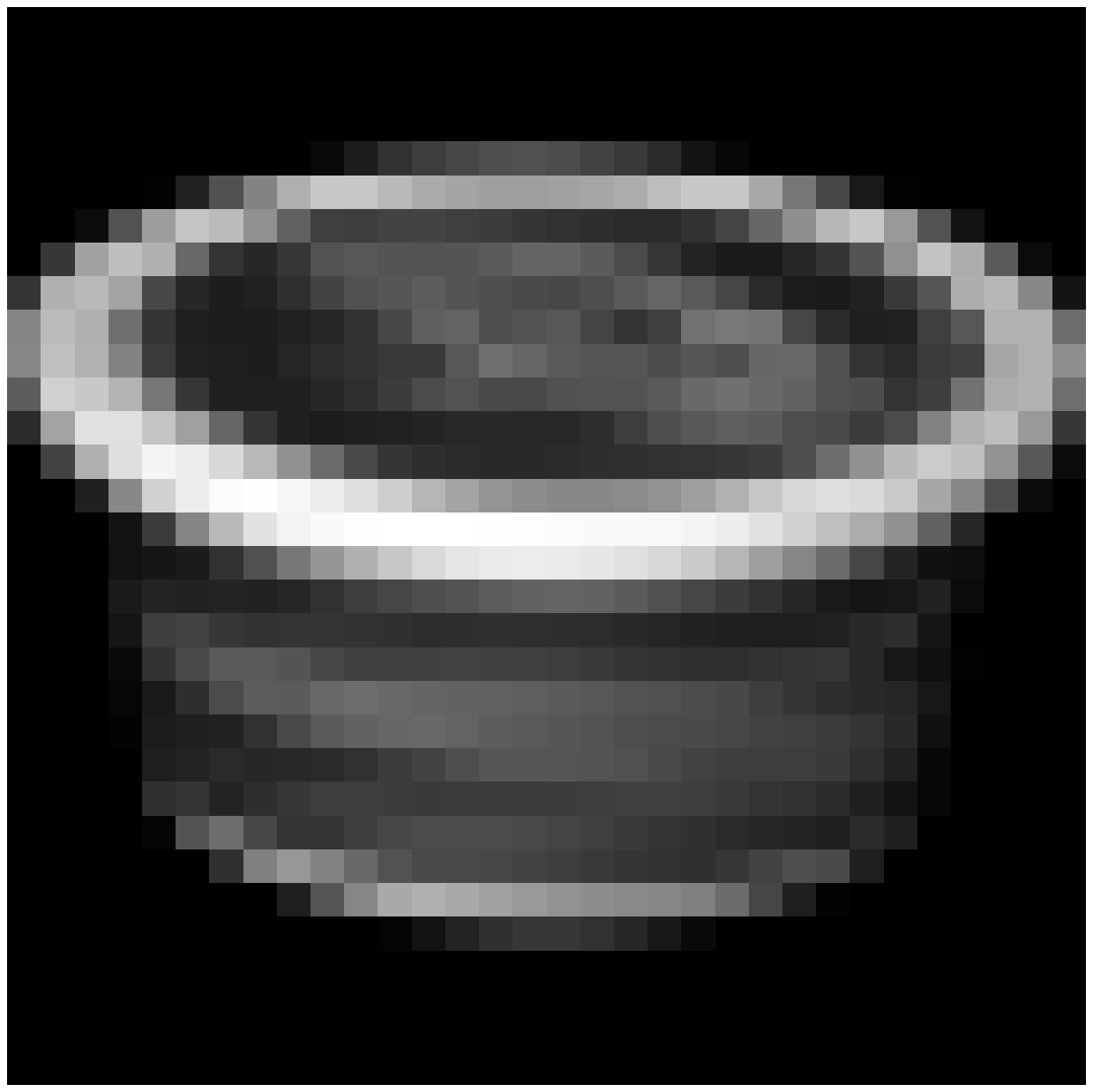} &
    \includegraphics[width=0.048\linewidth,bb=142 226 494 578,clip]{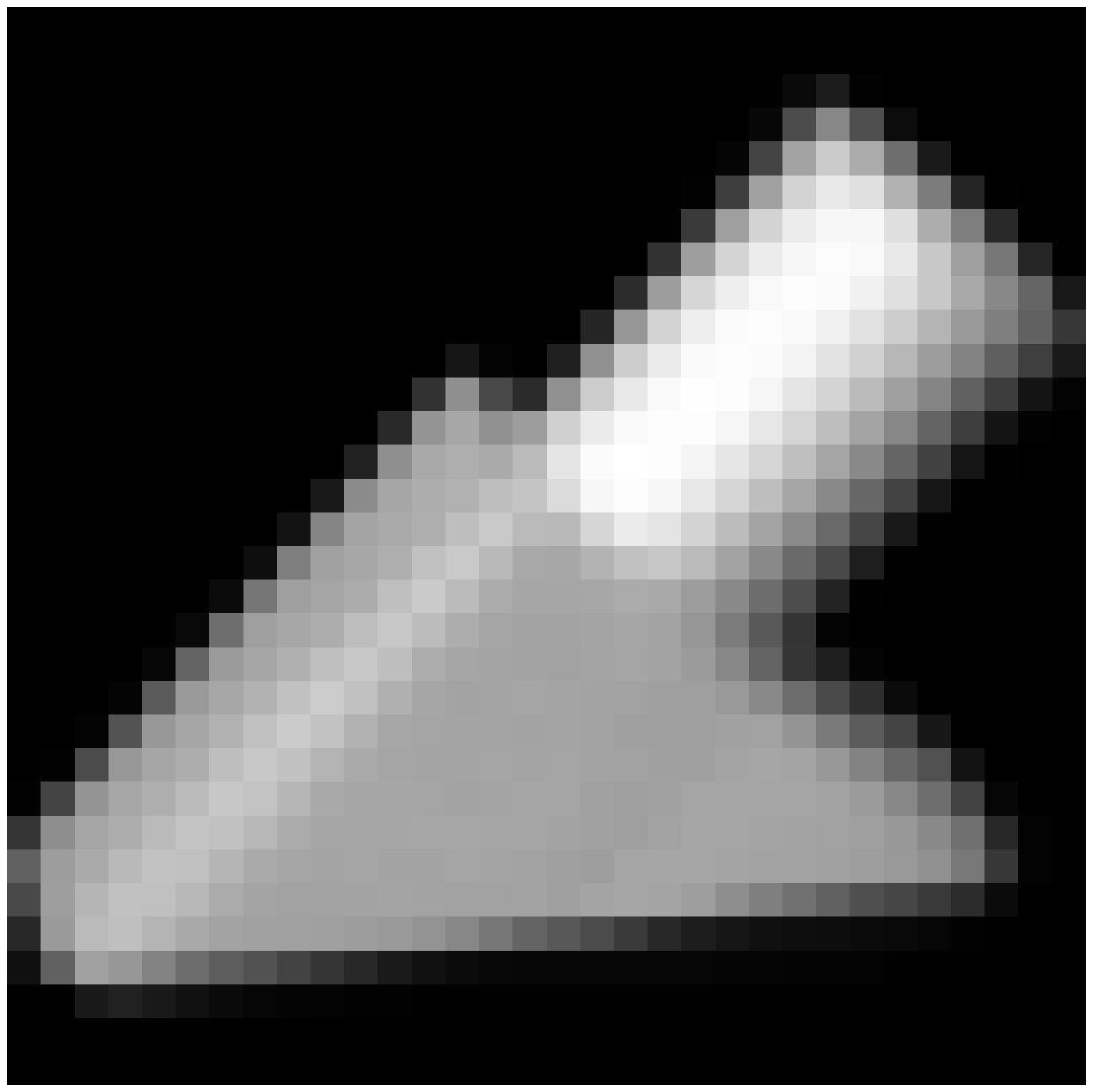} &
    \includegraphics[width=0.048\linewidth,bb=142 226 494 578,clip]{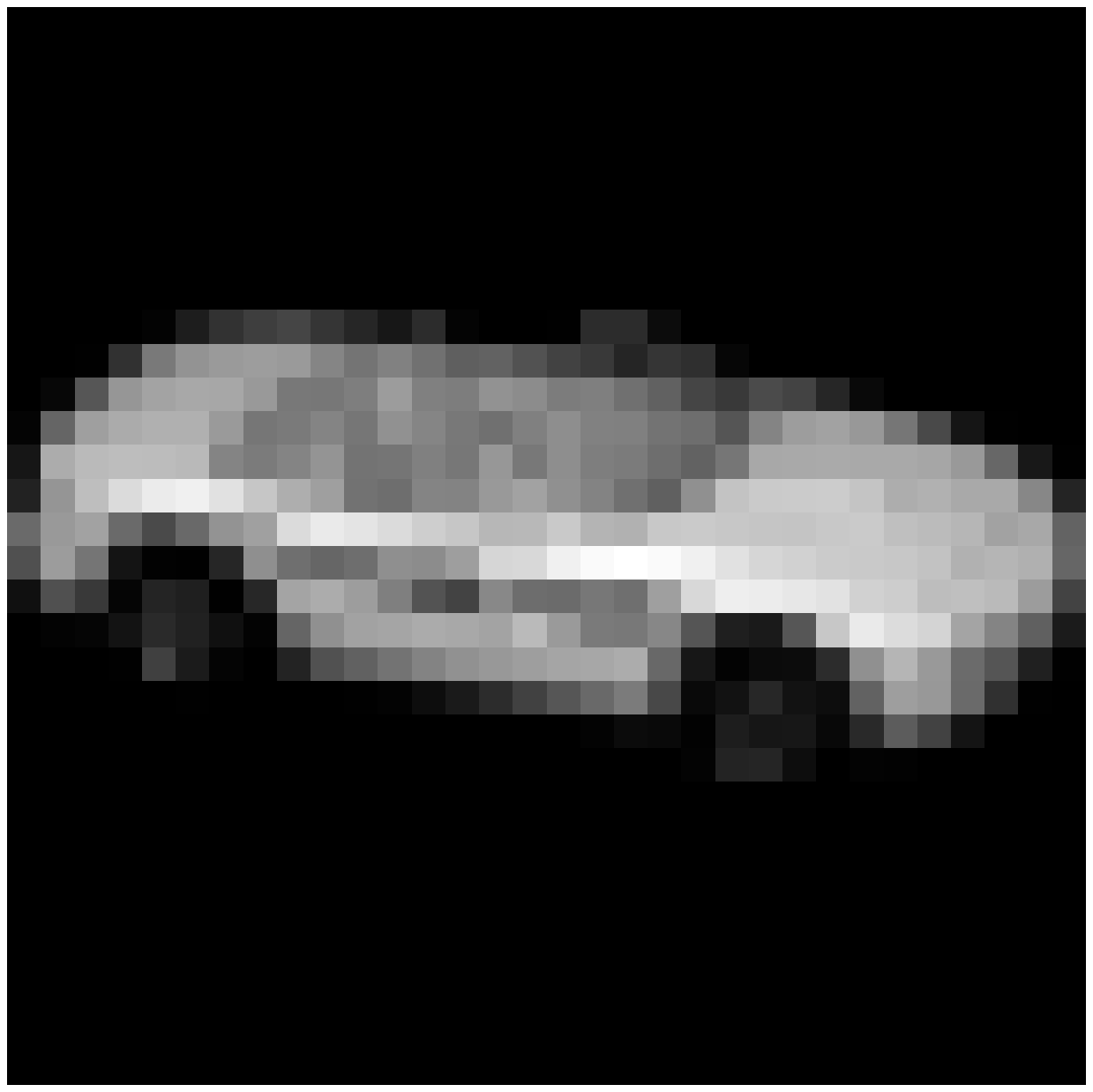} &
    \includegraphics[width=0.048\linewidth,bb=142 226 494 578,clip]{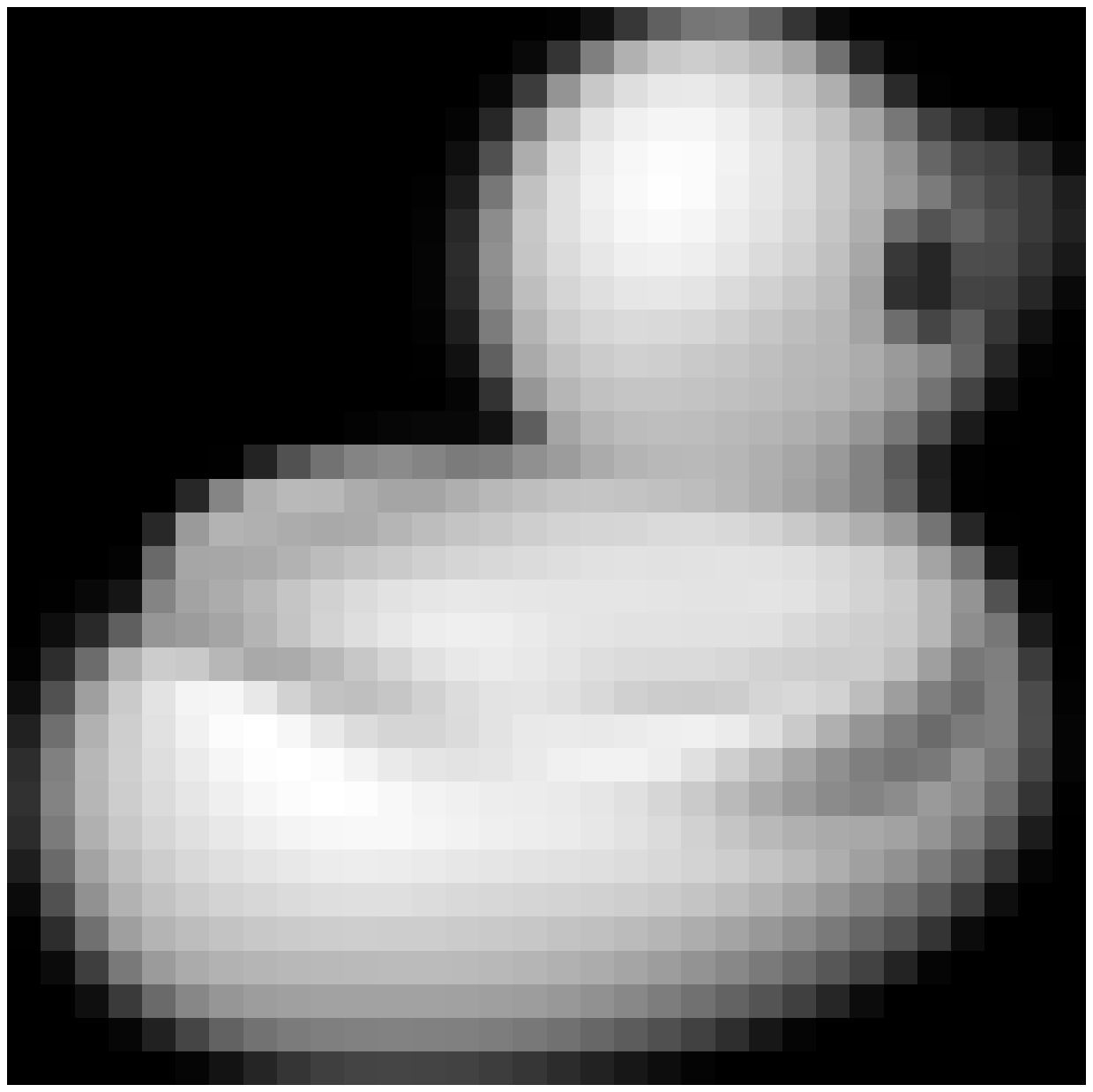} &
    \includegraphics[width=0.048\linewidth,bb=142 226 494 578,clip]{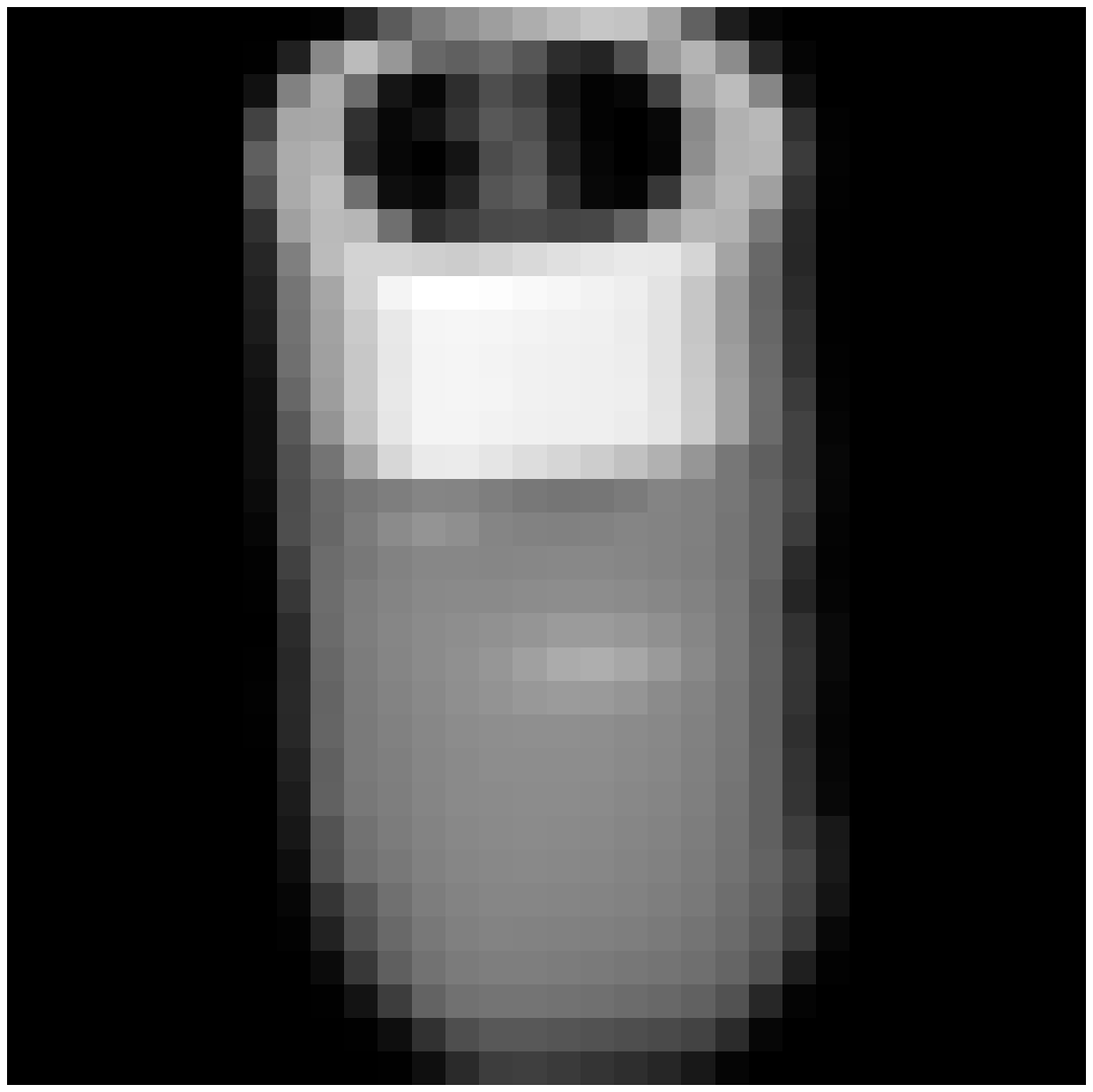} &
    \includegraphics[width=0.048\linewidth,bb=142 226 494 578,clip]{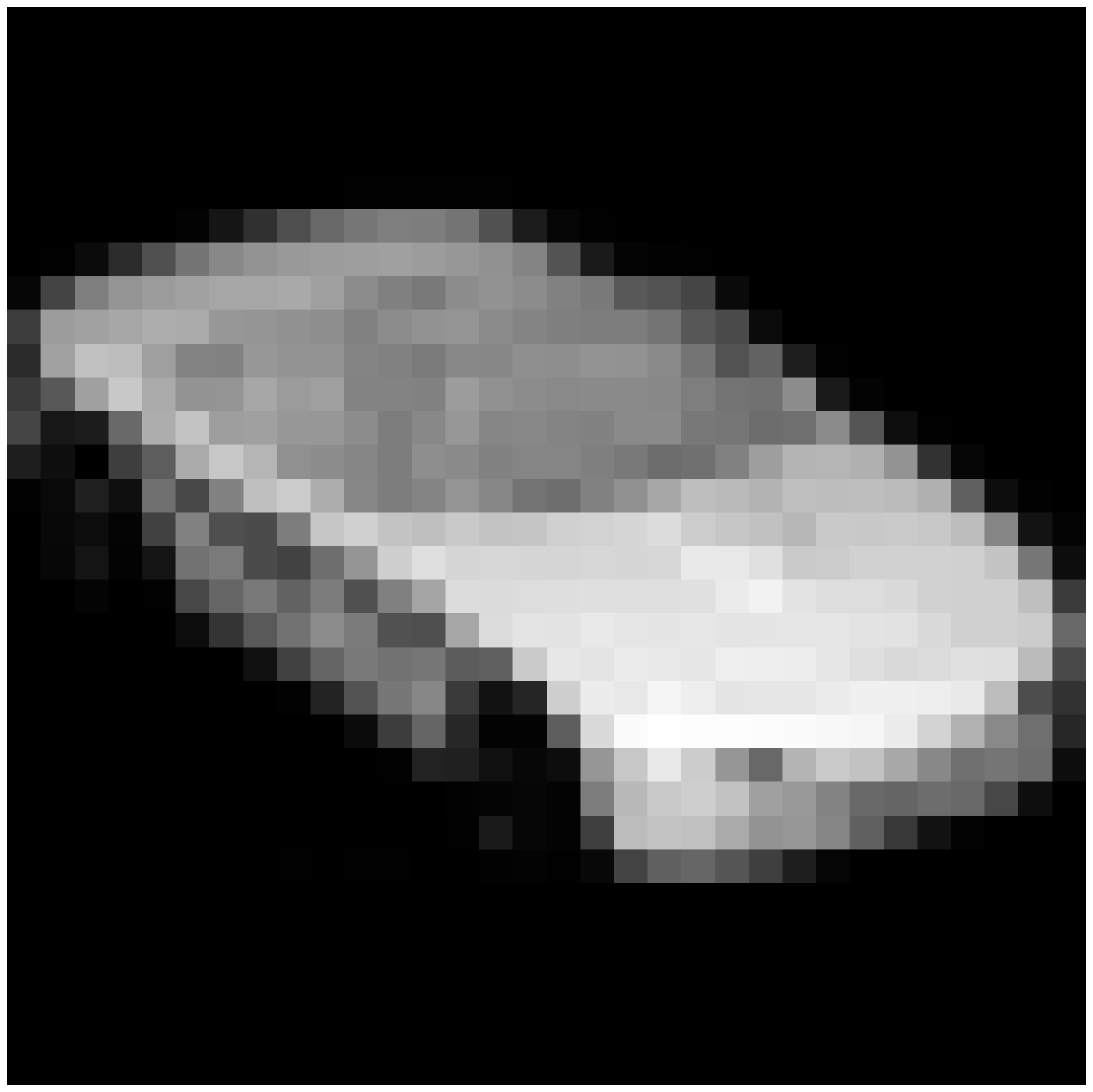} &
    \includegraphics[width=0.048\linewidth,bb=142 226 494 578,clip]{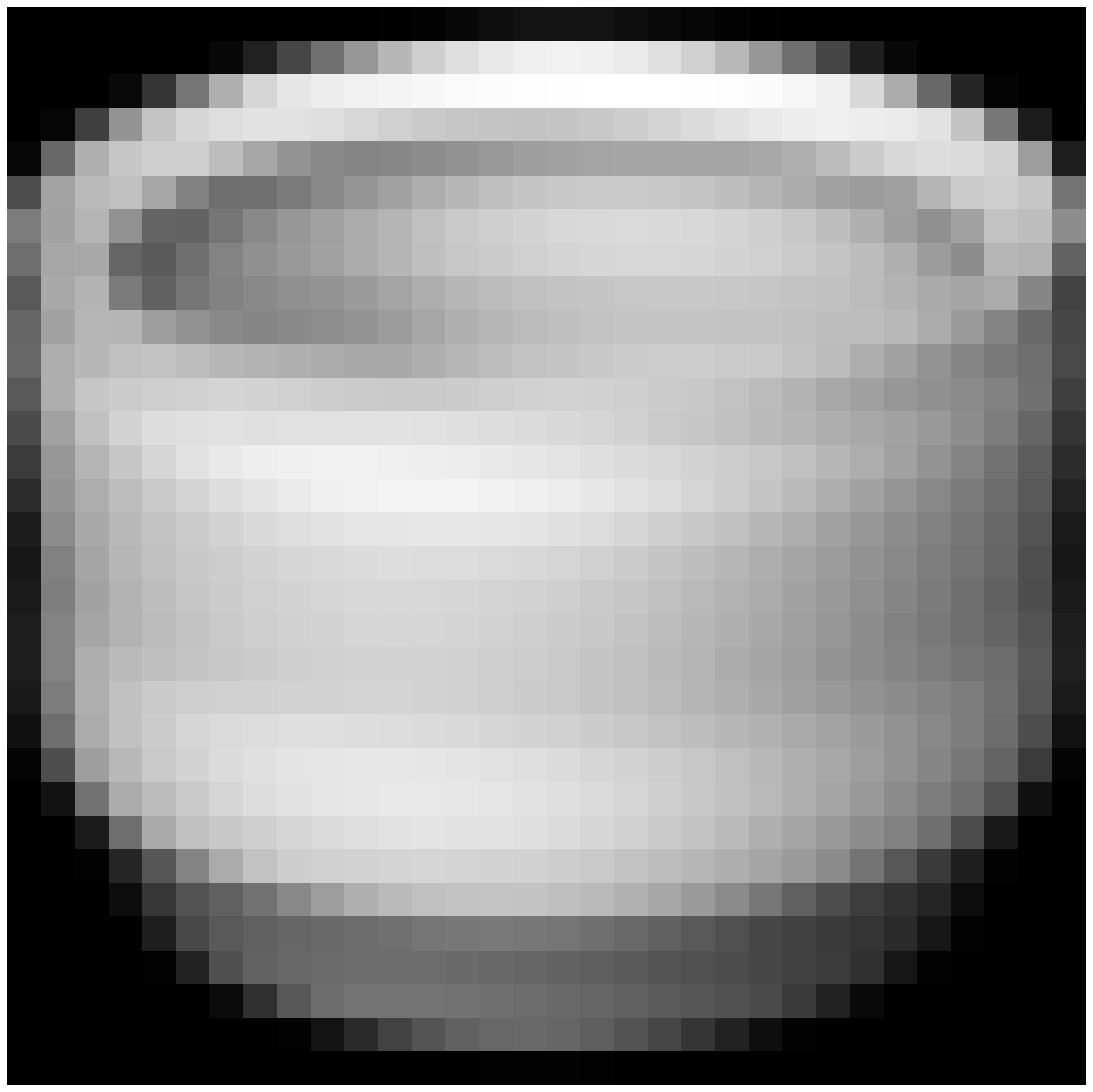} &
    \includegraphics[width=0.048\linewidth,bb=142 226 494 578,clip]{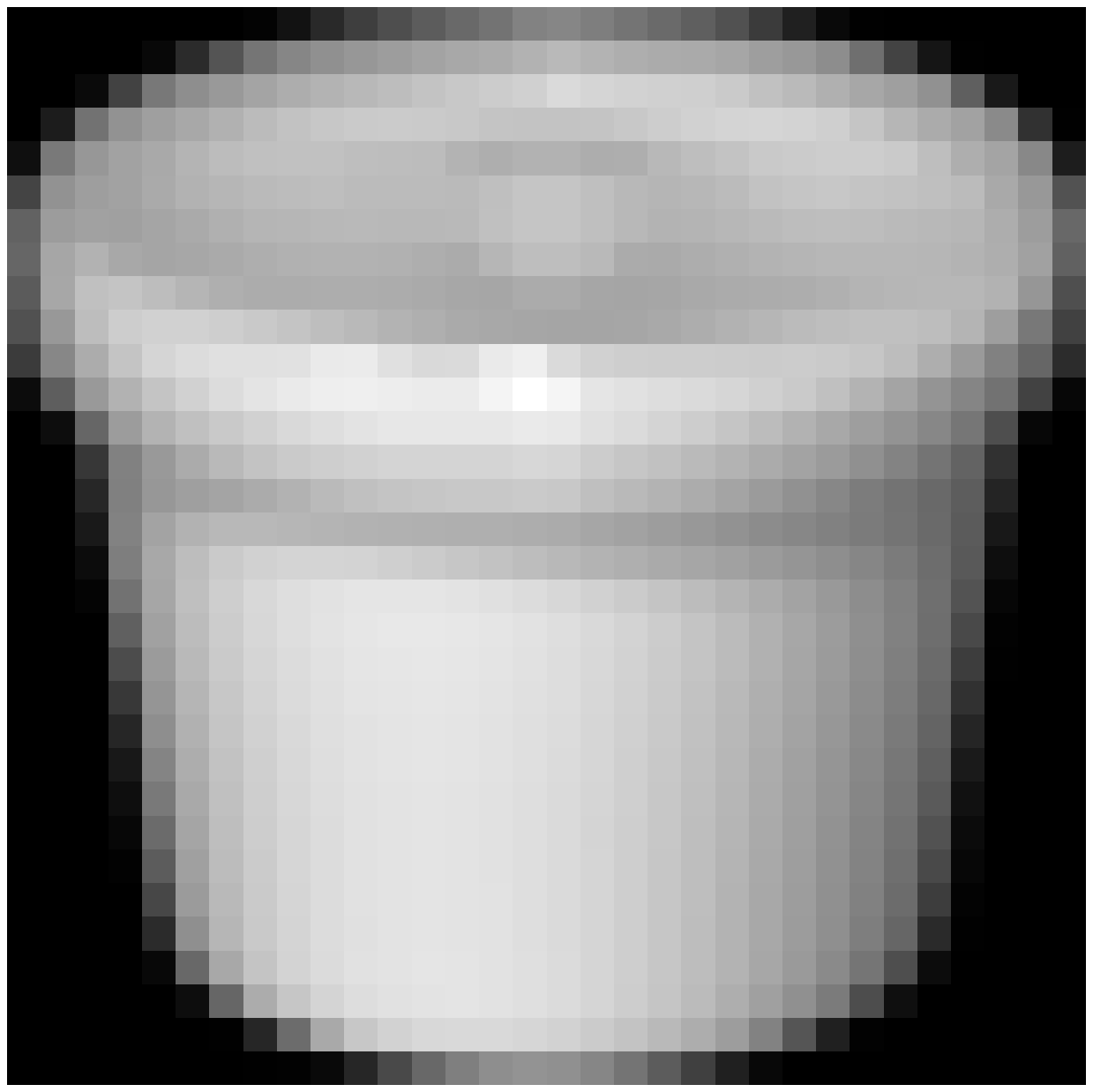} &
    \includegraphics[width=0.048\linewidth,bb=142 226 494 578,clip]{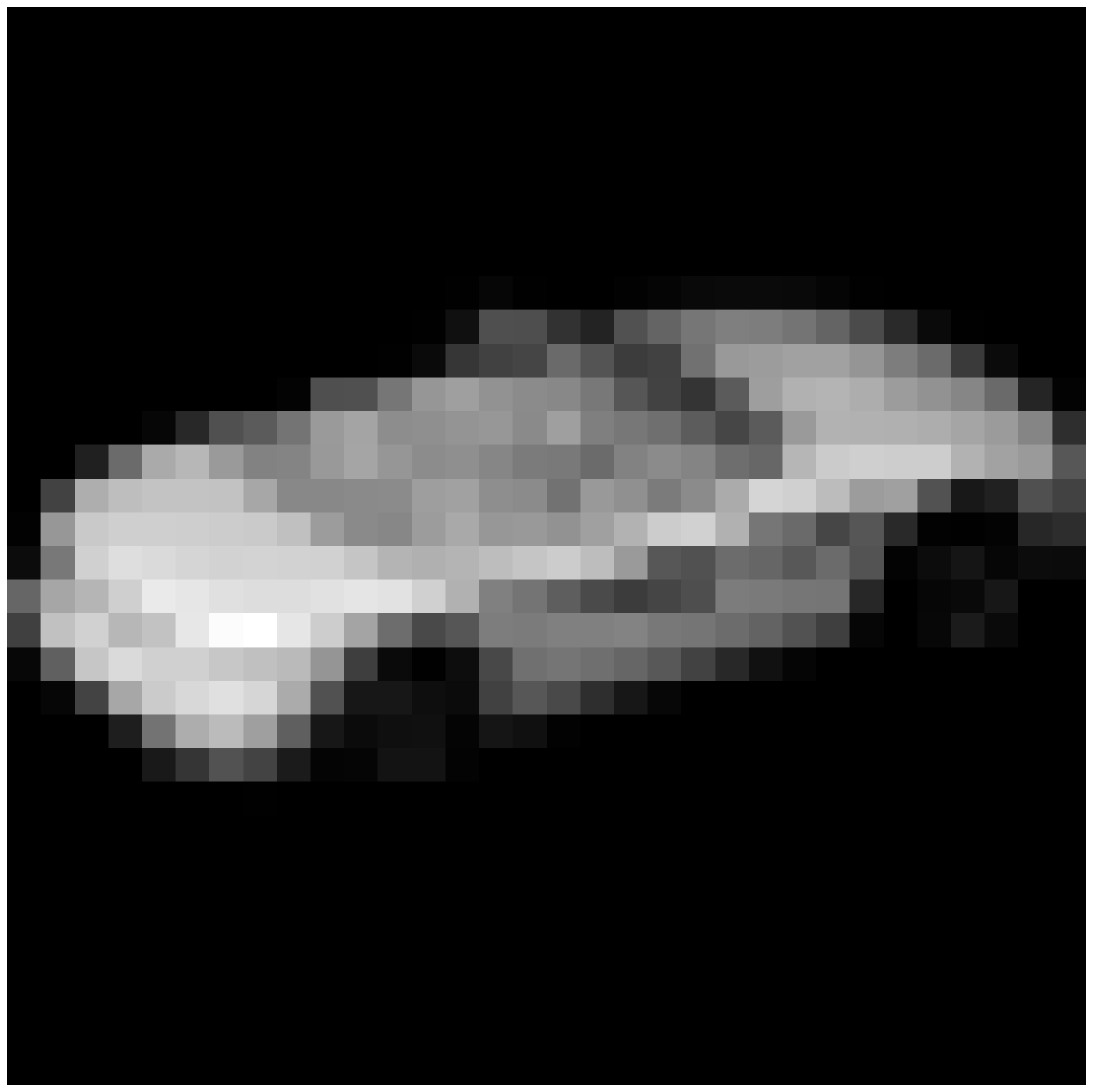} &
    \includegraphics[width=0.048\linewidth,bb=142 226 494 578,clip]{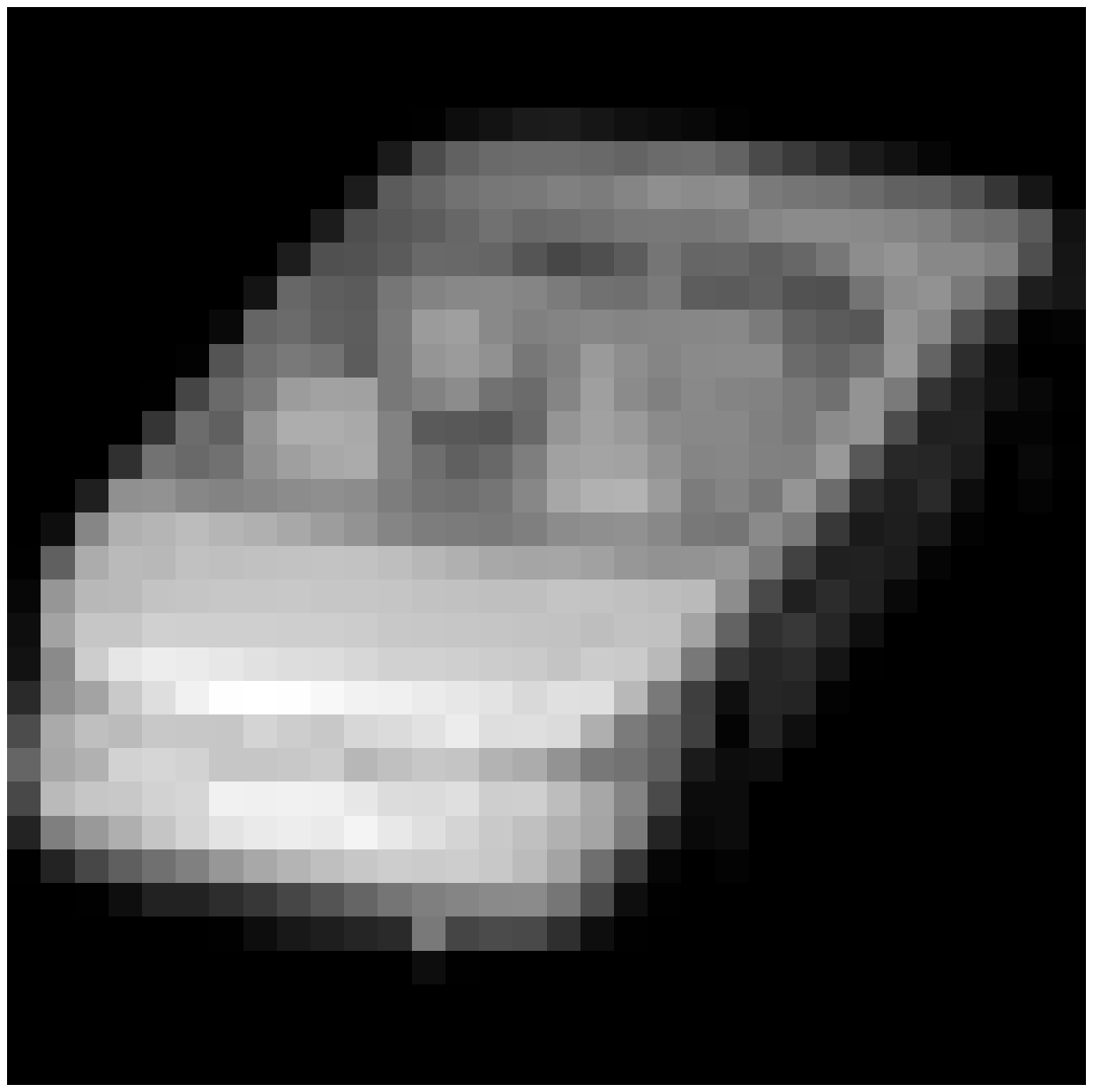} &
    \includegraphics[width=0.048\linewidth,bb=142 226 494 578,clip]{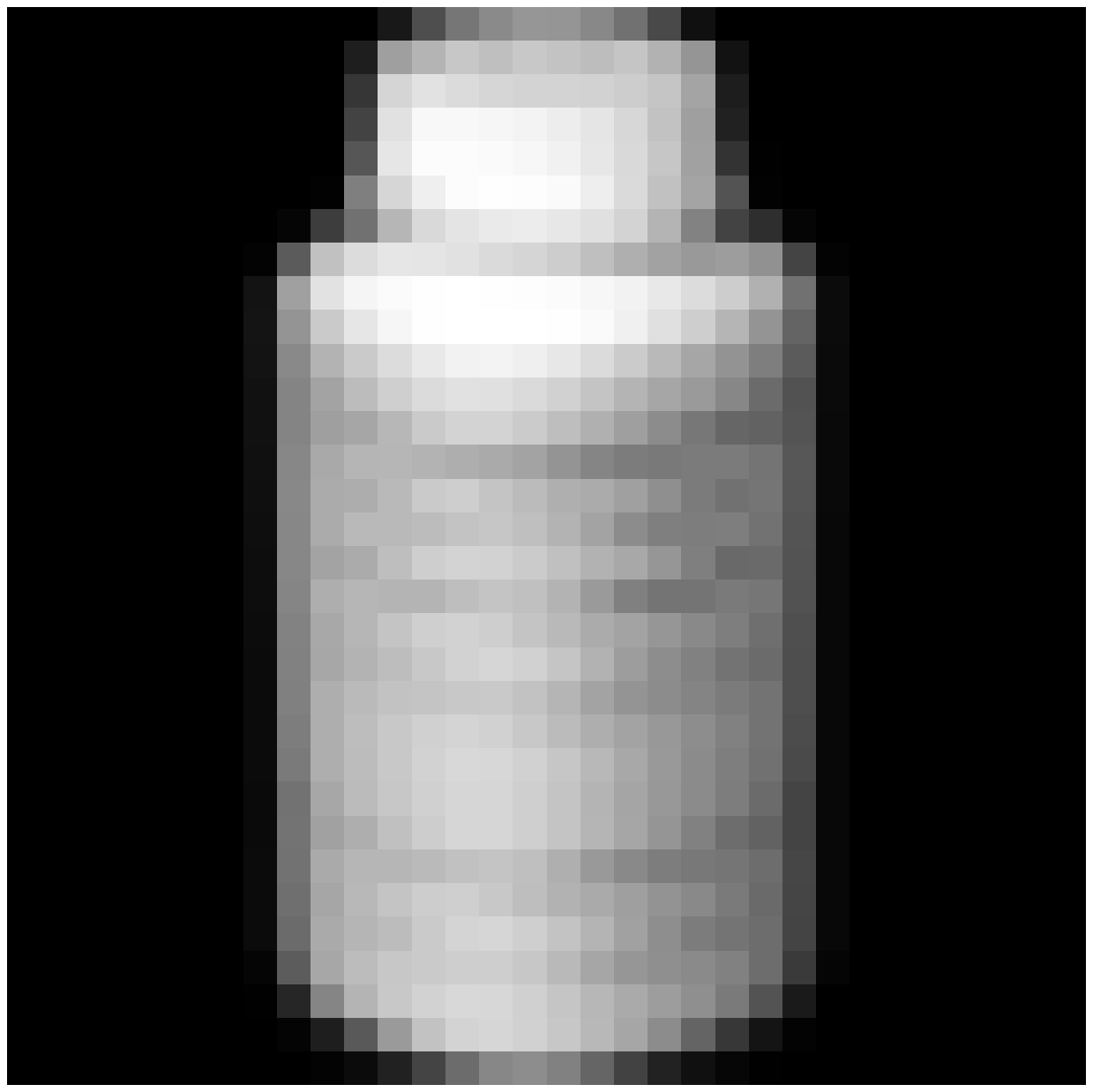} \\[-1ex]
    \rotatebox{90}{\tiny\caja{c}{c}{Lap $K$- \\ modes \\ homot.}} &
    \includegraphics[width=0.048\linewidth,bb=142 226 494 578,clip]{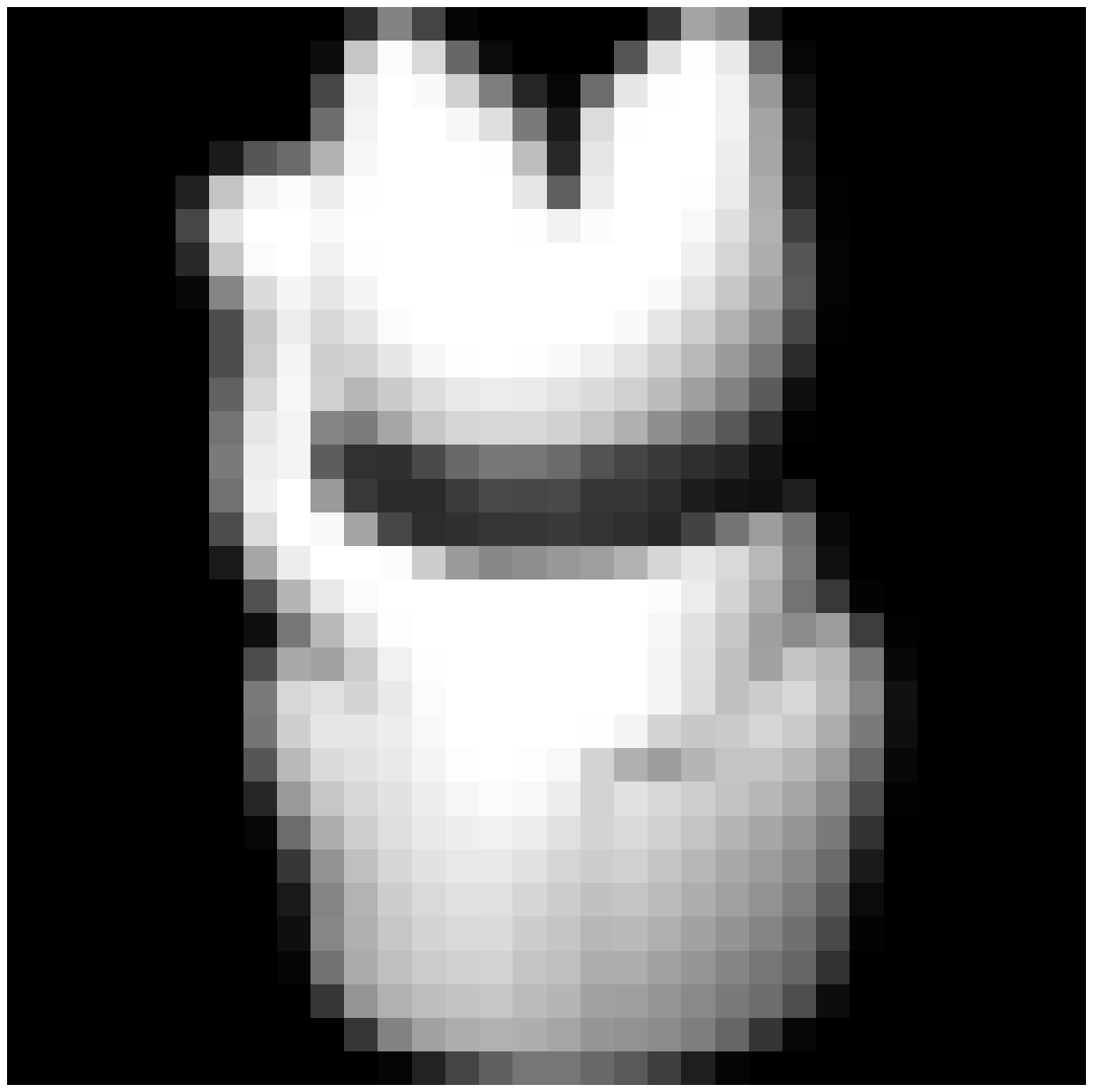} &
    \includegraphics[width=0.048\linewidth,bb=142 226 494 578,clip]{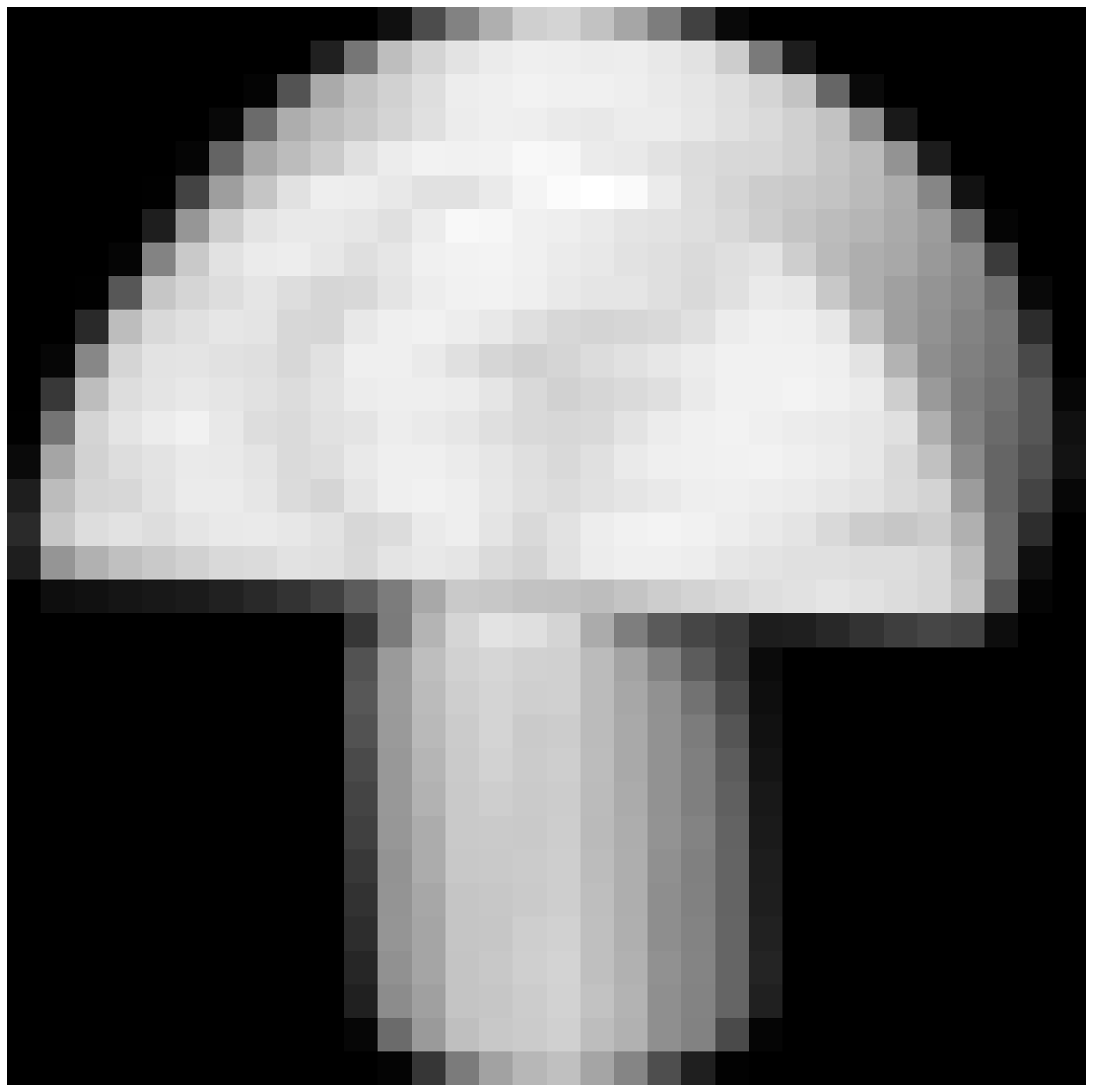} &
    \includegraphics[width=0.048\linewidth,bb=142 226 494 578,clip]{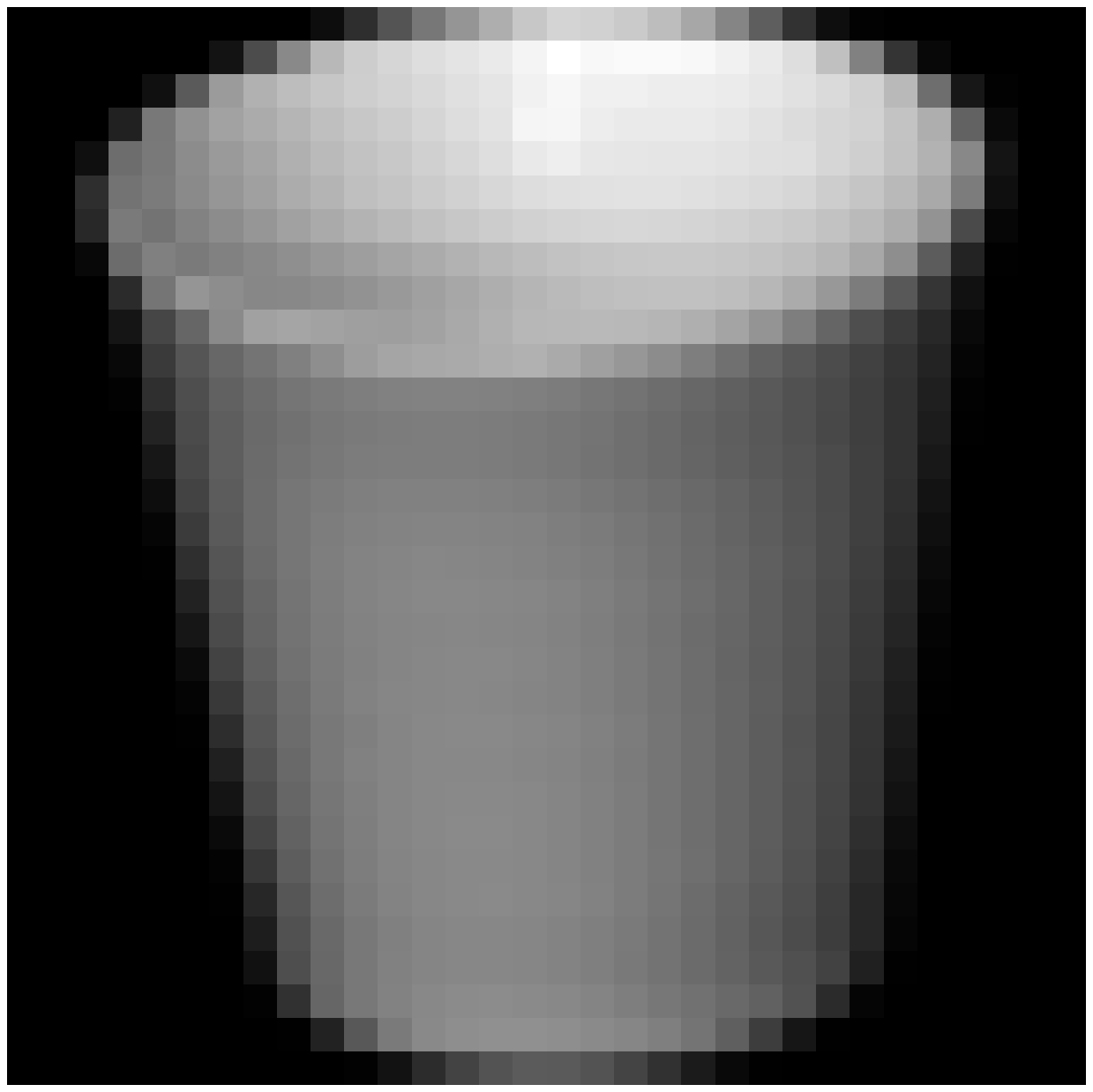} &
    \includegraphics[width=0.048\linewidth,bb=142 226 494 578,clip]{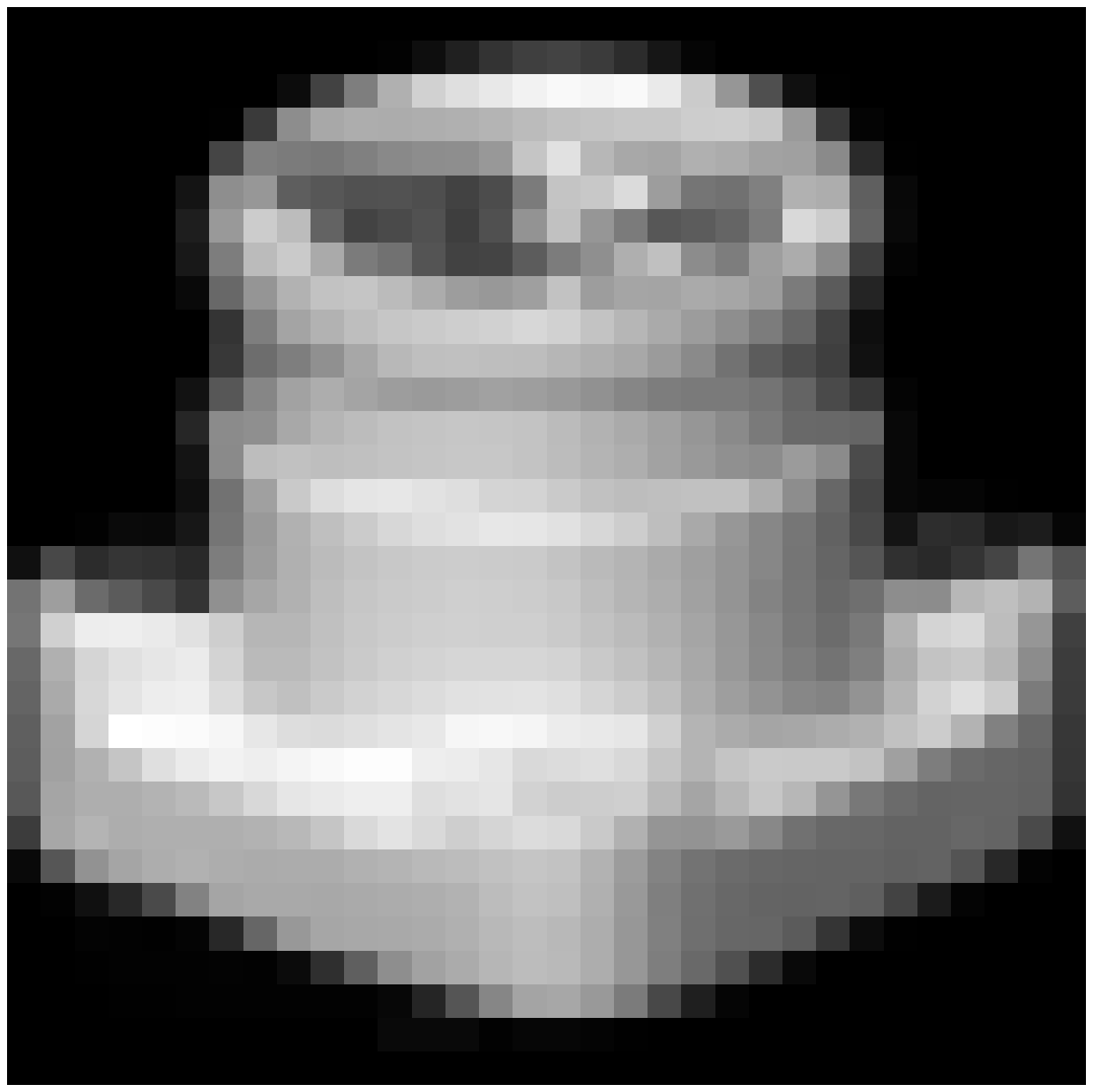} &
    \includegraphics[width=0.048\linewidth,bb=142 226 494 578,clip]{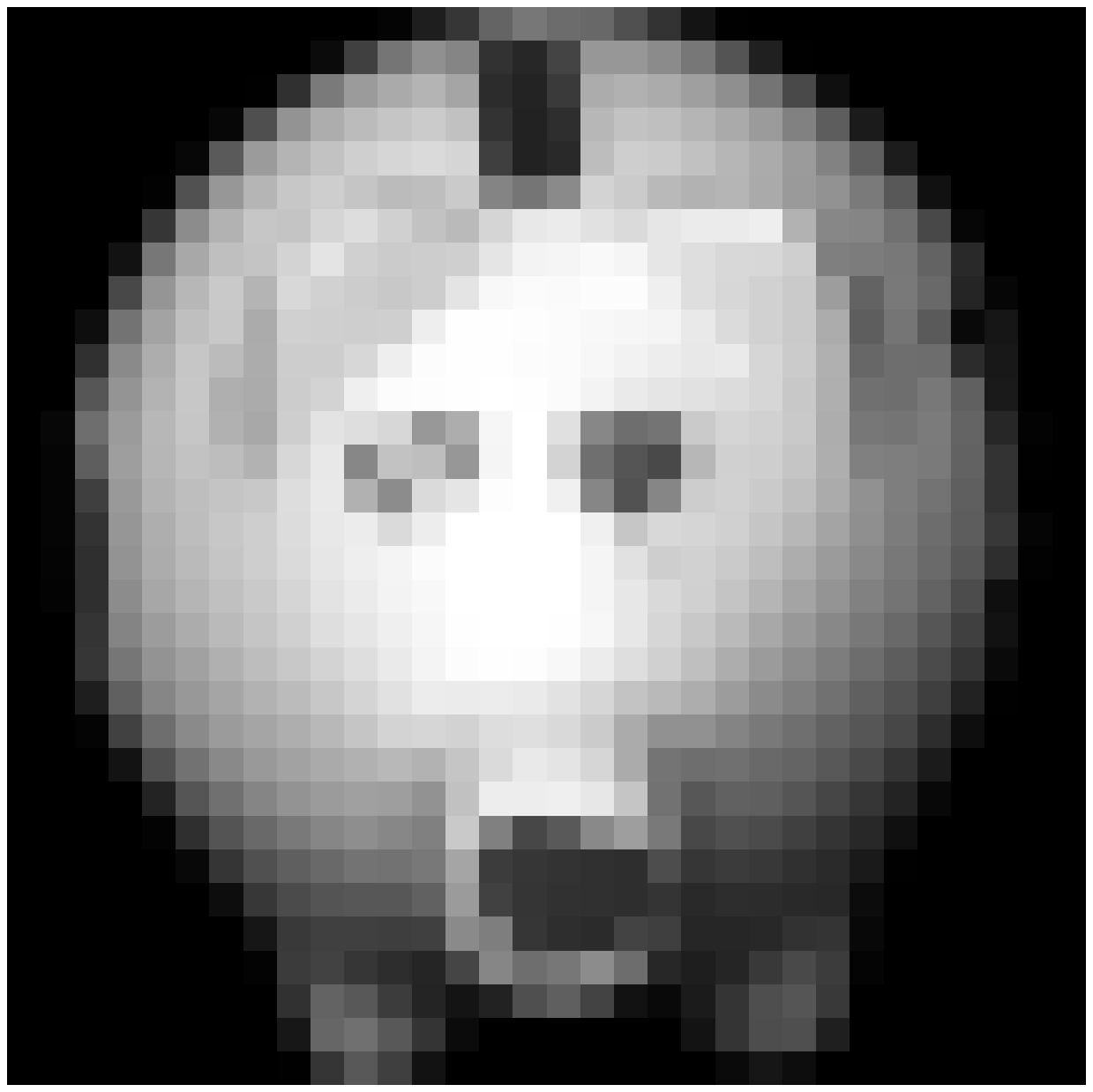} &
    \includegraphics[width=0.048\linewidth,bb=142 226 494 578,clip]{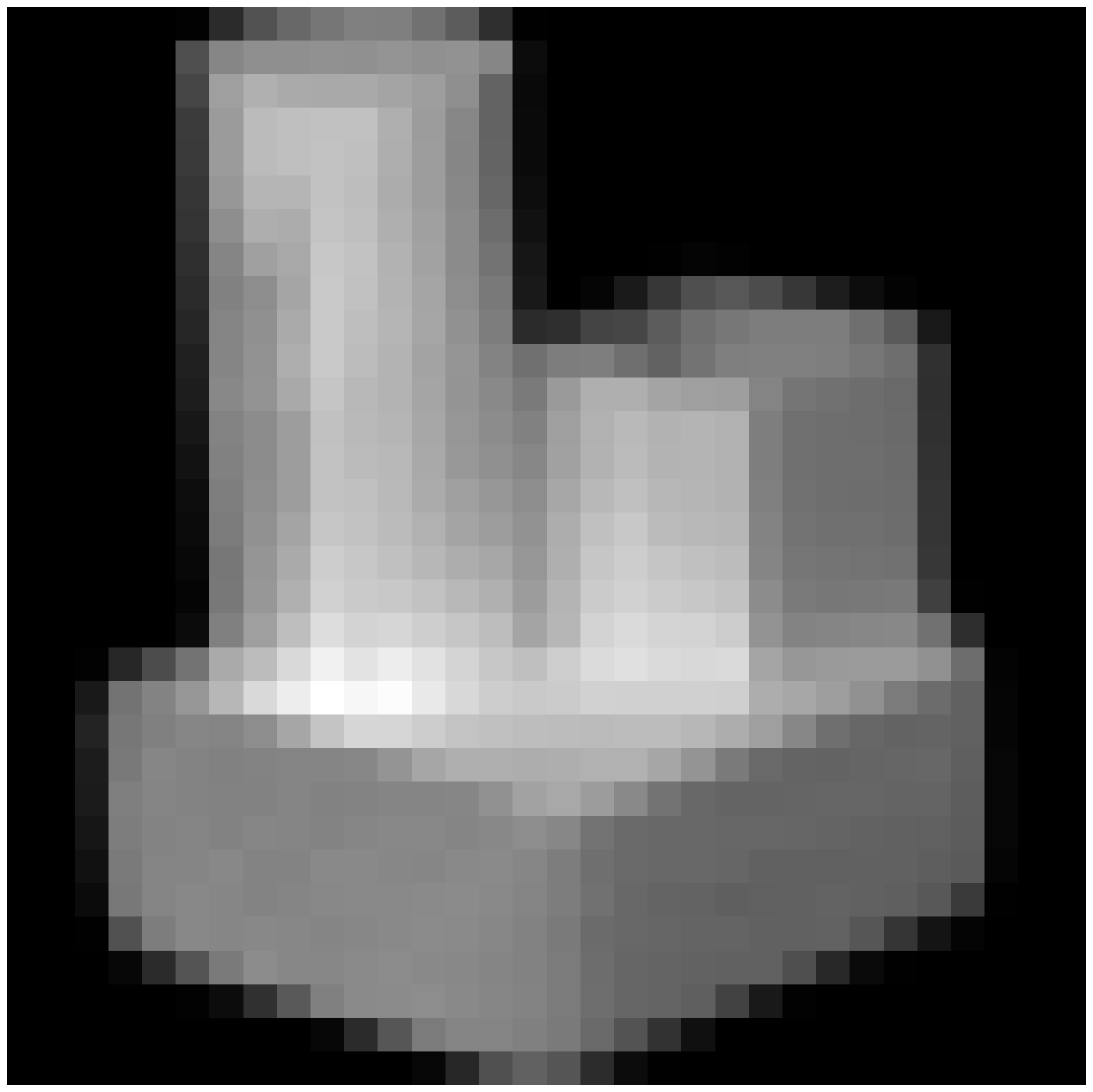} &
    \includegraphics[width=0.048\linewidth,bb=142 226 494 578,clip]{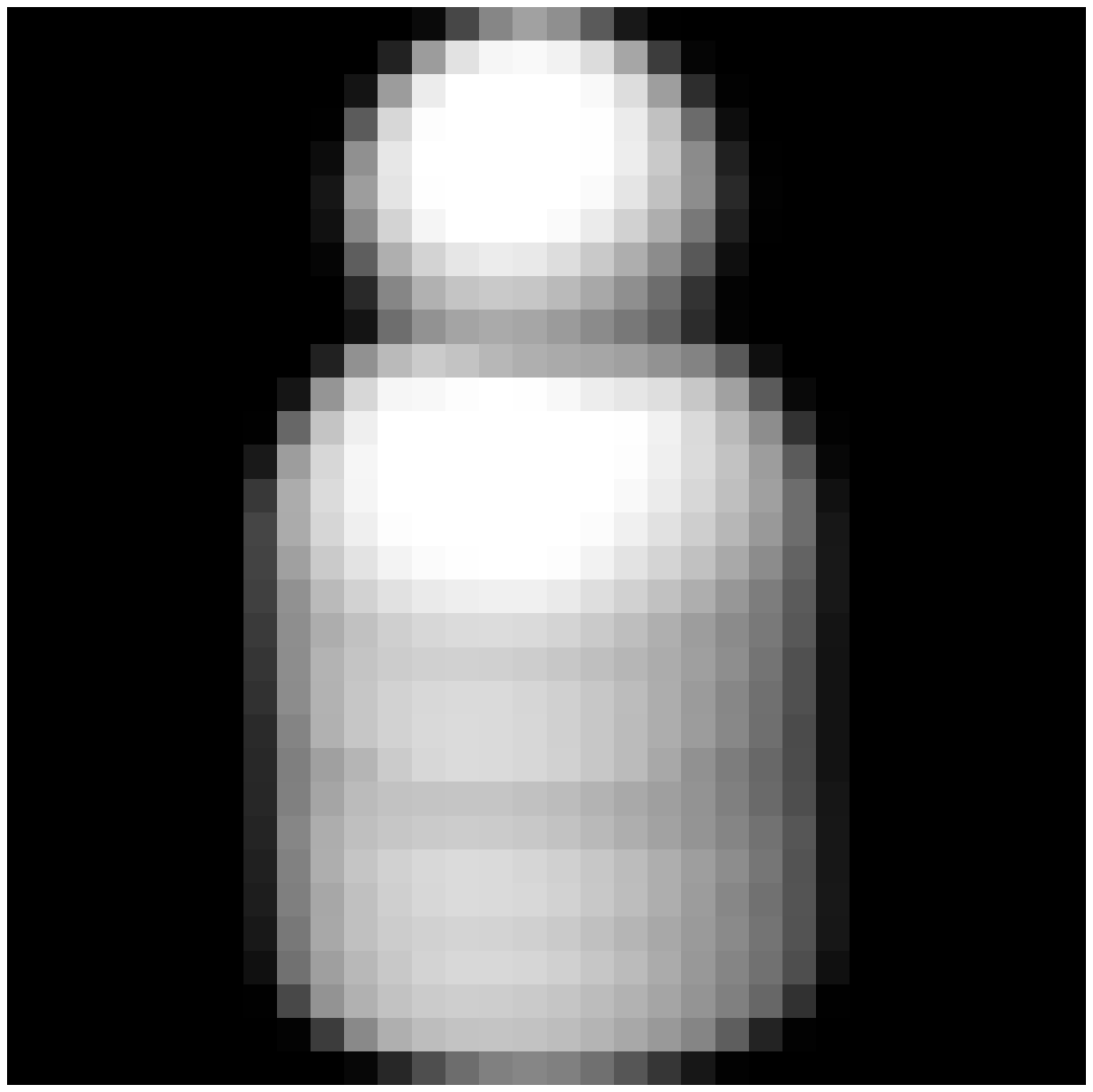} &
    \includegraphics[width=0.048\linewidth,bb=142 226 494 578,clip]{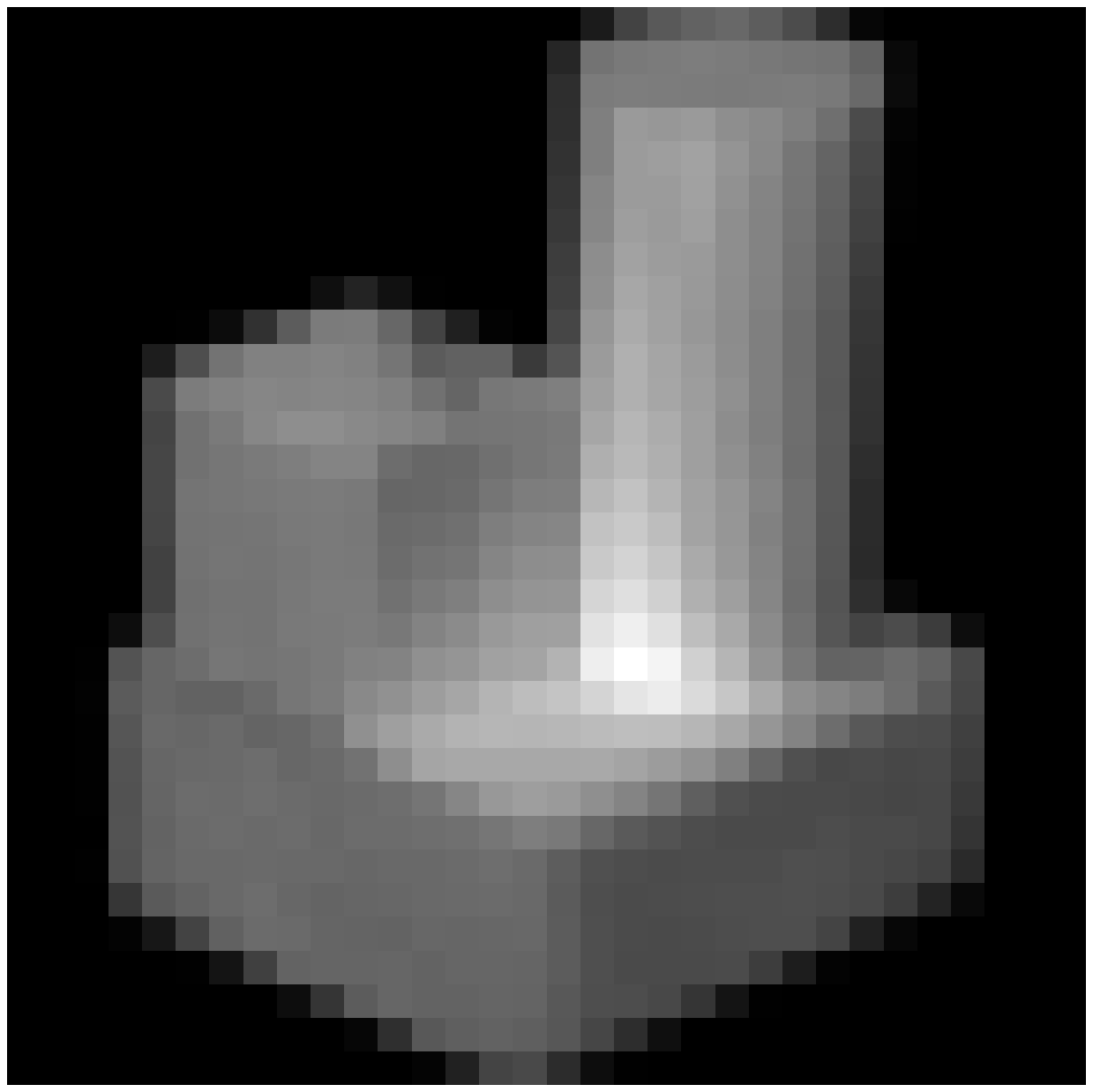} &
    \includegraphics[width=0.048\linewidth,bb=142 226 494 578,clip]{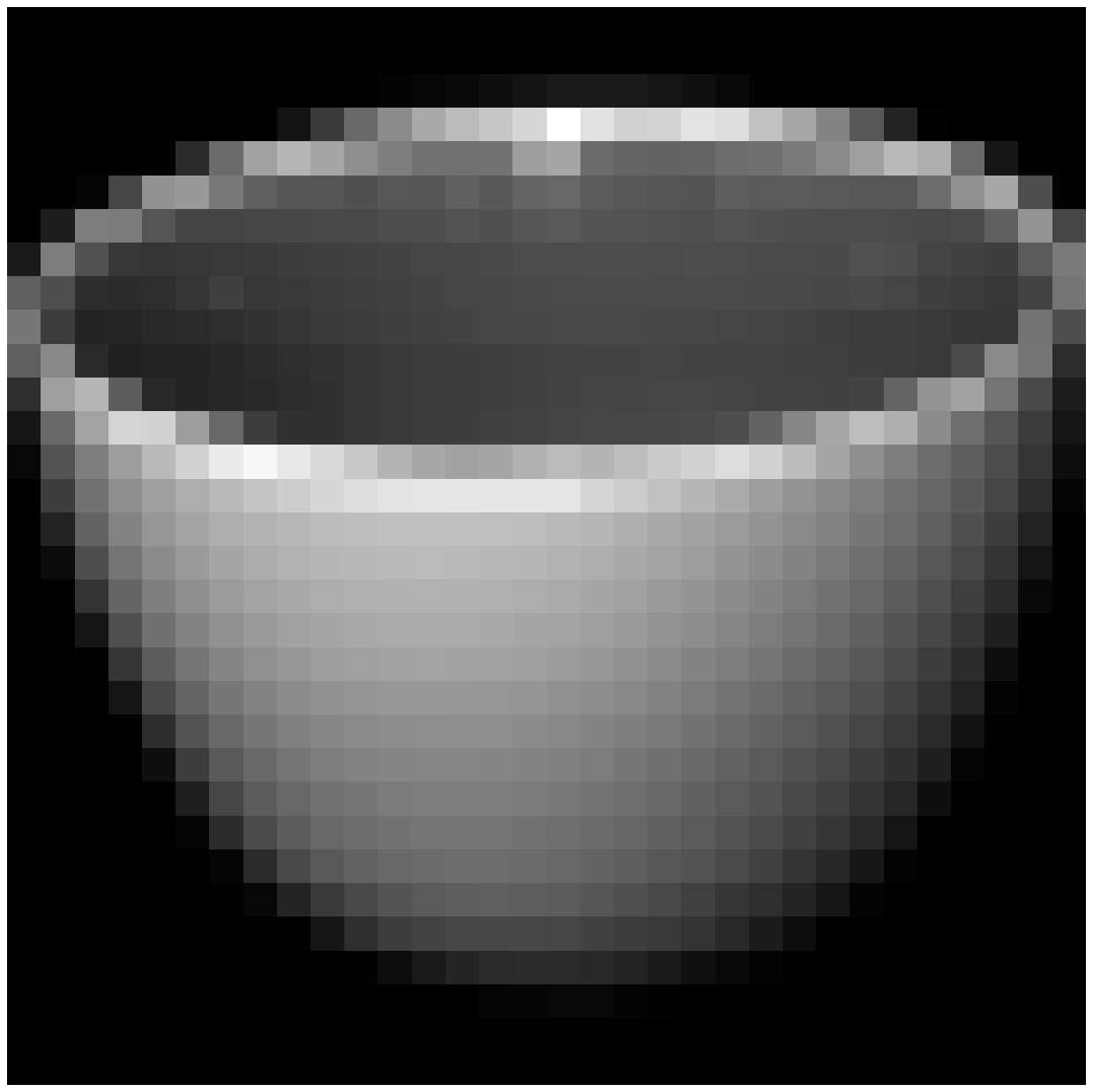} &
    \includegraphics[width=0.048\linewidth,bb=142 226 494 578,clip]{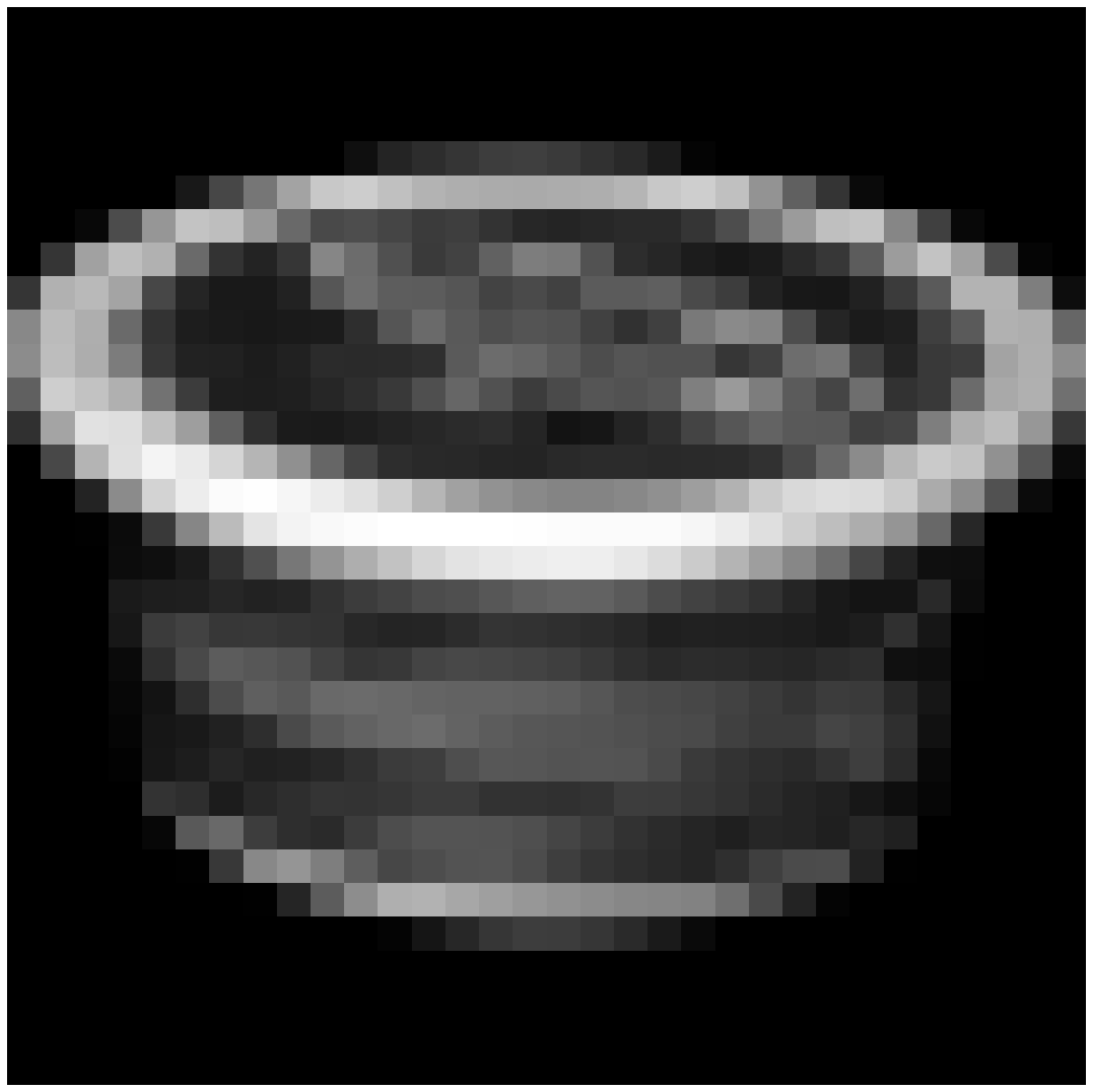} &
    \includegraphics[width=0.048\linewidth,bb=142 226 494 578,clip]{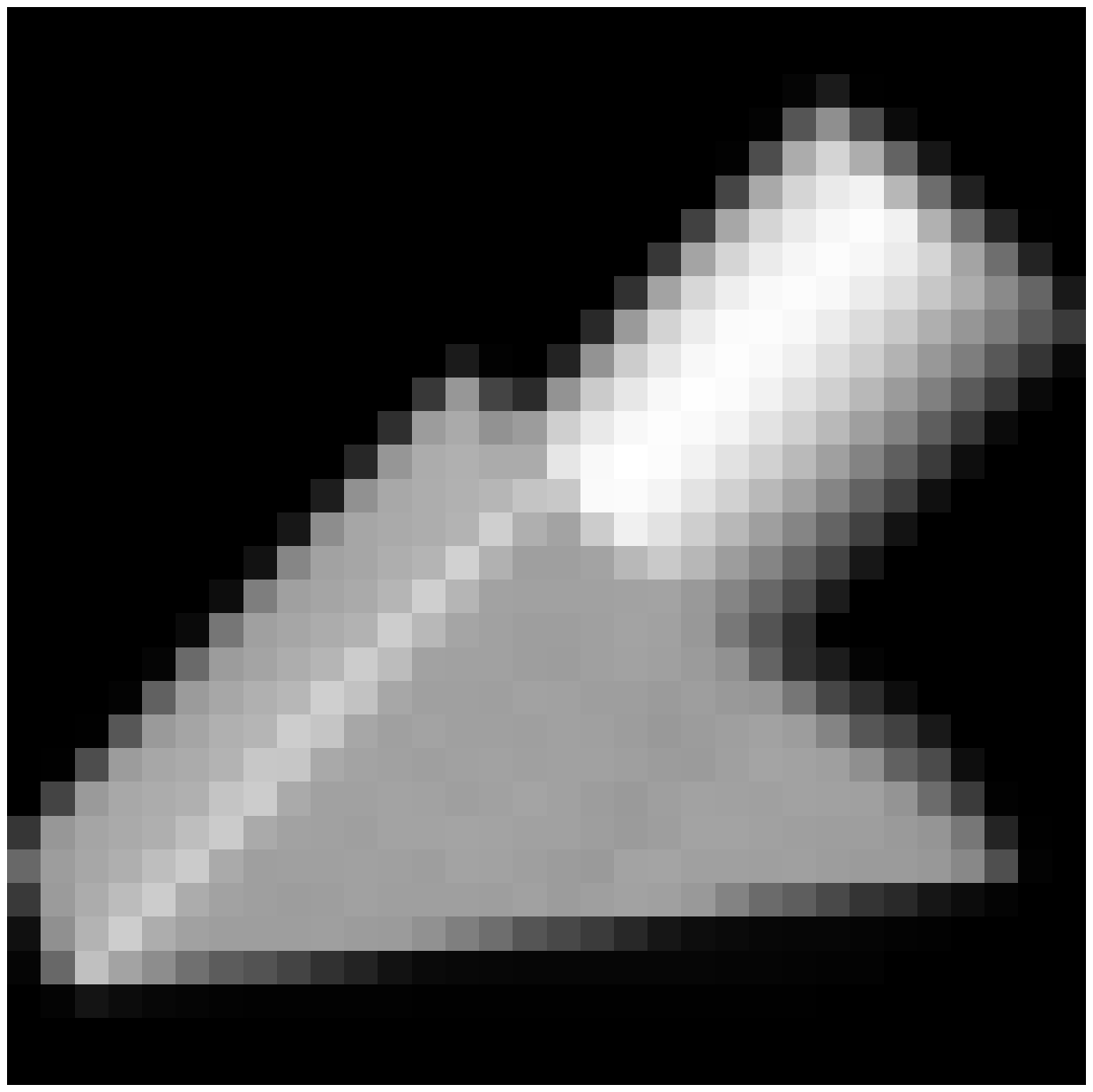} &
    \includegraphics[width=0.048\linewidth,bb=142 226 494 578,clip]{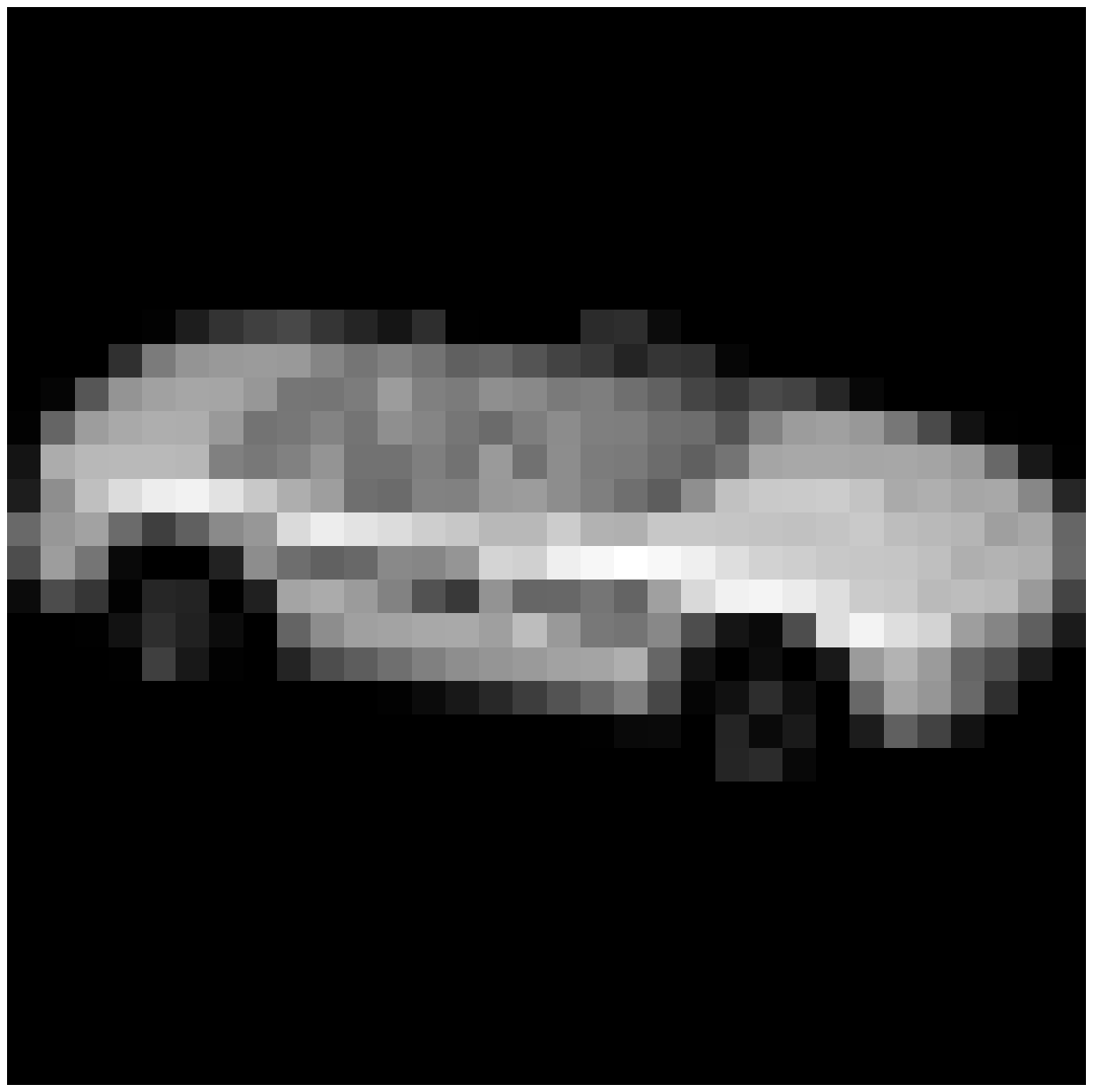} &
    \includegraphics[width=0.048\linewidth,bb=142 226 494 578,clip]{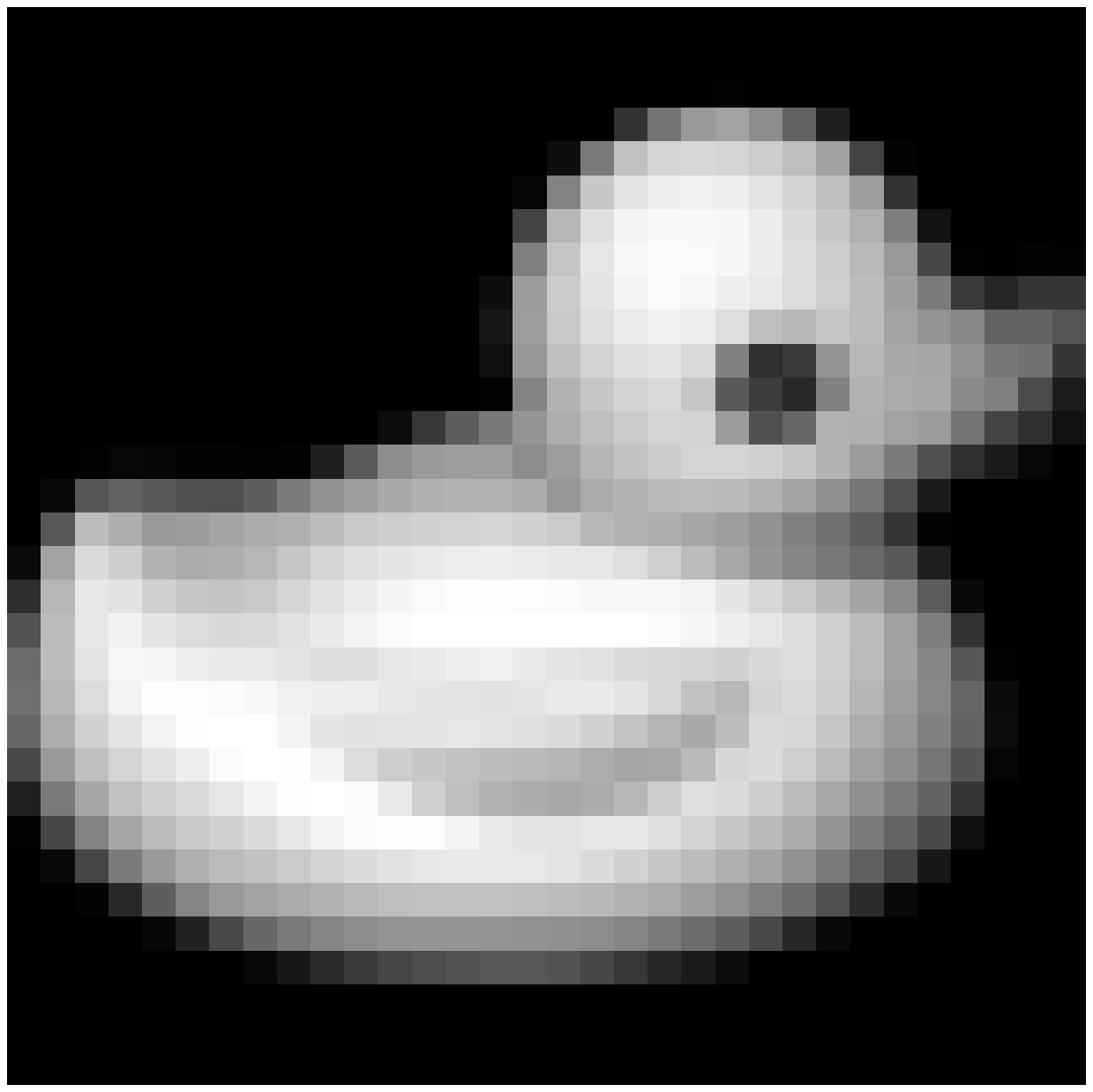} &
    \includegraphics[width=0.048\linewidth,bb=142 226 494 578,clip]{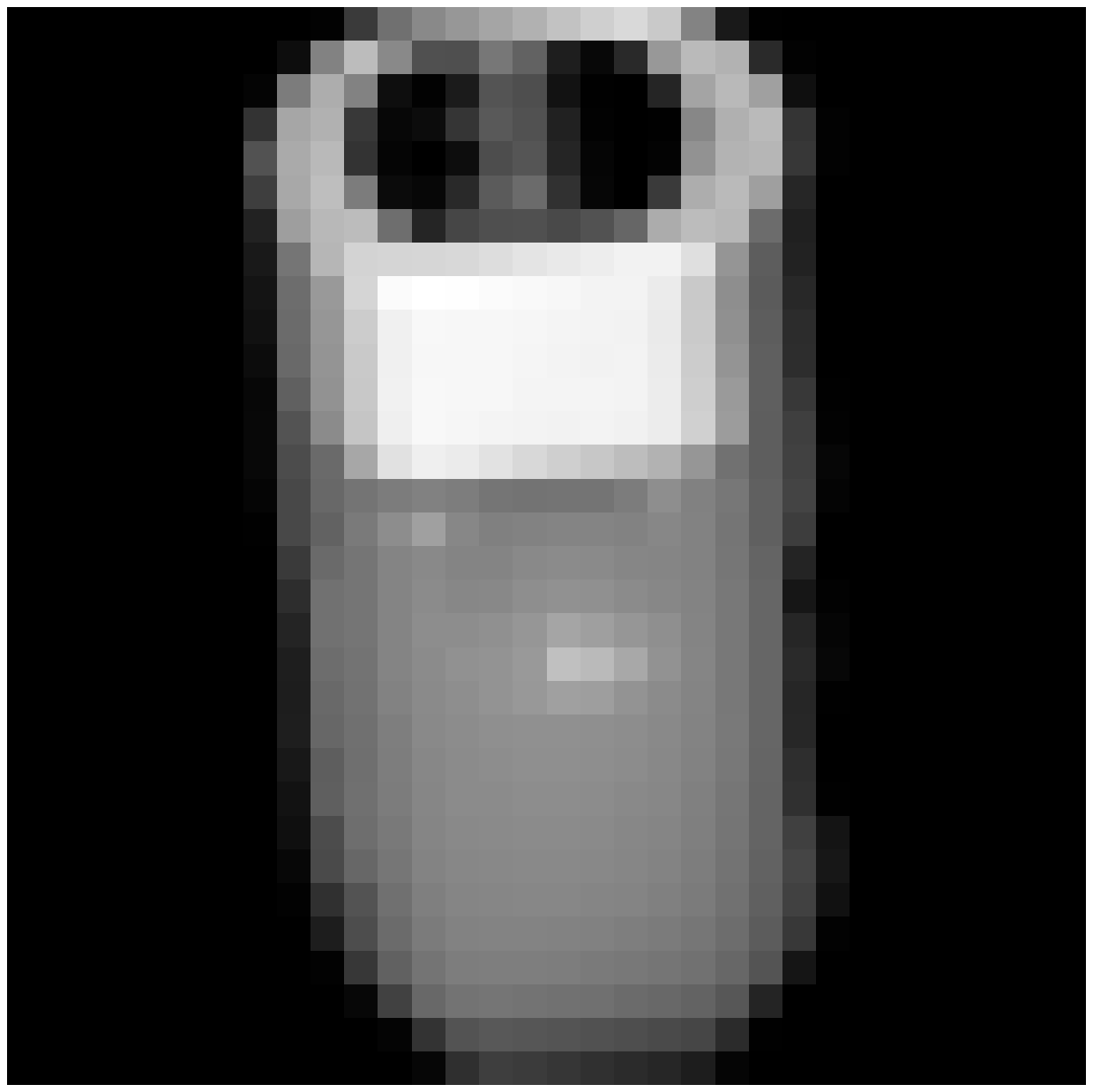} &
    \includegraphics[width=0.048\linewidth,bb=142 226 494 578,clip]{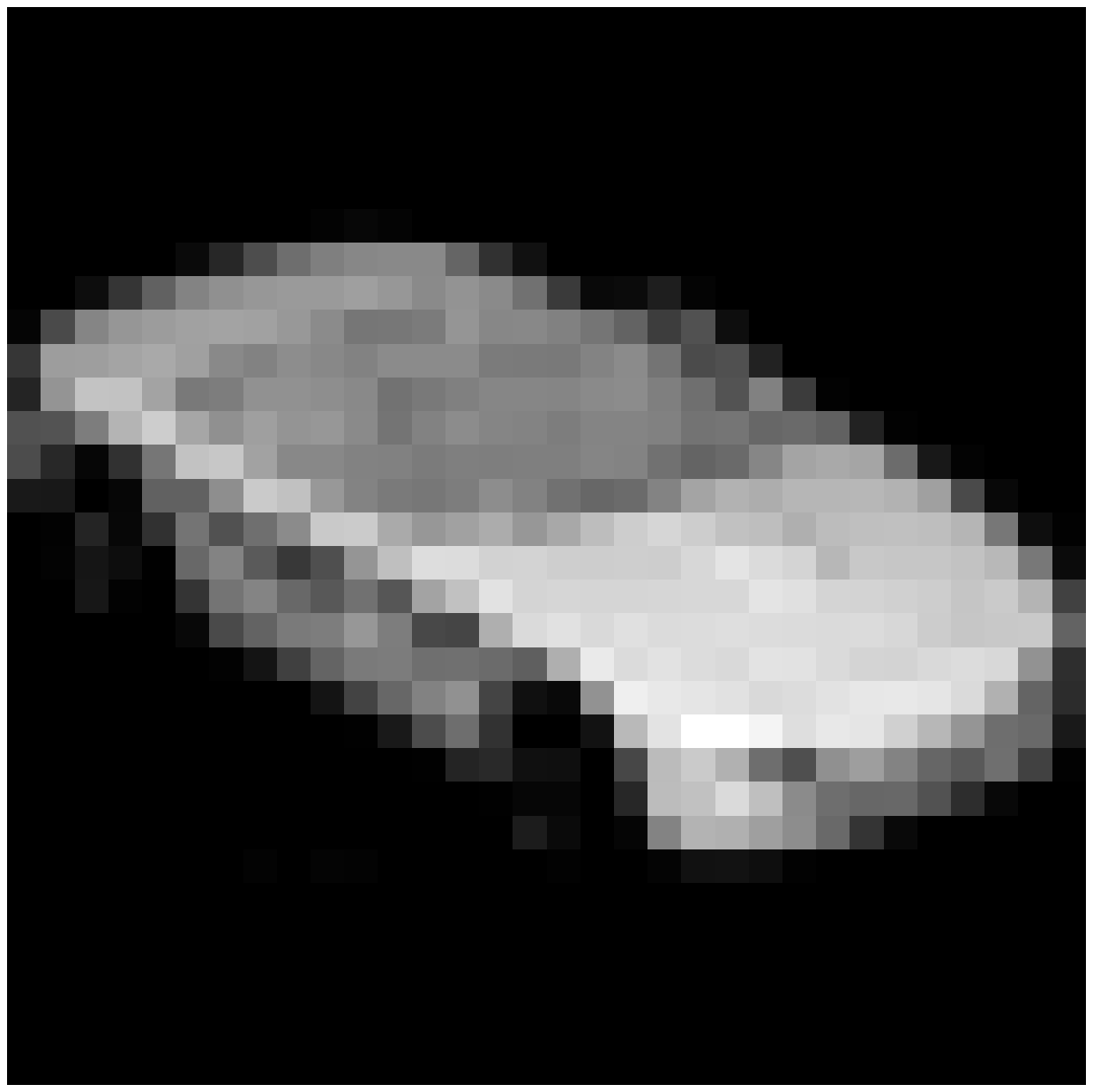} &
    \includegraphics[width=0.048\linewidth,bb=142 226 494 578,clip]{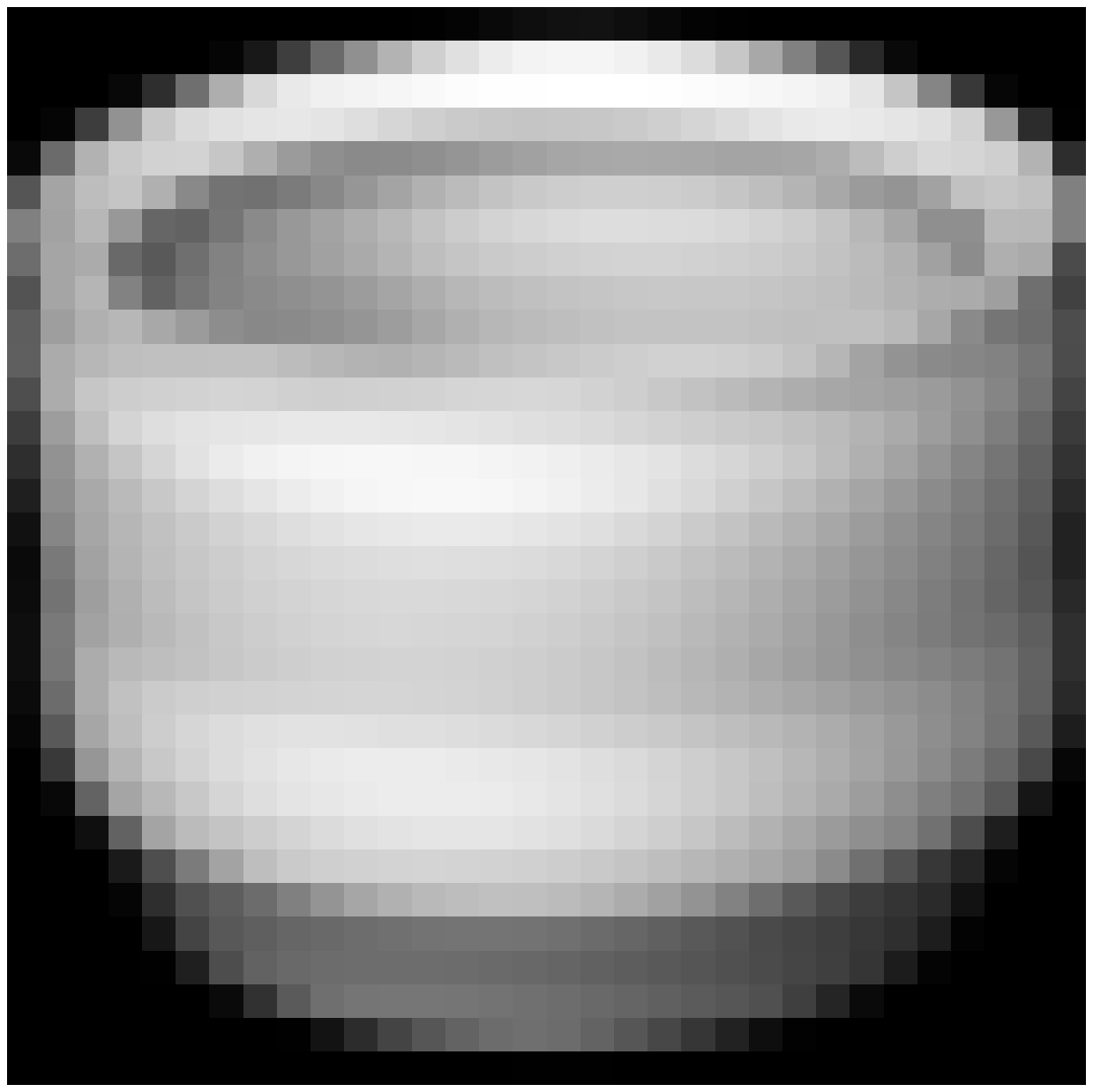} &
    \includegraphics[width=0.048\linewidth,bb=142 226 494 578,clip]{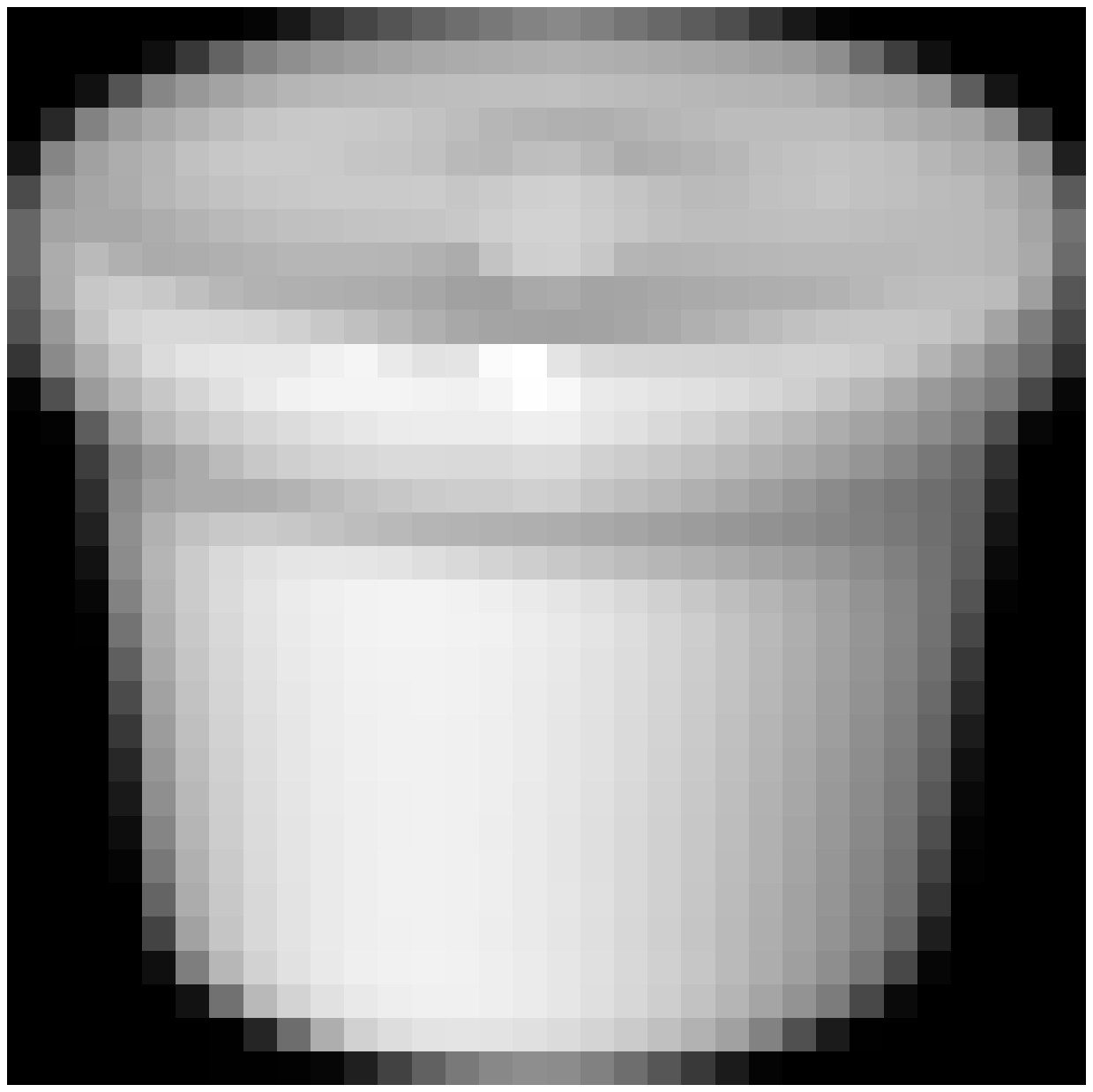} &
    \includegraphics[width=0.048\linewidth,bb=142 226 494 578,clip]{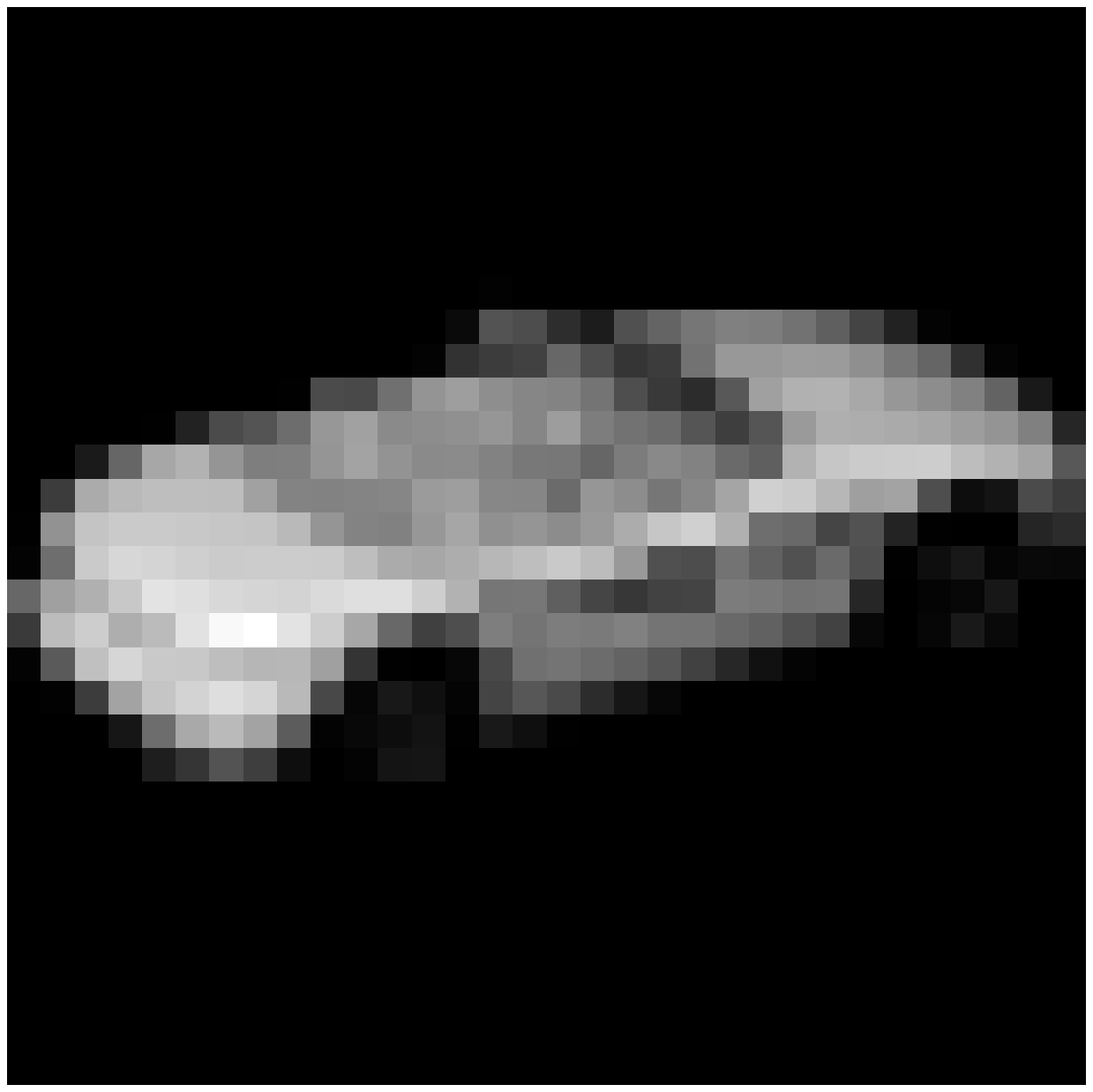} &
    \includegraphics[width=0.048\linewidth,bb=142 226 494 578,clip]{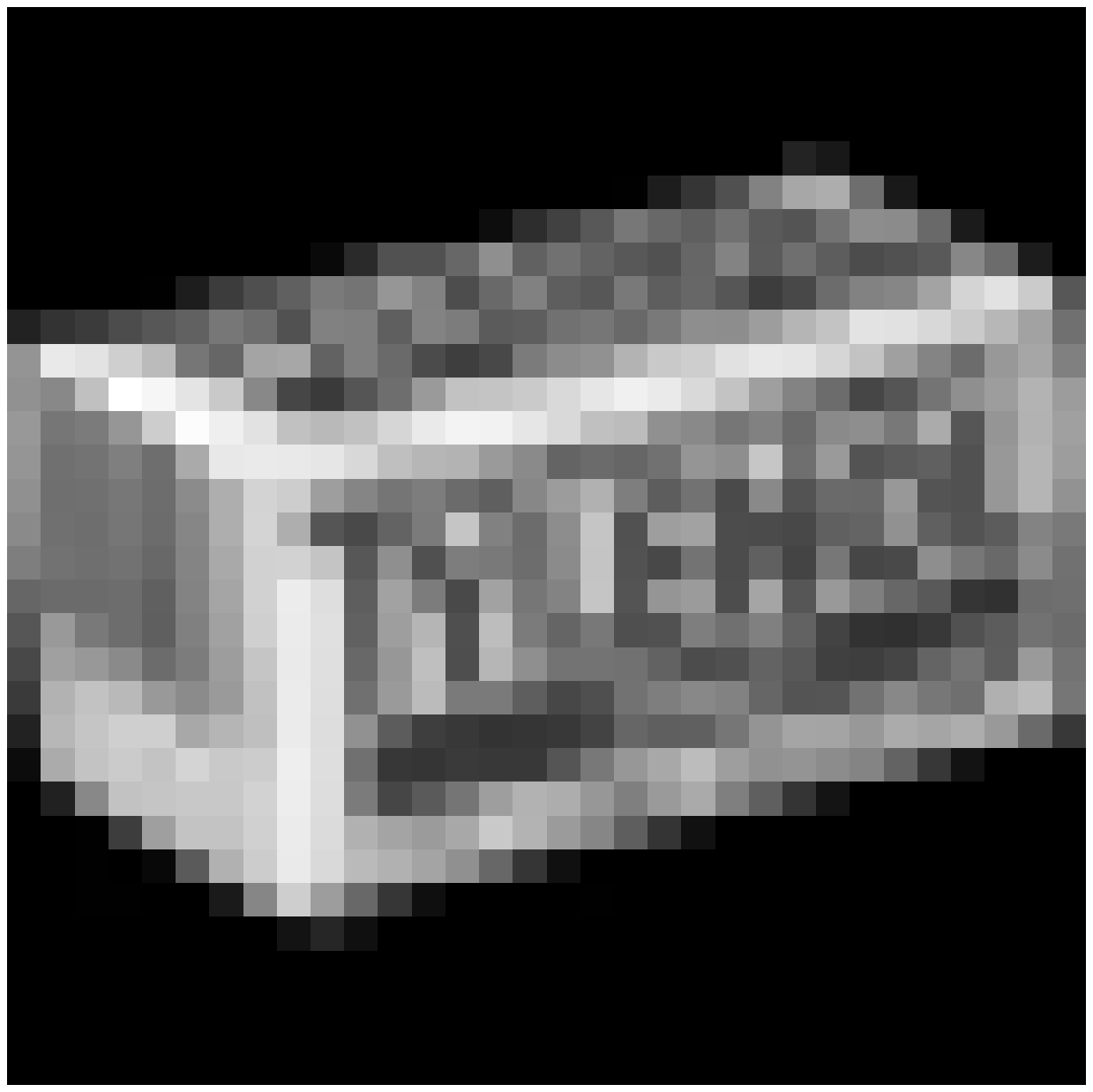} &
    \includegraphics[width=0.048\linewidth,bb=142 226 494 578,clip]{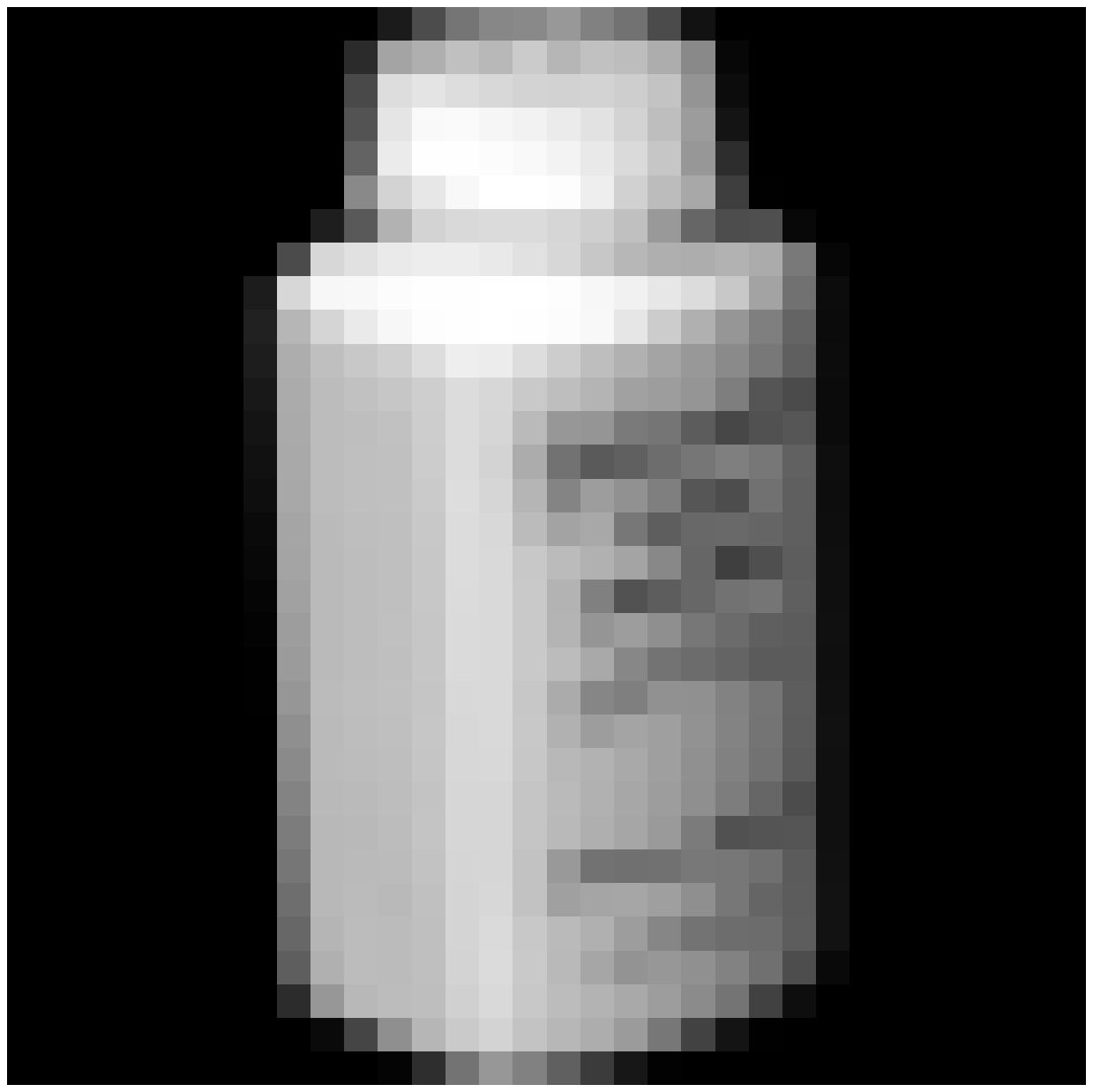} \\[-1ex]
    \rotatebox{90}{\tiny\hspace{1ex}\caja{c}{c}{Mean \\ shift}} &
    \includegraphics[width=0.048\linewidth,bb=142 226 494 578,clip]{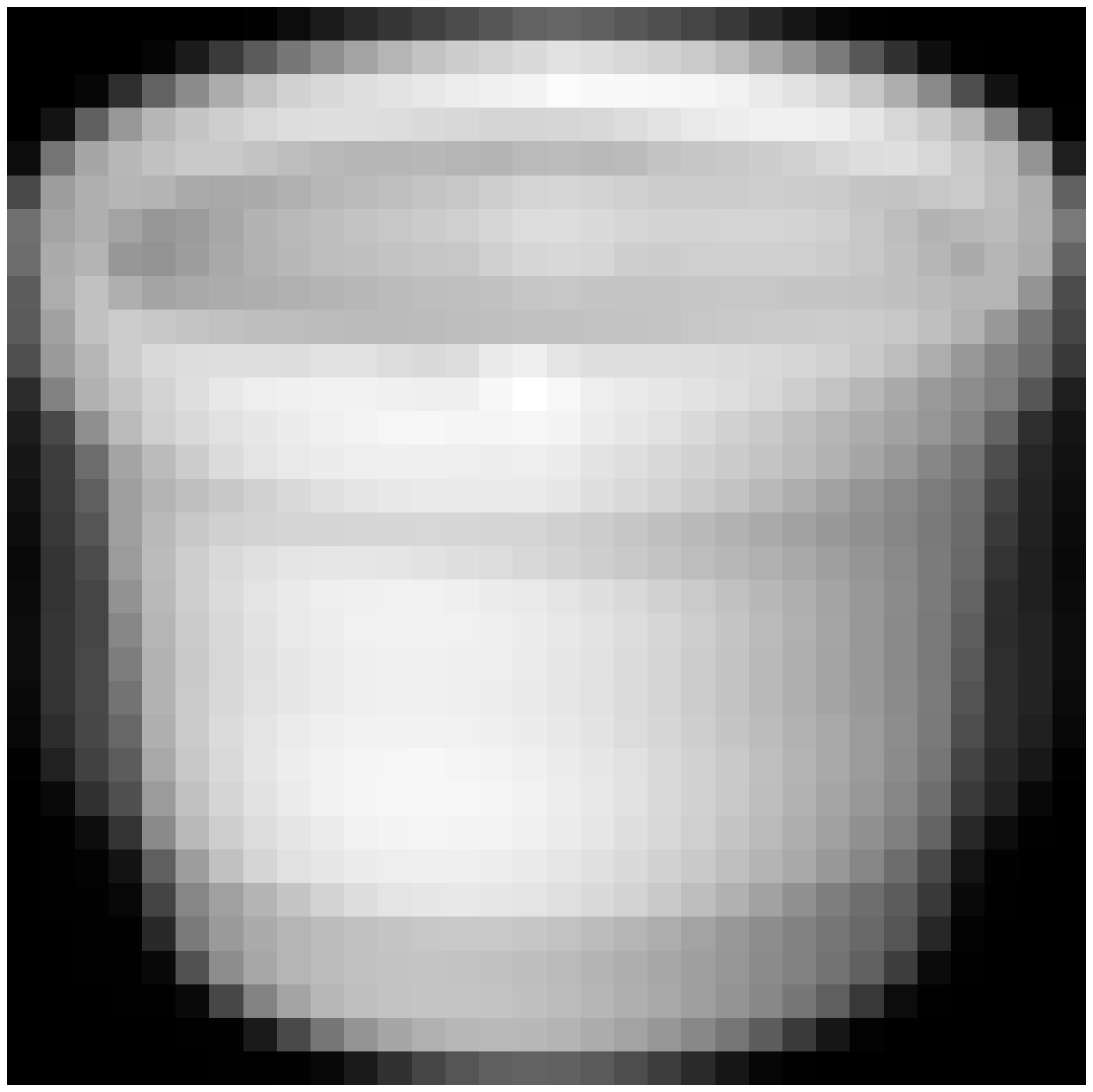} &
    \includegraphics[width=0.048\linewidth,bb=142 226 494 578,clip]{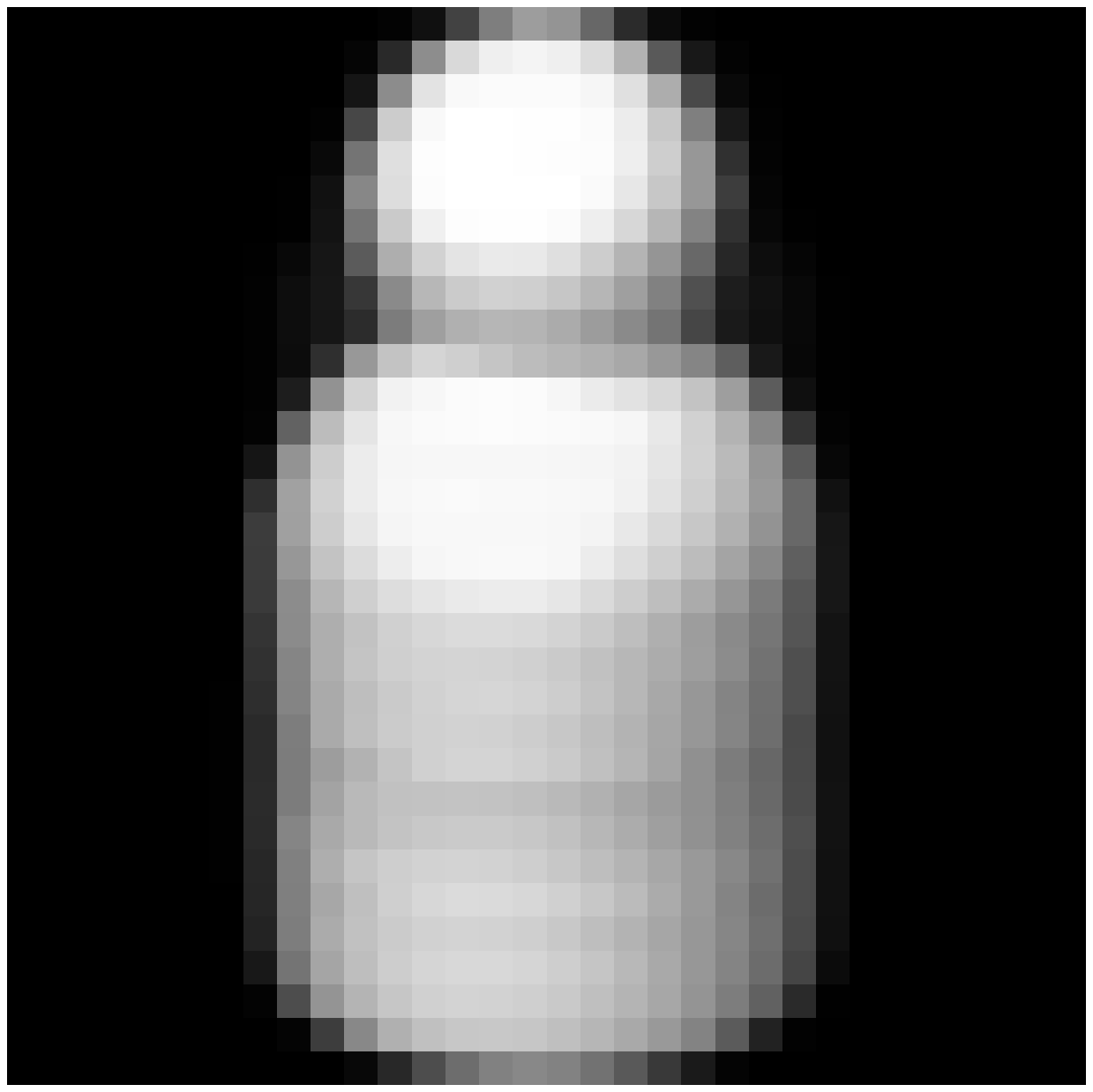} &
    \includegraphics[width=0.048\linewidth,bb=142 226 494 578,clip]{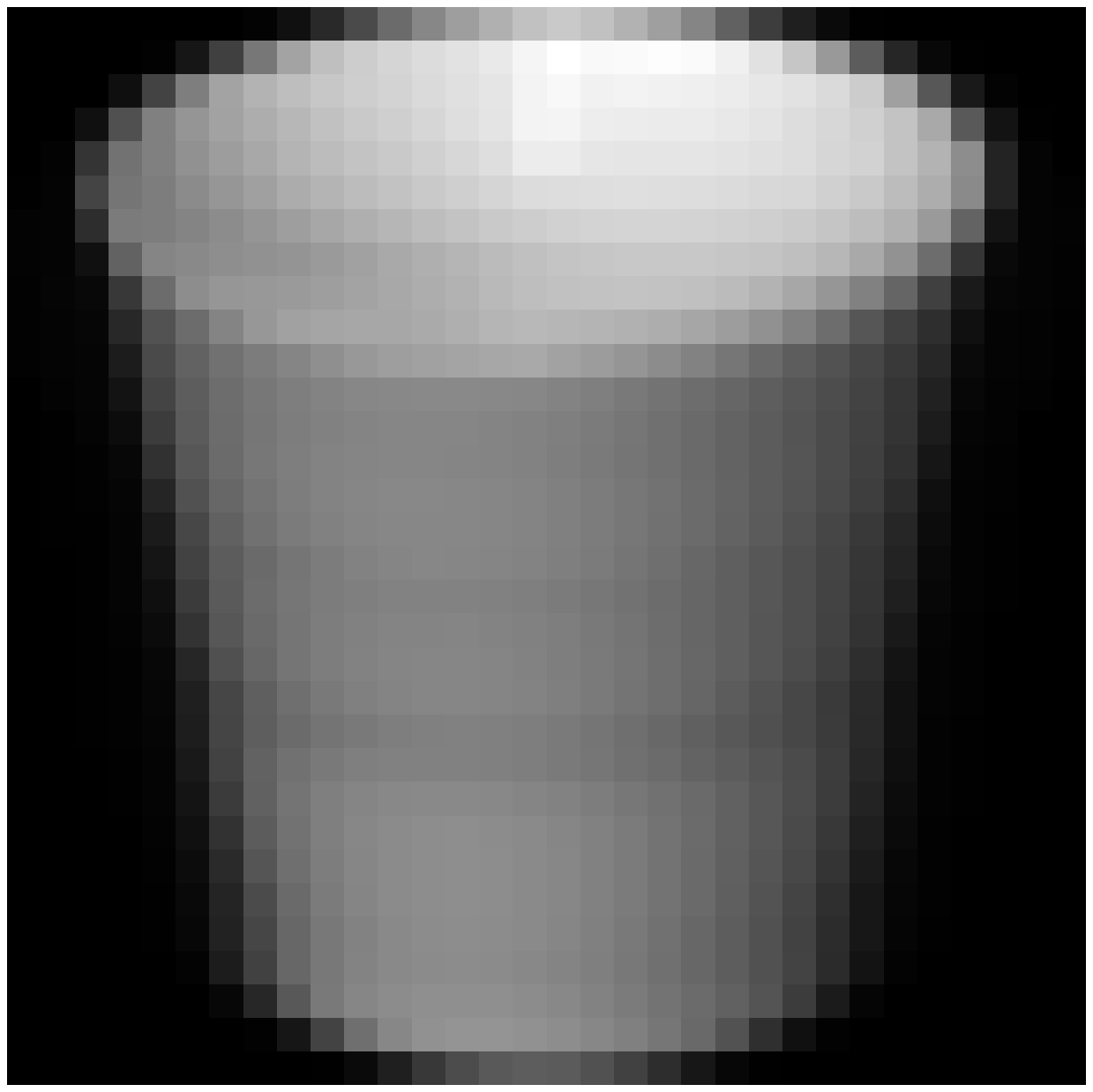} &
    \includegraphics[width=0.048\linewidth,bb=142 226 494 578,clip]{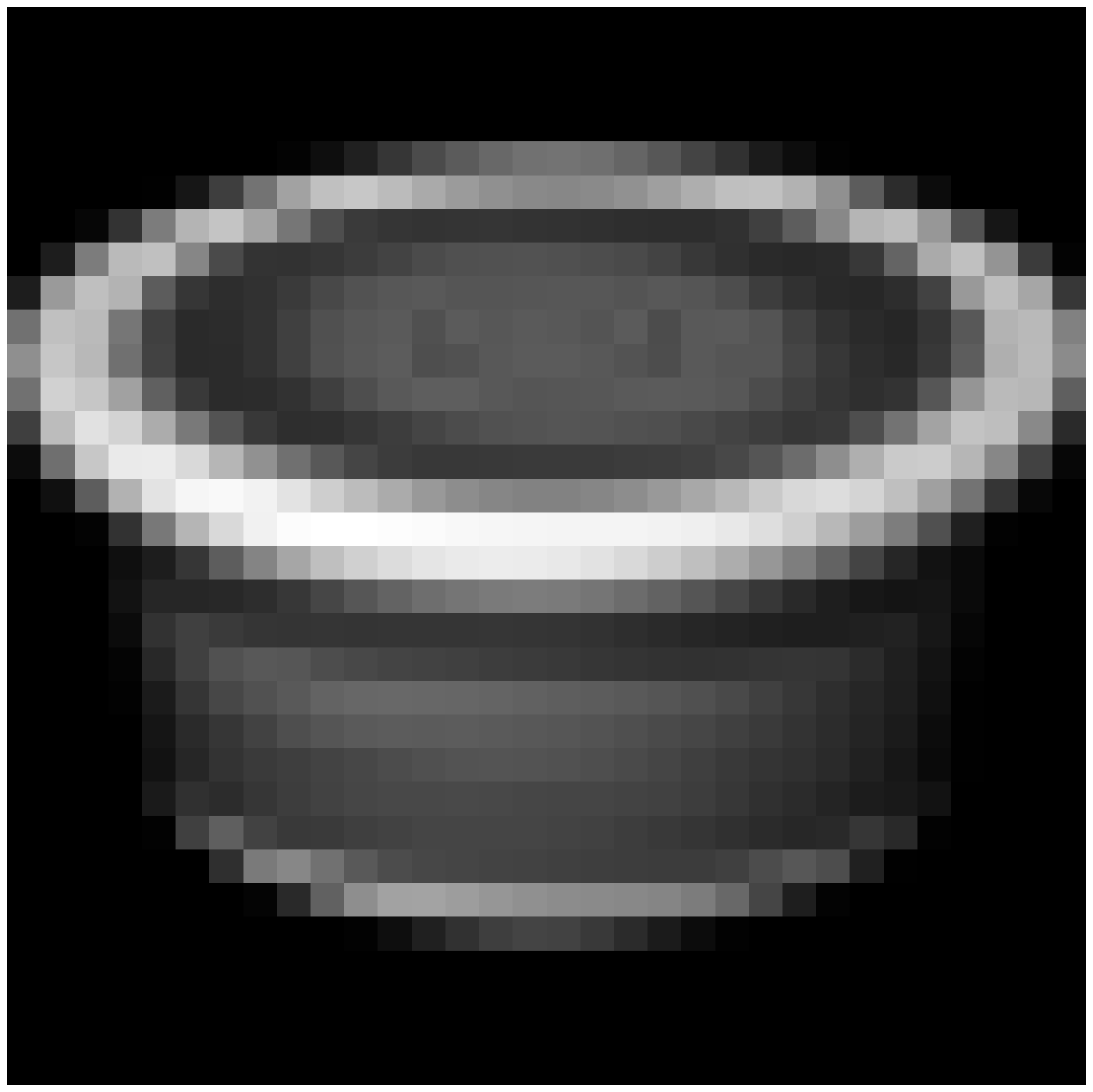} &
    \includegraphics[width=0.048\linewidth,bb=142 226 494 578,clip]{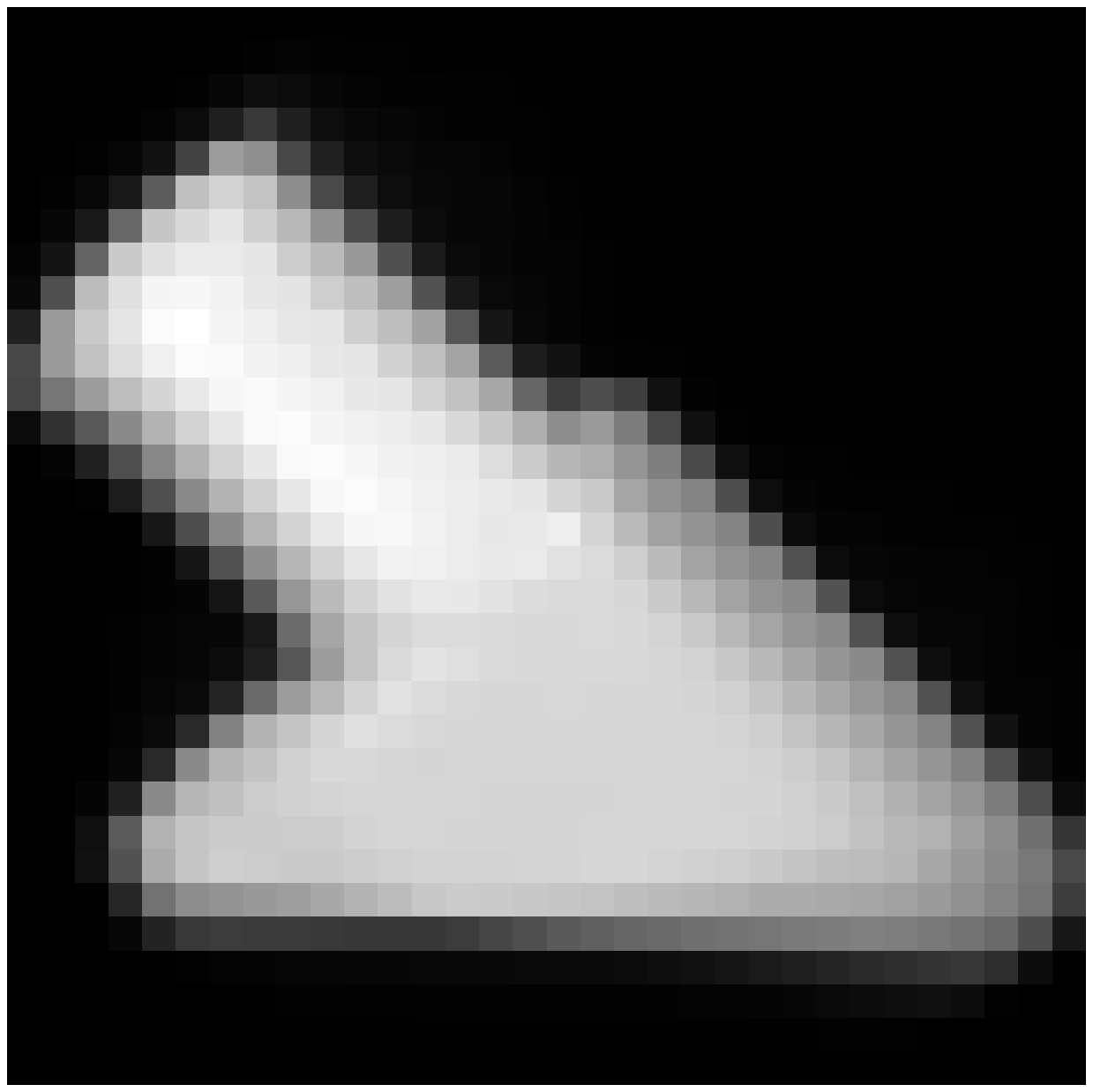} &
    \includegraphics[width=0.048\linewidth,bb=142 226 494 578,clip]{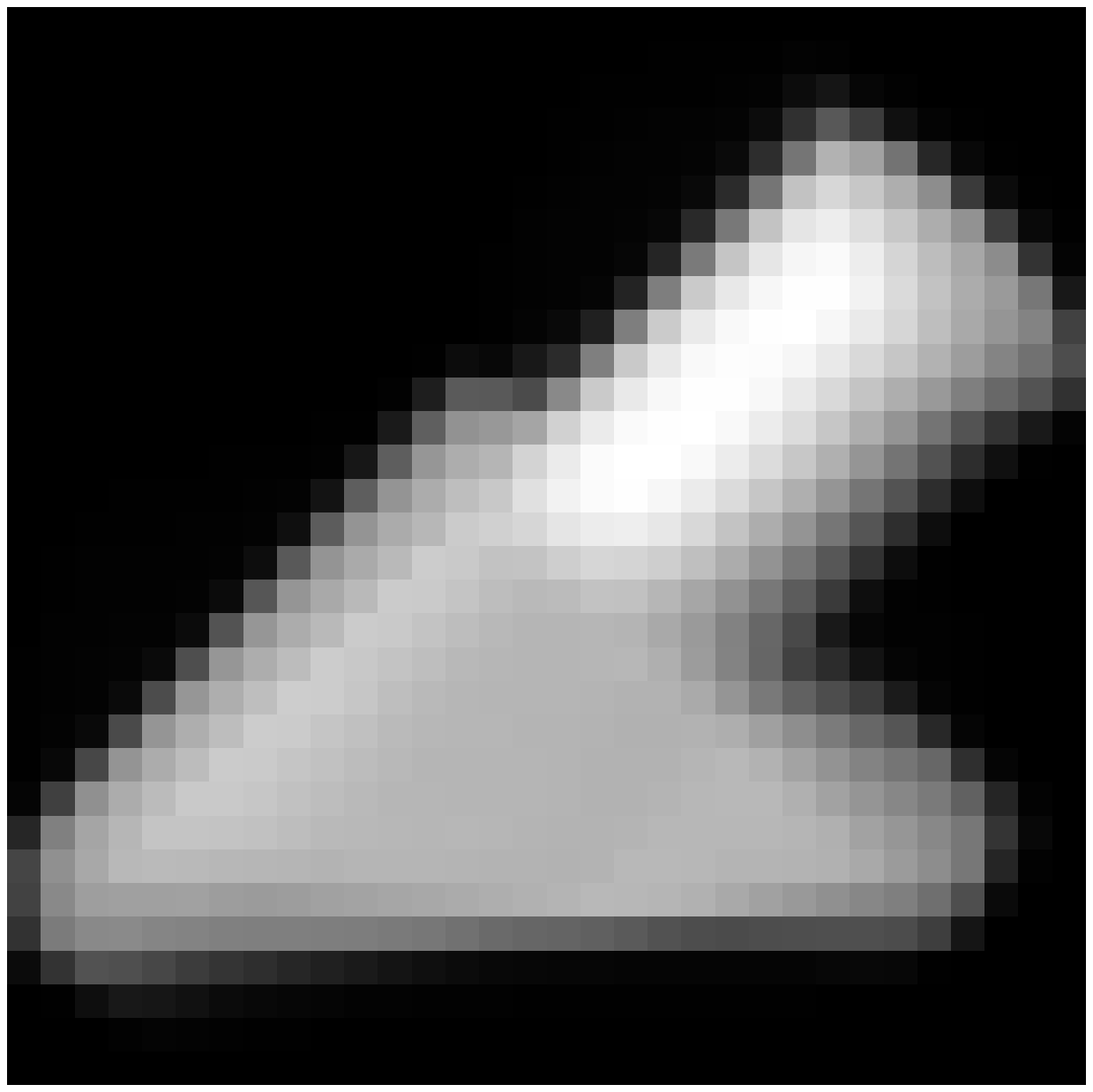} &
    \includegraphics[width=0.048\linewidth,bb=142 226 494 578,clip]{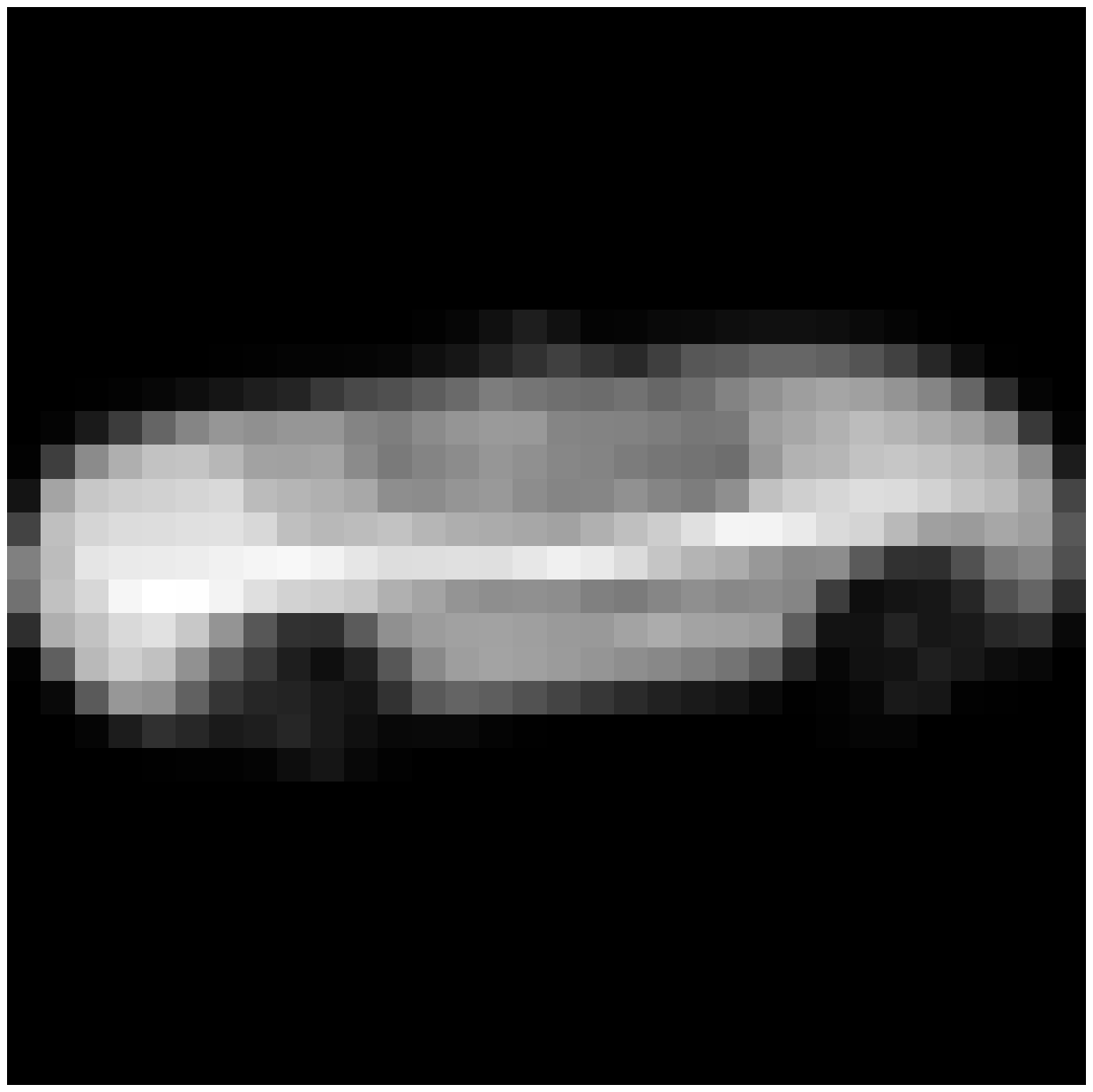} &
    \includegraphics[width=0.048\linewidth,bb=142 226 494 578,clip]{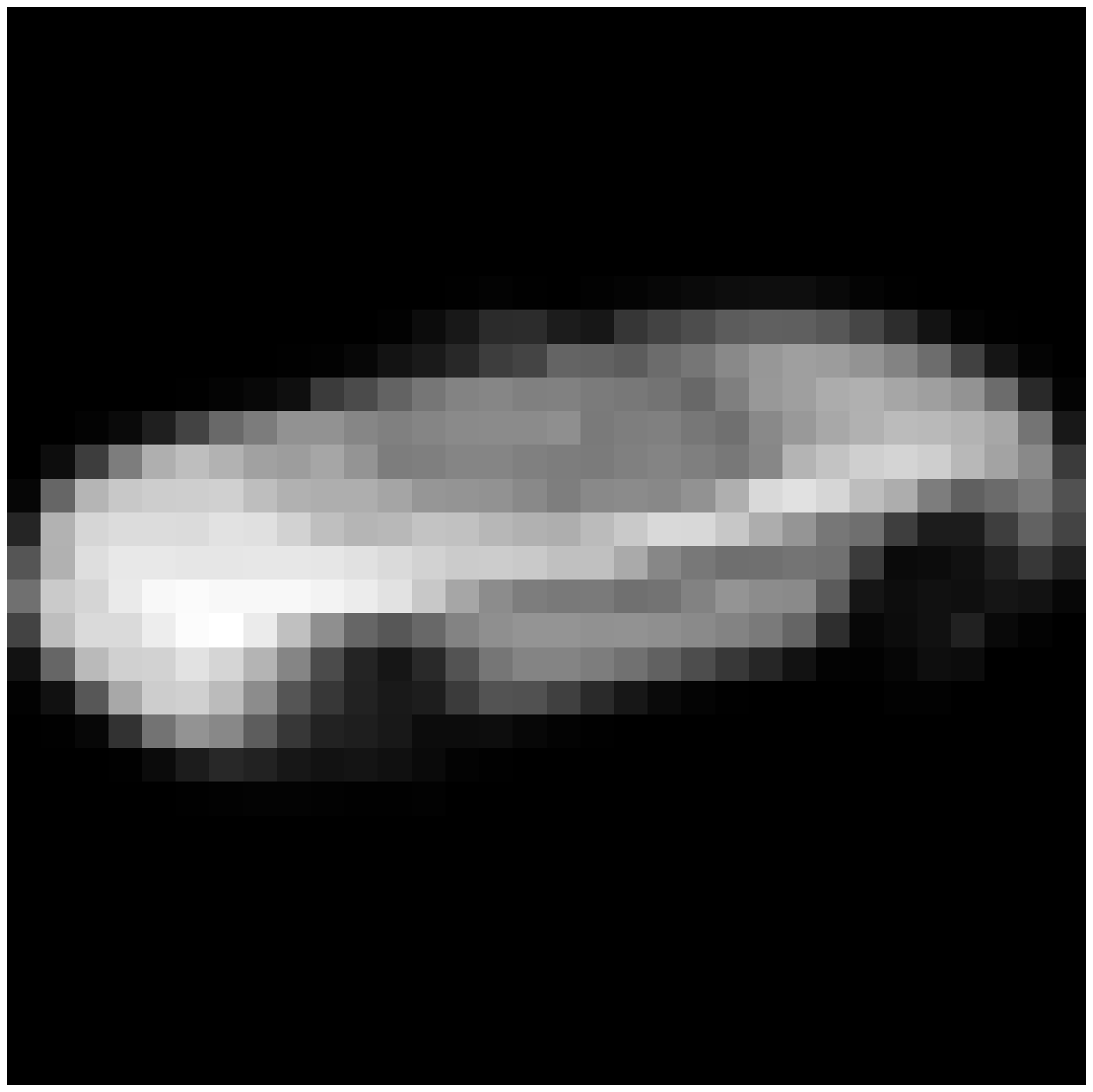} &
    \includegraphics[width=0.048\linewidth,bb=142 226 494 578,clip]{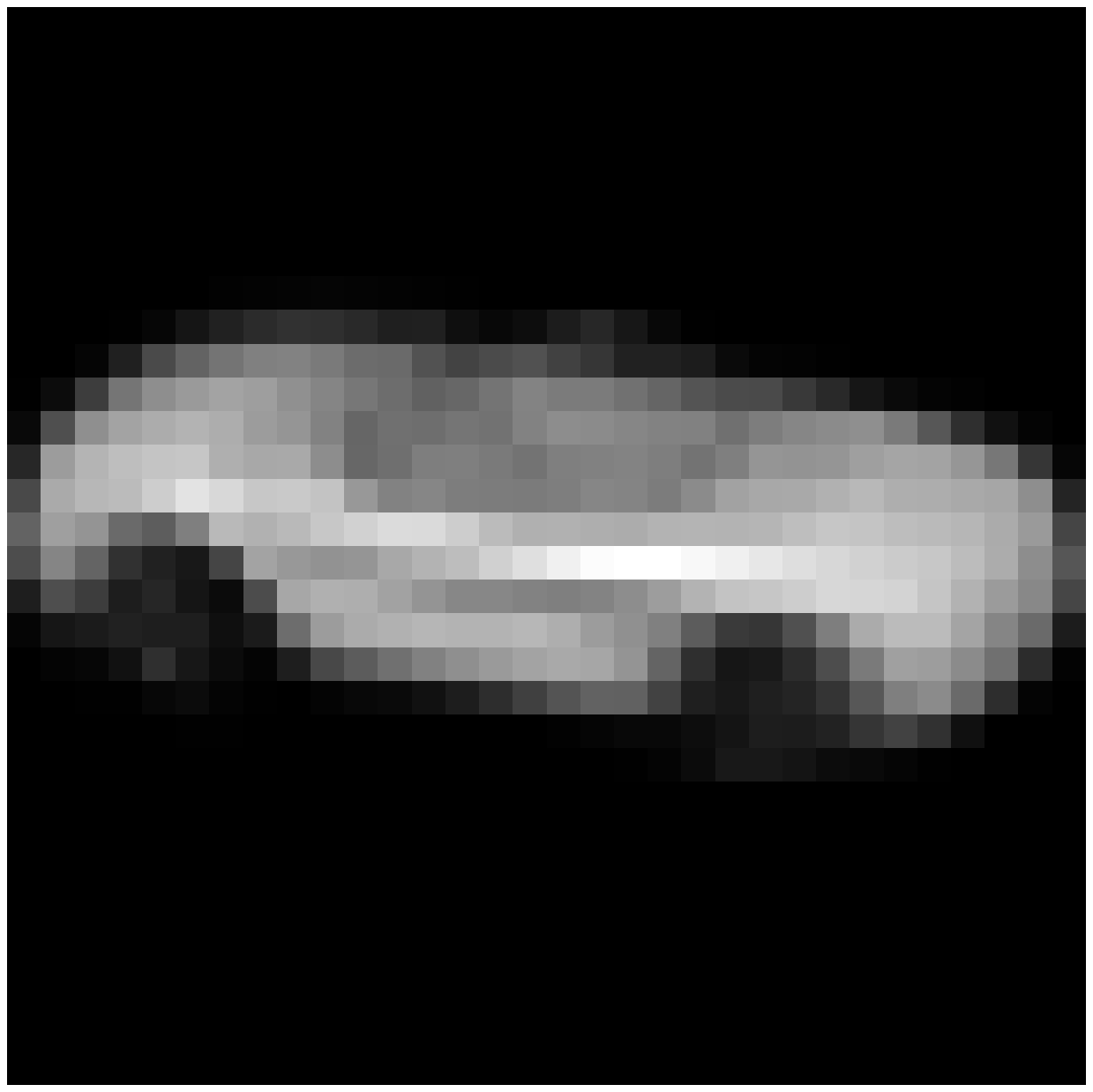} &
    \includegraphics[width=0.048\linewidth,bb=142 226 494 578,clip]{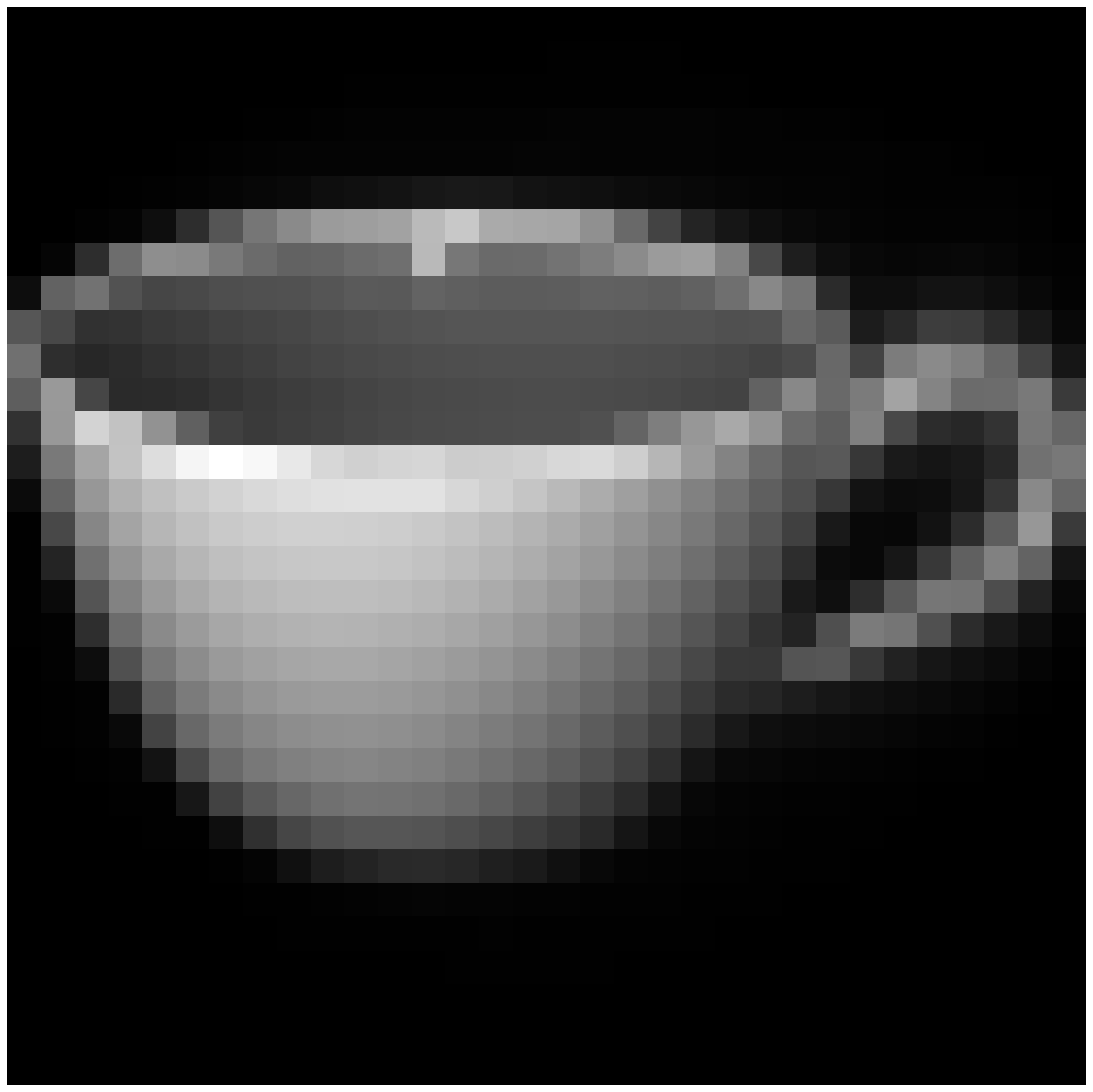} &
    \includegraphics[width=0.048\linewidth,bb=142 226 494 578,clip]{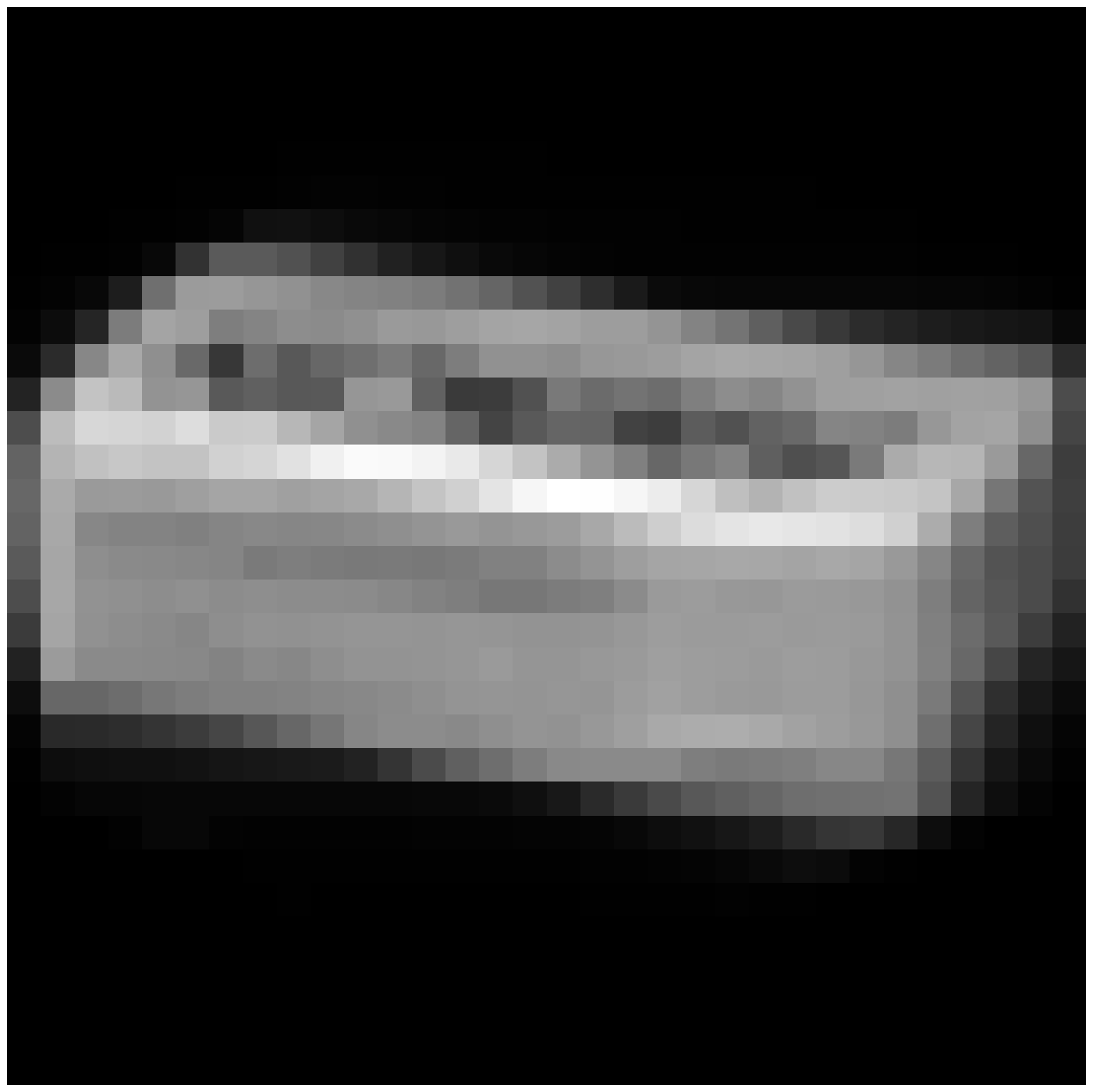} &
    \includegraphics[width=0.048\linewidth,bb=142 226 494 578,clip]{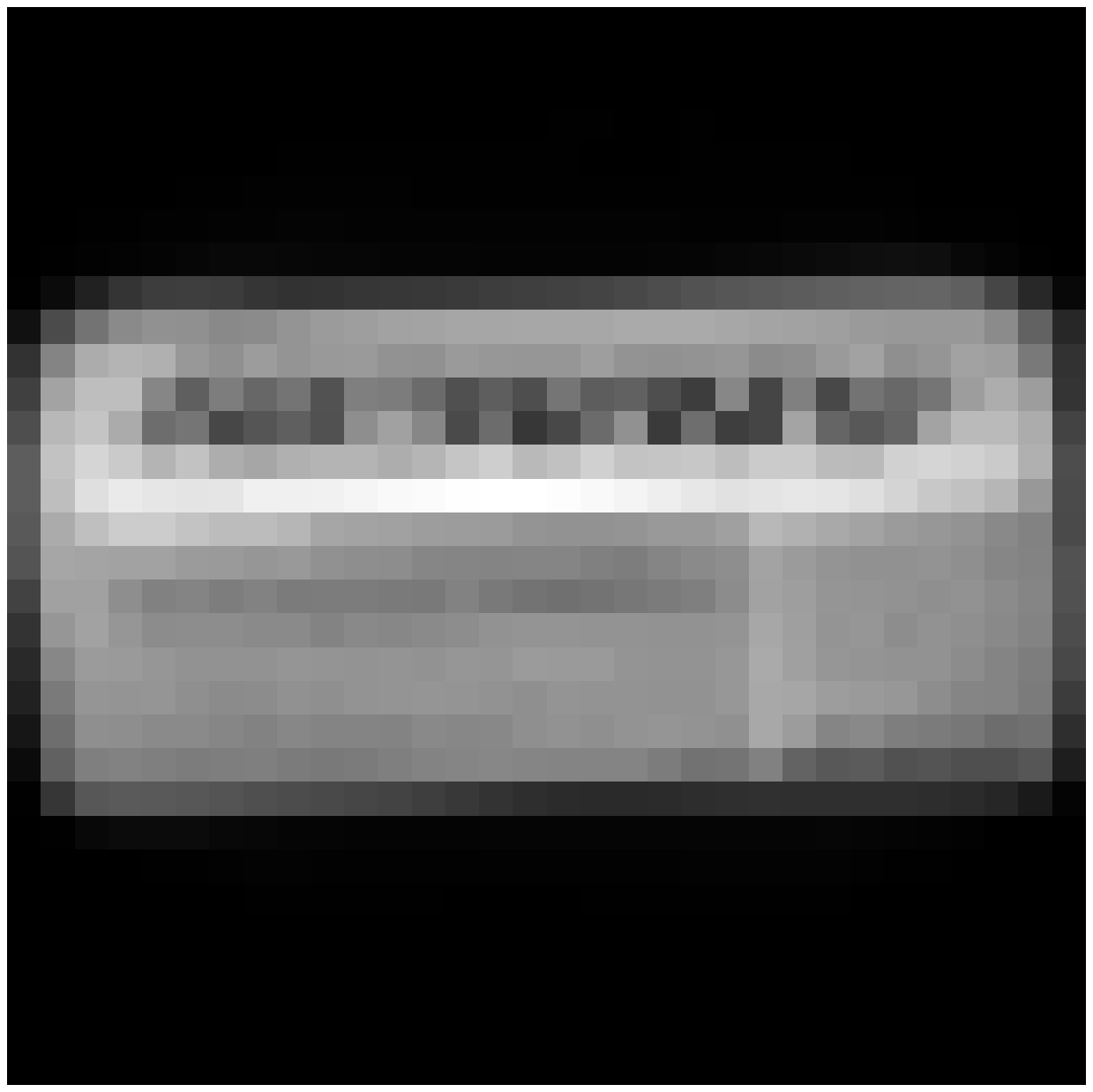} &
    \includegraphics[width=0.048\linewidth,bb=142 226 494 578,clip]{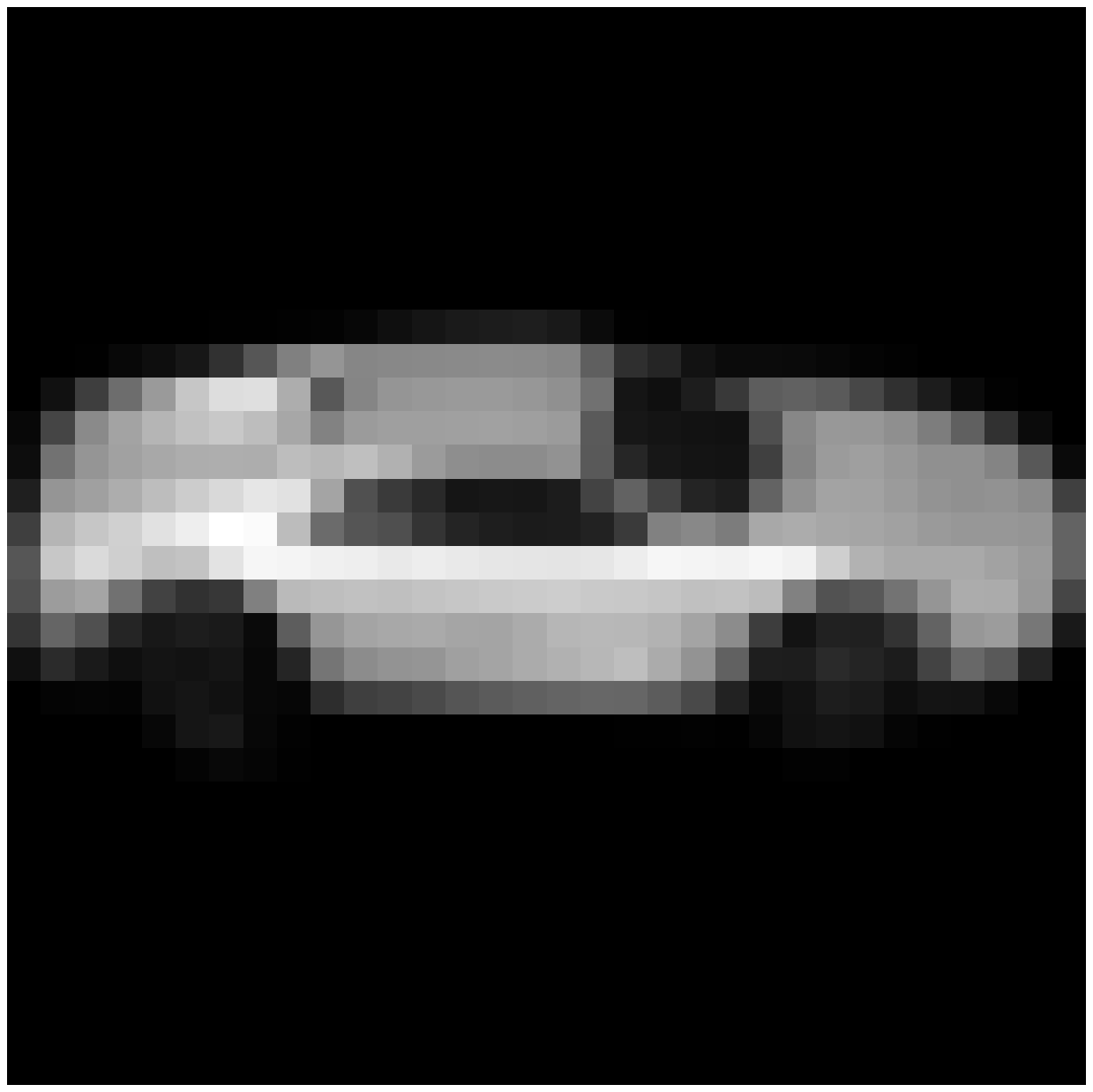} &
    \includegraphics[width=0.048\linewidth,bb=142 226 494 578,clip]{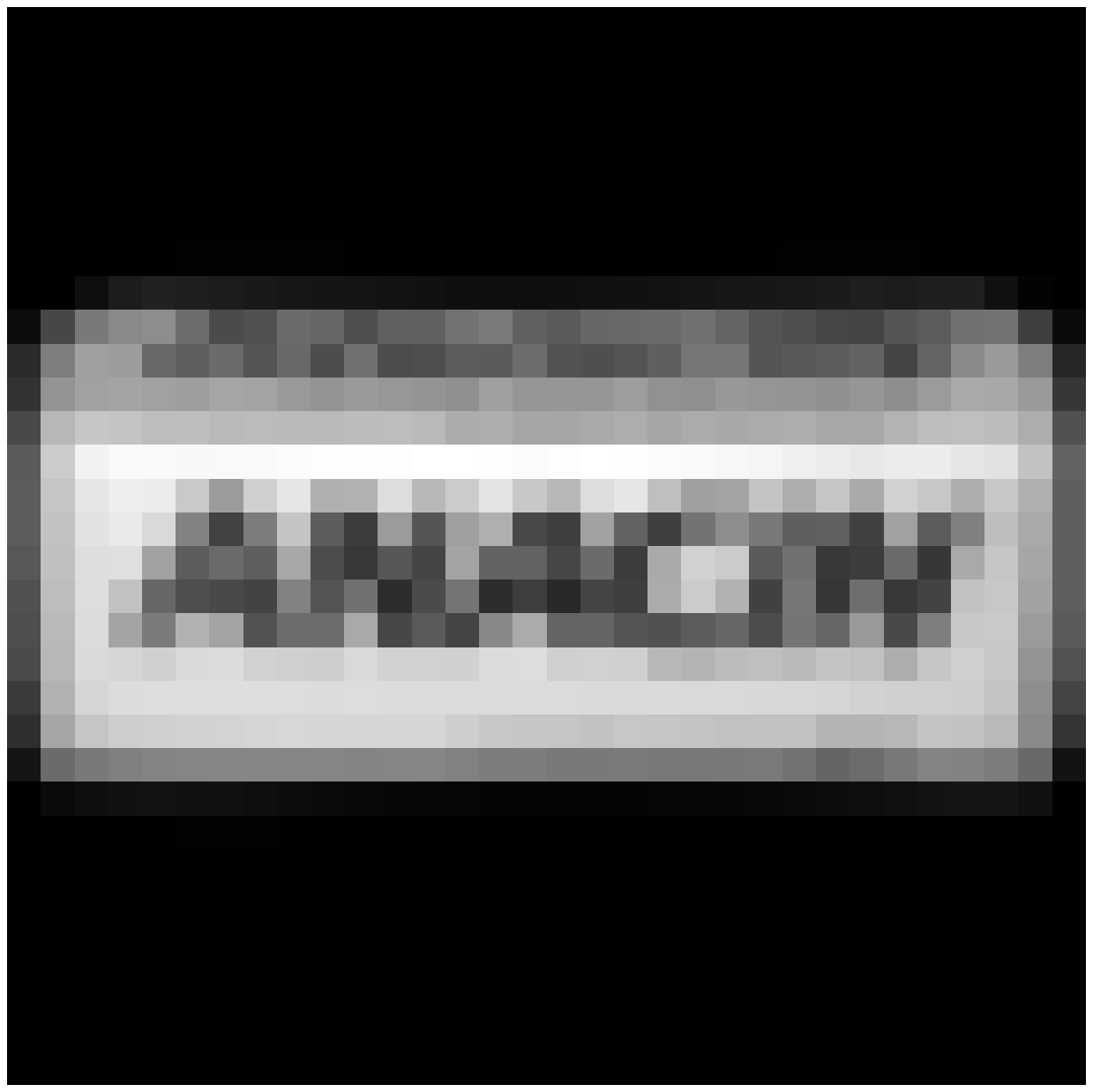} &
    \includegraphics[width=0.048\linewidth,bb=142 226 494 578,clip]{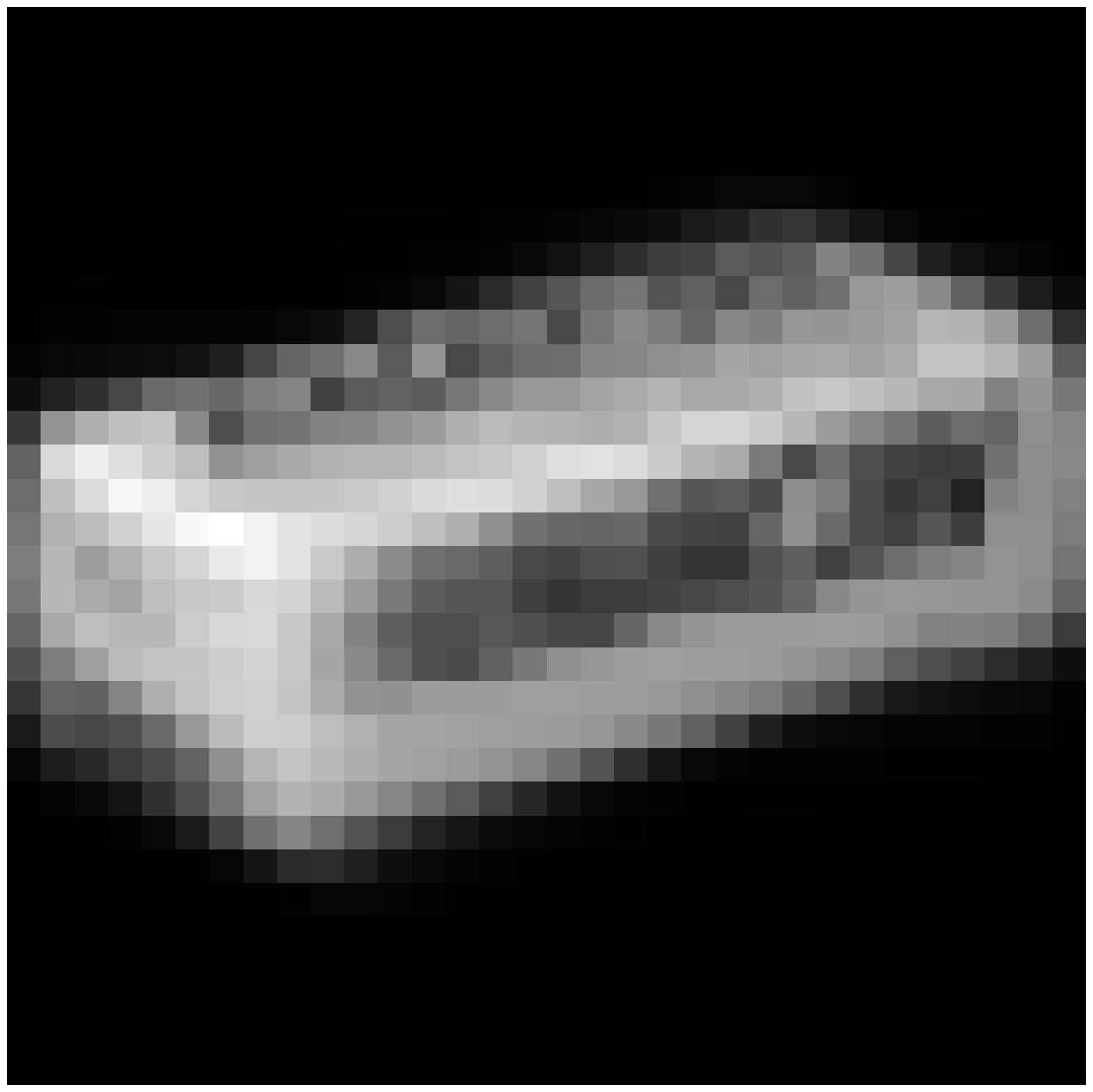} &
    \includegraphics[width=0.048\linewidth,bb=142 226 494 578,clip]{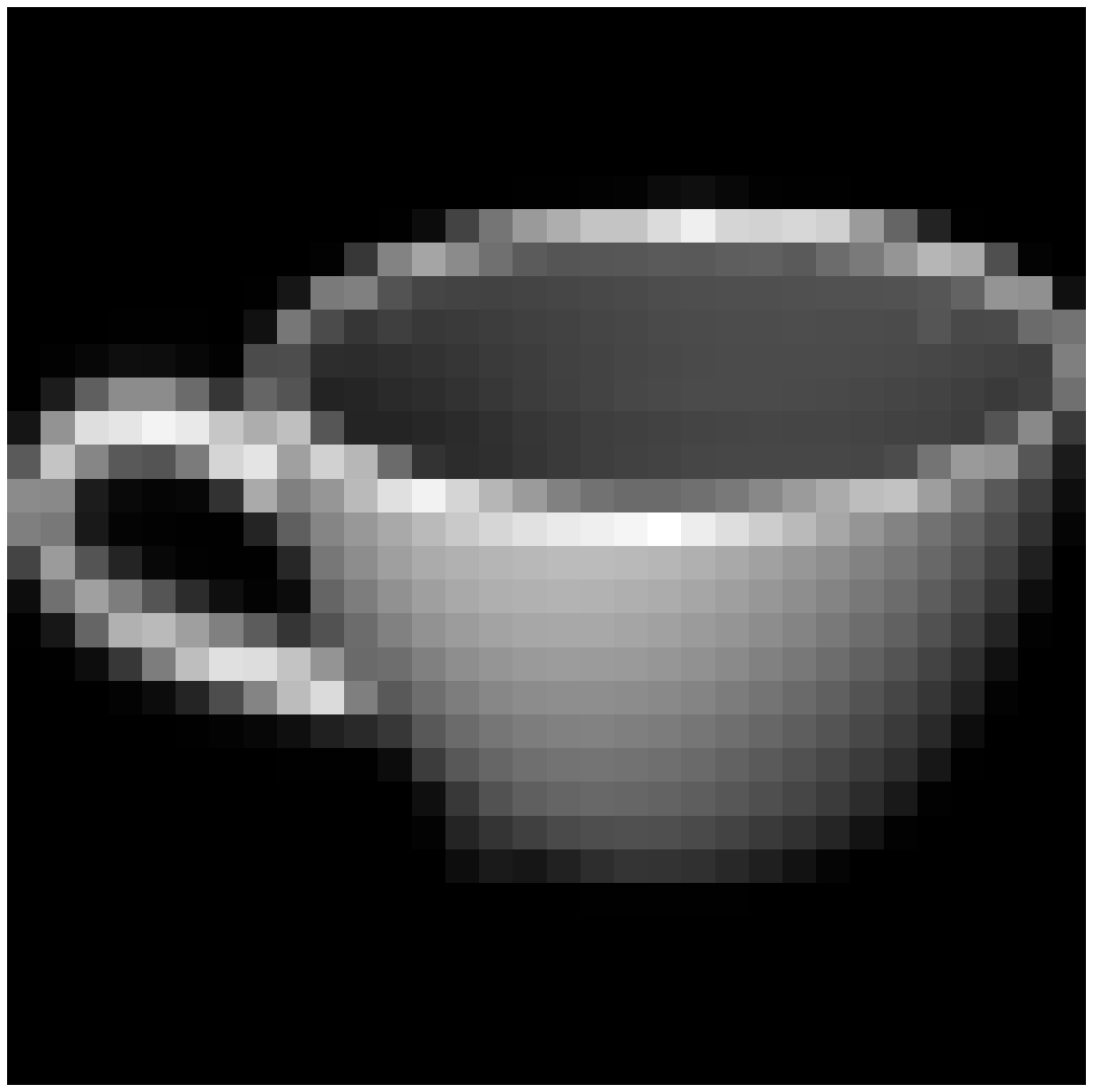} &
    \includegraphics[width=0.048\linewidth,bb=142 226 494 578,clip]{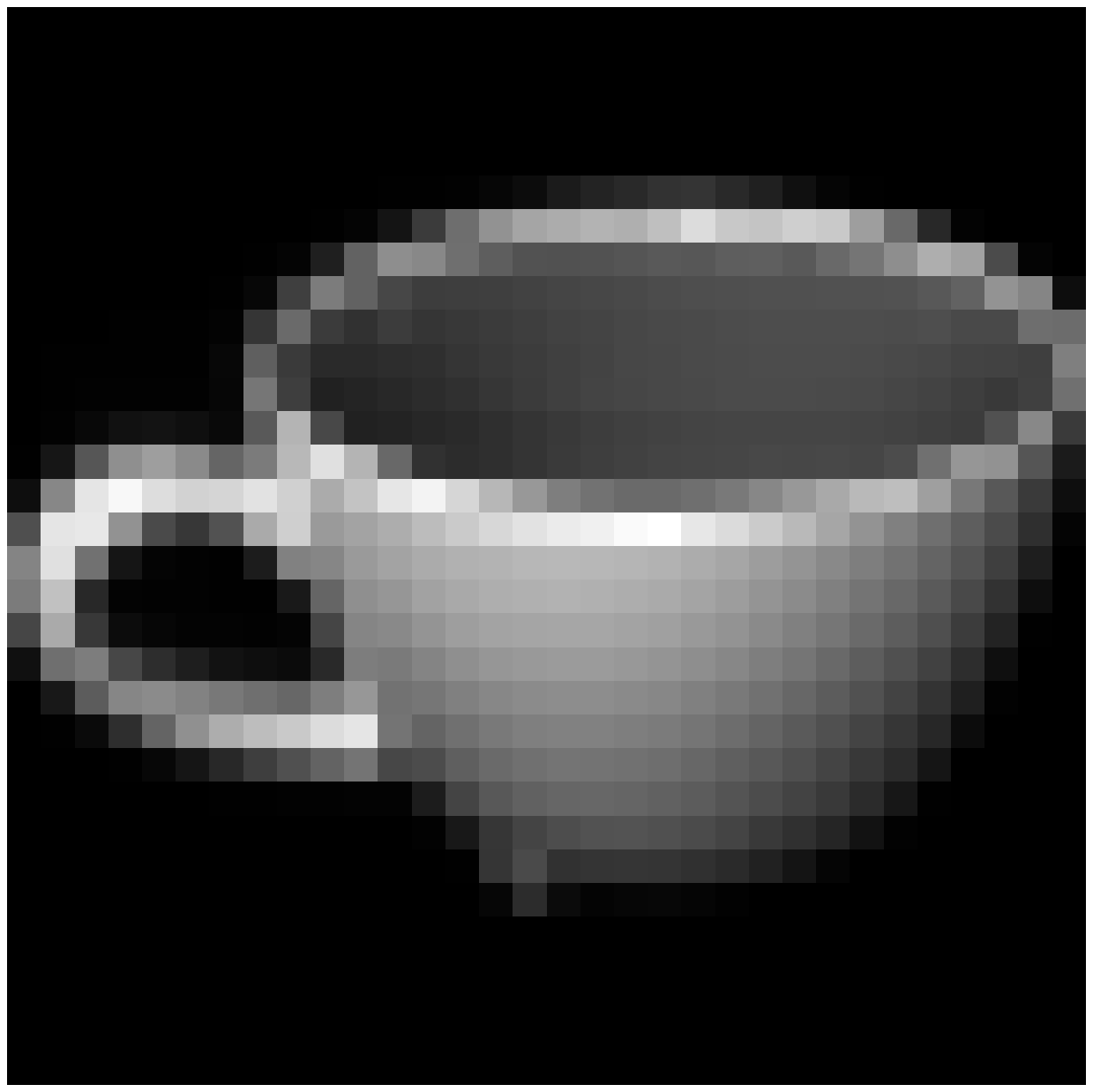} &
    \includegraphics[width=0.048\linewidth,bb=142 226 494 578,clip]{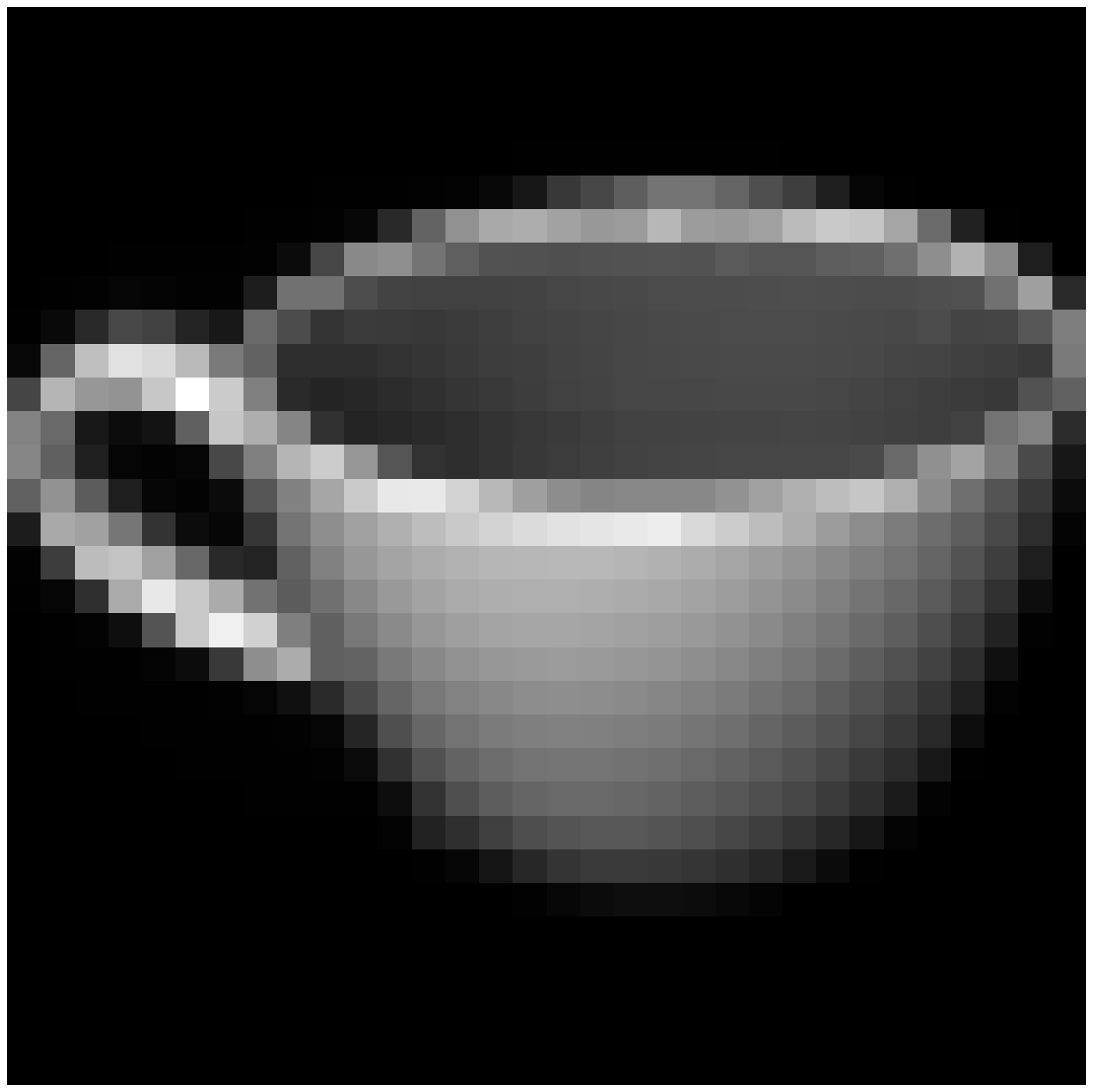} &
    \includegraphics[width=0.048\linewidth,bb=142 226 494 578,clip]{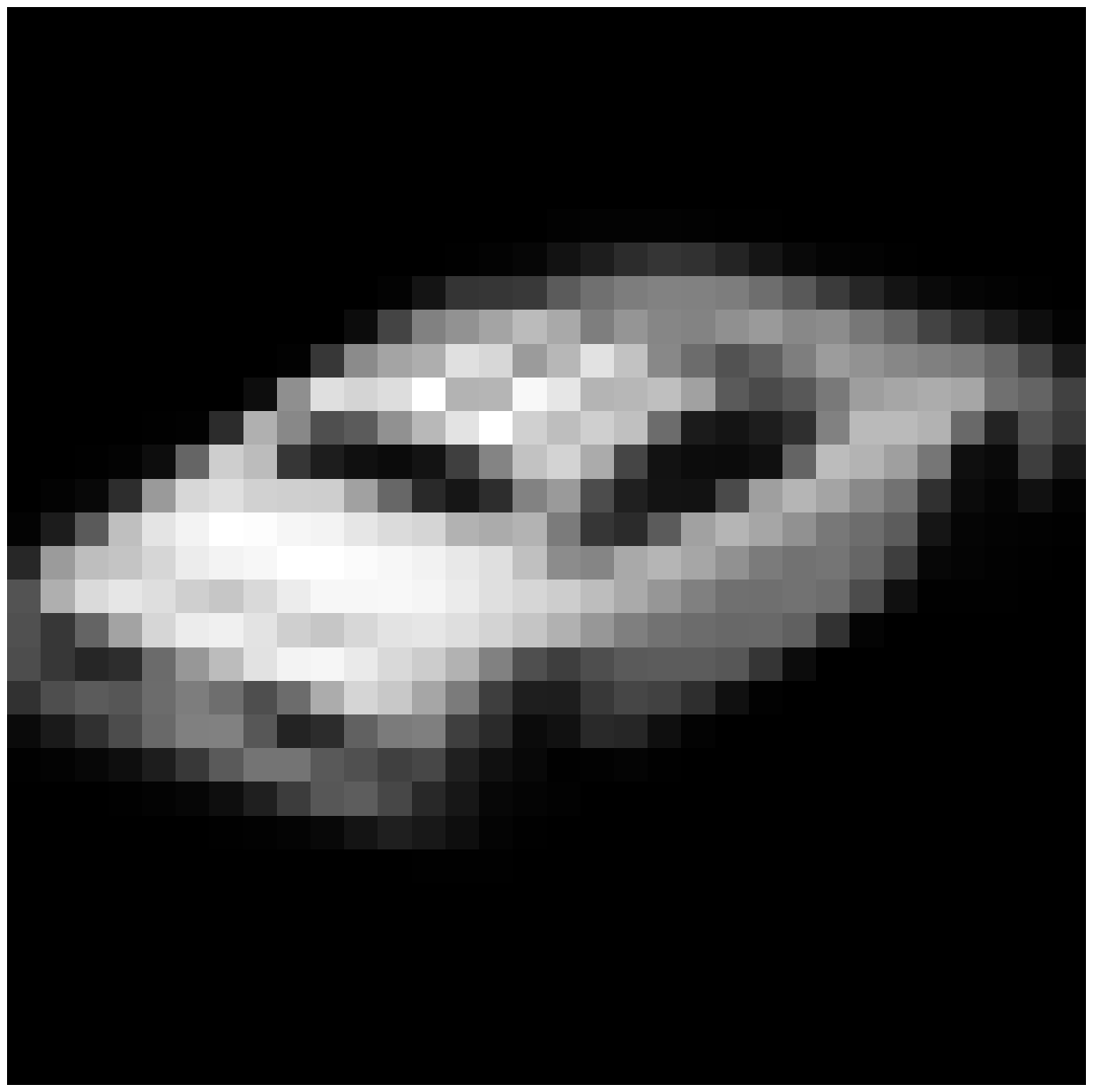} &
    \includegraphics[width=0.048\linewidth,bb=142 226 494 578,clip]{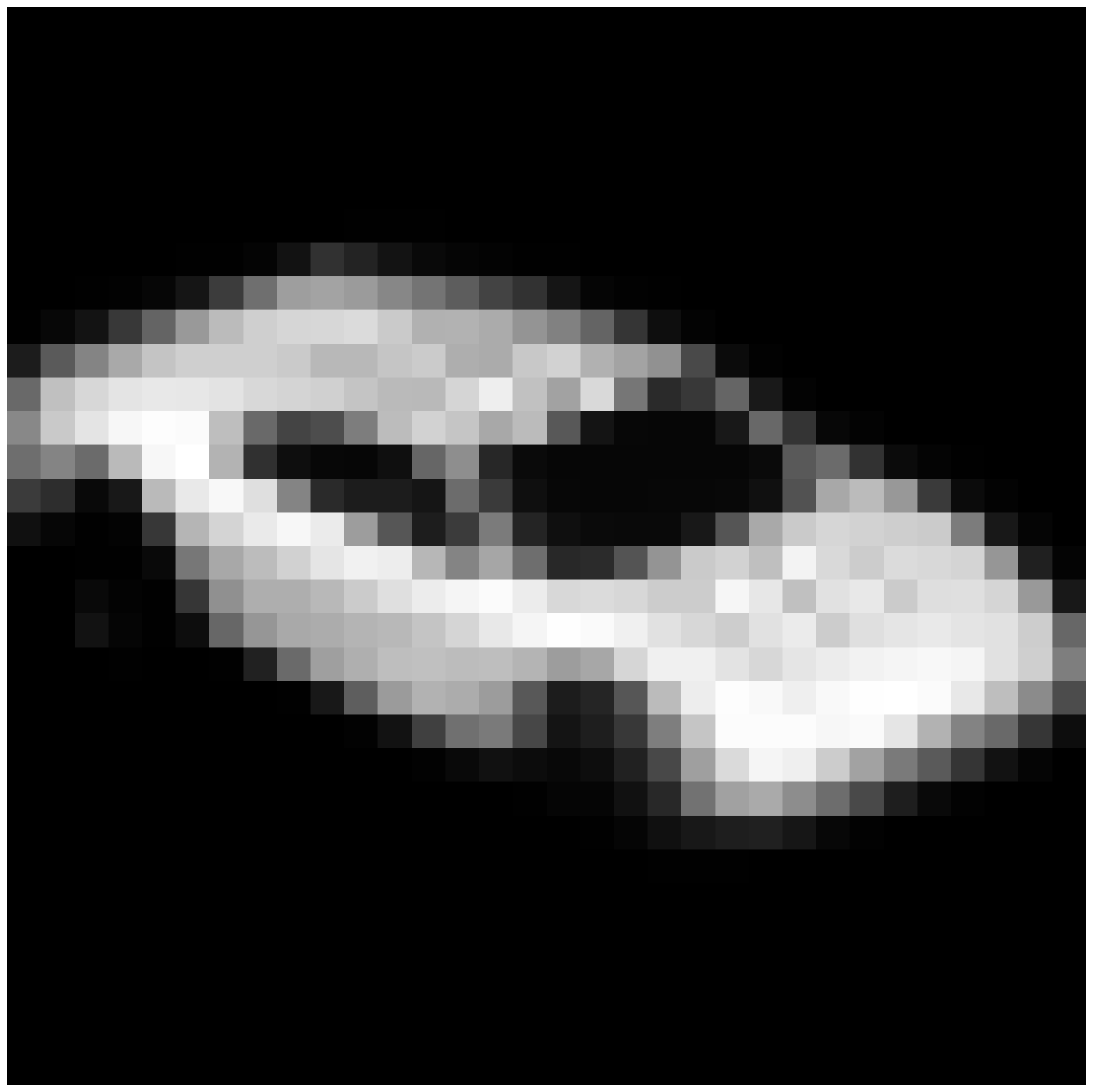}
  \end{tabular}
  \caption{Centroids found by different algorithms on COIL--20. \emph{First row}: $K$-means ($\lambda=0$, $\sigma\rightarrow\infty$, ACC: 64.8\%, NMI: 73.5\%). \emph{Second row}: $K$-modes ($\lambda=0$, $\sigma=0.3$, ACC: 65.5\%, NMI: 73.0\%). \emph{Third row}: Laplacian $K$-modes ($\lambda=0.01$, $\sigma=0.1$, ACC: 73.2\%, NMI: 83.8\%). \emph{Fourth row}: Laplacian $K$-modes with homotopy ($\lambda=0.01$, ACC: 81.5\%, NMI: 88.0\%) in $\sigma$. \emph{Fourth row}: mean-shift ($\sigma=0.1591$).}
  \label{f:coil}
\end{figure}

Another key advantage of Laplacian $K$-modes is that the centroids are interpretable patterns of the dataset. We show the centroids (each as an image) found by centroid-based algorithms (using optimal hyperparameters) on MNIST in Fig.~\ref{f:mnist} and COIL--20 in Fig.~\ref{f:coil}, all using the $K$-means initialization. For such high-dimensional problems with small $K$, GMS tends to have a majority of centroids associated with very few points that are outliers with unusual patterns, and we do not show them here. Not surprisingly, some $K$-means centroids are blurry images consisting of an average of digits/objects of different identity and style. This implies some centroids lie between different branches of data manifolds, thus in low-density areas and not prototypical. The $K$-modes centroids have cleaner shapes, but the identities of the different centroids somewhat overlap, so that some classes are represented by multiple centroids, at the expense of other classes, which may be represented by no centroids. This is because $K$-modes only takes into account the kde's, and classes with higher sample density will receive more centroids. However, this will result in a class being abruptly partitioned, and the smoothness term in Laplacian $K$-modes prevents this, so its centroids not only have prototypical shapes, but also cover more digit/object identities. On COIL-20, starting from the $K$-means solution, and by slowly varying the hyperparameter $\sigma$, the centroids gradually move to high-density areas under the homotopy algorithm, being more prototypical, while the assignments separate different objects better.

Applying our algorithm at an intermediate $\sigma$ achieves just the right amount of smoothing. This is clearly seen from the centroids obtained on MNIST. It allows the centroids to look like valid digit images, but at the same time to average out noise, unusual strokes or other idiosyncrasies of the dataset images (while not averaging digits of different identities or different styles, as $K$-means does). This yields centroids that are more representative even than individual images of the dataset.

\section{Conclusion}

Our Laplacian $K$-modes algorithm enjoys some of the best properties of a range of clustering algorithms. It is nonparametric and allows the user to work with a kernel density estimate that produces exactly $K$ clusters (as in $K$-means and $K$-modes), even in high dimension (unlike in mean-shift), and which can be nonconvex (as in mean-shift and spectral clustering). It also finds centroids that are valid patterns and lie in high-density areas, are representative of their cluster and neighborhood, yet they average out noise or idiosyncrasies that exist in individual data points. Computationally, our current alternating optimization scheme is simple, efficient and scales well. Experiments demonstrate the superior performance of Laplacian $K$-modes compared to well-known algorithms.


\begin{thebibliography}{33}
\providecommand{\natexlab}[1]{#1}
\providecommand{\url}[1]{\texttt{#1}}
\expandafter\ifx\csname urlstyle\endcsname\relax
  \providecommand{\doi}[1]{doi: #1}\else
  \providecommand{\doi}{doi: \begingroup \urlstyle{rm}\Url}\fi

\bibitem[Arora et~al.(2011)Arora, Gupta, Kapila, and Fazel]{Arora_11a}
R.~Arora, M.~Gupta, A.~Kapila, and M.~Fazel.
\newblock Clustering by left-stochastic matrix factorization.
\newblock In L.~Getoor and T.~Scheffer, editors, \emph{Proc. of the 28th Int.
  Conf. Machine Learning (ICML 2011)}, pages 761--768, Bellevue, WA, June~28 --
  July~2 2011.

\bibitem[Arthur and Vassilvitskii(2007)]{ArthurVassil07a}
D.~Arthur and S.~Vassilvitskii.
\newblock \texttt{k-means++}: The advantages of careful seeding.
\newblock In \emph{Proc. of the 18th ACM-SIAM Symposium on Discrete Algorithms
  (SODA 2007)}, pages 1027--1035, New Orleans, LA, Jan.~7--9 2007.

\bibitem[Beck and Teboulle(2009)]{BeckTeboul09a}
A.~Beck and M.~Teboulle.
\newblock A fast iterative shrinkage-thresholding algorithm for linear inverse
  problems.
\newblock \emph{SIAM J. Imaging Sciences}, 2\penalty0 (1):\penalty0 183--202,
  2009.

\bibitem[Belkin et~al.(2006)Belkin, Niyogi, and Sindhwani]{Belkin_06a}
M.~Belkin, P.~Niyogi, and V.~Sindhwani.
\newblock Manifold regularization: A geometric framework for learning from
  labeled and unlabeled examples.
\newblock \emph{J. Machine Learning Research}, 7:\penalty0 2399--2434, Nov.
  2006.

\bibitem[Bishop(2006)]{Bishop06a}
C.~M. Bishop.
\newblock \emph{Pattern Recognition and Machine Learning}.
\newblock Springer Series in Information Science and Statistics.
  Springer-Verlag, Berlin, 2006.

\bibitem[Cai et~al.(2011)Cai, He, Han, and Huang]{Cai_11b}
D.~Cai, X.~He, J.~Han, and T.~S. Huang.
\newblock Graph regularized nonnegative matrix factorization for data
  representation.
\newblock \emph{IEEE Trans. Pattern Analysis and Machine Intelligence},
  33\penalty0 (8):\penalty0 1548--1560, Aug. 2011.

\bibitem[Carreira-Perpi{\~n}{\'a}n(2000)]{Carreir00b}
M.~{\'A}. Carreira-Perpi{\~n}{\'a}n.
\newblock Mode-finding for mixtures of {Gaussian} distributions.
\newblock \emph{IEEE Trans. Pattern Analysis and Machine Intelligence},
  22\penalty0 (11):\penalty0 1318--1323, Nov. 2000.

\bibitem[Carreira-Perpi{\~n}{\'a}n(2006)]{Carreir06a}
M.~{\'A}. Carreira-Perpi{\~n}{\'a}n.
\newblock Acceleration strategies for {Gaussian} mean-shift image segmentation.
\newblock In C.~Schmid, S.~Soatto, and C.~Tomasi, editors, \emph{Proc. of the
  2006 IEEE Computer Society Conf. Computer Vision and Pattern Recognition
  (CVPR'06)}, pages 1160--1167, New York, NY, June~17--22 2006.

\bibitem[Carreira-Perpi{\~n}{\'a}n(2007)]{Carreir07a}
M.~{\'A}. Carreira-Perpi{\~n}{\'a}n.
\newblock Gaussian mean shift is an {EM} algorithm.
\newblock \emph{IEEE Trans. Pattern Analysis and Machine Intelligence},
  29\penalty0 (5):\penalty0 767--776, May 2007.

\bibitem[Carreira-Perpi{\~n}{\'a}n and Wang(2013)]{CarreirWang13a}
M.~{\'A}. Carreira-Perpi{\~n}{\'a}n and W.~Wang.
\newblock The {$K$}-modes algorithm for clustering.
\newblock Unpublished manuscript, arXiv:1304.6478, Apr.~23 2013.

\bibitem[Carreira-Perpi{\~n}{\'a}n and Zemel(2005)]{CarreirZemel05a}
M.~{\'A}. Carreira-Perpi{\~n}{\'a}n and R.~S. Zemel.
\newblock Proximity graphs for clustering and manifold learning.
\newblock In L.~K. Saul, Y.~Weiss, and L.~Bottou, editors, \emph{Advances in
  Neural Information Processing Systems (NIPS)}, volume~17, pages 225--232. MIT
  Press, Cambridge, MA, 2005.

\bibitem[Cheng(1995)]{Cheng95a}
Y.~Cheng.
\newblock Mean shift, mode seeking, and clustering.
\newblock \emph{IEEE Trans. Pattern Analysis and Machine Intelligence},
  17\penalty0 (8):\penalty0 790--799, Aug. 1995.

\bibitem[Chennubhotla and Jepson(2003)]{ChennubJepson03a}
C.~Chennubhotla and A.~Jepson.
\newblock {EigenCuts}: Half-lives of {EigenFlows} for spectral clustering.
\newblock In S.~Becker, S.~Thrun, and K.~Obermayer, editors, \emph{Advances in
  Neural Information Processing Systems (NIPS)}, volume~15, pages 705--712. MIT
  Press, Cambridge, MA, 2003.

\bibitem[Comaniciu and Meer(2002)]{ComanicMeer02a}
D.~Comaniciu and P.~Meer.
\newblock Mean shift: A robust approach toward feature space analysis.
\newblock \emph{IEEE Trans. Pattern Analysis and Machine Intelligence},
  24\penalty0 (5):\penalty0 603--619, May 2002.

\bibitem[Duchi et~al.(2008)Duchi, Shalev-Shwartz, Singer, and
  Chandra]{Duchi_08a}
J.~Duchi, S.~Shalev-Shwartz, Y.~Singer, and T.~Chandra.
\newblock Efficient projections onto the $\ell_1$-ball for learning in high
  dimensions.
\newblock In A.~{McCallum} and S.~Roweis, editors, \emph{Proc. of the 25th Int.
  Conf. Machine Learning (ICML'08)}, pages 272--279, Helsinki, Finland,
  July~5--9 2008.

\bibitem[Fukunaga and Hostetler(1975)]{FukunagHostet75a}
K.~Fukunaga and L.~D. Hostetler.
\newblock The estimation of the gradient of a density function, with
  application in pattern recognition.
\newblock \emph{IEEE Trans. Information Theory}, IT--21\penalty0 (1):\penalty0
  32--40, Jan. 1975.

\bibitem[Hastie et~al.(2009)Hastie, Tibshirani, and Friedman]{Hastie_09a}
T.~J. Hastie, R.~J. Tibshirani, and J.~H. Friedman.
\newblock \emph{The Elements of Statistical Learning---Data Mining, Inference
  and Prediction}.
\newblock Springer Series in Statistics. Springer-Verlag, second edition, 2009.

\bibitem[Hinton and Roweis(2003)]{HintonRoweis03a}
G.~Hinton and S.~T. Roweis.
\newblock Stochastic neighbor embedding.
\newblock In S.~Becker, S.~Thrun, and K.~Obermayer, editors, \emph{Advances in
  Neural Information Processing Systems (NIPS)}, volume~15, pages 857--864. MIT
  Press, Cambridge, MA, 2003.

\bibitem[Kaufman and Rousseeuw(1990)]{KaufmanRousseeuw90a}
L.~Kaufman and P.~J. Rousseeuw.
\newblock \emph{Finding Groups in Data: An Introduction to Cluster Analysis}.
\newblock Wiley Series in Probability and Mathematical Statistics. John Wiley
  \& Sons, 1990.

\bibitem[{LeCun} et~al.(1998){LeCun}, Bottou, Bengio, and Haffner]{Lecun_98a}
Y.~{LeCun}, L.~Bottou, Y.~Bengio, and P.~Haffner.
\newblock Gradient-based learning applied to document recognition.
\newblock \emph{Proc. IEEE}, 86\penalty0 (11):\penalty0 2278--2324, Nov. 1998.

\bibitem[Lee and Seung(1999)]{LeeSeung99a}
D.~D. Lee and H.~S. Seung.
\newblock Learning the parts of objects by non-negative matrix factorization.
\newblock \emph{Nature}, 401\penalty0 (6755):\penalty0 788--791, Oct.~21 1999.

\bibitem[Liu et~al.(2010)Liu, He, and Chang]{Liu_10c}
W.~Liu, J.~He, and S.-F. Chang.
\newblock Large graph construction for scalable semi-supervised learning.
\newblock In J.~F{\"u}rnkranz and T.~Joachims, editors, \emph{Proc. of the 27th
  Int. Conf. Machine Learning (ICML 2010)}, Haifa, Israel, June~21--25 2010.

\bibitem[Nene et~al.(1996)Nene, Nayar, and Murase]{Nene_96a}
S.~A. Nene, S.~K. Nayar, and H.~Murase.
\newblock Columbia object image library {(COIL-20)}.
\newblock Technical Report \mbox{CUCS--005--96}, Dept. of Computer Science,
  Columbia University, Feb. 1996.

\bibitem[Shi and Malik(2000)]{ShiMalik00a}
J.~Shi and J.~Malik.
\newblock Normalized cuts and image segmentation.
\newblock \emph{IEEE Trans. Pattern Analysis and Machine Intelligence},
  22\penalty0 (8):\penalty0 888--905, Aug. 2000.

\bibitem[Vladymyrov and Carreira-Perpi{\~n}{\'a}n(2013)]{VladymCarreir13a}
M.~Vladymyrov and M.~{\'A}. Carreira-Perpi{\~n}{\'a}n.
\newblock Entropic affinities: Properties and efficient numerical computation.
\newblock In S.~Dasgupta and D.~{McAllester}, editors, \emph{Proc. of the 30th
  Int. Conf. Machine Learning (ICML 2013)}, pages 477--485, Atlanta, GA,
  June~16--21 2013.

\bibitem[von Luxburg(2007)]{Luxbur07a}
U.~von Luxburg.
\newblock A tutorial on spectral clustering.
\newblock \emph{Statistics and Computing}, 17\penalty0 (4):\penalty0 395--416,
  Dec. 2007.

\bibitem[Wand and Jones(1994)]{WandJones94a}
M.~P. Wand and M.~C. Jones.
\newblock \emph{Kernel Smoothing}.
\newblock Number~60 in Monographs on Statistics and Applied Probability.
  Chapman \& Hall, London, New York, 1994.

\bibitem[Wang and Carreira-Perpi{\~n}{\'a}n(2013)]{WangCarreir13a}
W.~Wang and M.~{\'A}. Carreira-Perpi{\~n}{\'a}n.
\newblock Projection onto the probability simplex: An efficient algorithm with
  a simple proof, and an application.
\newblock Unpublished manuscript, arXiv:1309.1541, Sept.~3 2013.

\bibitem[Yang and Oja(2012)]{YangOja12a}
Z.~Yang and E.~Oja.
\newblock Clustering by low-rank doubly stochastic matrix decomposition.
\newblock In J.~Langford and J.~Pineau, editors, \emph{Proc. of the 29th Int.
  Conf. Machine Learning (ICML 2012)}, pages 831--838, Edinburgh, Scotland,
  June~26 -- July~1 2012.

\bibitem[Yu and Shi(2003)]{YuShi03a}
S.~X. Yu and J.~Shi.
\newblock Multiclass spectral clustering.
\newblock In \emph{Proc. 9th Int. Conf. Computer Vision (ICCV'03)}, pages
  313--319, Nice, France, Oct.~14--17 2003.

\bibitem[Yuan et~al.(2010)Yuan, Hu, and He]{Yuan_10a}
X.~Yuan, B.-G. Hu, and R.~He.
\newblock Agglomerative mean-shift clustering.
\newblock \emph{IEEE Trans. Knowledge and Data Engineering}, 24\penalty0
  (2):\penalty0 209--219, Feb. 2010.

\bibitem[Zelnik-Manor and Perona(2005)]{ZelnikPerona05a}
L.~Zelnik-Manor and P.~Perona.
\newblock Self-tuning spectral clustering.
\newblock In L.~K. Saul, Y.~Weiss, and L.~Bottou, editors, \emph{Advances in
  Neural Information Processing Systems (NIPS)}, volume~17, pages 1601--1608.
  MIT Press, Cambridge, MA, 2005.

\bibitem[Zhu et~al.(2003)Zhu, Ghahramani, and Lafferty]{Zhu_03a}
X.~Zhu, Z.~Ghahramani, and J.~Lafferty.
\newblock Semi-supervised learning using {Gaussian} fields and harmonic
  functions.
\newblock In T.~Fawcett and N.~Mishra, editors, \emph{Proc. of the 20th Int.
  Conf. Machine Learning (ICML'03)}, pages 912--919, Washington, DC,
  Aug.~21--24 2003.

\end{thebibliography}

\end{document}